\documentclass[runningheads]{llncs}

\usepackage[mobile]{eccv}

\usepackage{eccvabbrv}

\usepackage{graphicx}
\usepackage{booktabs}

\usepackage{amsmath,amsfonts}
\usepackage{algorithmic}
\usepackage{algorithm}
\usepackage{array}

\newcolumntype{C}[1]{>{\centering\arraybackslash}m{#1}}
\usepackage{textcomp}
\usepackage{stfloats}
\usepackage{url}
\usepackage{verbatim}
\usepackage{cite}

\usepackage{paralist}
\usepackage{amssymb}
\usepackage{bbm}
\usepackage{multicol}
\usepackage{caption}
\usepackage{subcaption}
\usepackage{tabularx}
\usepackage{xcolor}
\usepackage{multirow}
\usepackage{float}
\usepackage{forest}
\usepackage{makecell}

\usepackage[pagebackref]{hyperref}
\hypersetup{
  pdftitle={Moving6DPoSe: A Multimodal Database for Monocular 6D Pose Estimation and Segmentation of Moving Objects},
  pdfauthor={I. Bugueno-Cordova, J. Ruiz-del-Solar, R. Verschae},
  pdfsubject={Computer Vision and Pattern Recognition (cs.CV)},%
  pdfkeywords={Moving Object Perception, Multimodal Database, 6D Pose Estimation, Event-based Vision, Neuromorphic Vision},%
}

\usepackage[accsupp]{axessibility}  

\usepackage{hyperref}

\usepackage{orcidlink}

\begin{document}

\title{Moving6DPoSe: A Multimodal Database for Monocular 6D Pose Estimation and Segmentation of Moving Objects}

\titlerunning{Abbreviated paper title}

\author{Ignacio Bugueno-Cordova\inst{1,3}\orcidlink{0000-0002-0133-0330} \and
Javier Ruiz-del-Solar\inst{1,2}\orcidlink{0000-0003-2965-633X} \and
Rodrigo Verschae\inst{3}\orcidlink{0000-0002-1661-3309}}

\authorrunning{I. Bugueno-Cordova et al.}

\institute{
Department of Electrical Engineering, University of Chile, Santiago, Chile 
\and
Advanced Mining Technology Center (AMTC), University of Chile, Santiago, Chile 
\and
The Iniciativa de Datos e Inteligencia Artificial, University of Chile, Santiago, Chile\\
\email{
i.bugueno@ieee.org,
jruizd@ing.uchile.cl, 
rodrigo@verschae.org}
}

\maketitle

\begin{abstract}
Estimating the 6D pose of moving objects remains challenging due to motion blur and the limited temporal resolution of conventional frame-based cameras. Existing event-based datasets further provide limited sensing modalities, annotations, and motion scenarios. We introduce Moving6DPoSe, a multimodal database comprising two complementary subsets: Moving6DPoSe-R with real-world recordings and Moving6DPoSe-S with synthetic sequences generated from the same objects. The dataset contains 16 scanned objects and 1,702 real and synthetic rosbags spanning multiple motion scenarios, with annotations for semantic segmentation, object detection, and monocular 6D pose estimation. We further provide baseline results for all three tasks across frame and event-based modalities. Experimental results show that event-based representations achieve more robust moving-object segmentation than conventional RGB images, while monocular orientation estimation remains challenging, highlighting the potential of Moving6DPoSe for moving-object perception research.
\end{abstract}

\section{Introduction}
\label{sec:introduction}

Estimating the six-degree-of-freedom (6D) pose of moving objects remains challenging due to motion blur, large appearance changes, and the limited temporal resolution of conventional frame-based cameras. Accurate pose estimation is required for applications such as robotic manipulation~\cite{xie2026dynamicvla}, high-speed robotic table tennis~\cite{durr2026ace}, and human-robot handover~\cite{wang2022evcatcher}. Although recent monocular and RGB-D methods achieve strong performance for static objects~\cite{xiang2018posecnn,labbe2022megapose,bundlesdfwen2023,foundationposewen2024}, they remain less effective under fast object motion~\cite{hoque2021,he2021,marullo2023,guan2024survey,ordoumpozanis2025}.

Event cameras provide high temporal resolution, high dynamic range, and asynchronous sensing, making them well suited to dynamic scenes~\cite{delbruck2008,gallego2019wf}. Recent event-based datasets support tasks such as grasping, tracking, and 6D pose estimation~\cite{li2020-egrasping,cao2022-neurograsp,hay2025pose,kang2026event6d,li2026eventbasedmotionappearance}. However, they remain limited in sensing modalities, annotations, or motion diversity.

We present \texttt{Moving6DPoSe}, a multimodal database for semantic segmentation, object detection, and monocular 6D pose estimation of moving objects. The database contains paired real (Moving6DPoSe-R) and synthetic (Moving6DPoSe-S) subsets built from the same 16 scanned objects. It provides synchronized RGB, stereo RGB, and event data, together with 3D object models, calibration parameters, and annotations for all three tasks.

The main contributions are: 
\begin{inparaenum}[(i)]
\item a multimodal database for moving-object segmentation, detection, and monocular 6D pose estimation;
\item paired real and synthetic subsets built from the same 16 scanned objects;
\item a sensing platform combining RGB, stereo RGB, and dual-resolution event cameras; and
\item ground-truth annotations, scanned 3D object models, and baseline results for all tasks.
\end{inparaenum}
The database, calibration parameters, and baseline implementations will be publicly released.

\section{Related work}
\label{sec:related-work}

Event-based vision has recently attracted increasing attention for 6D object pose estimation and tracking, particularly in scenarios involving high-speed motion. A few event-based datasets have been proposed, differing in sensing modalities, application domains, and motion characteristics.

E-Grasping~\cite{li2020-egrasping} is designed for robotic grasp detection and includes 91 objects captured using a Dynamic and Active Pixel Vision Sensor (DAVIS) mounted on a robotic gripper. NeuroGrasp~\cite{cao2022-neurograsp} extends this concept by introducing the first multimodal RGB-Event grasping dataset containing moving objects. More recently, E-POSE~\cite{hay2025pose} provides 306 annotated sequences for event-based 6D pose estimation using 13 YCB objects under different motion speeds, clutter levels, and illumination conditions. Event6D~\cite{kang2026event6d} and Li \etal~\cite{li2026eventbasedmotionappearance} address event-based 6D pose tracking using synthetic and real sequences. MTevent~\cite{awasthi2025mtevent} targets long-range robotic perception with annotations for 6D pose estimation and moving-object detection.

Table~\ref{tab:6d_pose_databases} summarizes the main differences among existing event-based datasets. Moving6DPoSe provides paired real and synthetic subsets of household objects captured under free-fall, pendular, and throwing motions, together with annotations for segmentation, detection, and 6D pose estimation, scanned 3D object models, and synchronized RGB, stereo RGB, and event data.

\begin{table}[!h]
\centering
\caption{Summary of event-based databases and benchmarks related to object grasping, tracking, and 6D pose estimation.}
\label{tab:6d_pose_databases}
\begin{tabular}{|l|c|c|c|c|c|}
\hline
\textbf{Database} & \textbf{Cite} & \textbf{Year} & \textbf{Sensors} & \textbf{Tasks} & \textbf{Objects} \\
\hline
E-Grasping & \cite{li2020-egrasping} & 2020 & Events & Grasp detection & 91 \\
\hline
NeuroGrasp & \cite{cao2022-neurograsp} & 2022 & RGB, Events & Grasp pose estimation & 154 \\
\hline
E-POSE & \cite{hay2025pose} & 2025 & Events & 6D pose estimation & 13 \\
\hline
MTevent & \cite{awasthi2025mtevent} & 2025 & RGB, Stereo Events & \makecell[t]{Moving object detection\\6D pose estimation} & 16 \\
\hline
Li et al. & \cite{li2026eventbasedmotionappearance} & 2026 & RGB-D, Events & 6D pose tracking & 14 \\
\hline
Event6D & \cite{kang2026event6d} & 2026 & RGB-D, Events & 6D pose tracking & 4 \\
\hline
Moving6DPoSe &
\makecell[t]{Ours} &
\makecell[t]{2026} &
\makecell[t]{RGB\\Stereo RGB (Depth)\\Events} &
\makecell[t]{Segmentation\\Detection\\3D reconstruction\\6D pose estimation} &
\makecell[t]{16} \\
\hline
\end{tabular}
\end{table}

\section{Data generation principles of frame and event sensors}
\label{sec:preliminaries}

\subsubsection{Frame-based cameras} 
These sensors capture absolute-intensity images at fixed intervals, producing sequences sampled uniformly over time. Each frame integrates the illumination of the scene during the exposure period, which can introduce motion blur for fast-moving objects. The data can be represented as:
$
    \mathcal{F}(t_N) = \{ f_k \}_{k=0}^{N}, \quad f_k = I(x,y,t_k),
$
where $I(x,y,t_k)$ denotes the intensity image in pixel $(x,y)$ and time $t_k = t_0 + k\Delta t_{frame}$, with $\Delta t_{frame}$ being the fixed sampling interval.

\subsubsection{Event-based cameras} 
Unlike frame-based sensors, event cameras capture changes in brightness asynchronously at each pixel \cite{delbruck2008,gallego2019wf}. An event is triggered when the logarithmic intensity change exceeds a threshold, enabling microsecond temporal resolution and low latency with small motion blur. Events are defined as follows:
$
    \mathcal{E}(t_N) = \{ e_k \}_{k=1}^{N} = \{ (x_k, y_k, t_k, p_k) \}_{k=1}^{N},
$
where $(x_k, y_k)$ are the coordinates of the pixels, $t_k$ is the timestamp, and $p_k \in \{ -1, +1 \}$ indicates the polarity (sign) of the brightness change.

\subsubsection{Active Pixel Sensor (APS)}
\begin{figure}[!tb]
    \centering
    \includegraphics[width=0.6\linewidth]{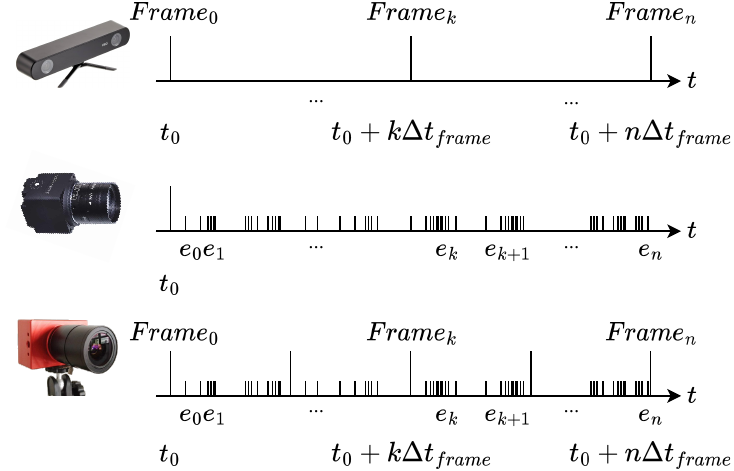}
    \caption{Illustration of data generation in frame and event-based vision sensors. Top: frame-based camera (ROG Eye S and ZED-2); Middle: event-based camera (Prophesee EVK4); Bottom: event and frame-based camera (DAVIS346).}
    \label{fig:frames_events_times}
\end{figure}

This hybrid sensor integrates both modes by combining a conventional frame-based active pixel sensor with an event-based sensor within the same pixel array \cite{gallego2019wf}. The data output can be modeled as
$
    \mathcal{H}(t_N) = \{ f_k, e_k \}_{k=1}^{N},
$
where $f_k$ represents the grayscale frames and $e_k$ asynchronous events. 
Figure~\ref{fig:frames_events_times} illustrates the data generation principles of each sensor.

\section{Moving6DPoSe Database}
\label{sec:database}
Moving6DPoSe is a multimodal database for moving-object analysis, supporting segmentation, detection, and 6D pose estimation. It comprises two complementary subsets: Moving6DPoSe-R, containing real-world recordings, and Moving6DPoSe-S, containing synthetic sequences generated from the same object models.

\begin{figure}[!ht]
    \centering
    \includegraphics[width=\linewidth]{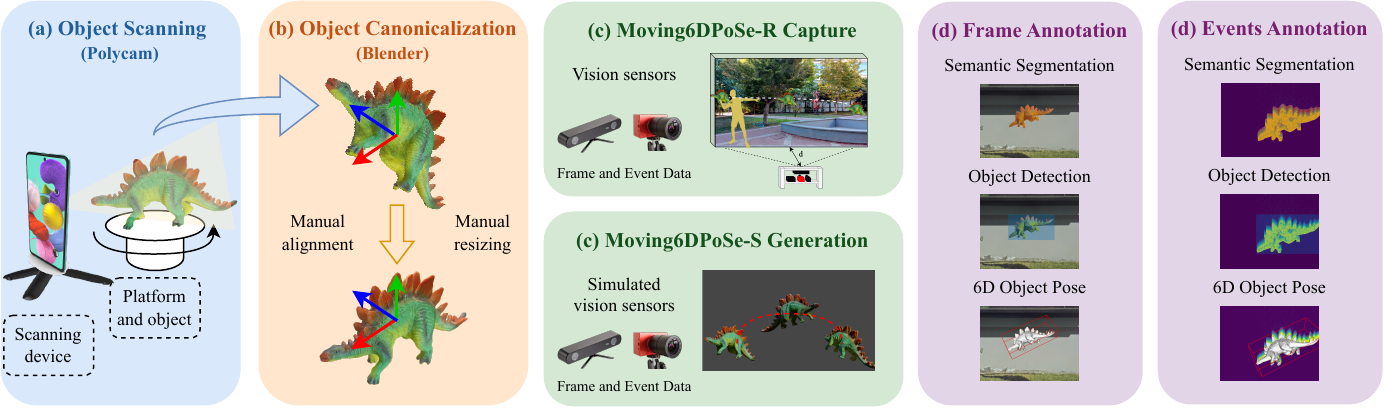}
    
    \caption{Moving6DPoSe database collection and annotation. (a) Object scanning; (b) Object canonicalization, involving the alignment of each object category to the canonical space; (c) Moving6DPoSe-R dataset experimental capture and Moving6DPoSe-S dataset synthetic generation; and (d) Frame and event-based data annotation, using different labeling tools for segmentation, detection, and 6D pose of moving objects.} 
    \label{fig:database_proposal}
\end{figure}

As illustrated in Figure~\ref{fig:database_proposal}, the database construction pipeline consists of four stages: 3D object scanning, model canonicalization, real and synthetic data acquisition, and annotation generation.

\subsection{Objects}
\label{sec:objects}

\begin{figure*}[!t]
\centering
    \subcaptionbox*{}
    [.10\textwidth]{\includegraphics[width=\linewidth]{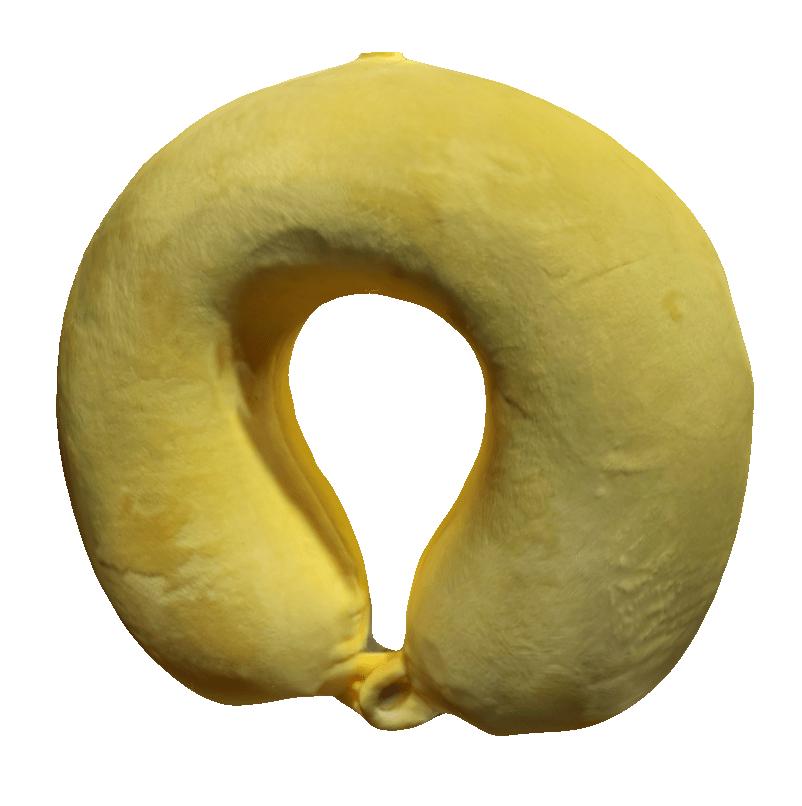} }
    \subcaptionbox*{}
    [.10\textwidth]{\includegraphics[width=\linewidth]{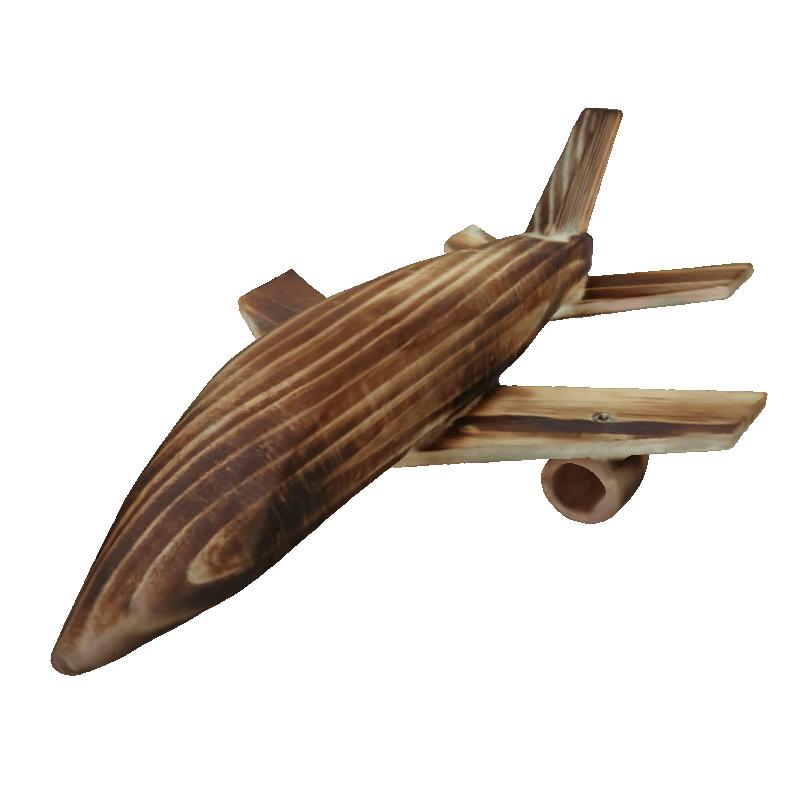} }
    \subcaptionbox*{}
    [.10\textwidth]{\includegraphics[width=\linewidth]{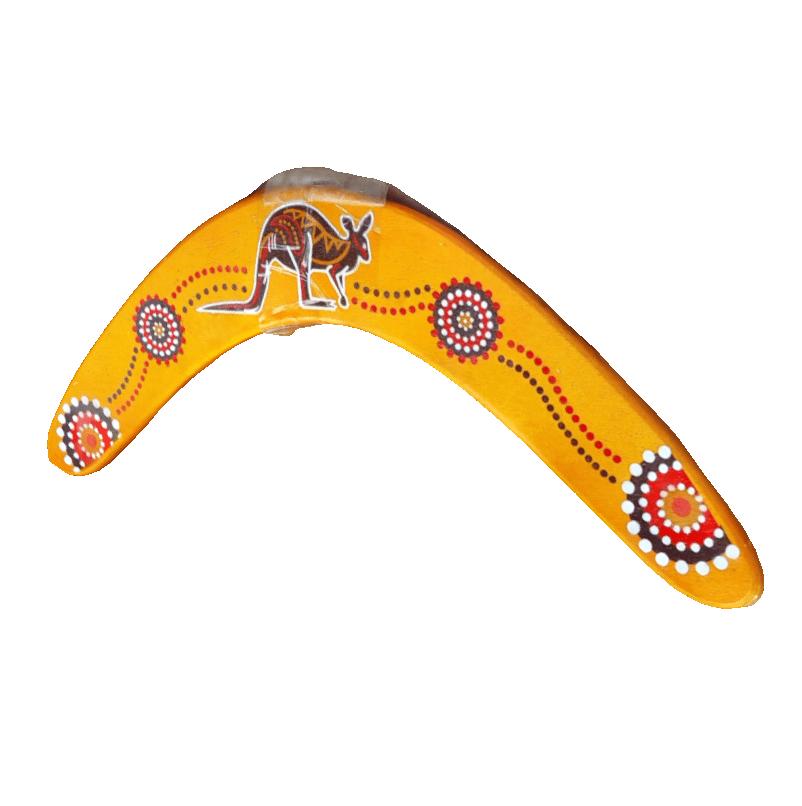} }
    \subcaptionbox*{}
    [.10\textwidth]{\includegraphics[width=\linewidth]{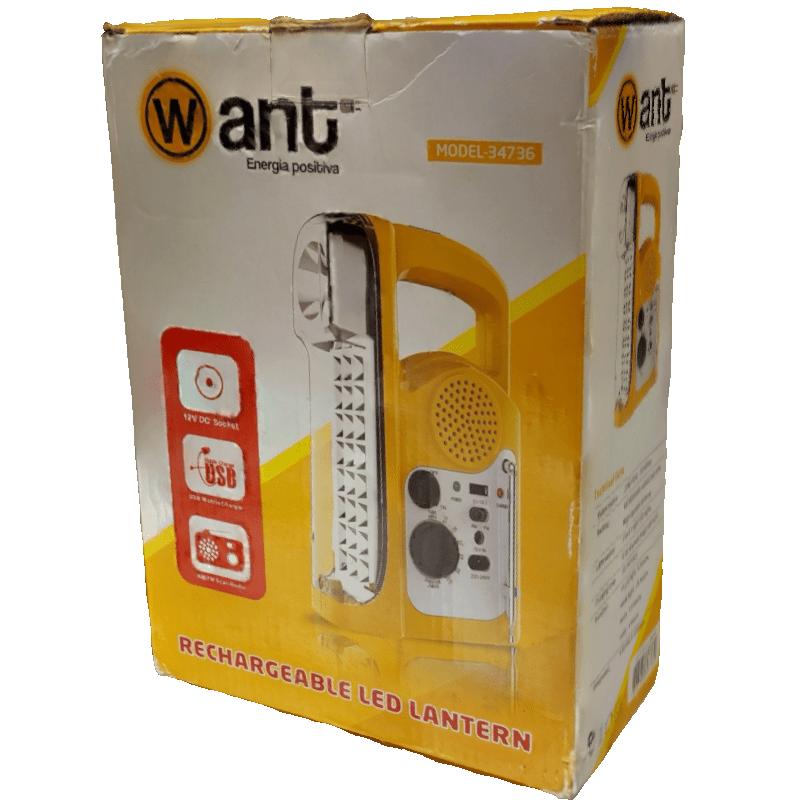} }
    \subcaptionbox*{}
    [.10\textwidth]{\includegraphics[width=\linewidth]{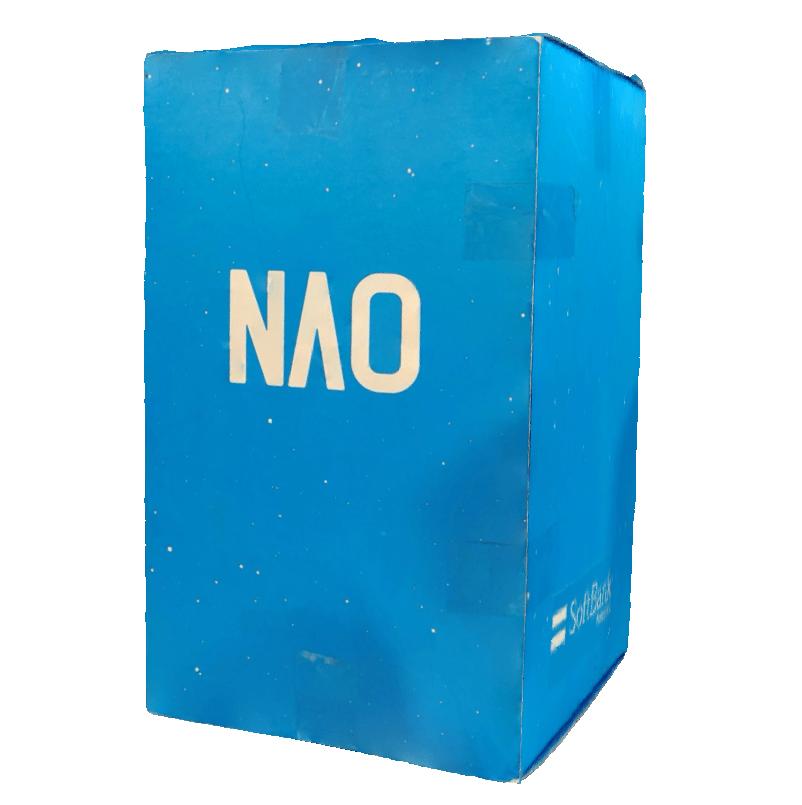} }
    \subcaptionbox*{}
    [.10\textwidth]{\includegraphics[width=\linewidth]{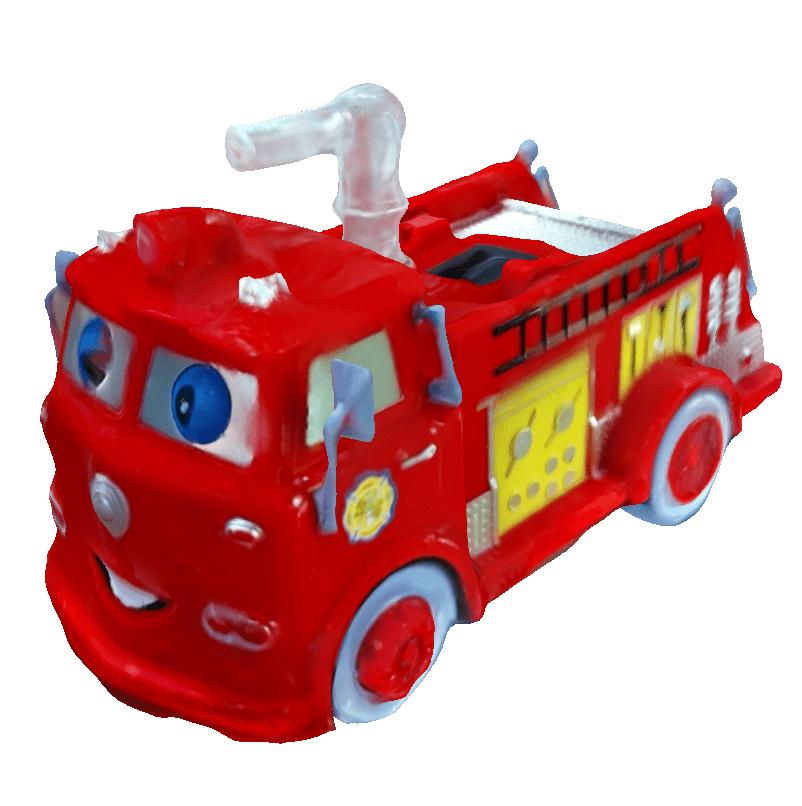} }
    \subcaptionbox*{}
    [.10\textwidth]{\includegraphics[width=\linewidth]{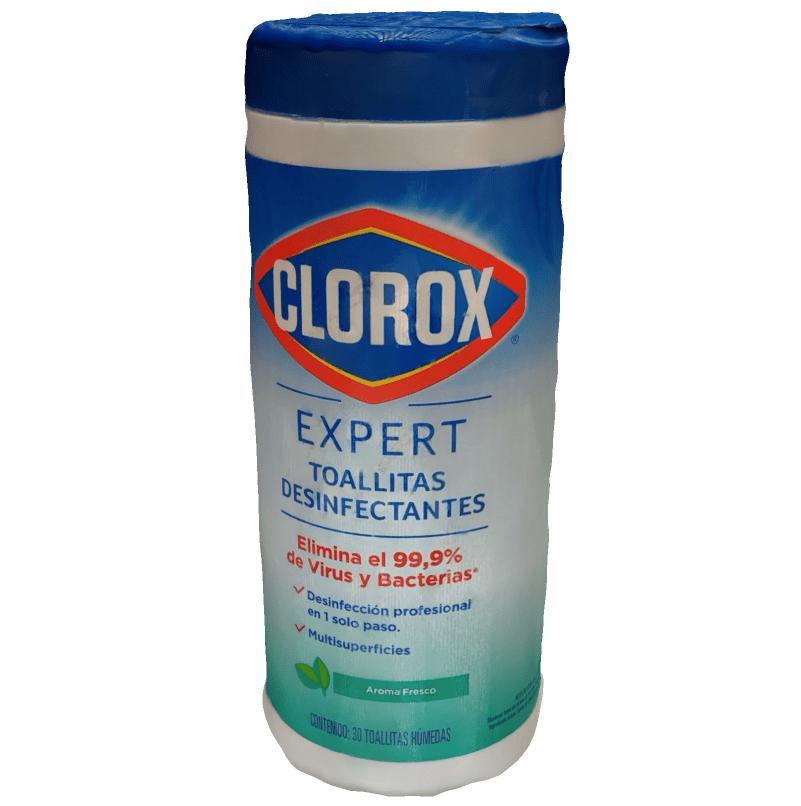} }
    \subcaptionbox*{}
    [.10\textwidth]{\includegraphics[width=\linewidth]{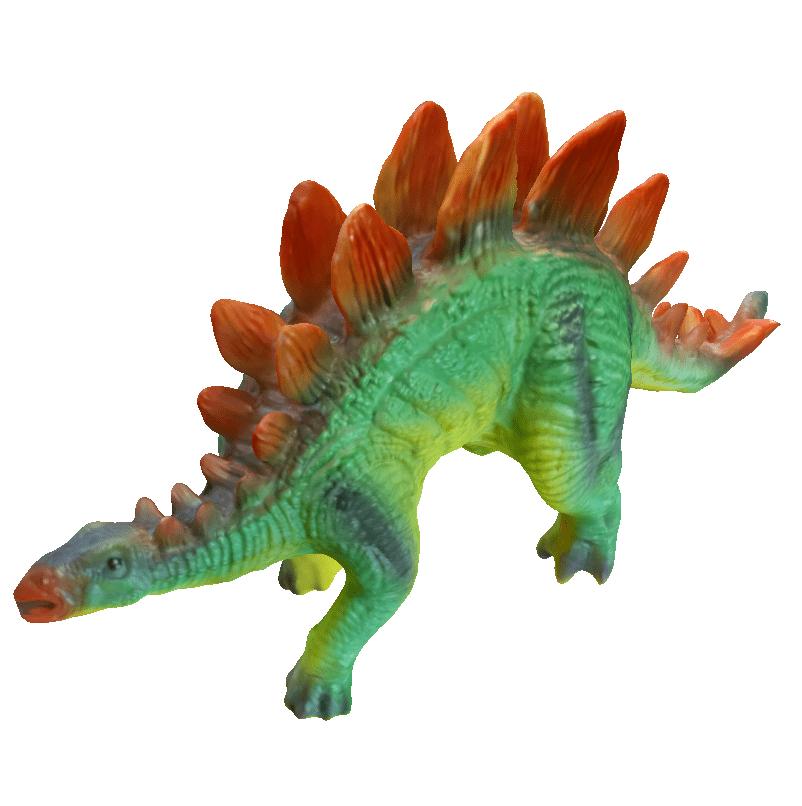} }
    
    \subcaptionbox*{}
    [.10\textwidth]{\includegraphics[width=\linewidth]{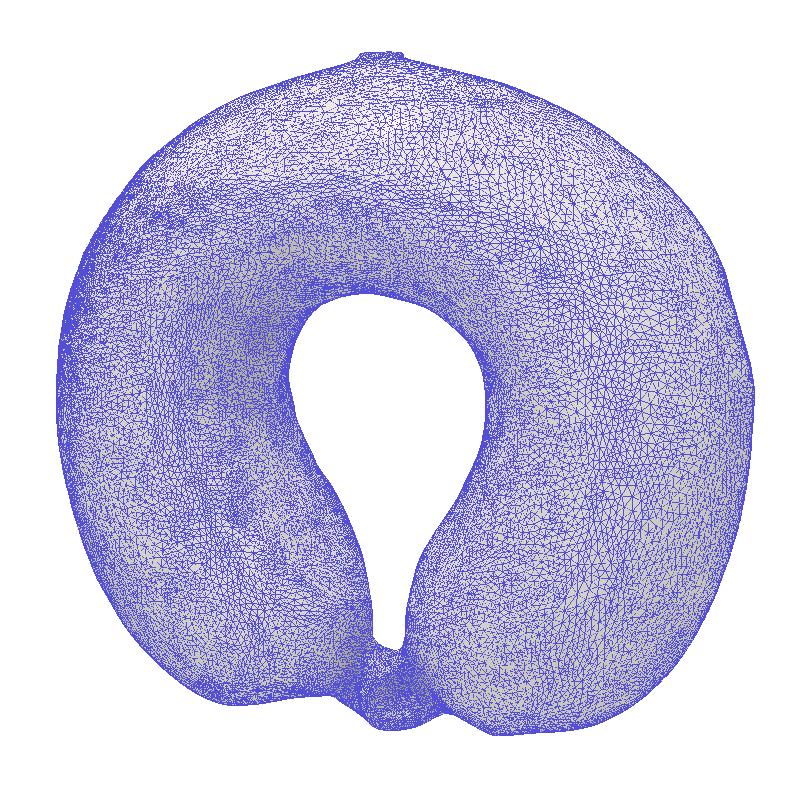} }
    \subcaptionbox*{}
    [.10\textwidth]{\includegraphics[width=\linewidth]{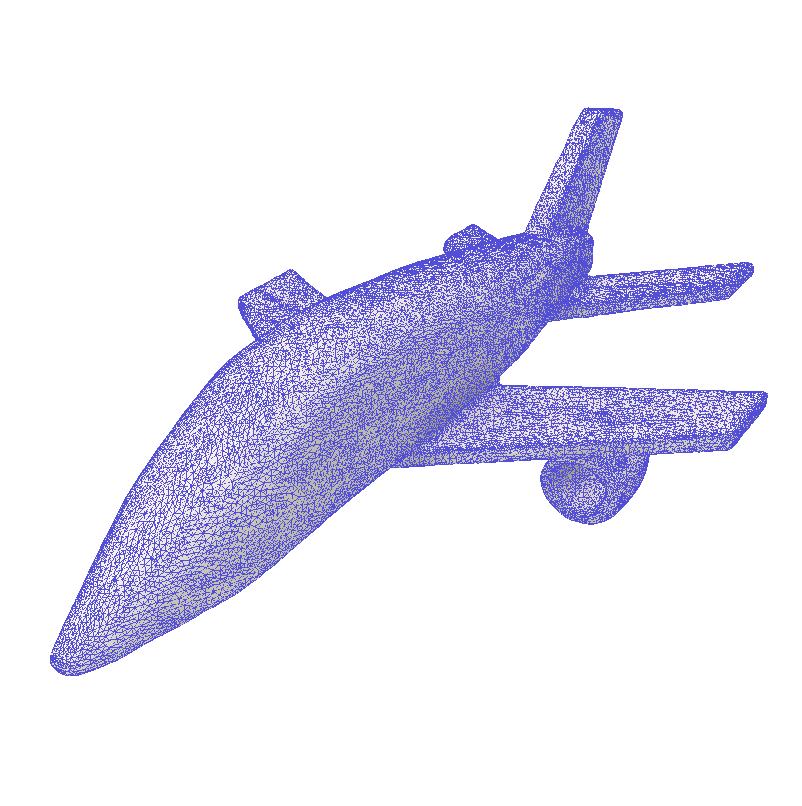} }
    \subcaptionbox*{}
    [.10\textwidth]{\includegraphics[width=\linewidth]{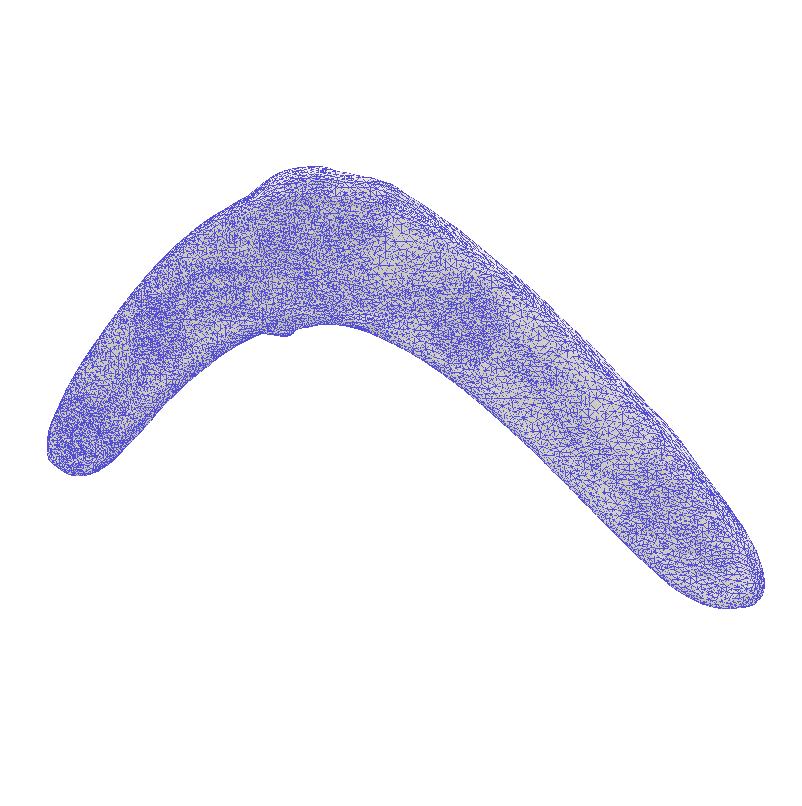} }
    \subcaptionbox*{}
    [.10\textwidth]{\includegraphics[width=\linewidth]{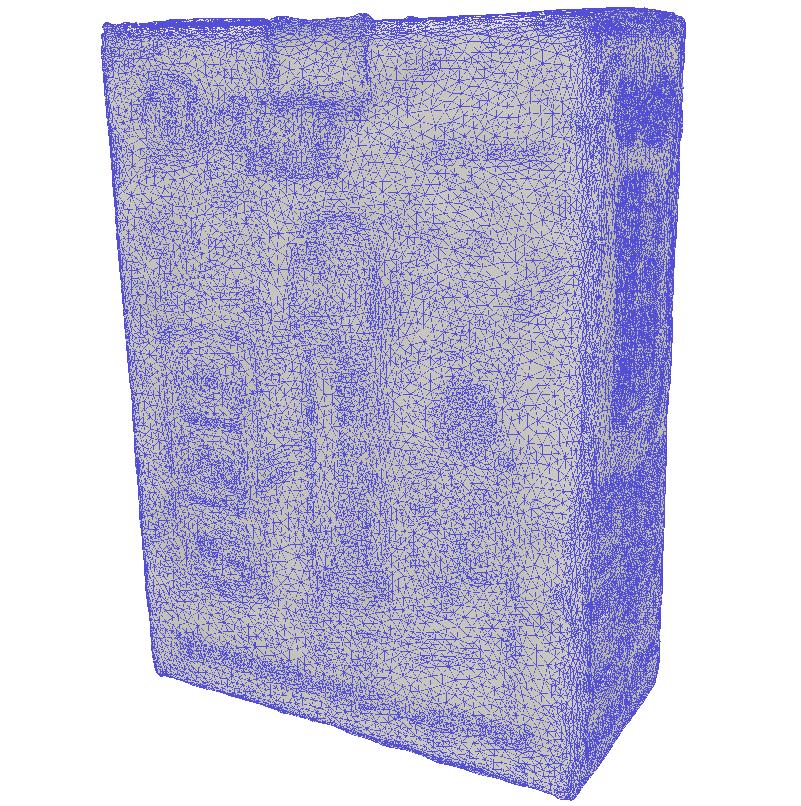} }
    \subcaptionbox*{}
    [.10\textwidth]{\includegraphics[width=\linewidth]{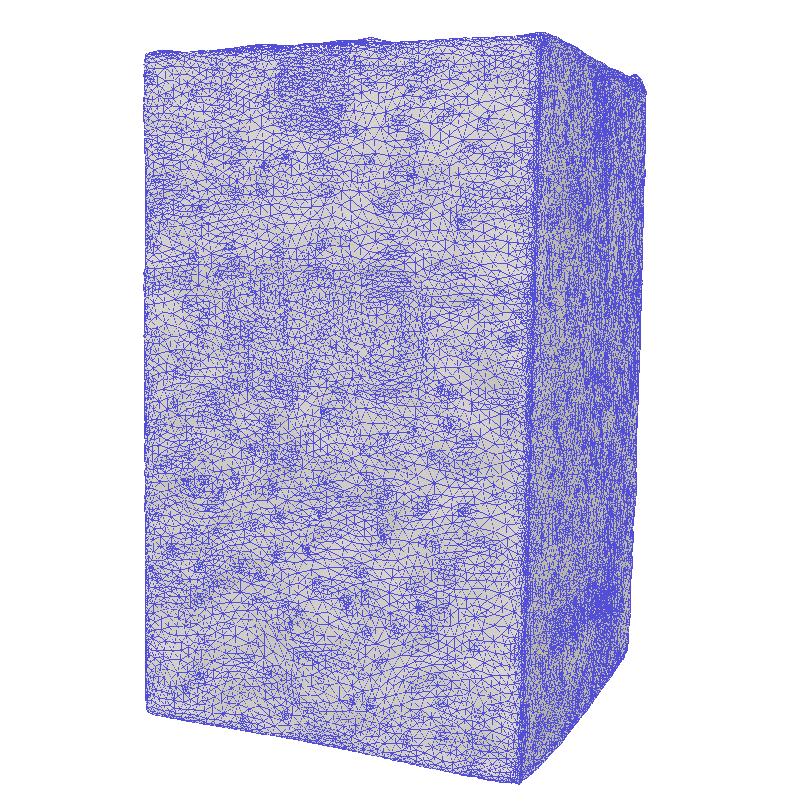} }
    \subcaptionbox*{}
    [.10\textwidth]{\includegraphics[width=\linewidth]{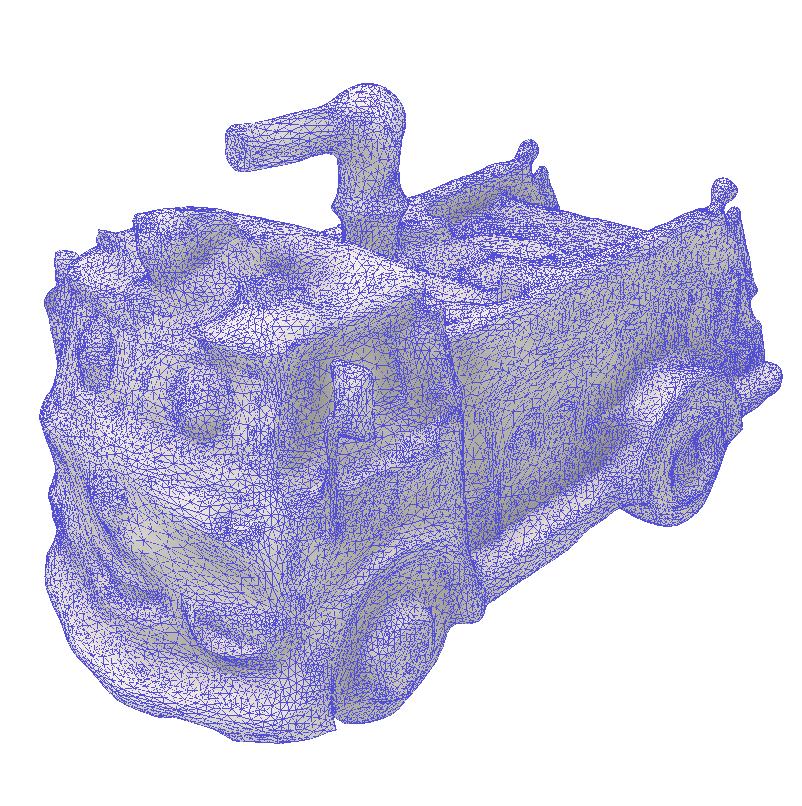} }
    \subcaptionbox*{}
    [.10\textwidth]{\includegraphics[width=\linewidth]{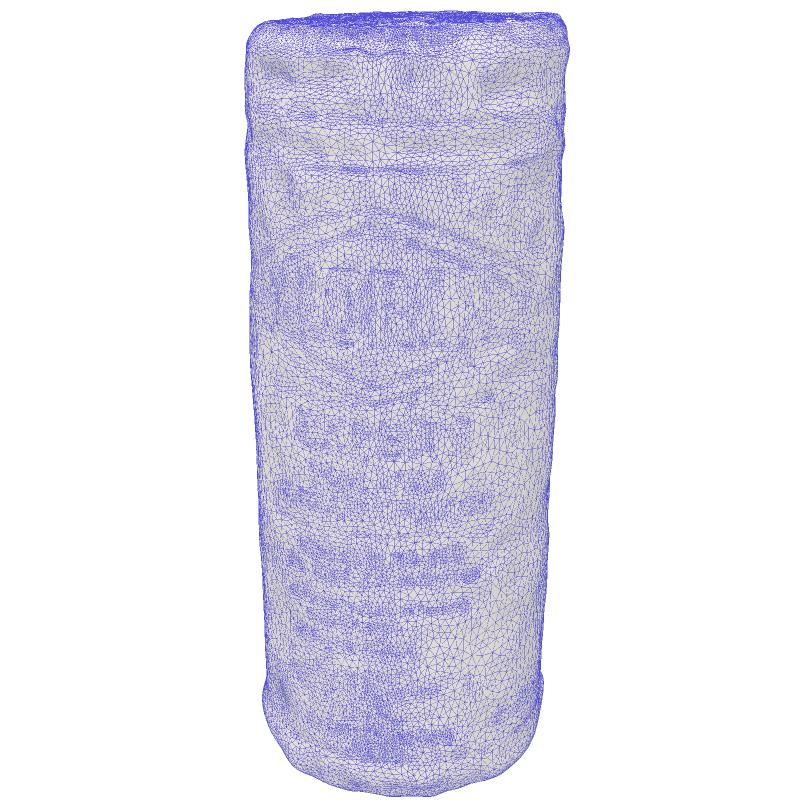} }
    \subcaptionbox*{}
    [.10\textwidth]{\includegraphics[width=\linewidth]{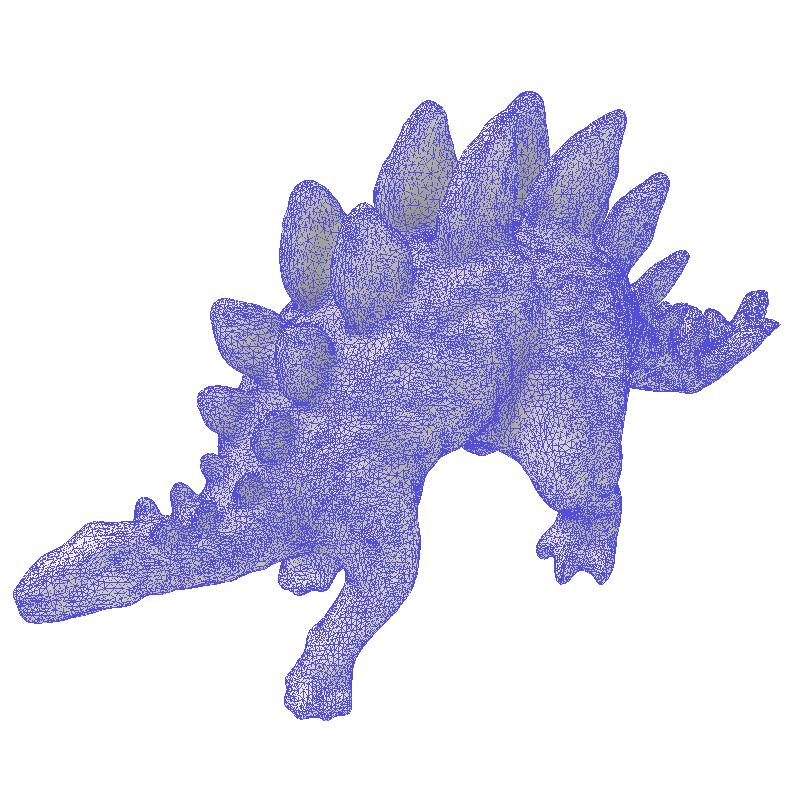} }

    \subcaptionbox*{}
    [.10\textwidth]{\includegraphics[width=\linewidth]{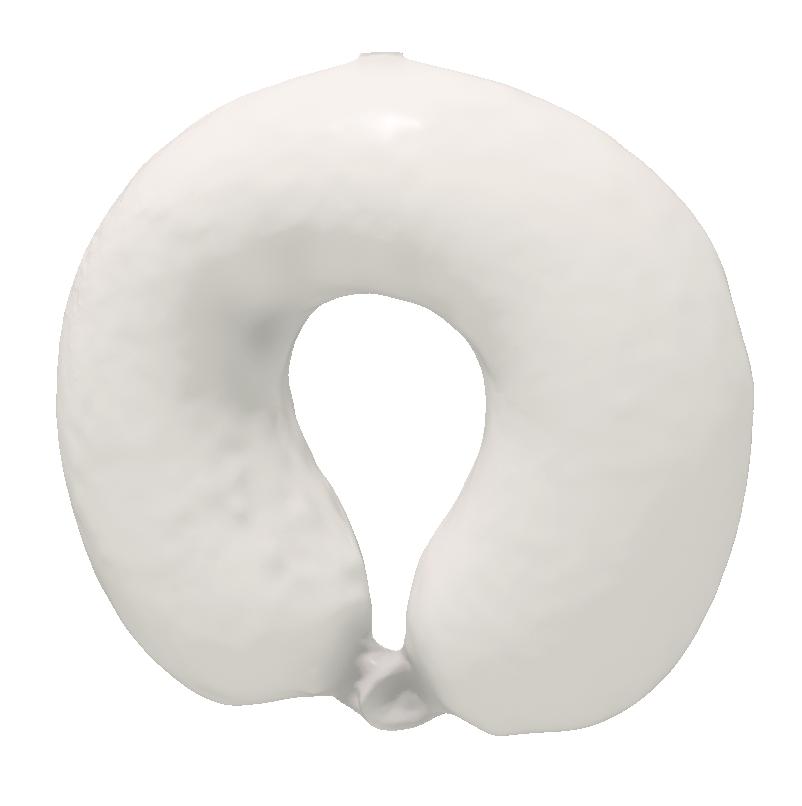} }
    \subcaptionbox*{}
    [.10\textwidth]{\includegraphics[width=\linewidth]{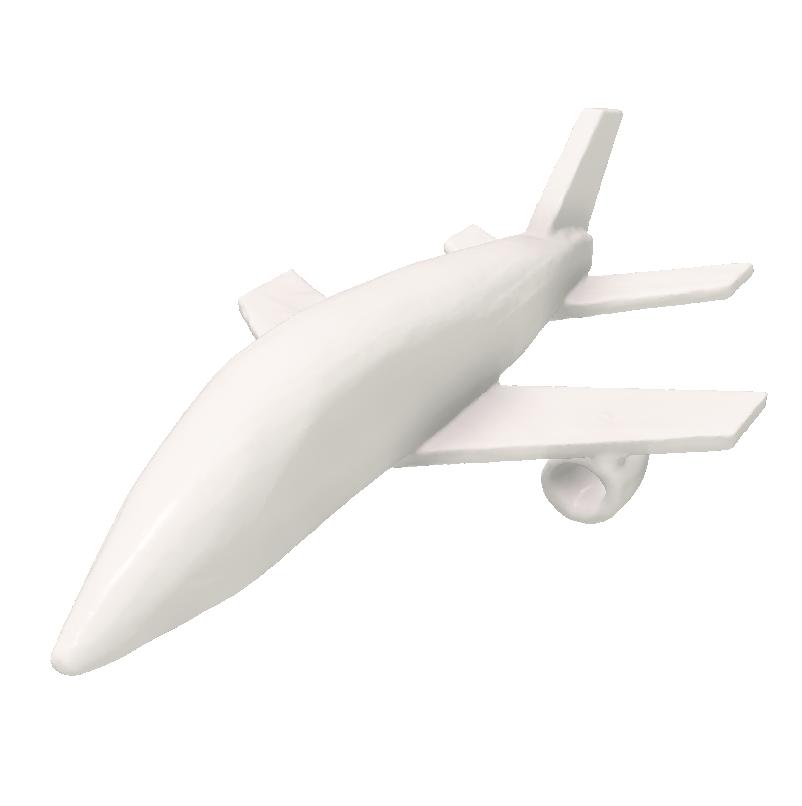} }
    \subcaptionbox*{}
    [.10\textwidth]{\includegraphics[width=\linewidth]{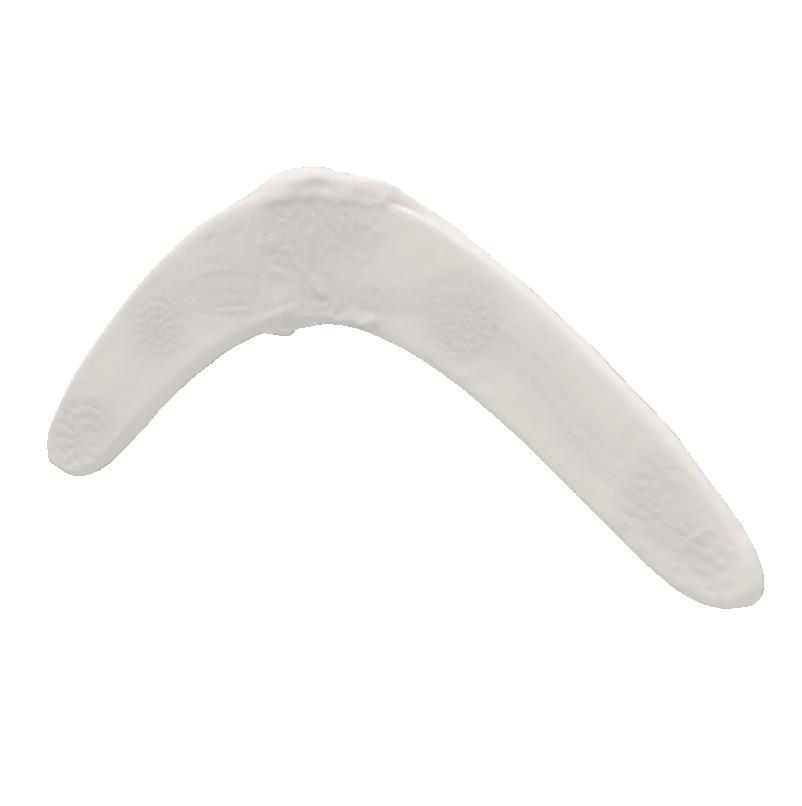} }
    \subcaptionbox*{}
    [.10\textwidth]{\includegraphics[width=\linewidth]{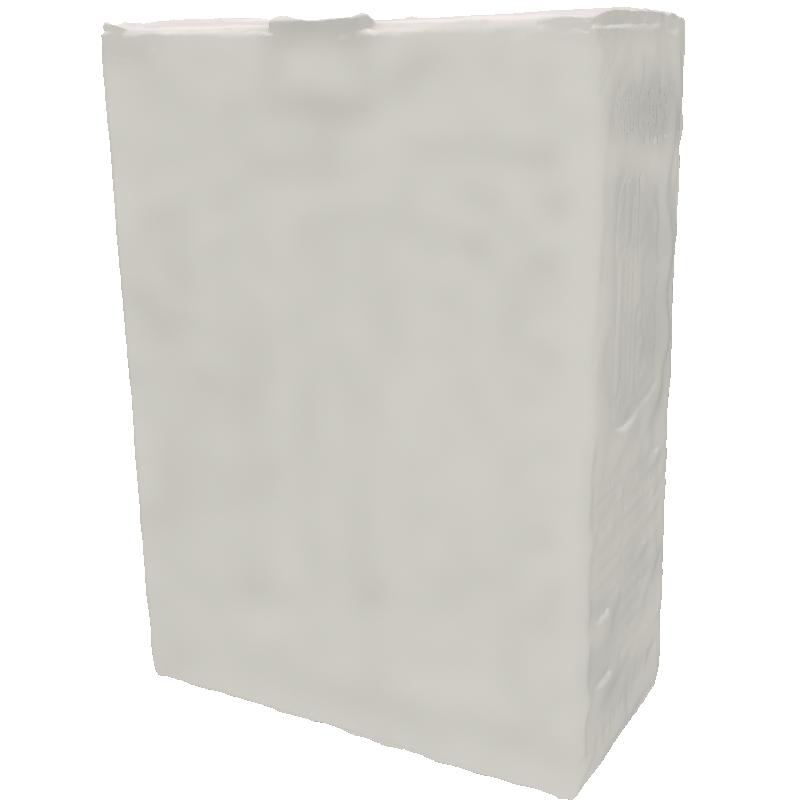} }
    \subcaptionbox*{}
    [.10\textwidth]{\includegraphics[width=\linewidth]{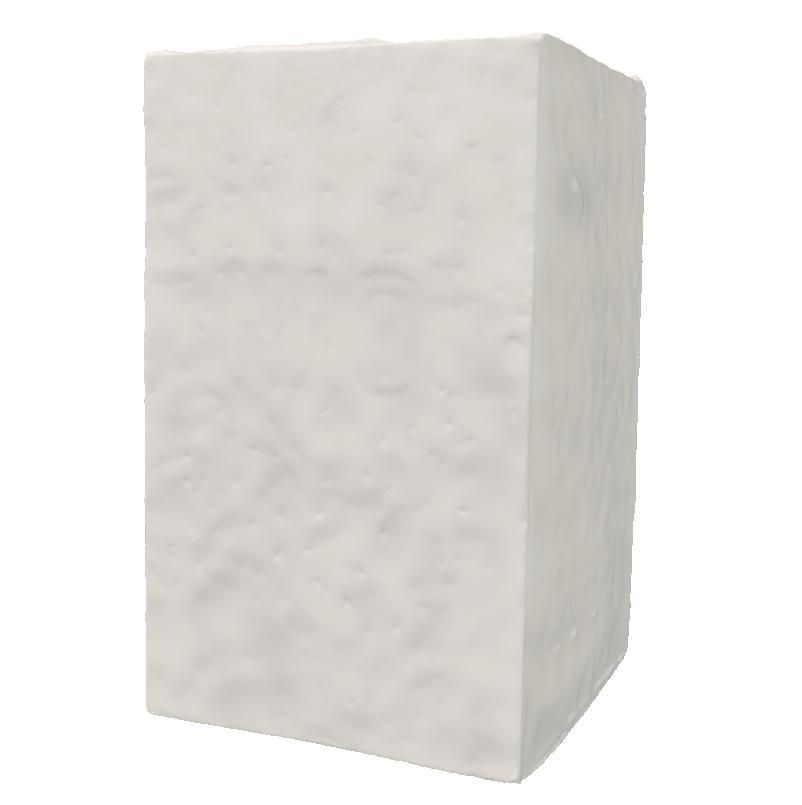} }
    \subcaptionbox*{}
    [.10\textwidth]{\includegraphics[width=\linewidth]{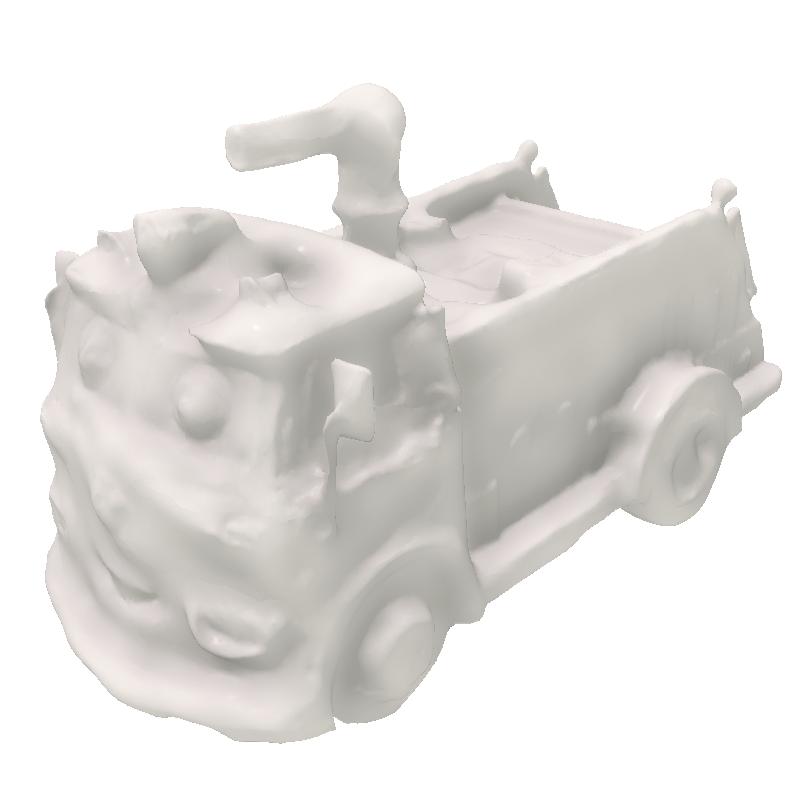} }
    \subcaptionbox*{}
    [.10\textwidth]{\includegraphics[width=\linewidth]{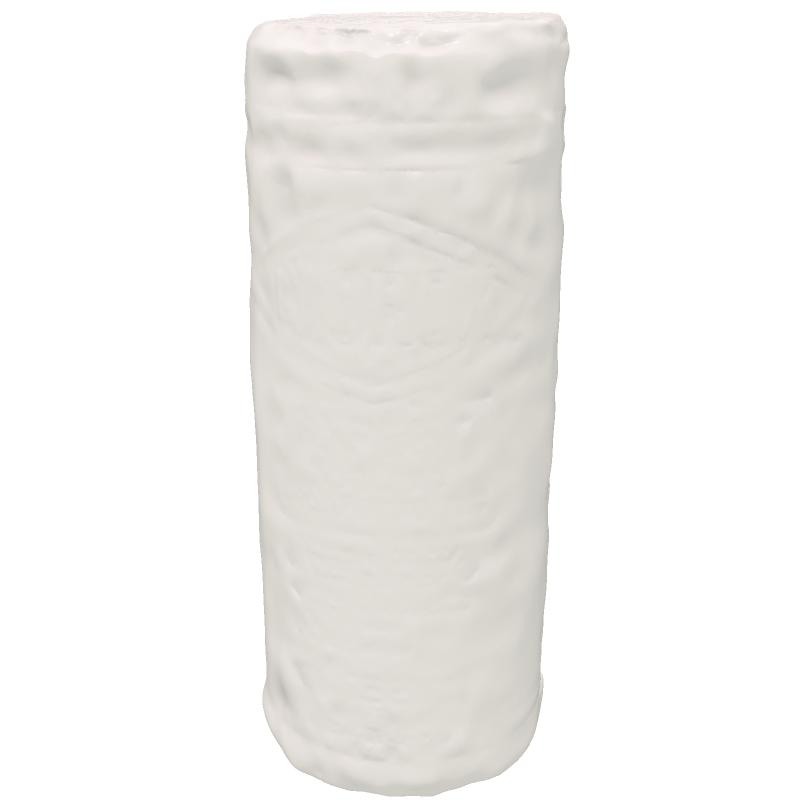} }
    \subcaptionbox*{}
    [.10\textwidth]{\includegraphics[width=\linewidth]{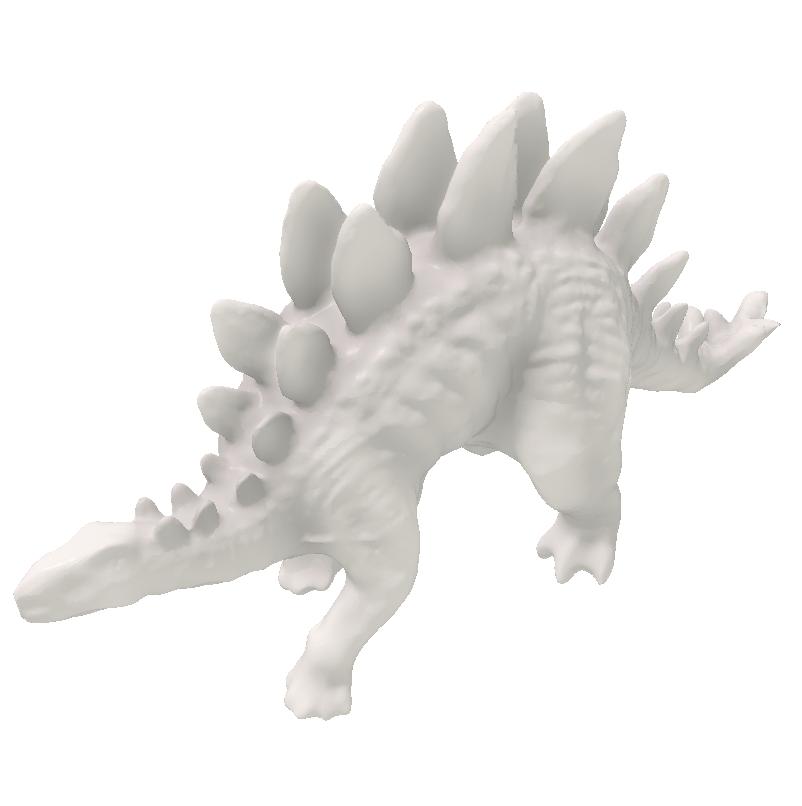} }

    \subcaptionbox*{}
    [.10\textwidth]{\includegraphics[width=\linewidth]{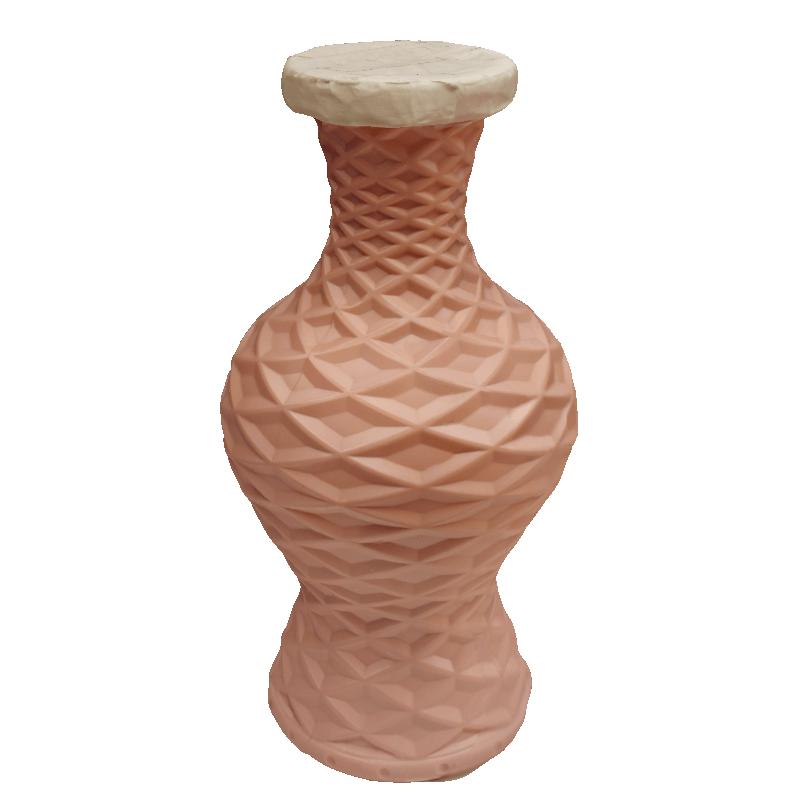} }
    \subcaptionbox*{}
    [.10\textwidth]{\includegraphics[width=\linewidth]{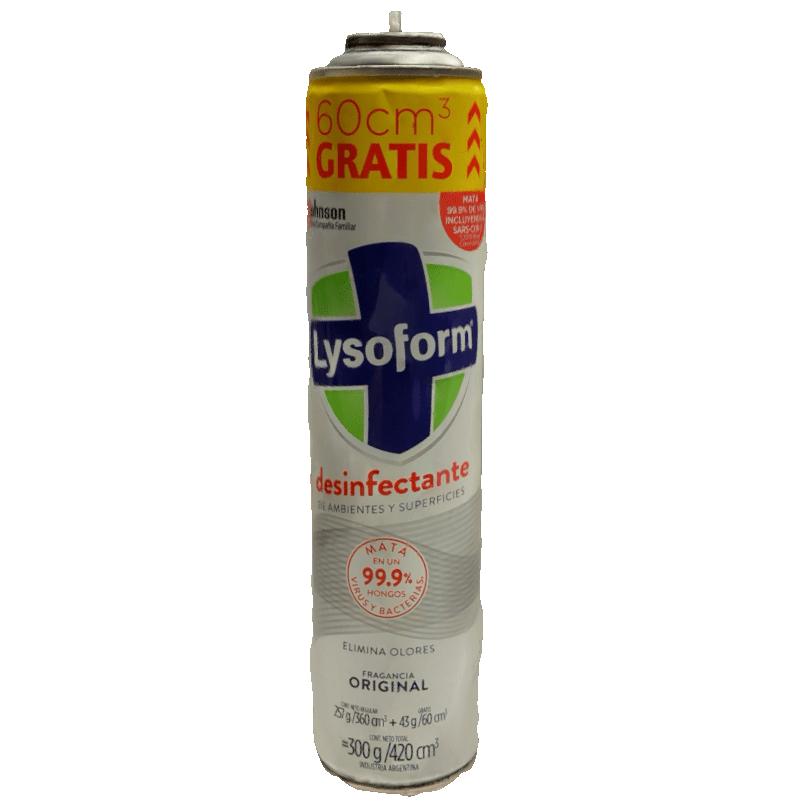} }
    \subcaptionbox*{}
    [.10\textwidth]{\includegraphics[width=\linewidth]{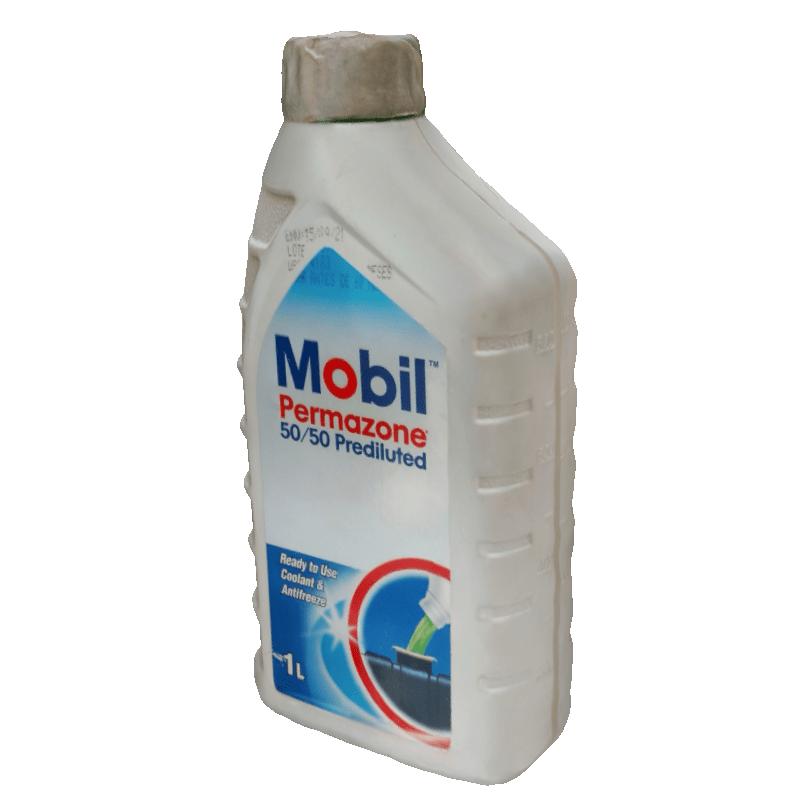} }
    \subcaptionbox*{}
    [.10\textwidth]{\includegraphics[width=\linewidth]{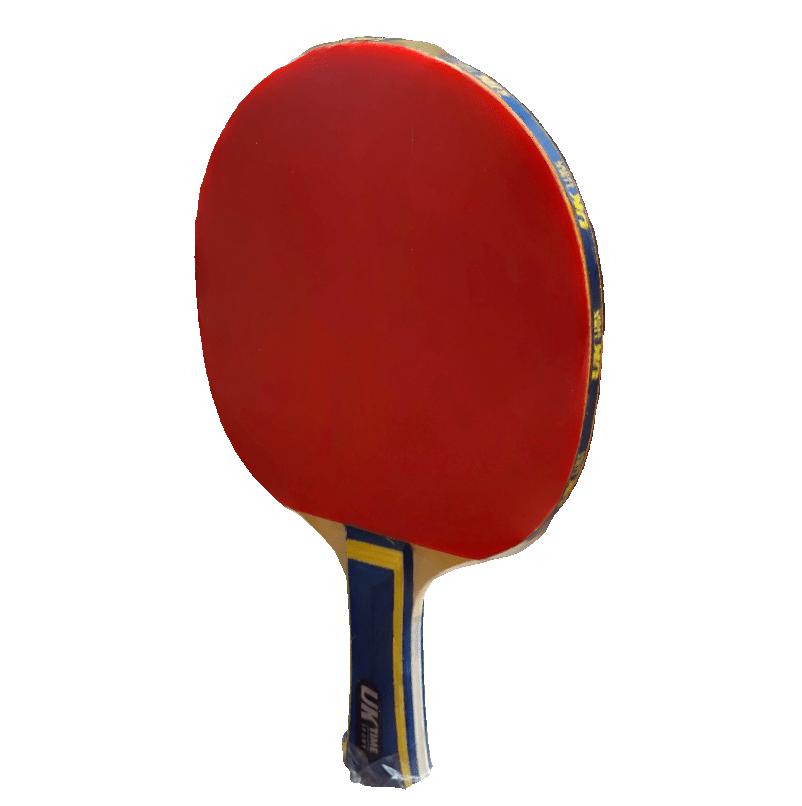} }
    \subcaptionbox*{}
    [.10\textwidth]{\includegraphics[width=\linewidth]{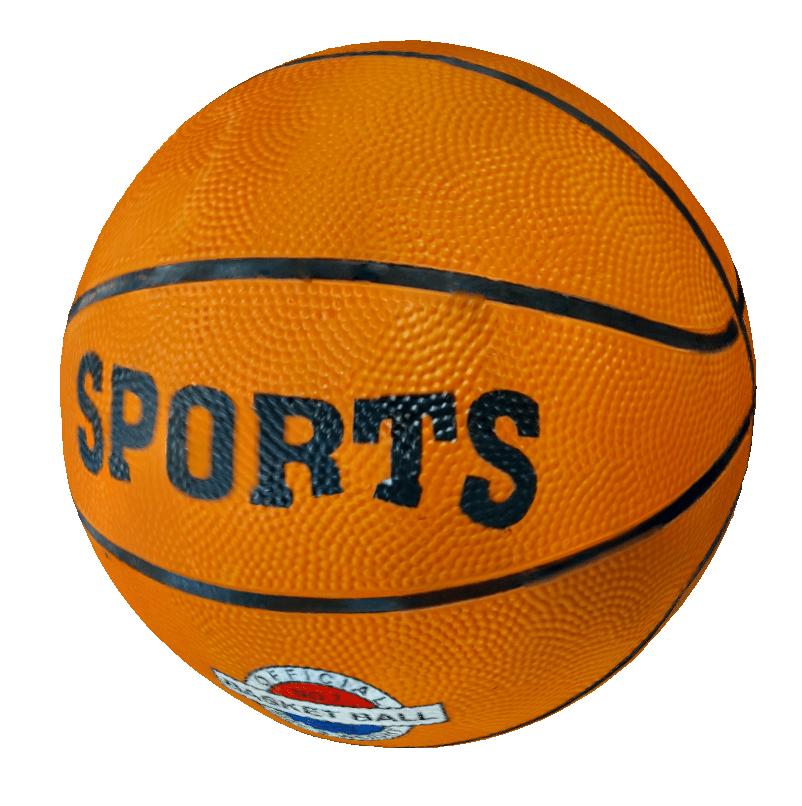} }
    \subcaptionbox*{}
    [.10\textwidth]{\includegraphics[width=\linewidth]{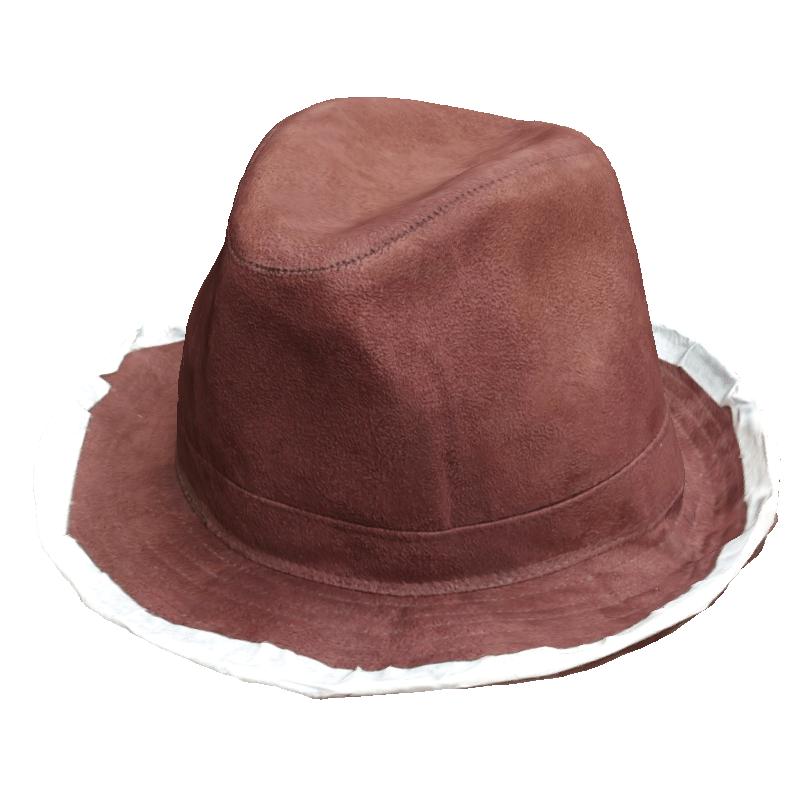} }
    \subcaptionbox*{}
    [.10\textwidth]{\includegraphics[width=\linewidth]{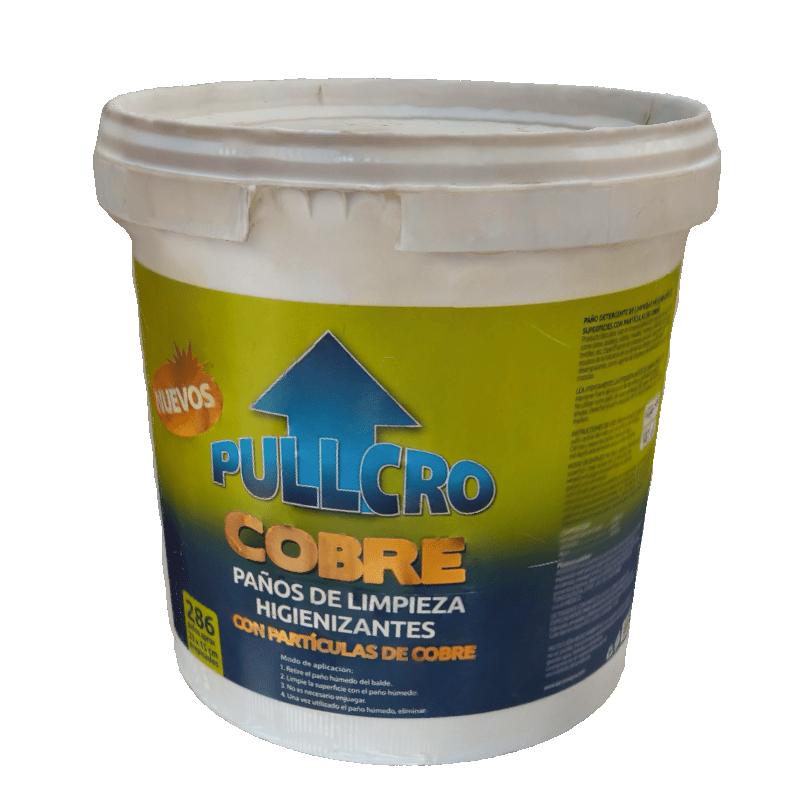} }
    \subcaptionbox*{}
    [.10\textwidth]{\includegraphics[width=\linewidth]{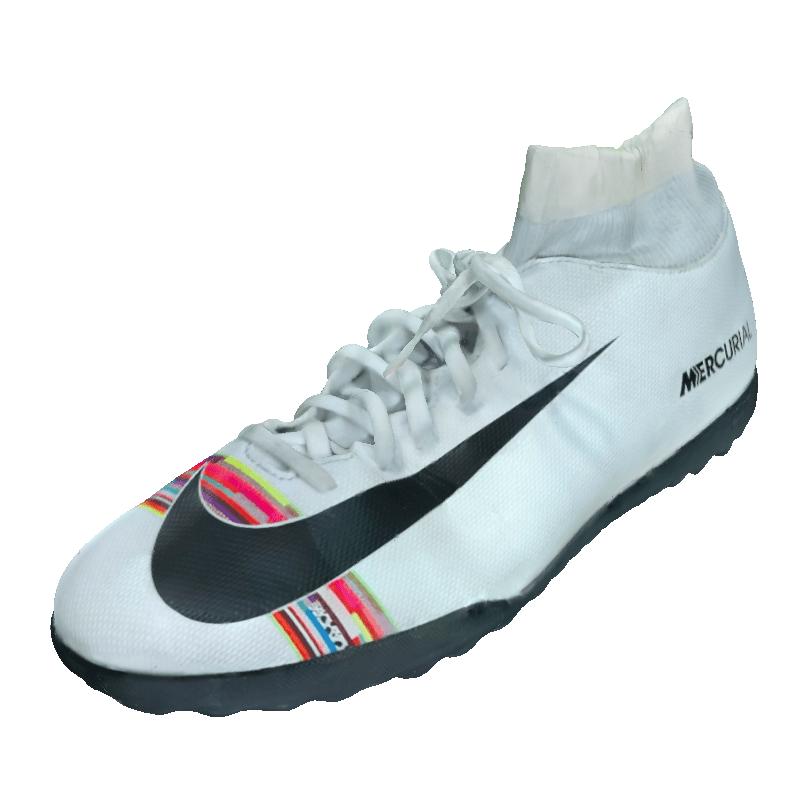} }

    \subcaptionbox*{}
    [.10\textwidth]{\includegraphics[width=\linewidth]{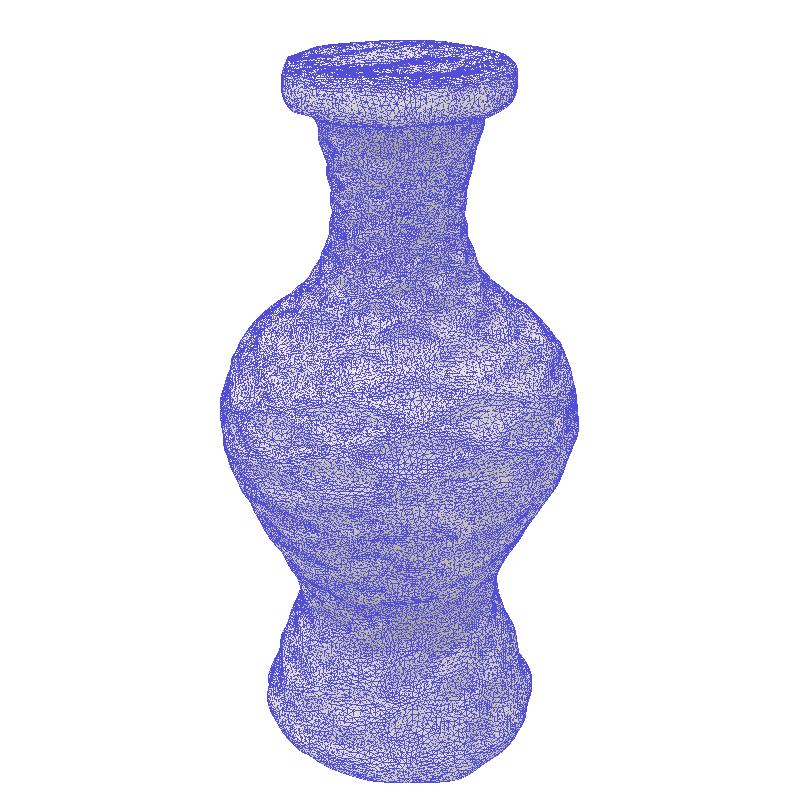} }
    \subcaptionbox*{}
    [.10\textwidth]{\includegraphics[width=\linewidth]{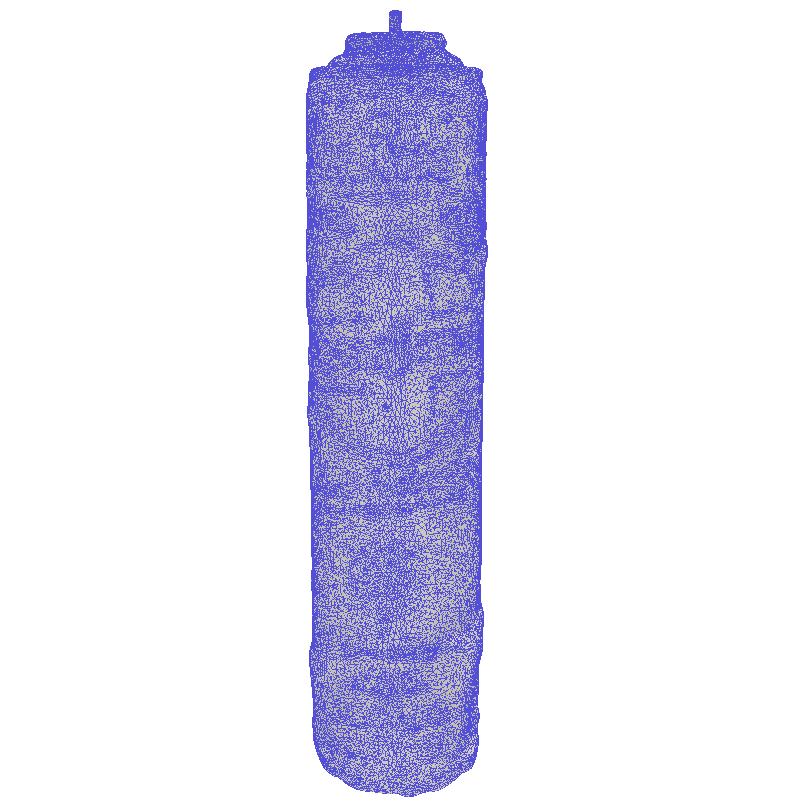} }
    \subcaptionbox*{}
    [.10\textwidth]{\includegraphics[width=\linewidth]{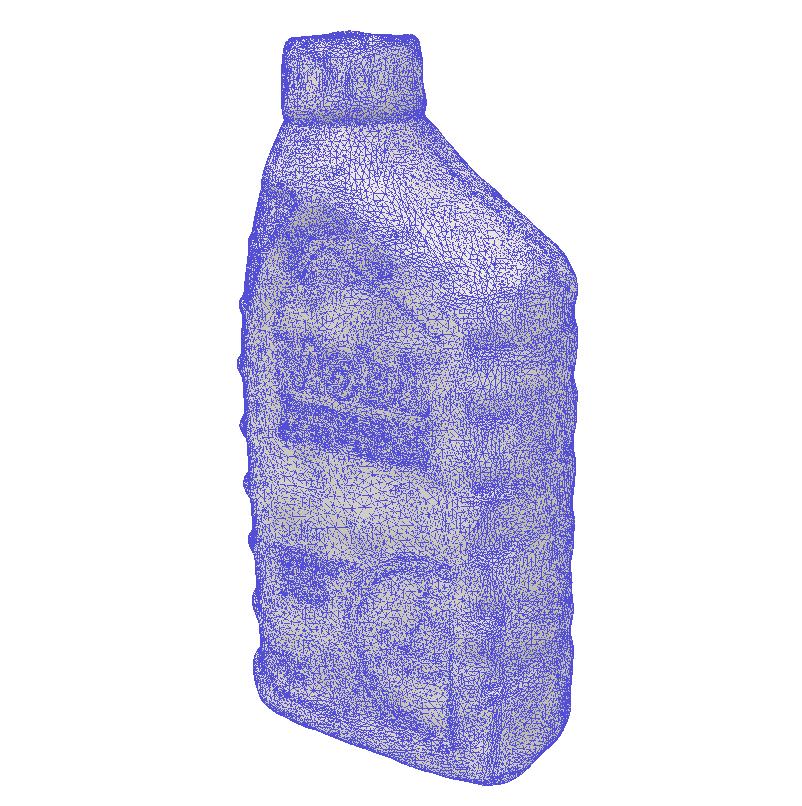} }
    \subcaptionbox*{}
    [.10\textwidth]{\includegraphics[width=\linewidth]{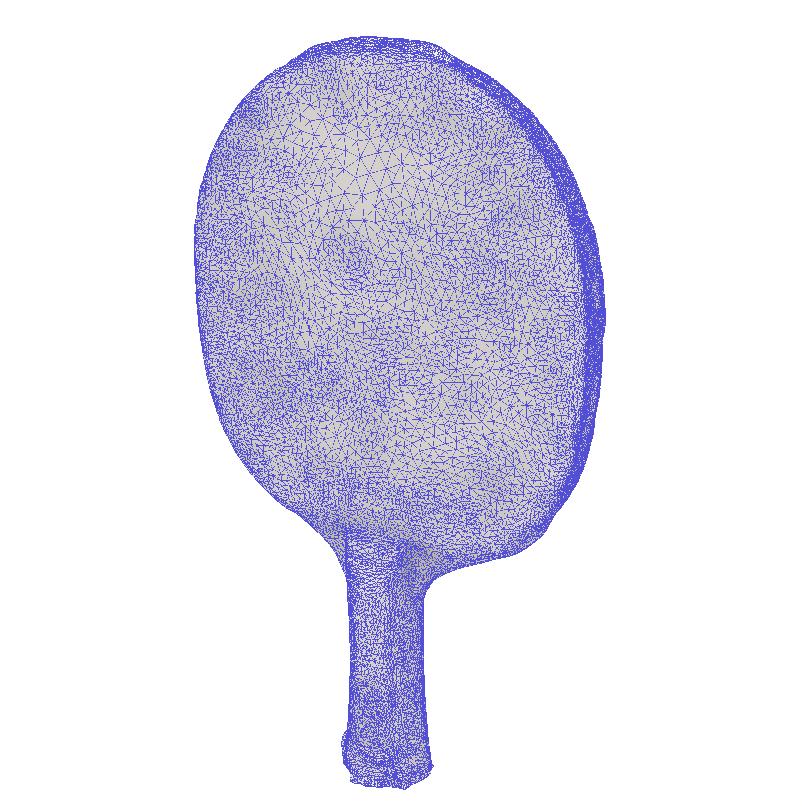} }
    \subcaptionbox*{}
    [.10\textwidth]{\includegraphics[width=\linewidth]{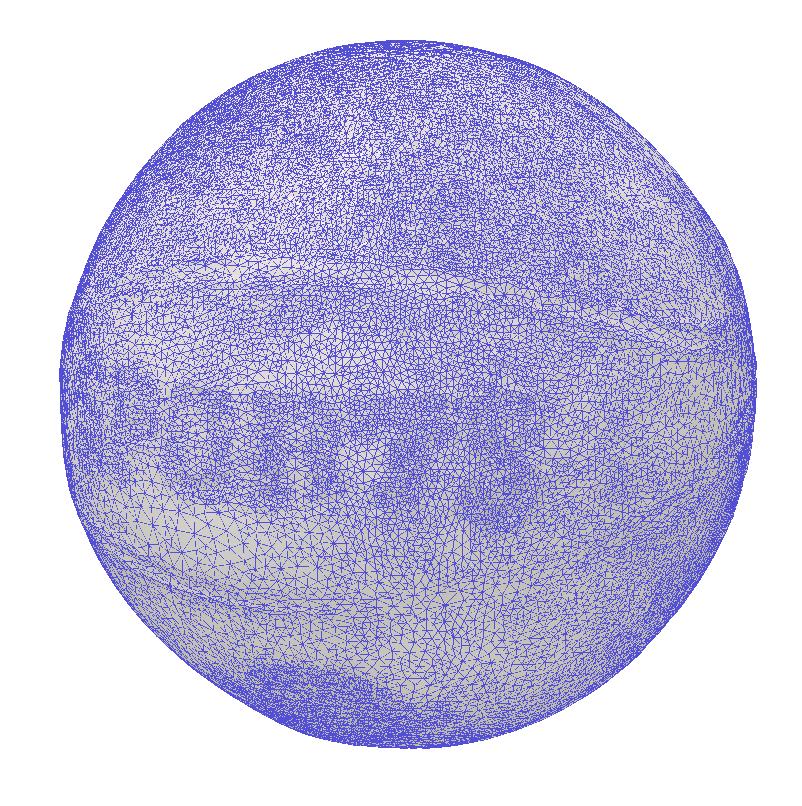} }
    \subcaptionbox*{}
    [.10\textwidth]{\includegraphics[width=\linewidth]{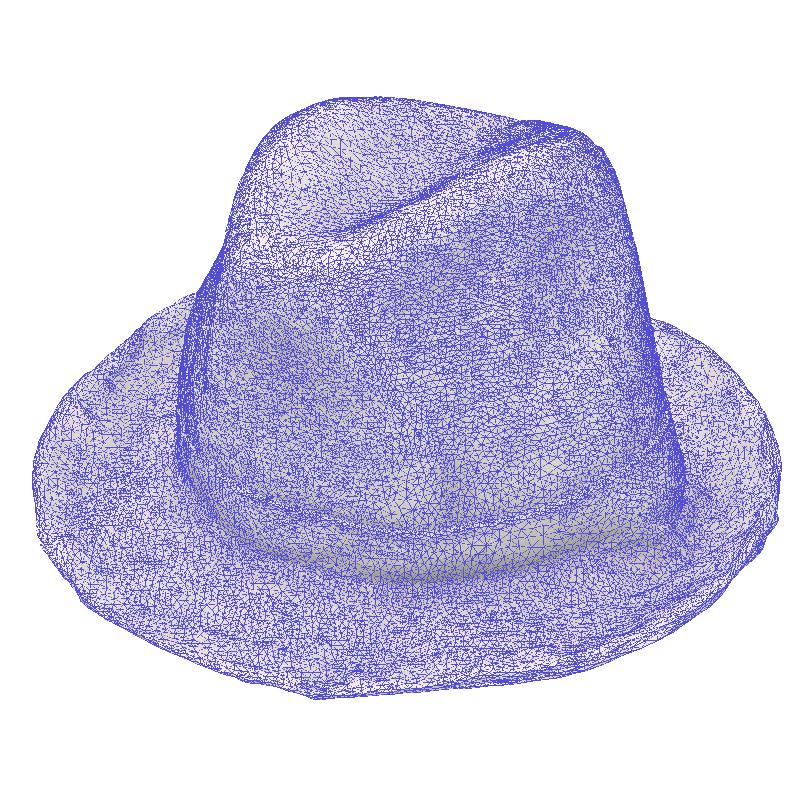} }
    \subcaptionbox*{}
    [.10\textwidth]{\includegraphics[width=\linewidth]{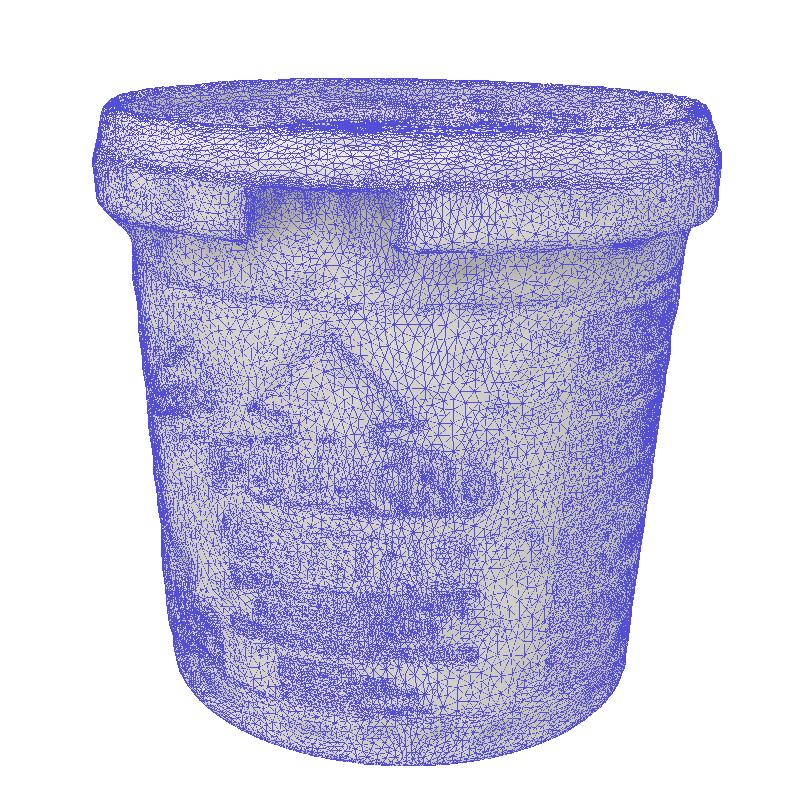} }
    \subcaptionbox*{}
    [.10\textwidth]{\includegraphics[width=\linewidth]{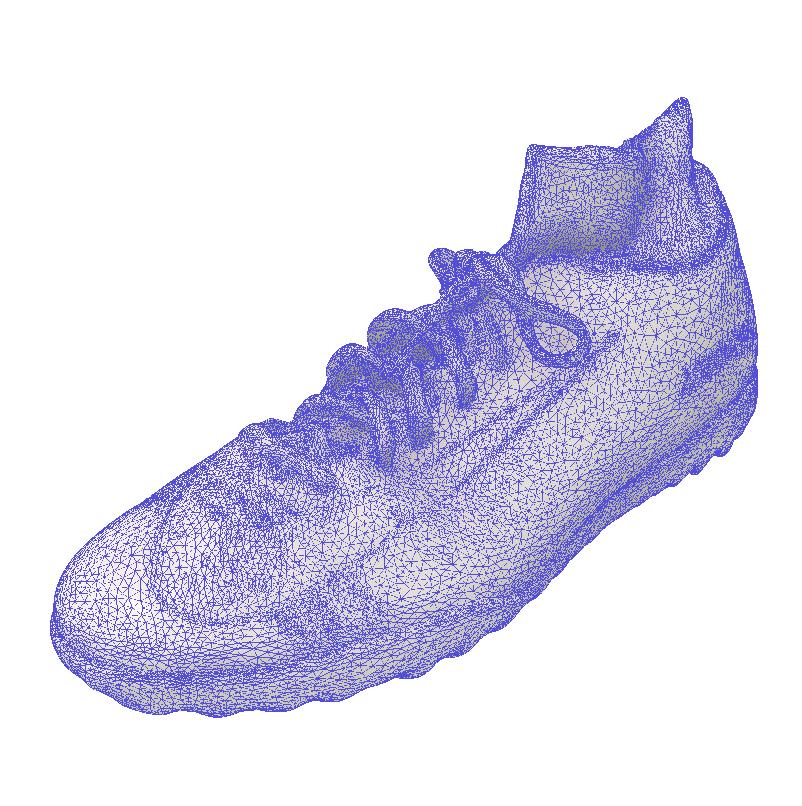} }

    \subcaptionbox*{}
    [.10\textwidth]{\includegraphics[width=\linewidth]{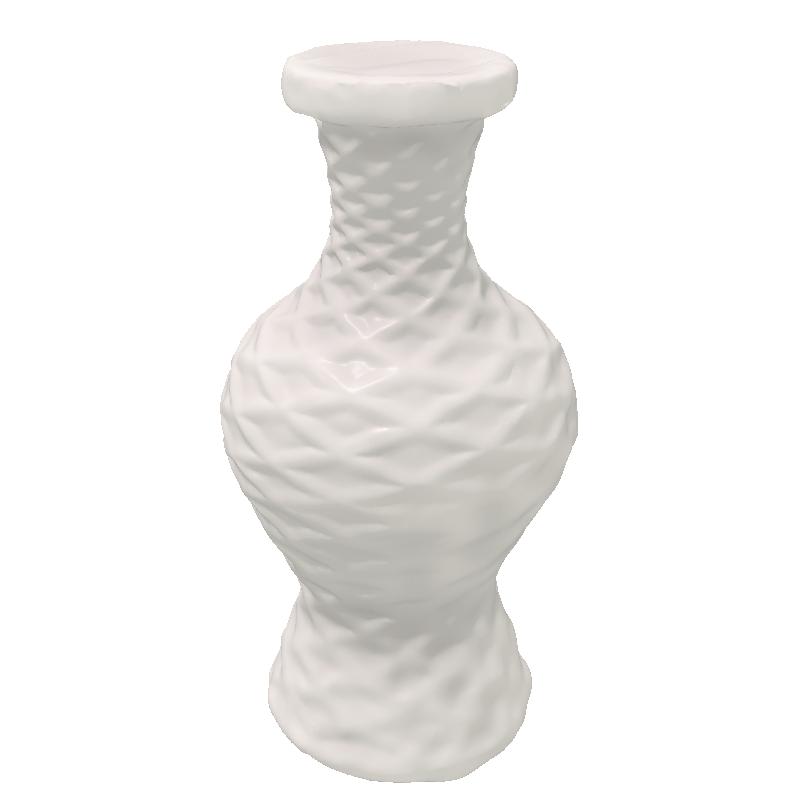} }
    \subcaptionbox*{}
    [.10\textwidth]{\includegraphics[width=\linewidth]{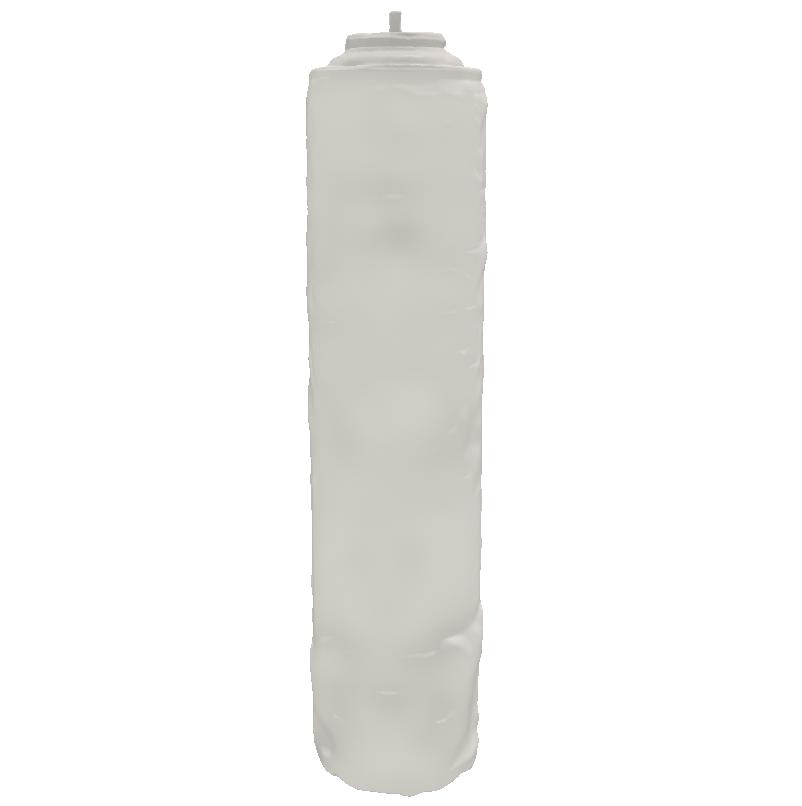} }
    \subcaptionbox*{}
    [.10\textwidth]{\includegraphics[width=\linewidth]{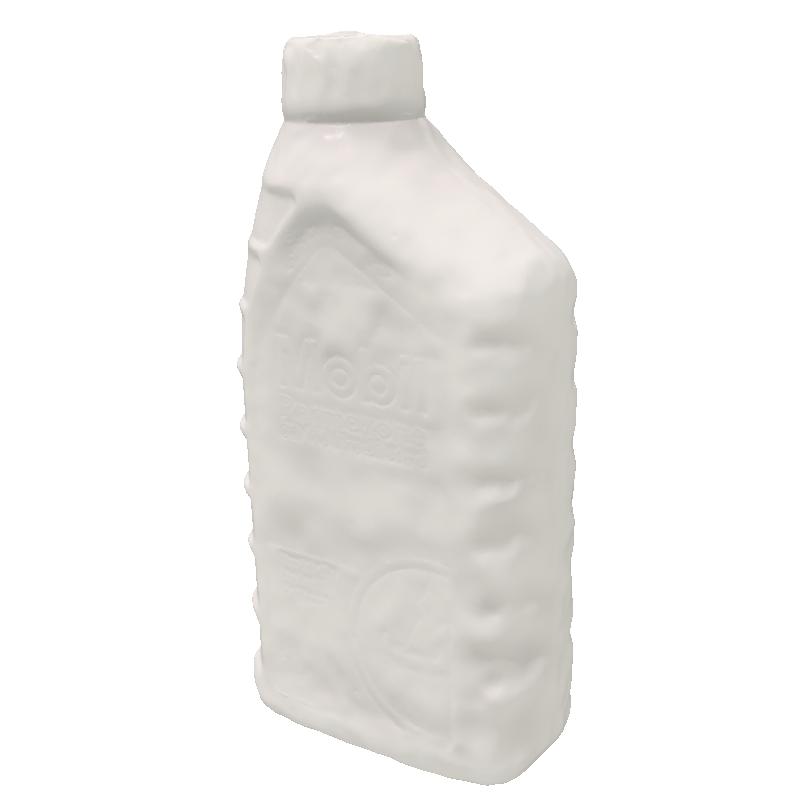} }
    \subcaptionbox*{}
    [.10\textwidth]{\includegraphics[width=\linewidth]{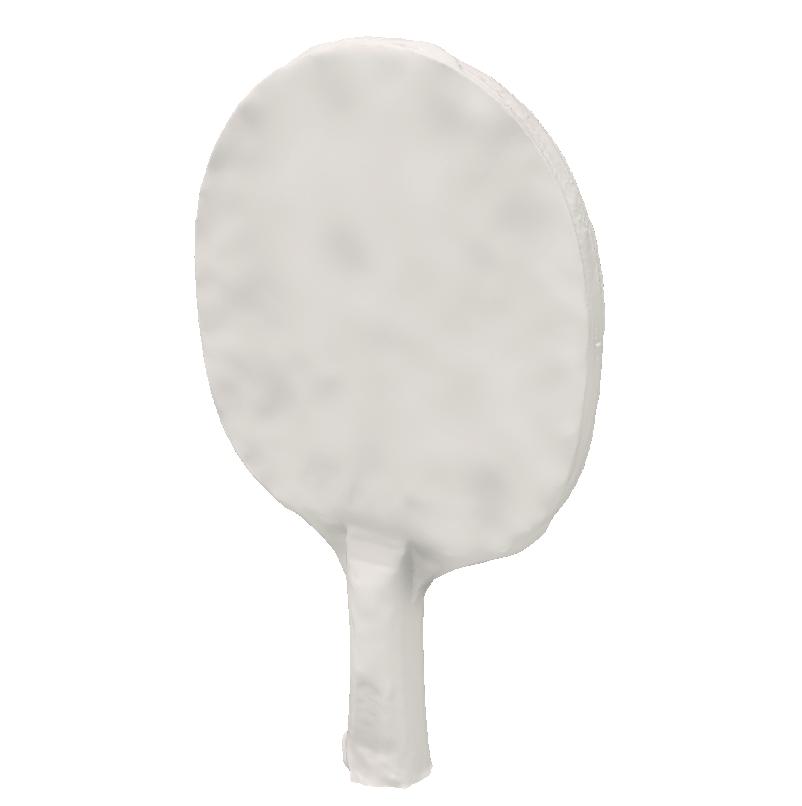} }
    \subcaptionbox*{}
    [.10\textwidth]{\includegraphics[width=\linewidth]{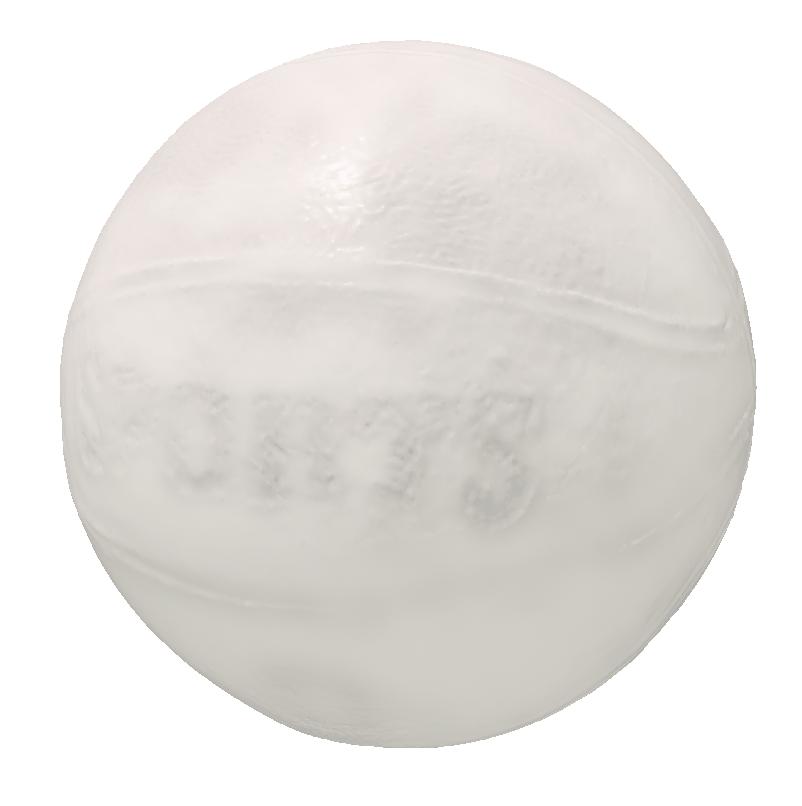} }
    \subcaptionbox*{}
    [.10\textwidth]{\includegraphics[width=\linewidth]{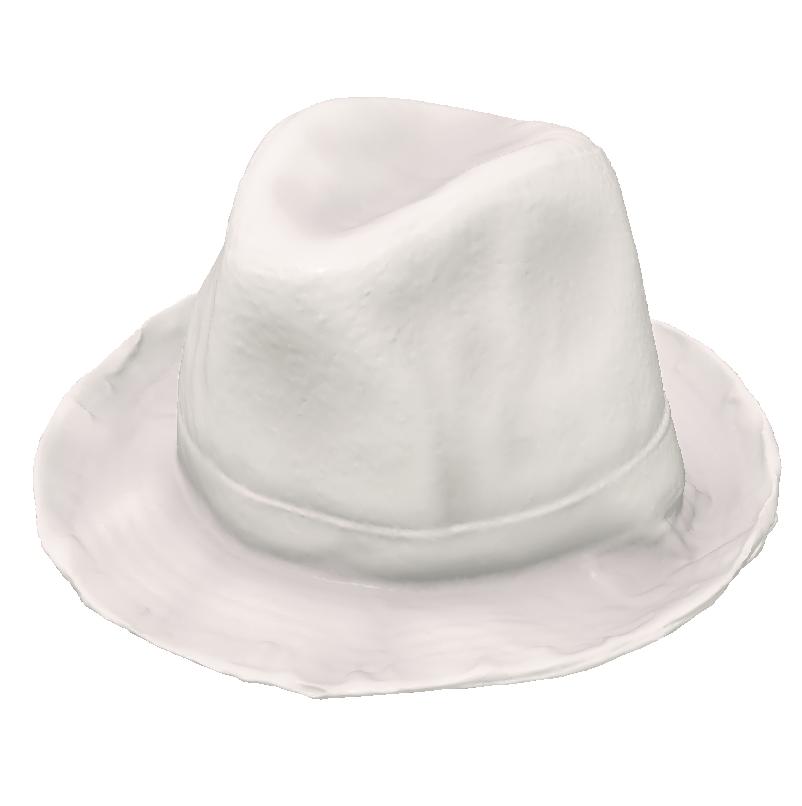} }
    \subcaptionbox*{}
    [.10\textwidth]{\includegraphics[width=\linewidth]{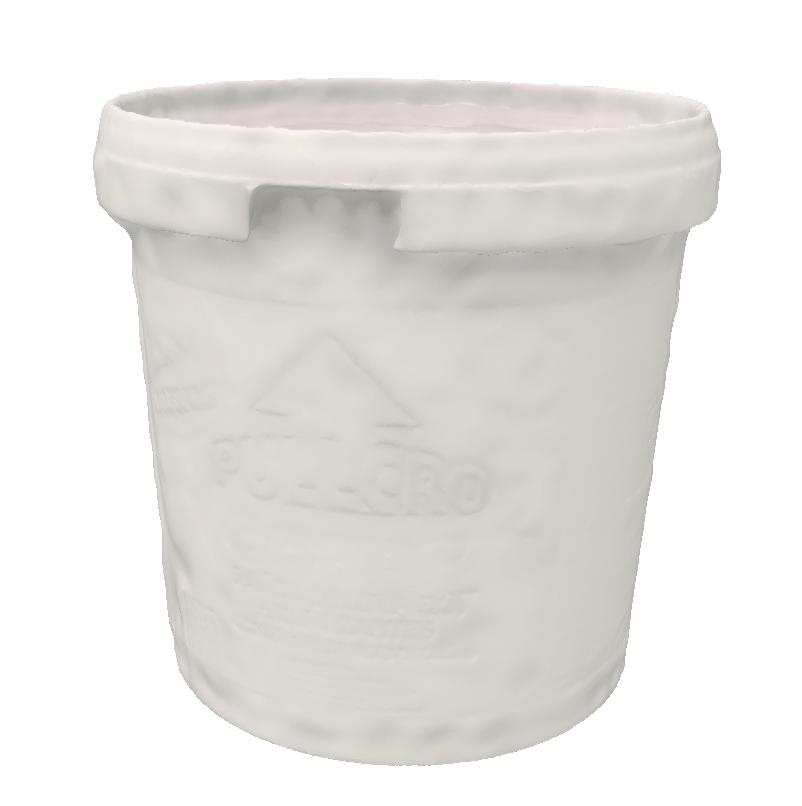} }
    \subcaptionbox*{}
    [.10\textwidth]{\includegraphics[width=\linewidth]{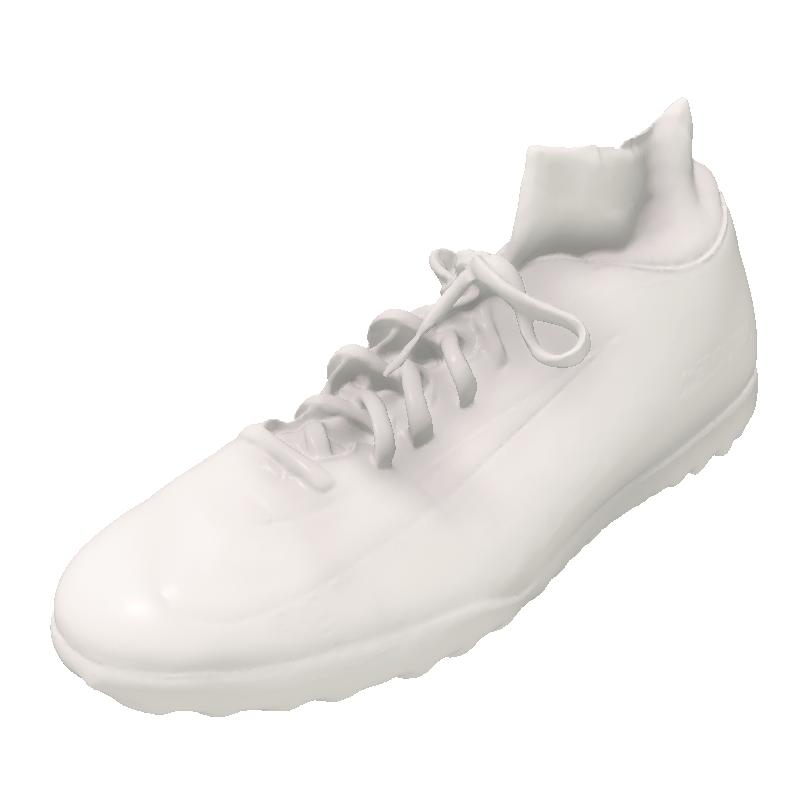} }
    
    \caption{Objects selected for the Moving6DPoSe database. The first and fourth rows show the real scanned models captured using PolyCam. The second and fifth rows display their corresponding 3D point clouds. The third and sixth rows present the rendered depth models used for generating 6D pose ground truth annotations.}
    \label{fig:database-objects}
\end{figure*}

Sixteen different objects are selected to build the dataset, providing varying shapes and textures for depth estimation and spatio-temporal analysis. The objects are as follows: 
\begin{inparaenum}[1)]
    \item Yellow pillow,
    \item Wooden airplane,
    \item Wooden boomerang,
    \item Yellow box,
    \item Blue box,
    \item Red truck,
    \item Metallic cylinder,
    \item Plastic cylinder,
    \item Plastic vase,
    \item Plastic cup,
    \item Plastic dinosaur,
    \item Table tennis racket,
    \item Ball,
    \item Hat,
    \item Plastic container,
    \item Sports shoe.
\end{inparaenum}
The objects are shown in the first and fourth rows of Figure \ref{fig:database-objects}.

\subsection{3D Scanning platform and method}

A 3D model was generated for each object using a custom scanning setup consisting of a Samsung Galaxy A71 smartphone and a rotating platform (Figure~\ref{fig:database_proposal}). The device features a 64 MP primary camera, ultrawide and macro cameras, and a dedicated depth sensor, enabling the acquisition of multiple views and depth cues during a full object rotation. Object meshes and point clouds were reconstructed using PolyCam \cite{polycam2024} from the captured images.

The resulting 3D models were used to generate Moving6DPoSe-S and to support object segmentation, 2D/3D reconstruction, and 6D pose annotation. Examples of the reconstructed models are shown in Figure~\ref{fig:database-objects}.
 
\subsection{Moving6DPoSe-R: Real-world Dataset}

The Moving6DPoSe-R dataset consists of real-world recordings of objects that undergo controlled and uncontrolled motions. These sequences were captured using multiple synchronized cameras to provide RGB images, event-based data, depth maps, and ground truth annotations for each frame. This data set was designed for the estimation of 6D poses under realistic dynamic conditions.

\subsubsection{Sensors and calibration}

To acquire this dataset, we used four vision sensors: an RGB webcam (ASUS ROG Eye S), a stereo RGB camera (ZED-2), an APS-based event camera (DAVIS346~\cite{brandli2014, mueggler2014}), and an event-based camera (Prophesee EVK4). The ZED-2 sensor was used primarily to generate depth maps of the scenes, refining the ground truth for the estimation of 6D poses, while the other cameras provided complementary frame-based and event-based modalities. Table~\ref{tab:vision-sensors-summary} summarizes each sensor's data modality, physical dimensions, field of view, spatial resolution, and data rates.

\begin{table*}[!t]
    \caption{Summary table of relevant technical characteristics of vision sensors. Description: \textit{Resolution} refers to the spatial resolution of the sensor; $\textit{Rate}\textit{G}$ refers to the data generation rate; $\textit{Rate}\textit{S}$ refers to the data storage rate.}
    \label{tab:vision-sensors-summary}
    \centering
    \begin{tabular}{|l|l|l|c|c|c|}
    \hline
    \textbf{Sensor} & \textbf{Data} & \textbf{Field-of-view} & \textbf{Resolution} & $\textbf{Rate}_\textbf{G}$ & $\textbf{Rate}_\textbf{S}$\\ \hline
    
        ROG Eye S & RGB Frame & 78.0° & 640x360 & \begin{tabular}[c]{@{}c@{}}Fixed\\(30 Hz)\end{tabular} & \begin{tabular}[c]{@{}c@{}}Fixed\\(30 Hz)\end{tabular} \\ \hline
        
        \multirow{4}{*}{ZED-2} & \multirow{2}{*}{RGB frame per lens} & \multirow{4}{*}{\begin{tabular}[c]{@{}l@{}}H: 92-103°\\V: 61-71°\end{tabular}} & \multirow{2}{*}{640x480} & \multirow{4}{*}{\begin{tabular}[c]{@{}c@{}}Fixed\\(15 Hz)\end{tabular}} & \multirow{4}{*}{\begin{tabular}[c]{@{}c@{}}Fixed\\(15 Hz)\end{tabular}} \\
        & & & & & \\ \cline{2-2} \cline{4-4}
        & \multirow{2}{*}{Stereo RGB frame} & & \multirow{2}{*}{1280x480} & & \\
        & & & & & \\ \hline
        
        \multirow{4}{*}{DAVIS346} & \multirow{2}{*}{Gray frame} & \multirow{4}{*}{\begin{tabular}[c]{@{}l@{}}H: 29.9-113°\\V:22.7-99.7°\\D: 36.9-215°\end{tabular}} & \multirow{2}{*}{346x260} & \multirow{2}{*}{\begin{tabular}[c]{@{}c@{}}Fixed\\(30 Hz)\end{tabular}} & \multirow{2}{*}{\begin{tabular}[c]{@{}c@{}}Fixed\\(30 Hz)\end{tabular}} \\
        & & & & & \\ \cline{2-2} \cline{4-6}
        & \multirow{2}{*}{Events} & & \multirow{2}{*}{346x260} & \multirow{2}{*}{\begin{tabular}[c]{@{}c@{}}Variable\\(1 MHz)\end{tabular}} & \multirow{2}{*}{\begin{tabular}[c]{@{}c@{}}Fixed\\(30 Hz)\end{tabular}} \\
        & & & & & \\ \hline
        
        EVK4 & Events & \begin{tabular}[c]{@{}l@{}}H: 41.4°\\V: 23.6°\\D: 47.0°\end{tabular} & 1280x720 & \begin{tabular}[c]{@{}c@{}}Variable\\(10 KHz)\end{tabular} & \begin{tabular}[c]{@{}c@{}}Variable\\(10 KHz)\end{tabular} \\ \hline
    \end{tabular}
\end{table*}

\begin{figure}[!tb]
    \centering
    \includegraphics[width=0.35\linewidth]{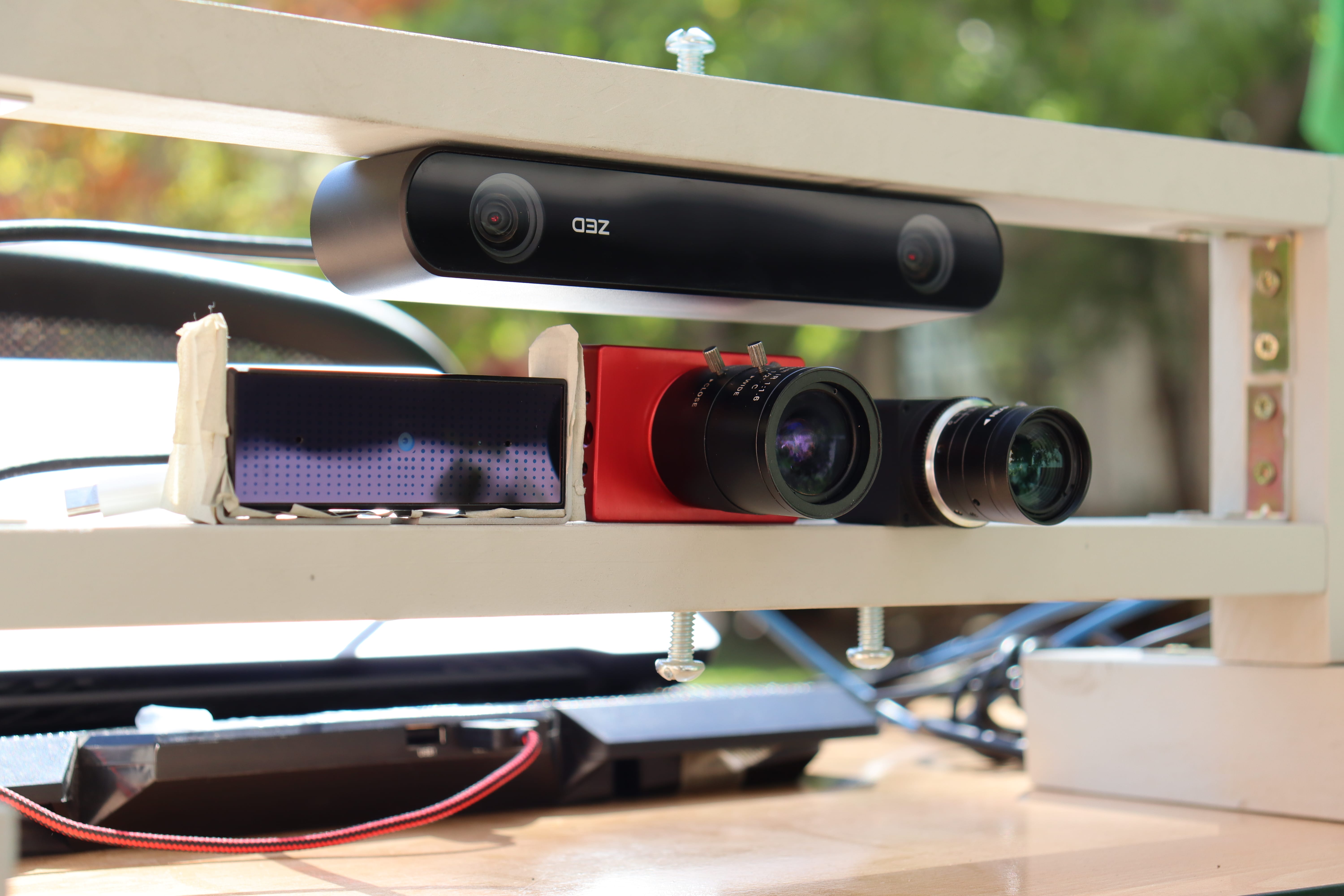}
    \includegraphics[width=0.55\linewidth]{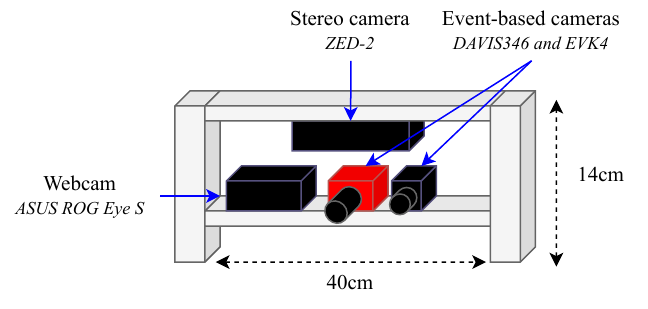}

    \caption{Platform designed for mounting the vision sensors. This platform is used to acquire the Moving6DPoSe-R dataset.}
    \label{fig:plataform_design}
\end{figure}

A custom-made capture platform (Figure~\ref{fig:plataform_design}) was designed to mount these sensors, considering their physical dimensions and respective field of view. This ensured precise alignment and minimized occlusions between modalities during data collection.

To calibrate the parameters of each vision sensor, including frame-based cameras (ROG Eye S and ZED2) and event-based cameras (DAVIS346 and Prophesee EVK4), we used the checkerboard process to calculate the camera matrix, distortion coefficients, and rectification/projection matrix. 
For frame-based cameras and the DAVIS346 camera, the OpenCV calibration pipeline was used to obtain camera parameters.
For the EVK4, an animated 8x6 checkerboard was required to generate event images suitable for the sensor calibration.
Finally, Table~\ref{tab:calibration-parameters} reports the intrinsic parameters $(fx, fy, cx, cy)$ obtained for each sensor. 

\begin{table}[!tb]
\centering
\small
\caption{Intrinsic parameters of frame-based and event-based vision sensor calibration.}
\label{tab:calibration-parameters}
\begin{tabular}{|l|c|c|c|c|}
    \hline
    Sensor & fx & fy & cx & cy \\ \hline
    ROG Eye S & 603.08 & 604.74 & 324.80 & 181.85 \\ \hline
    ZED-2 & 306.49 & 306.49 & 274.94 & 155.09 \\ \hline
    DAVIS346 & 292.93 & 293.92 & 183.34 & 127.46 \\ \hline
    EVK4 HD & 1985.33 & 1985.33 & 645.32 & 363.26 \\ \hline
\end{tabular}
\end{table}

\subsubsection{Scenarios}
\label{sec:scenarios}

\begin{figure*}[!ht]
\captionsetup[subfigure]{skip=5pt}
\centering
    \subcaptionbox{Scenario 1\label{fig:ddbb-r-scn2}}
    [.22\textwidth]{\includegraphics[width=\linewidth]{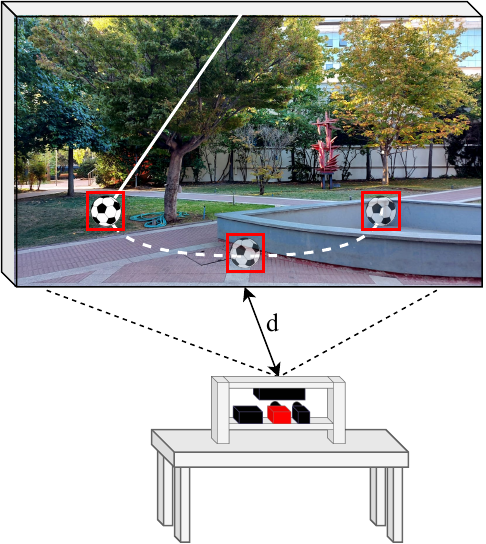} }
    \subcaptionbox{Scenario 2\label{fig:ddbb-r-scn3}}
    [.22\textwidth]{\includegraphics[width=\linewidth]{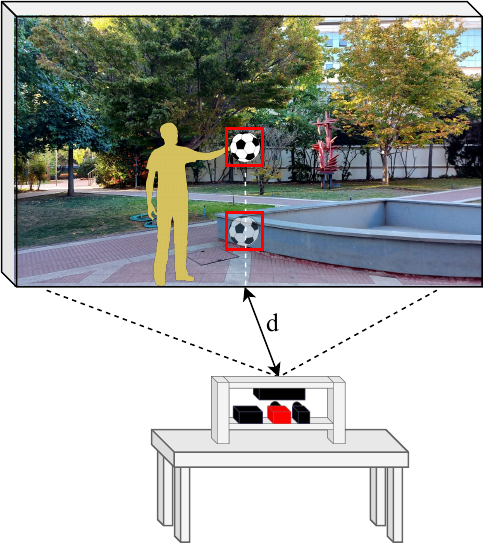} }
    \subcaptionbox{Scenario 3\label{fig:ddbb-r-scn4}}
    [.225\textwidth]{\includegraphics[width=\linewidth]{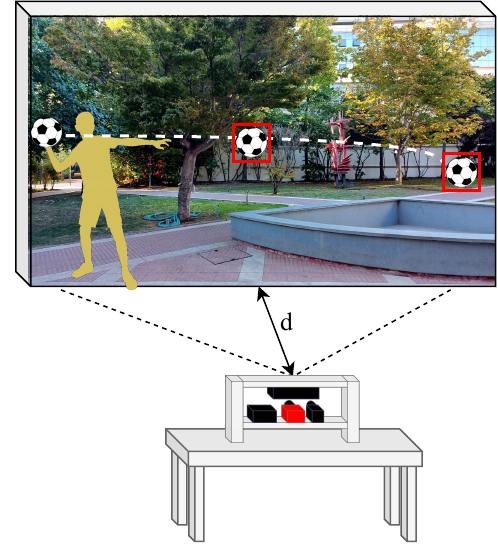} }
    \subcaptionbox{Scenario 4\label{fig:ddbb-r-scn5}}
    [.225\textwidth]{\includegraphics[width=\linewidth]{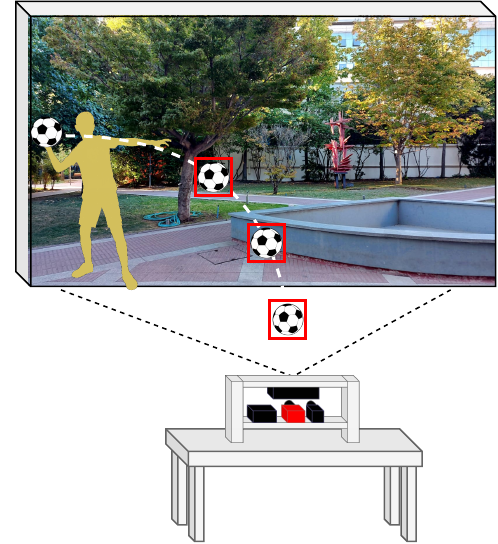} }
    
    \caption{Moving6DPoSe-R scenarios: (a) Scenario 1: An object attached to a rope is thrown on a pendulum with a controlled parabolic motion; (b) Scenario 2: An object is released in free fall; (c) Scenario 3: An object is thrown; (d) Scenario 4: An object is thrown to the platform.}
    \label{fig:ddbb-r-scenarios}
\end{figure*}

To provide diverse object trajectories and motion dynamics, five experimental scenarios were designed, as illustrated in Figure~\ref{fig:ddbb-r-scenarios}. 

\begin{itemize}
    \item Scenario 1 captures controlled pendulum-like movements by attaching objects to a rope and releasing them, resulting in parabolic trajectories (Figure \ref{fig:ddbb-r-scn2}).
    \item Scenario 2 consists of free-fall sequences in which objects are released vertically downward under gravity (Figure \ref{fig:ddbb-r-scn3}).
    \item Scenario 3 involves manually throwing objects along uncontrolled trajectories to capture high-speed translational motion (Figure \ref{fig:ddbb-r-scn4}).
    \item Scenario 4 captures objects thrown directly onto the platform with an uncontrolled trajectory (Figure \ref{fig:ddbb-r-scn5}).
\end{itemize}

These scenarios were designed to cover a wide range of motion dynamics, enabling the evaluation of pose estimation methods under realistic and challenging conditions.

\subsection{Moving6DPoSe-S: Synthetic Dataset}
\label{sec:db-sync}

Collecting representative samples for training and evaluating 6D pose estimation models can be challenging, especially for dynamic motions that are difficult to replicate consistently in real environments. Generating a synthetic dataset enables the creation of large volumes of annotated data across various scenarios, with controlled variations in object motion and appearance.

This section describes the elaboration of Moving6DPoSe-S, which uses rendered 3D models of scanned objects and simulates their motion in free-fall and throwing trajectories.

\subsubsection{Simulation and data capture of moving objects}

\begin{figure}[!h]
\captionsetup[subfigure]{skip=5pt}
\centering
    \subcaptionbox{Scenario 2\label{fig:ddbb-s-scn3}}
    [.32\linewidth]{\includegraphics[width=\linewidth]{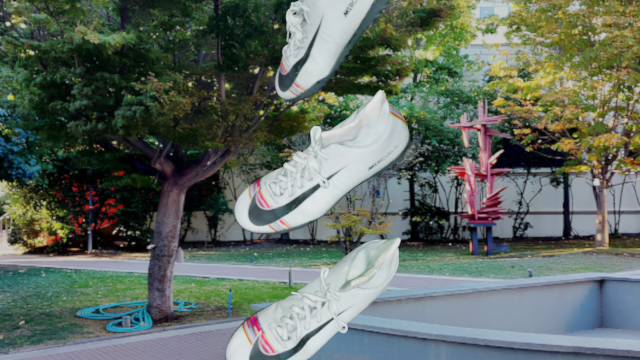} }
    \subcaptionbox{Scenario 3\label{fig:ddbb-s-scn4}}
    [.32\linewidth]{\includegraphics[width=\linewidth]{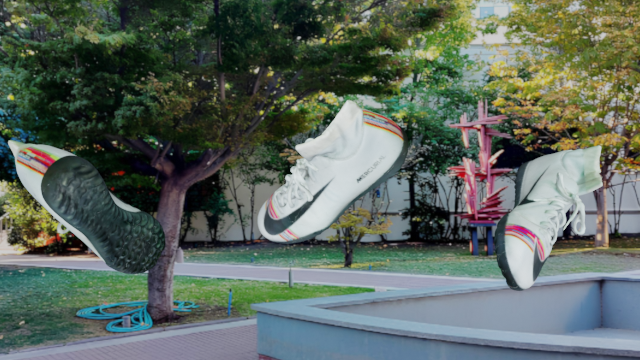} }
    \subcaptionbox{Scenario 4\label{fig:ddbb-s-scn5}}
    [.32\linewidth]{\includegraphics[width=\linewidth]{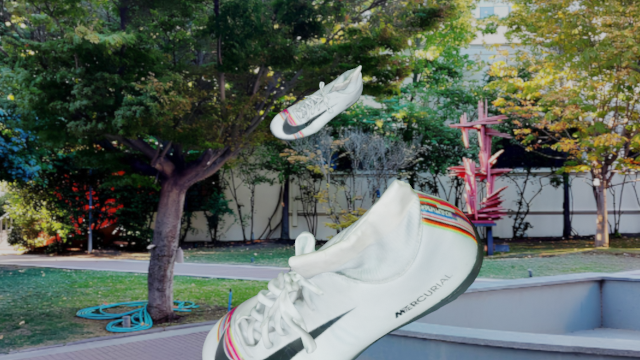} }

    \caption{Moving6DPoSe-S simulations: (a) Scenario 3: An object is released in free fall; (b) Scenario 4: An object is thrown; (e) Scenario 5: An object is thrown to the platform.
    }
    \label{fig:ddbb-s-scenarios}
\end{figure}

Moving6DPoSe-S is generated in Blender, which provides physics-based simulation and photorealistic rendering for the scanned object models. Four motion scenarios are simulated, as illustrated in Figure~\ref{fig:ddbb-s-scenarios}: (a) object rotating about its own axis, (b) free fall, (c) parabolic throwing trajectories, and (d) object motion towards the camera.

For each object, 24 random initial orientations are sampled as quaternions for every simulated trajectory. Blender renders the sequences at 1000~fps, generating RGB and grayscale images, depth maps, segmentation masks, and per-frame 6D pose annotations. 

\subsubsection{Event-based camera simulation}

Since Blender does not natively support event-based cameras, the V2E simulator \cite{hu2021v2e} is used to generate event streams from high frame rate intensity image sequences rendered in Blender. 
This pipeline generates event data from high-temporal-resolution frames rendered in Blender at 1000 Hz for each simulated moving object sequence, maintaining temporal consistency with the ground truth annotations. Figure~\ref{fig:moving6dpose-s} illustrates examples of the rendered frames and the corresponding event representations.
As a result, this pipeline produces both frames and event streams for each simulated moving object sequence, combining Blender-rendered data with V2E-generated events. 

\subsection{Dataset Statistics}
\label{sec:dataset_statistics}

Table~\ref{tab:dataset_statistics} summarizes the number of sequences in the real and synthetic subsets for each motion category. The synthetic subset contains 24 simulated recordings per object for the free-fall, throwing, and frontal scenarios (384 sequences per category across the 16 scanned objects). No dedicated synthetic pendular sequences are generated, as this motion is physically equivalent to the throwing scenario. Additionally, real and synthetic subsets include rotation-only sequences, in which each object rotates about its own axis on a virtual turntable, providing additional viewpoint and appearance variations.

\begin{table}[!t]
\centering
\caption{Summary of the Moving6DPoSe dataset. The table reports the number of sequences available in the real and synthetic subsets for each motion category.}
\label{tab:dataset_statistics}
\small
\begin{tabular}{lC{1.8cm}C{1.8cm}C{1.8cm}C{1.8cm}C{1.2cm}}
\toprule
\textbf{Subset} &
\textbf{Free-fall} &
\textbf{Throwing} &
\textbf{Pendular} &
\textbf{Frontal} &
\textbf{Total} \\
\midrule
Moving6DPoSe-R & 145 & 221 & 82 & 102 & 550 \\
Moving6DPoSe-S & 384 & 384 & -- & 384 & 1152 \\
\bottomrule
\end{tabular}
\end{table}

\section{Moving6DPoSe Data Annotation}
\label{sec:annotation} 

To support the different perception tasks provided by Moving6DPoSe, including segmentation, detection, 3D reconstruction, and 6D pose estimation, separate annotation pipelines were developed for the real and synthetic datasets.

\subsection{Moving6DPoSe-R dataset annotation}

\begin{figure}[!tb]
\centering
    \includegraphics[width=\linewidth]{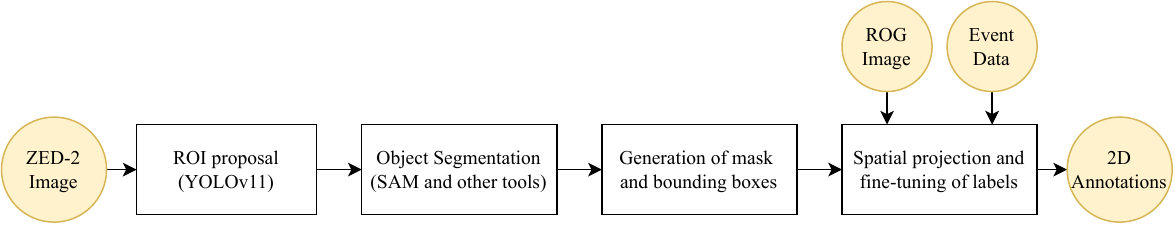}
    \caption{Annotation pipeline for Moving6DPoSe-R. ZED-2 images are processed with YOLOv11 for ROI proposals, followed by SAM-based segmentation and mask generation. Labels are then projected onto monocular frame and event data for 2D annotation.}
    \label{fig:ddbb-r-labeling}
\end{figure}

For frame-based data, object proposals were first obtained using YOLO \cite{redmon2016yolo} on ZED-2 images and refined with the Segment Anything Model (SAM) \cite{kirillov2023segany} to generate segmentation masks and bounding boxes. The ZED-2 sensor was selected as the reference modality because it provides synchronized RGB and depth information, enabling more reliable segmentation and localization. The resulting annotations were projected onto the ROG Eye S webcam and the DAVIS346 and Prophesee EVK4 event cameras using calibration-based intersensor transformations estimated from chessboard observations. Manual refinement was performed when necessary, particularly for event-based data using \textit{Segments.ai} \cite{segmentsai}. The annotation pipeline is illustrated in Figure~\ref{fig:ddbb-r-labeling}.

\begin{figure}[!tb]
    \centering
    \includegraphics[width=0.4\linewidth,trim={0 3cm 0 3cm},clip]{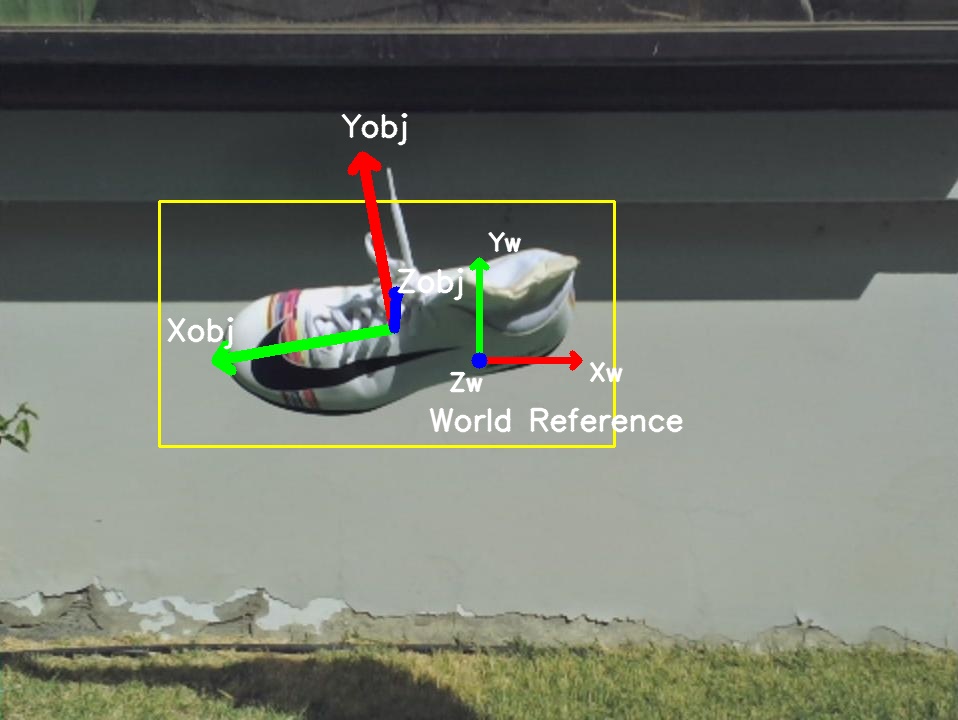}
    \includegraphics[width=0.4\linewidth,trim={0 3cm 0 3cm},clip]{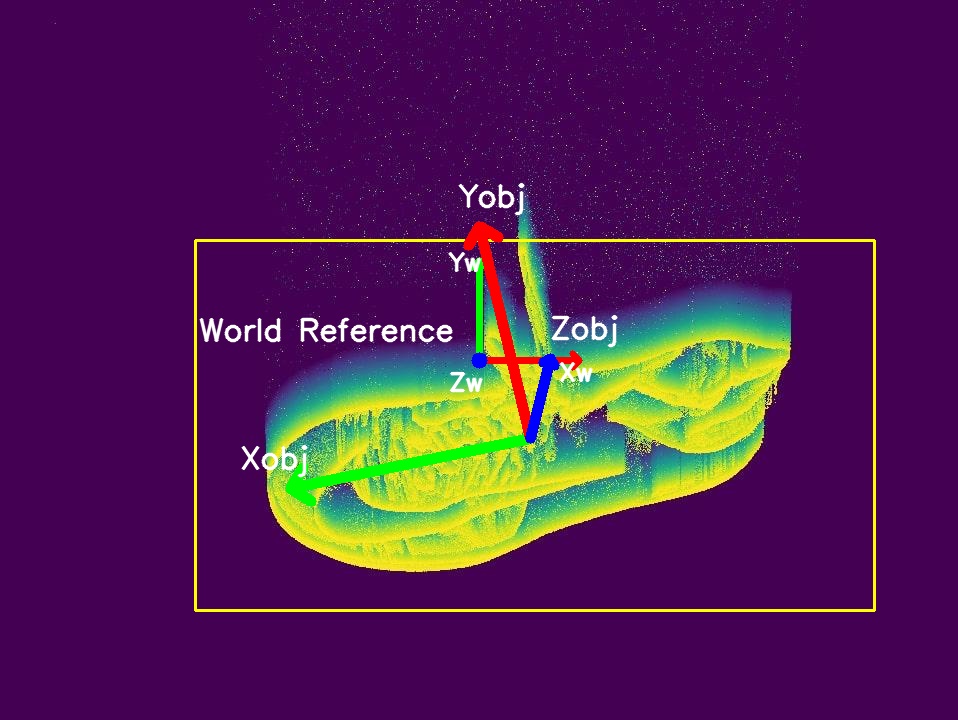}
    
    \caption{Examples of 6D pose annotations for frame-based (left) and event-based (right) data in the Moving6DPoSe-R dataset.}
    \label{fig:moving6dpose_annotation}
\end{figure}

\begin{figure}[!tb]
\centering
    \subcaptionbox*{}%
    [.09\textwidth]{\includegraphics[width=\linewidth]{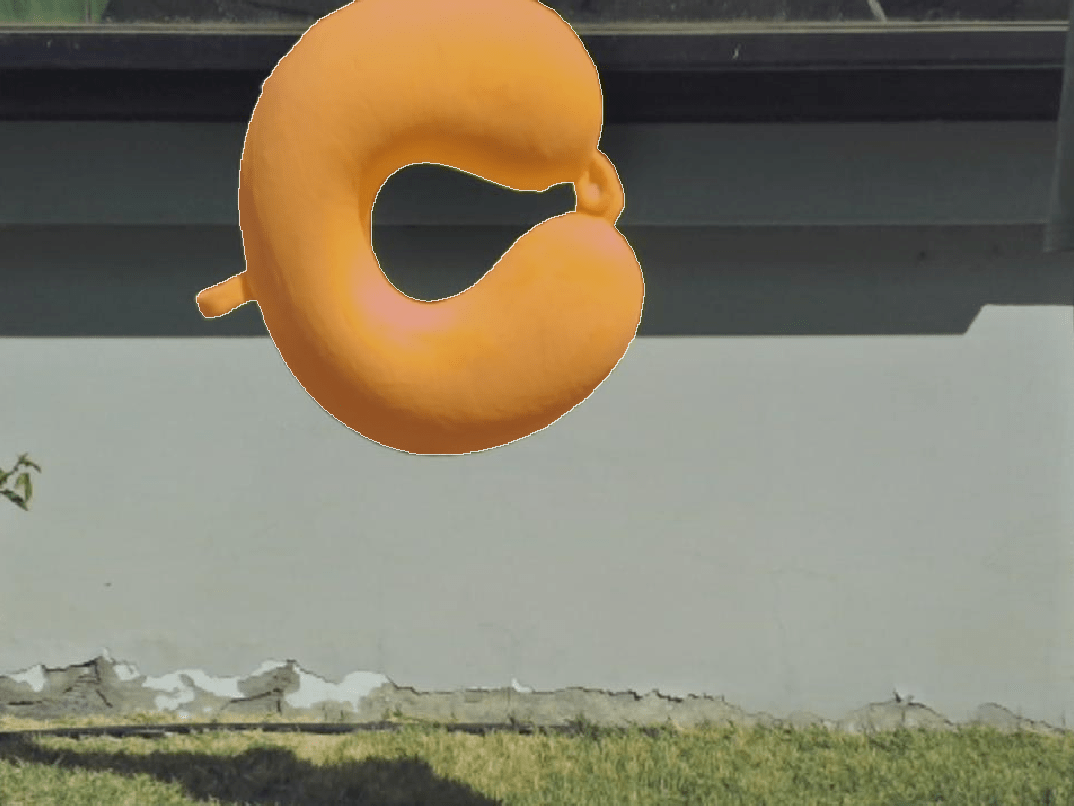 } }
    \subcaptionbox*{}%
    [.09\textwidth]{\includegraphics[width=\linewidth]{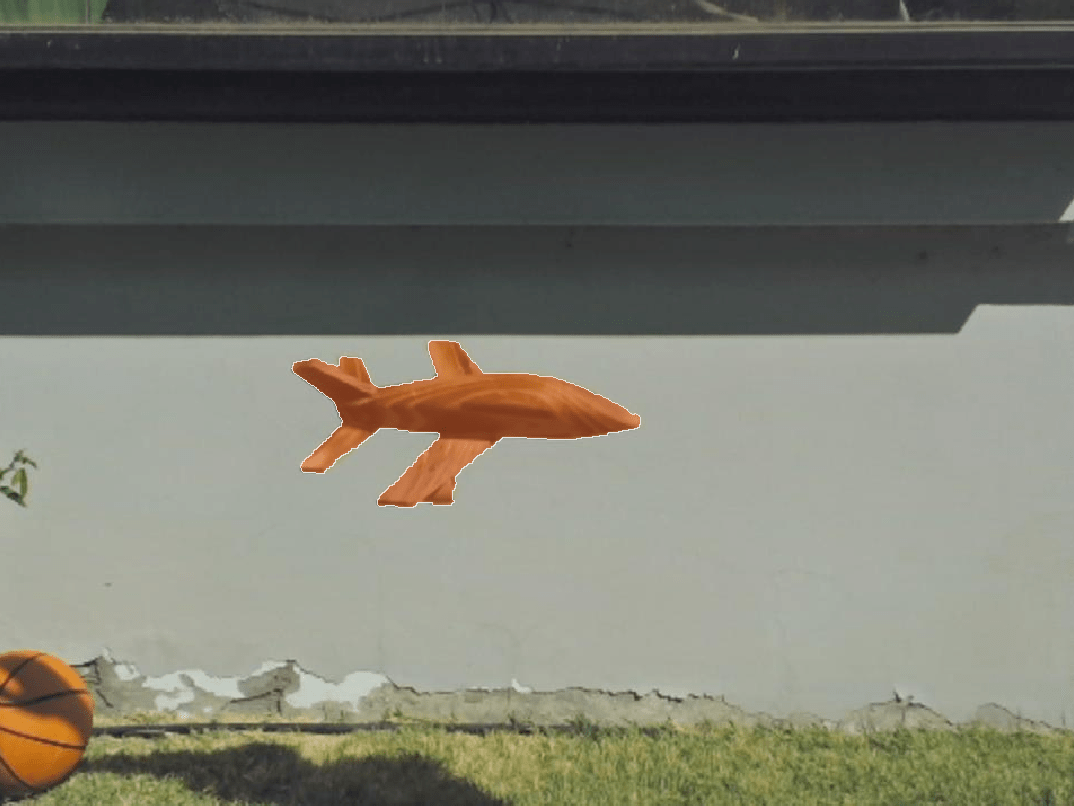 } }
    \subcaptionbox*{}%
    [.09\textwidth]{\includegraphics[width=\linewidth]{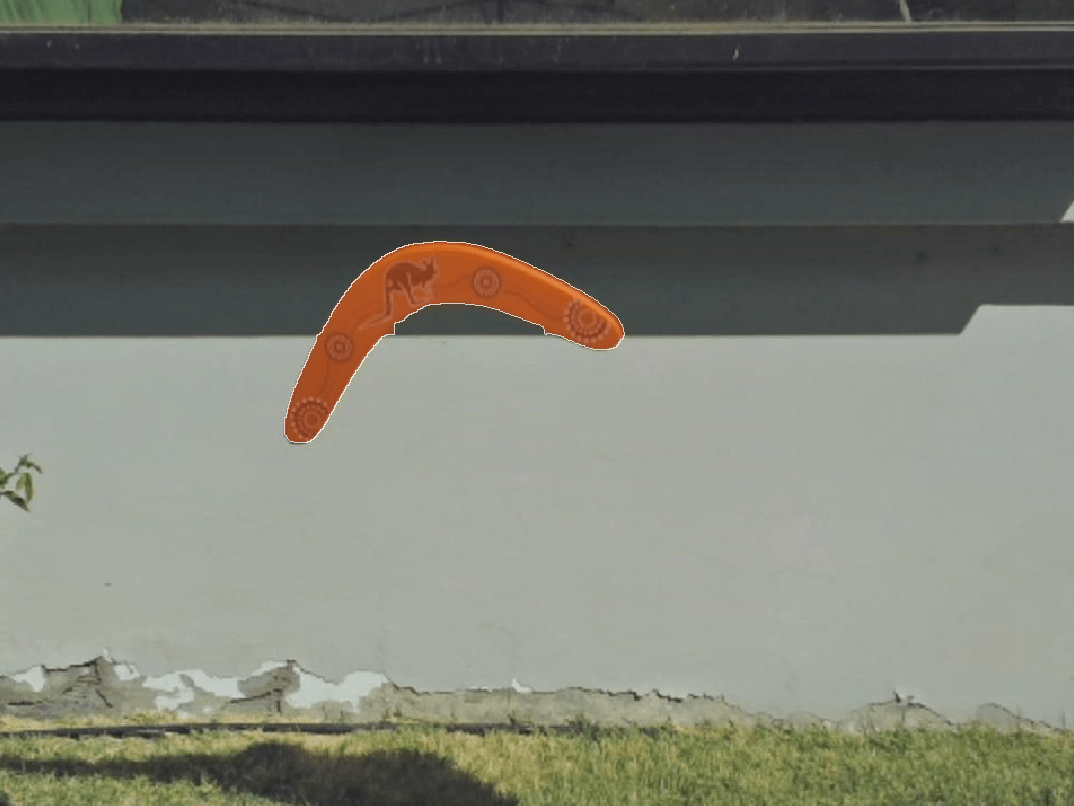 } }
    \subcaptionbox*{}%
    [.09\textwidth]{\includegraphics[width=\linewidth]{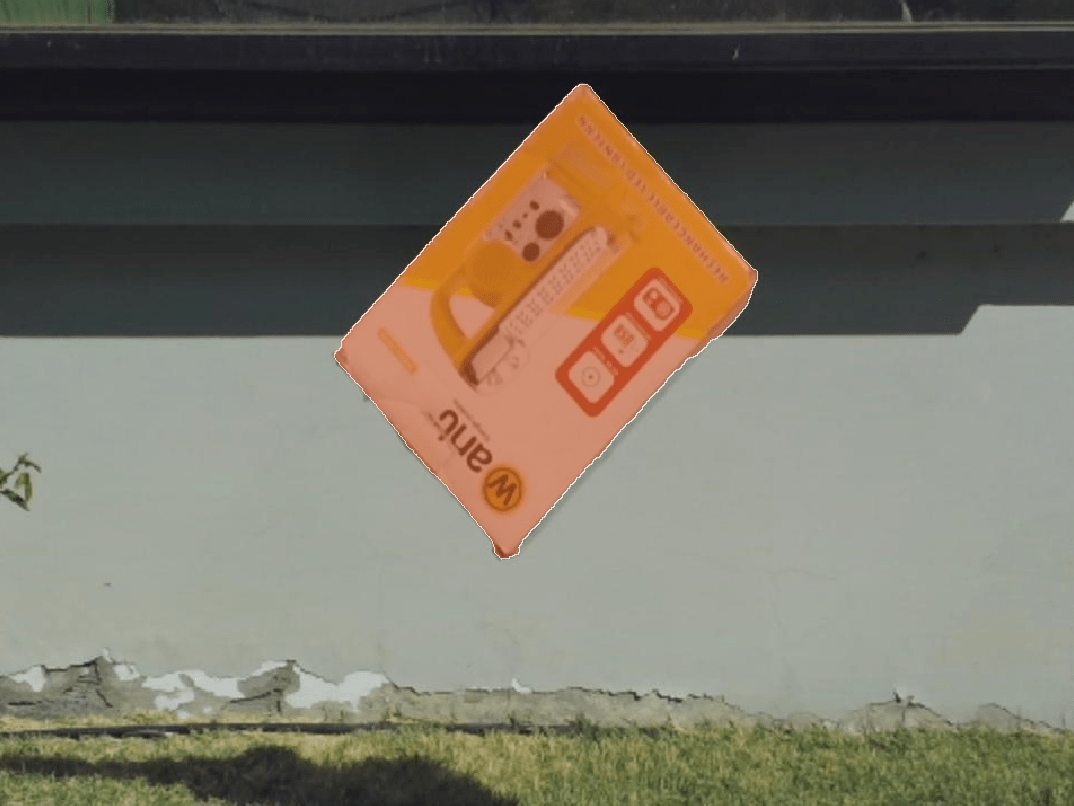 } }
    \subcaptionbox*{}%
    [.09\textwidth]{\includegraphics[width=\linewidth]{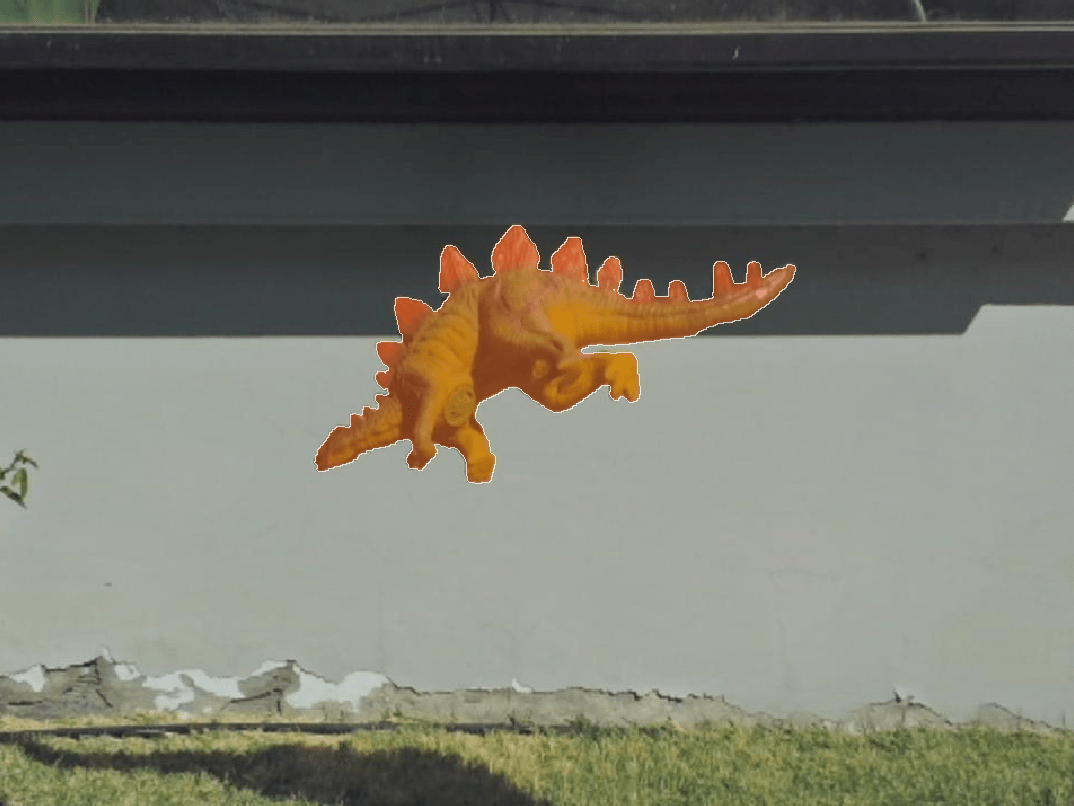 } }
    \subcaptionbox*{}%
    [.09\textwidth]{\includegraphics[width=\linewidth]{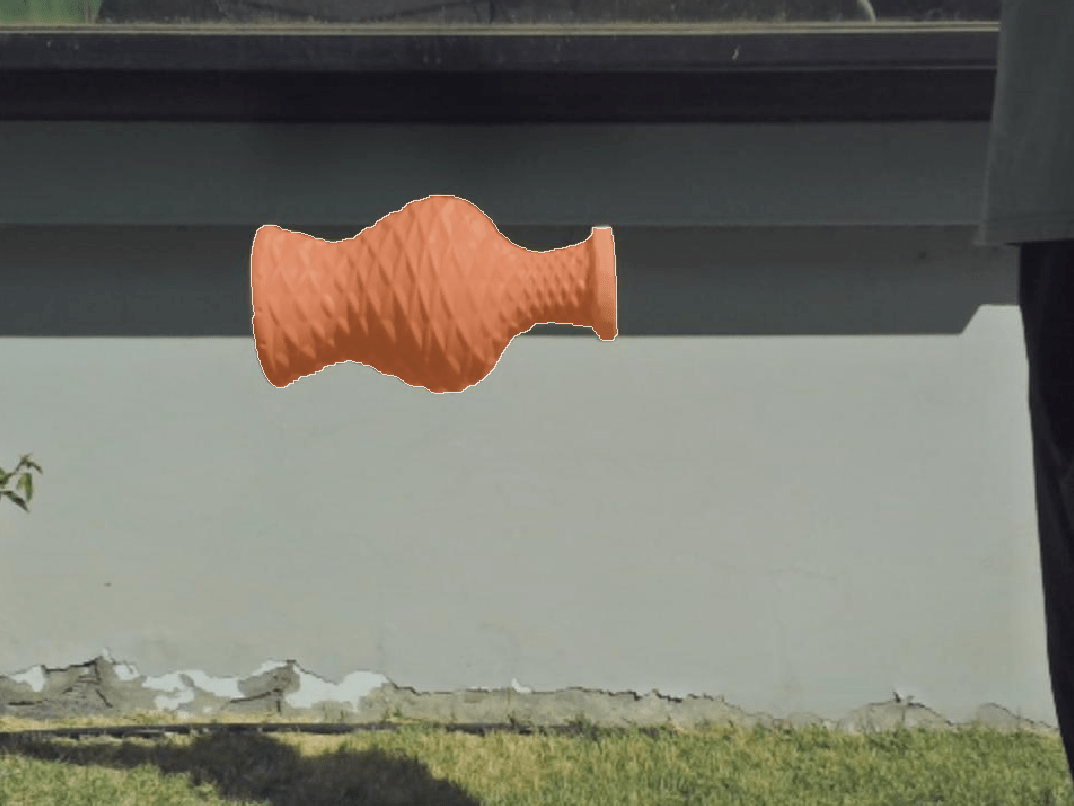 } }
    \subcaptionbox*{}%
    [.09\textwidth]{\includegraphics[width=\linewidth]{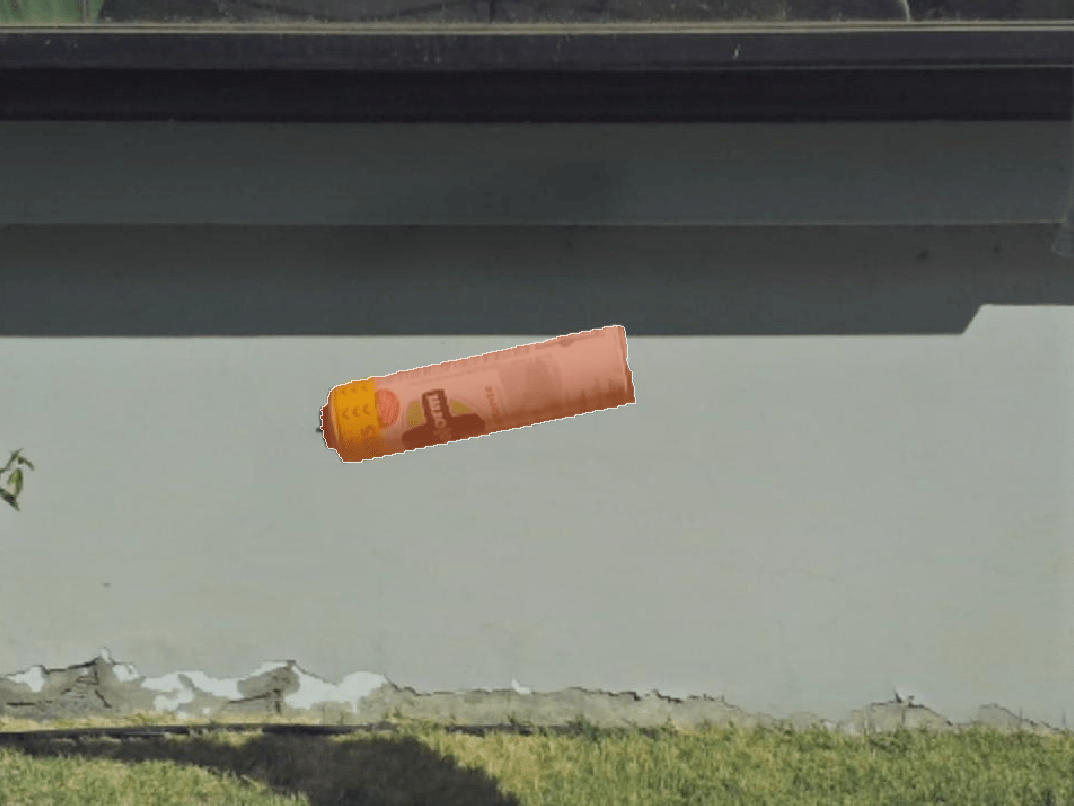 } }
    \subcaptionbox*{}%
    [.09\textwidth]{\includegraphics[width=\linewidth]{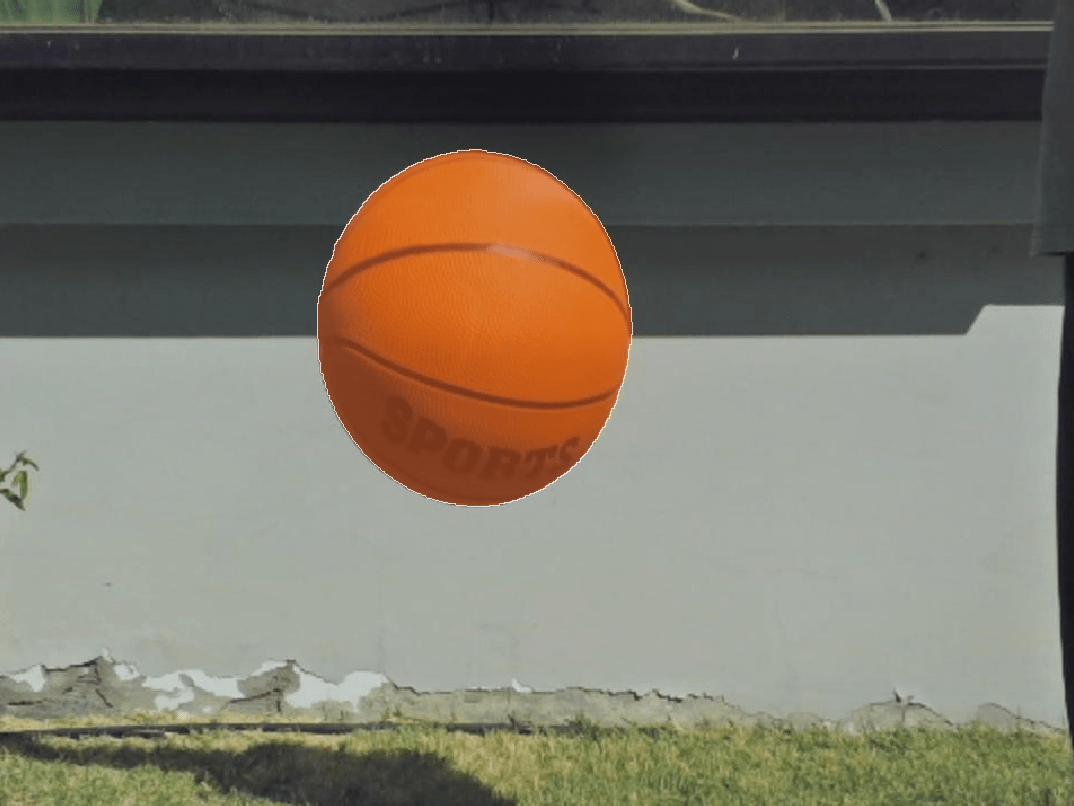 } }
    \subcaptionbox*{}%
    [.09\textwidth]{\includegraphics[width=\linewidth]{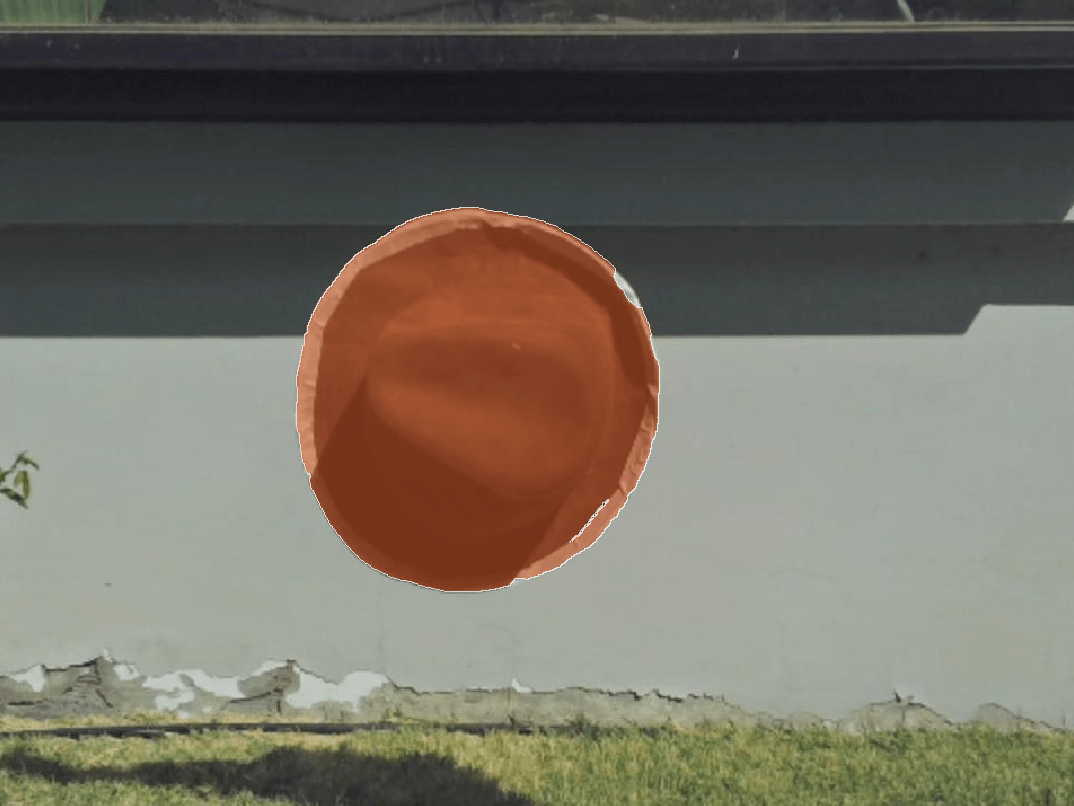 } }
    \subcaptionbox*{}%
    [.09\textwidth]{\includegraphics[width=\linewidth]{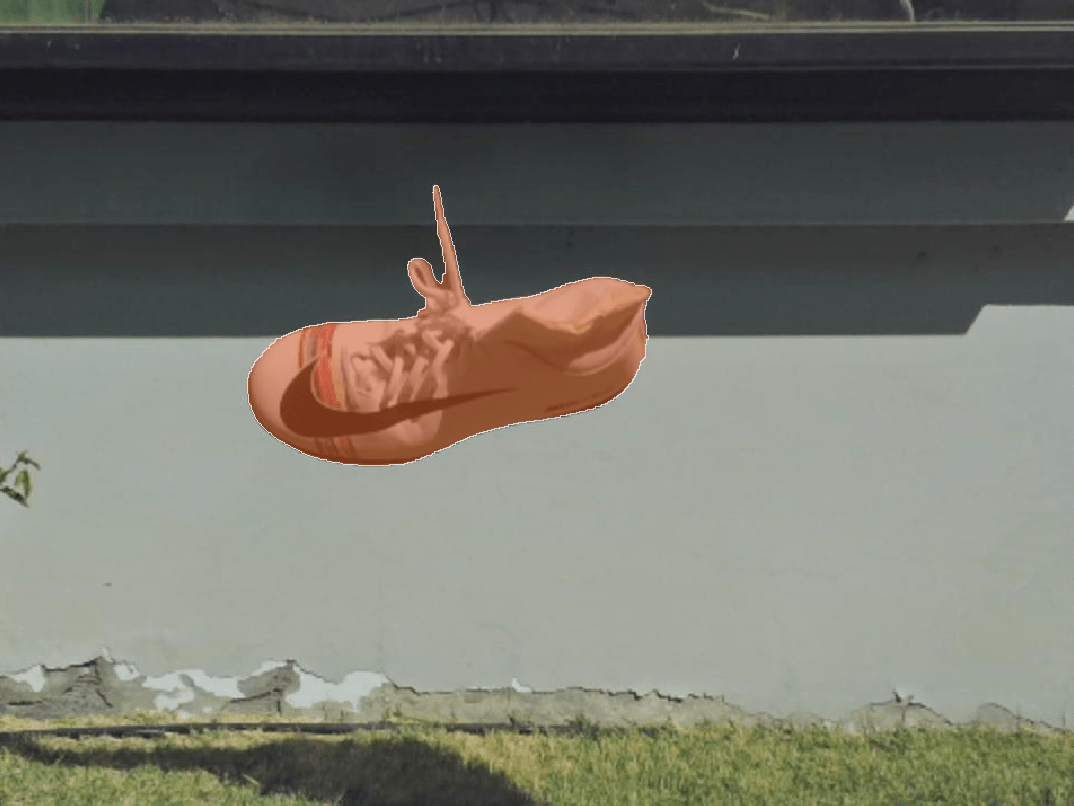 } }

    \vspace{-3mm}

    \subcaptionbox*{}%
    [.09\textwidth]{\includegraphics[width=\linewidth]{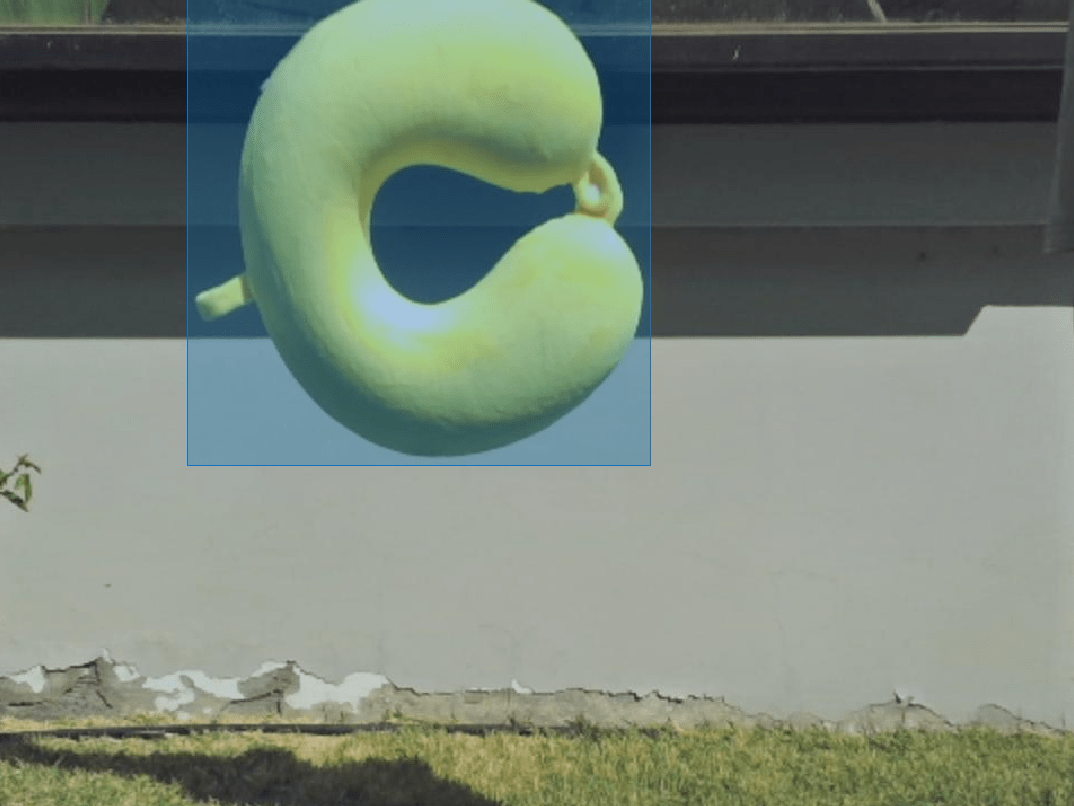 } }
    \subcaptionbox*{}%
    [.09\textwidth]{\includegraphics[width=\linewidth]{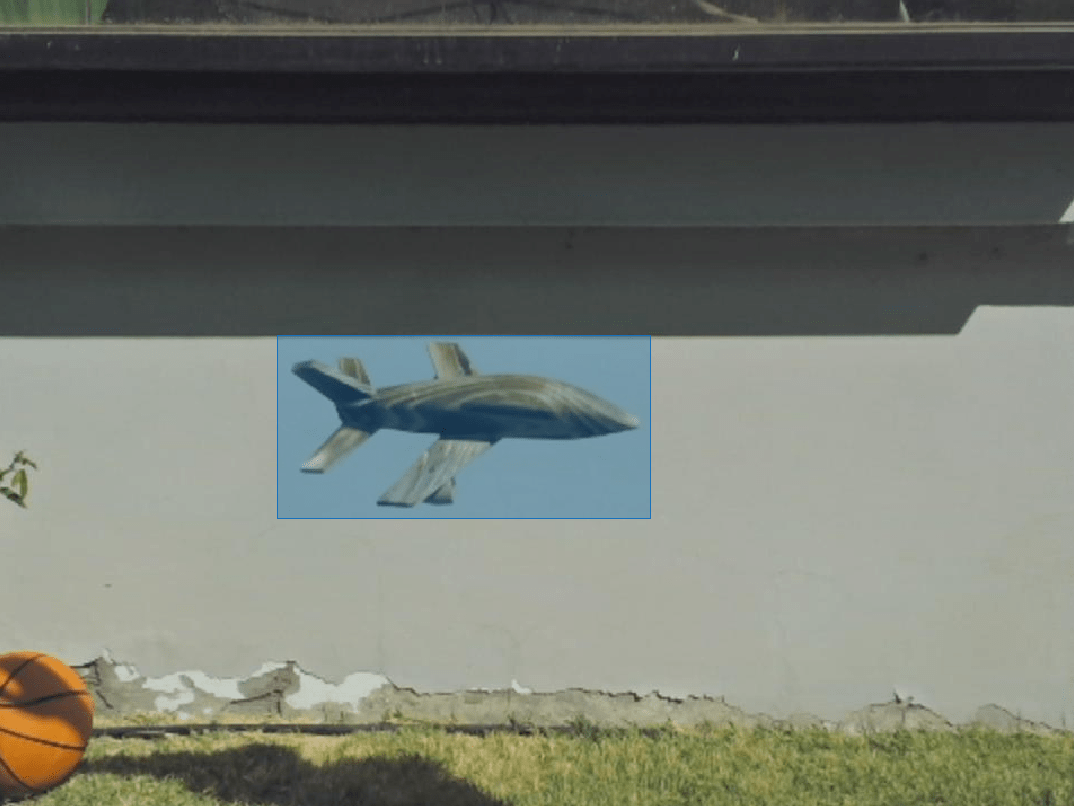 } }
    \subcaptionbox*{}%
    [.09\textwidth]{\includegraphics[width=\linewidth]{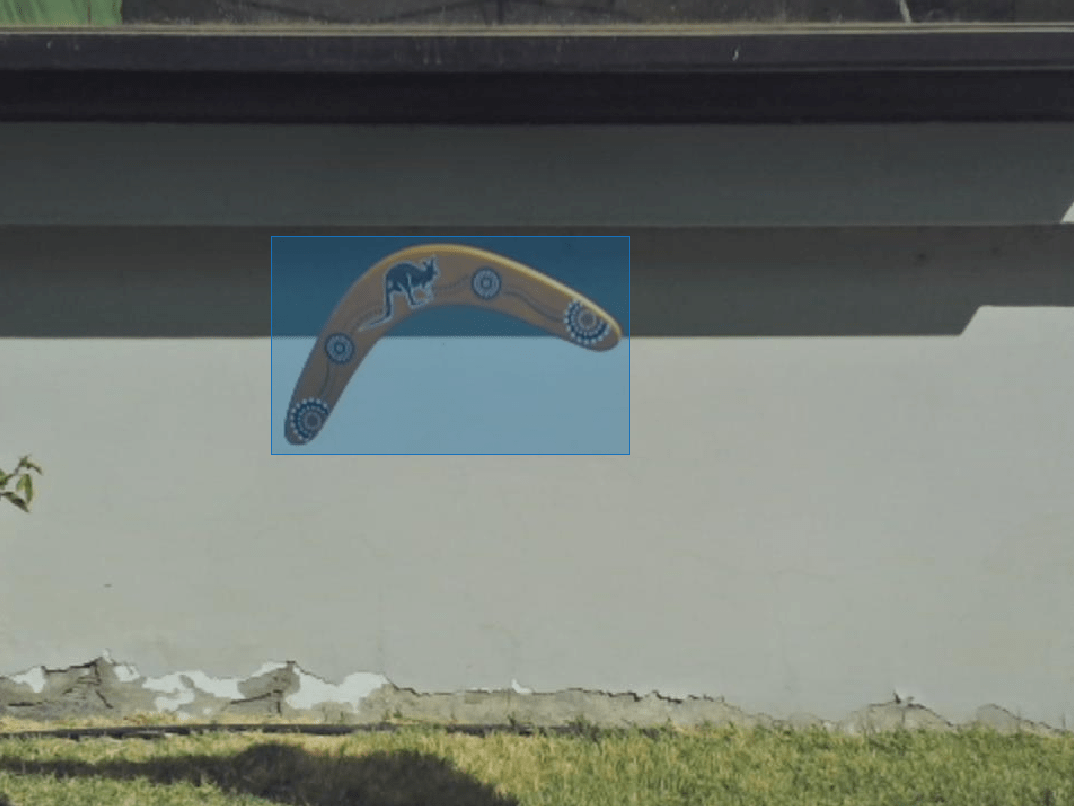 } }
    \subcaptionbox*{}%
    [.09\textwidth]{\includegraphics[width=\linewidth]{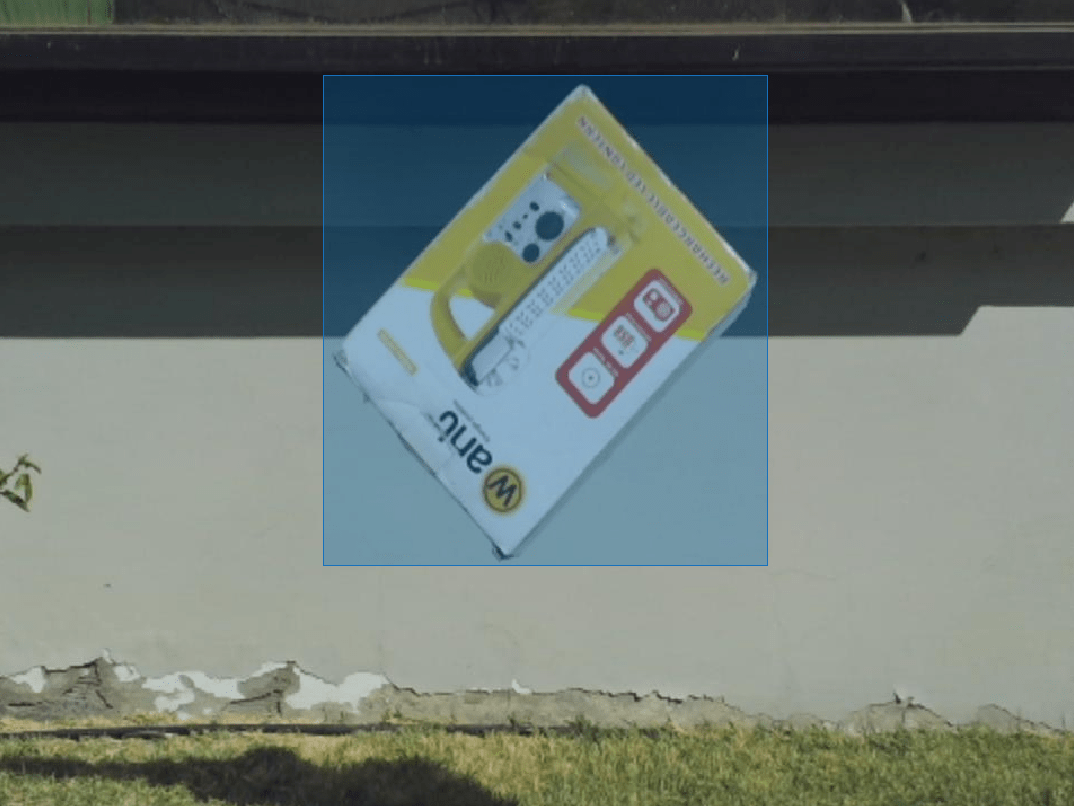 } }
    \subcaptionbox*{}%
    [.09\textwidth]{\includegraphics[width=\linewidth]{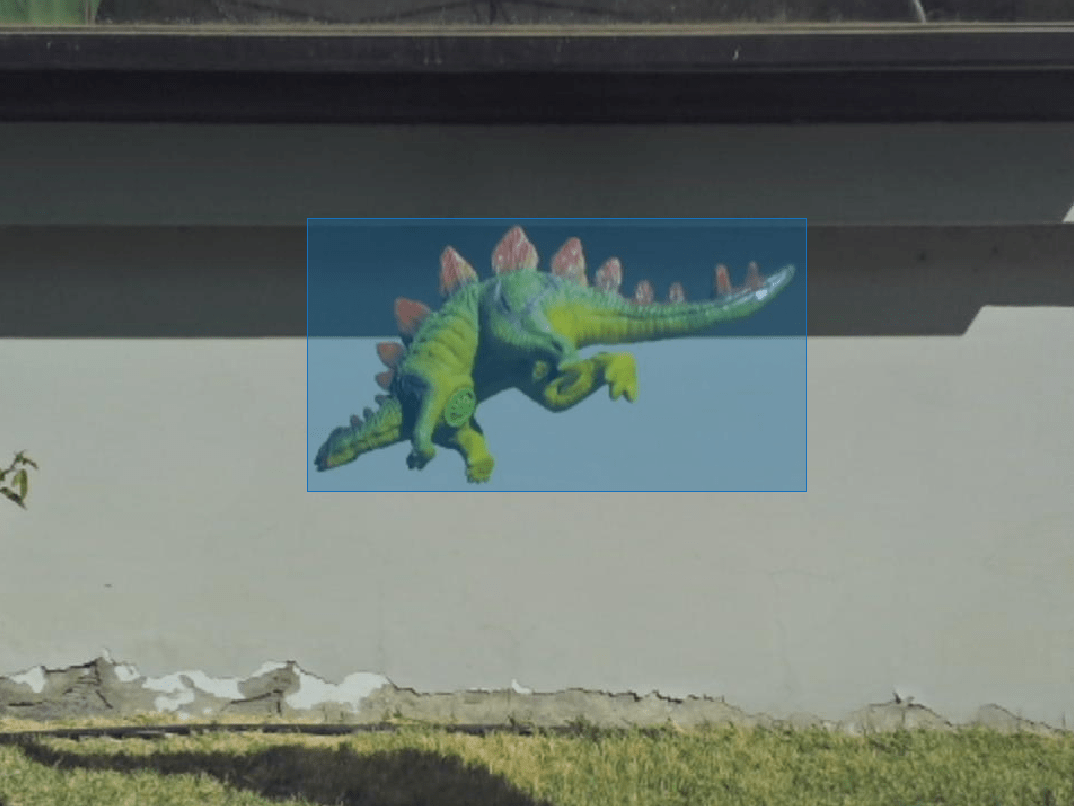 } }
    \subcaptionbox*{}%
    [.09\textwidth]{\includegraphics[width=\linewidth]{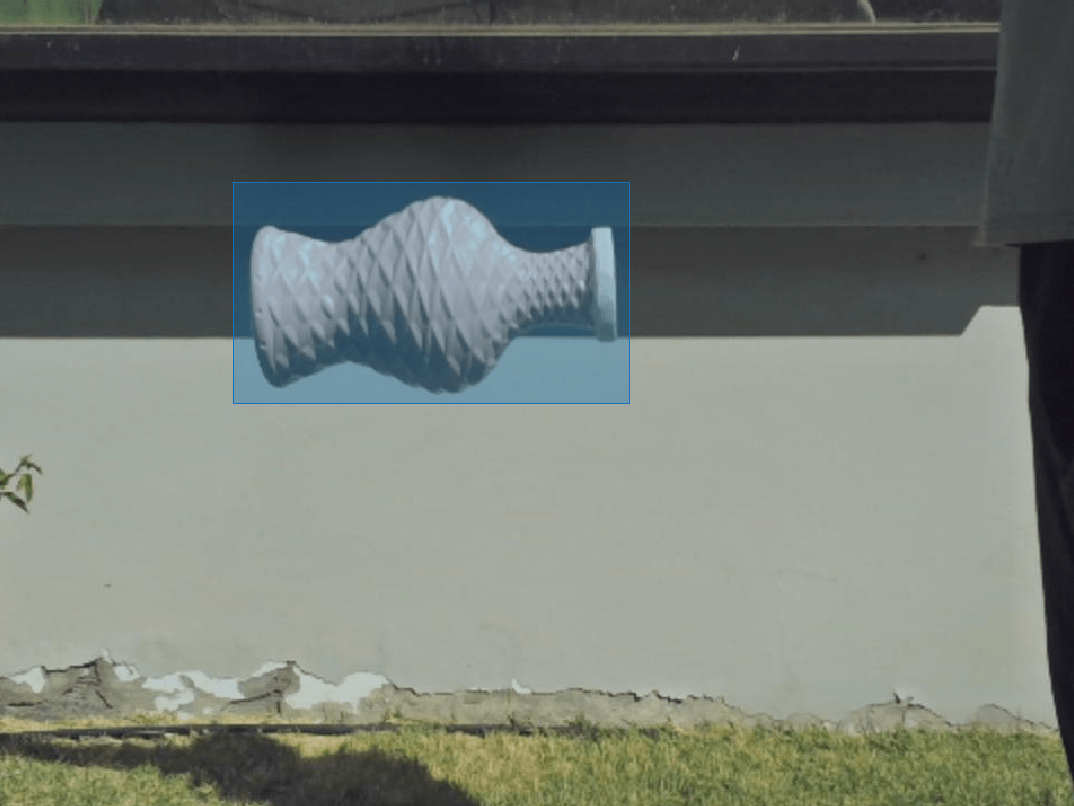 } }
    \subcaptionbox*{}%
    [.09\textwidth]{\includegraphics[width=\linewidth]{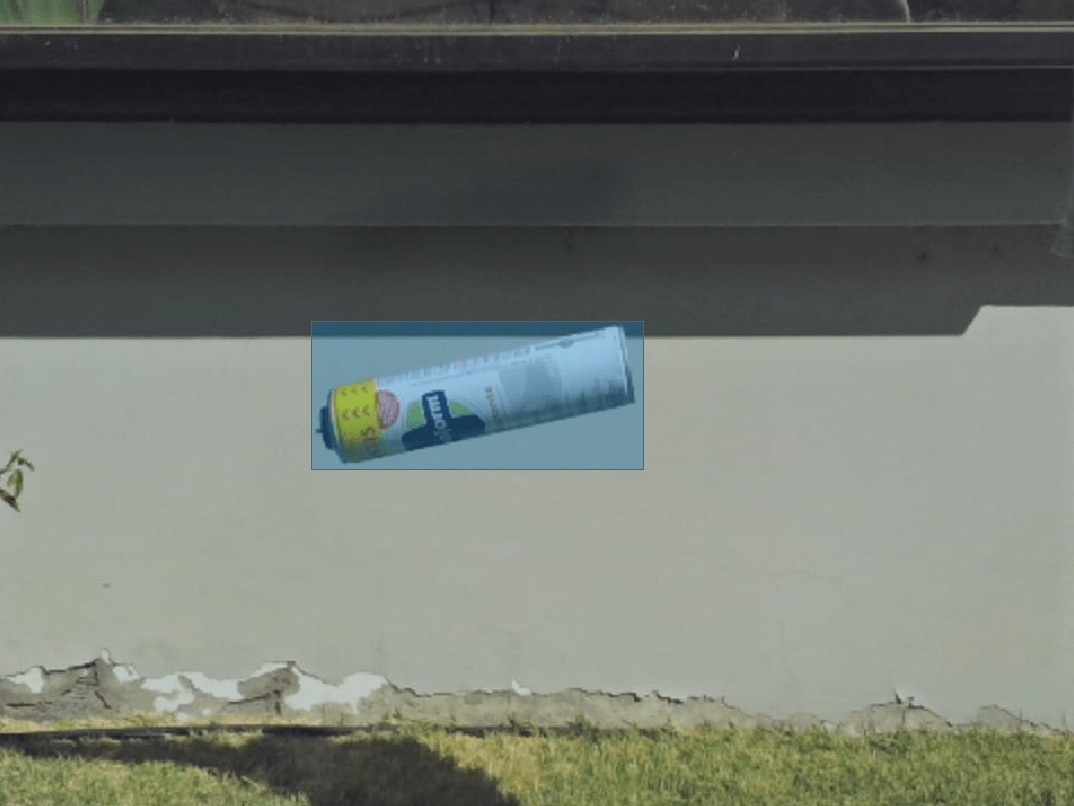 } }
    \subcaptionbox*{}%
    [.09\textwidth]{\includegraphics[width=\linewidth]{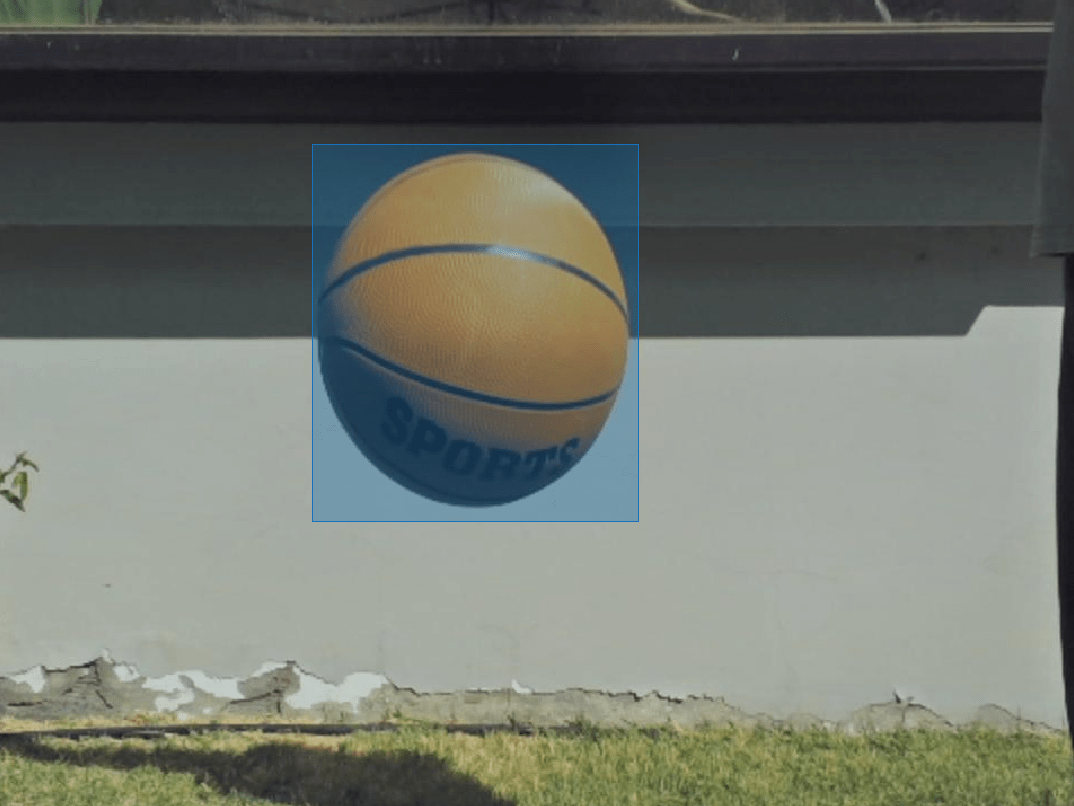 } }
    \subcaptionbox*{}%
    [.09\textwidth]{\includegraphics[width=\linewidth]{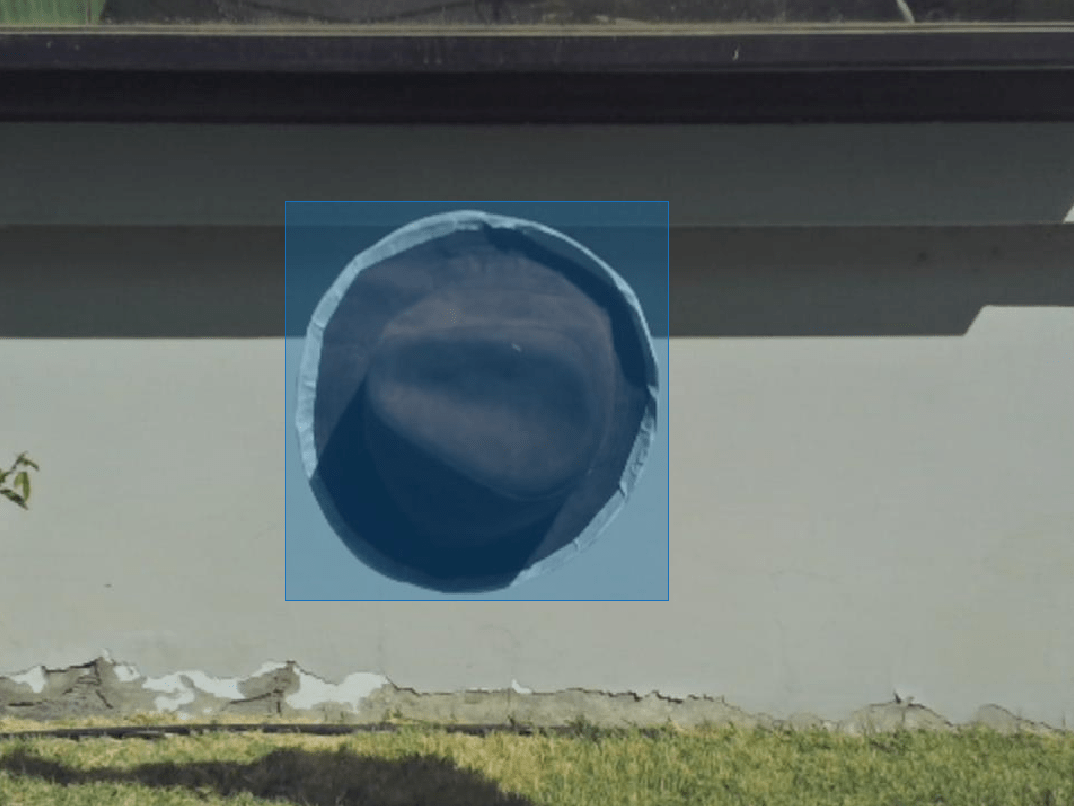 } }
    \subcaptionbox*{}%
    [.09\textwidth]{\includegraphics[width=\linewidth]{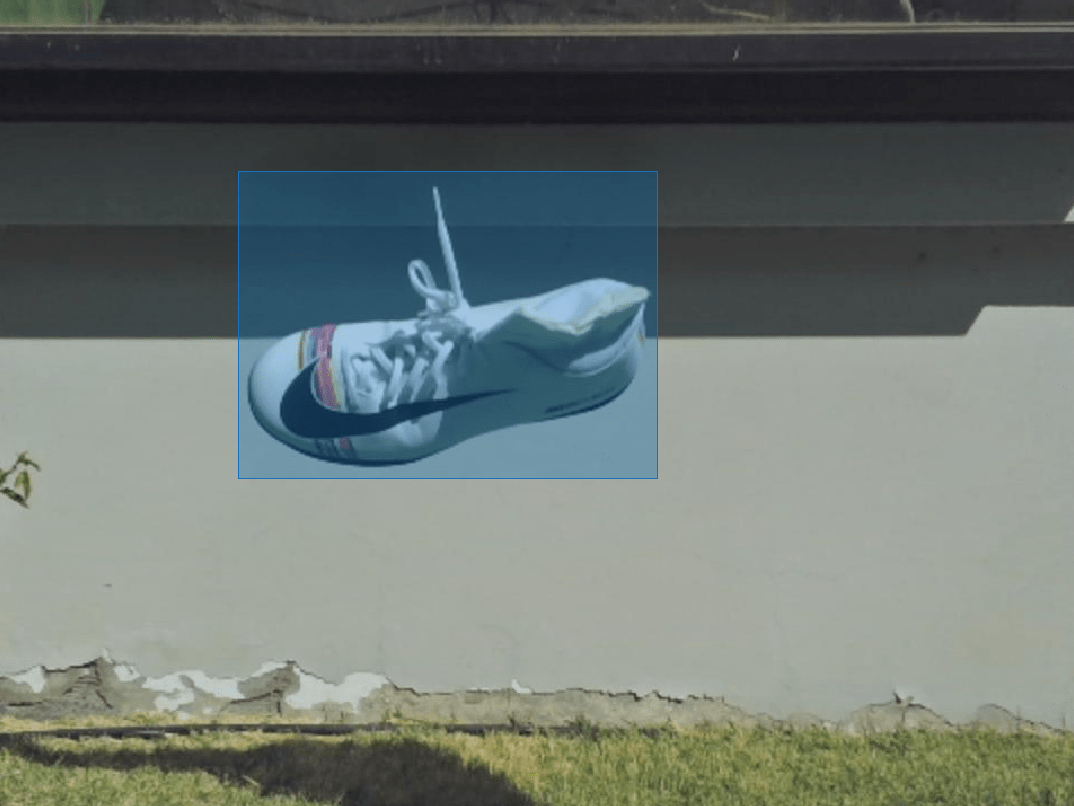 } }

    \vspace{-3mm}

    \subcaptionbox*{}%
    [.09\textwidth]{\includegraphics[width=\linewidth]{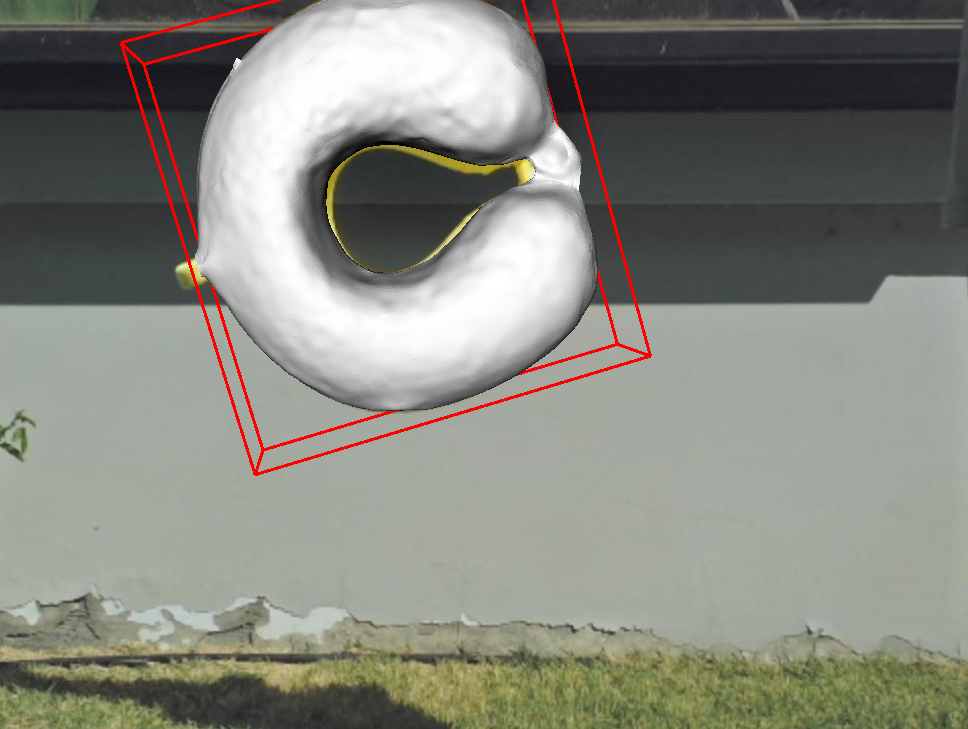 } }
    \subcaptionbox*{}%
    [.09\textwidth]{\includegraphics[width=\linewidth]{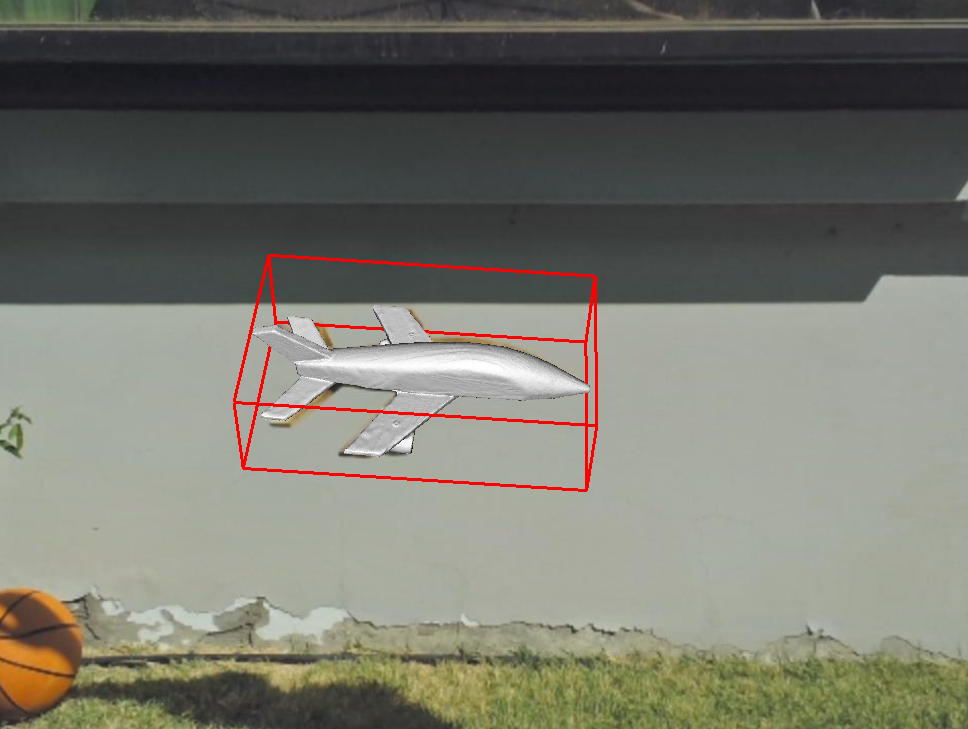 } }
    \subcaptionbox*{}%
    [.09\textwidth]{\includegraphics[width=\linewidth]{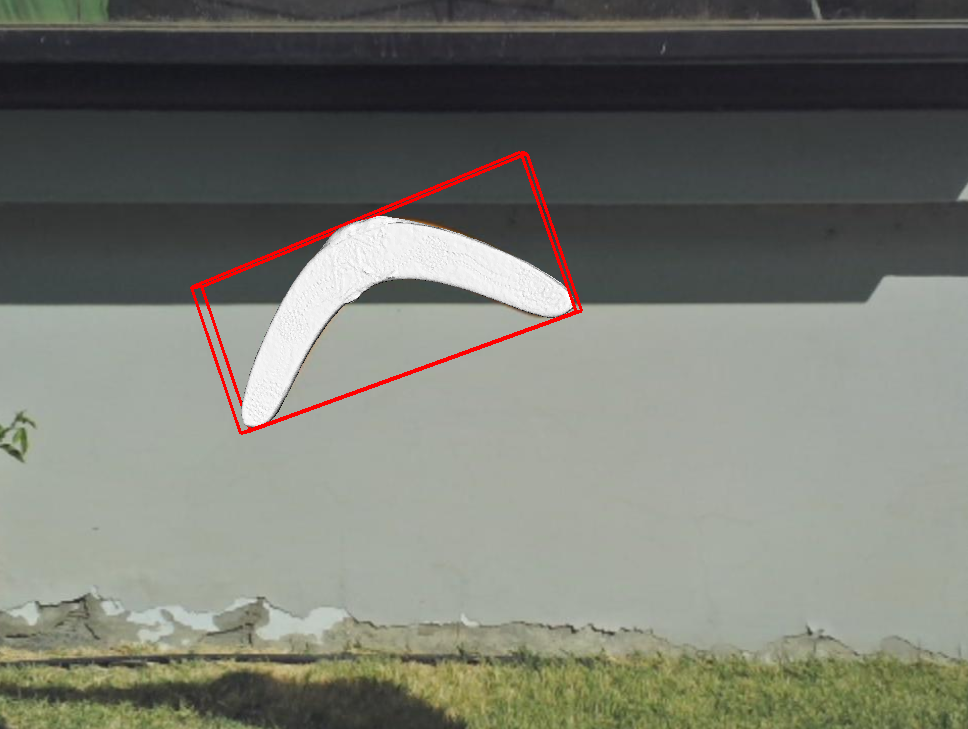 } }
    \subcaptionbox*{}%
    [.09\textwidth]{\includegraphics[width=\linewidth]{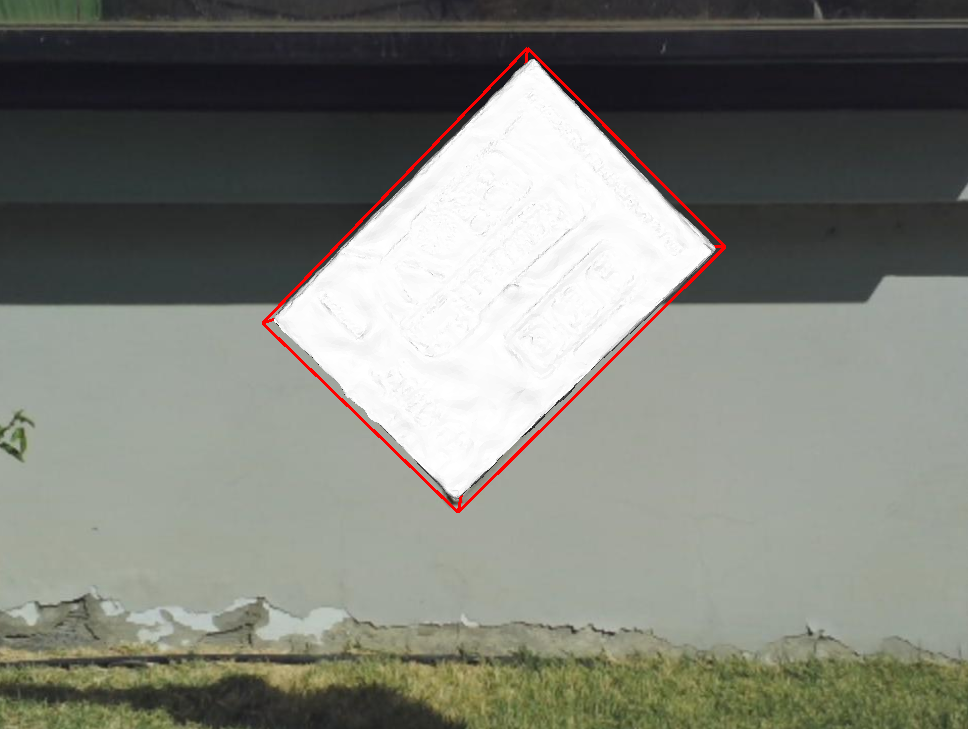 } }
    \subcaptionbox*{}%
    [.09\textwidth]{\includegraphics[width=\linewidth]{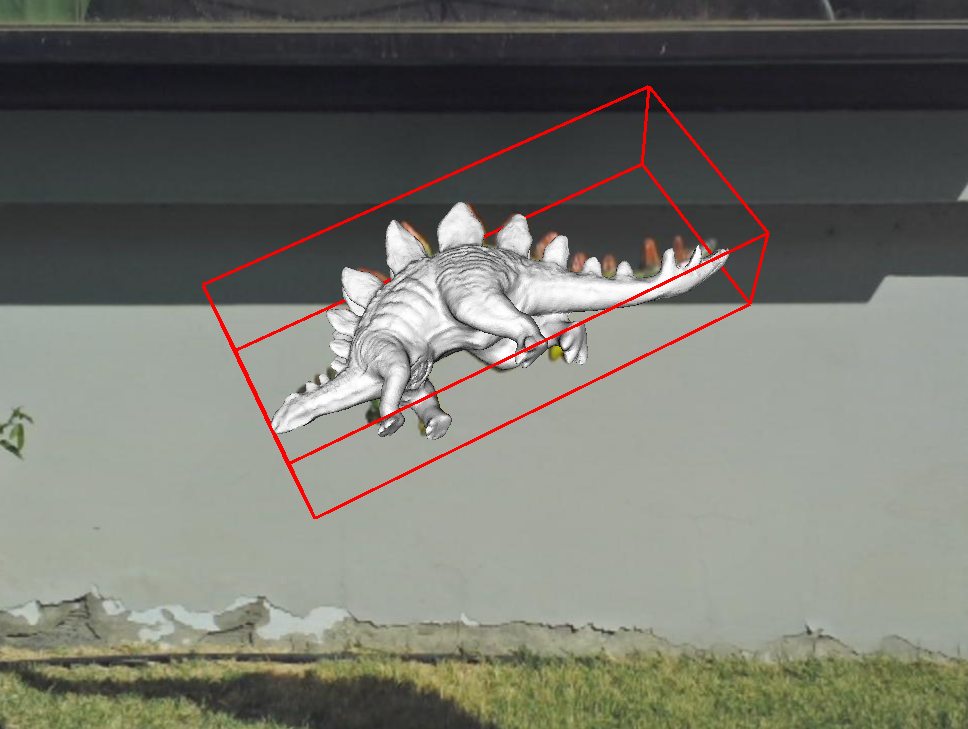 } }
    \subcaptionbox*{}%
    [.09\textwidth]{\includegraphics[width=\linewidth]{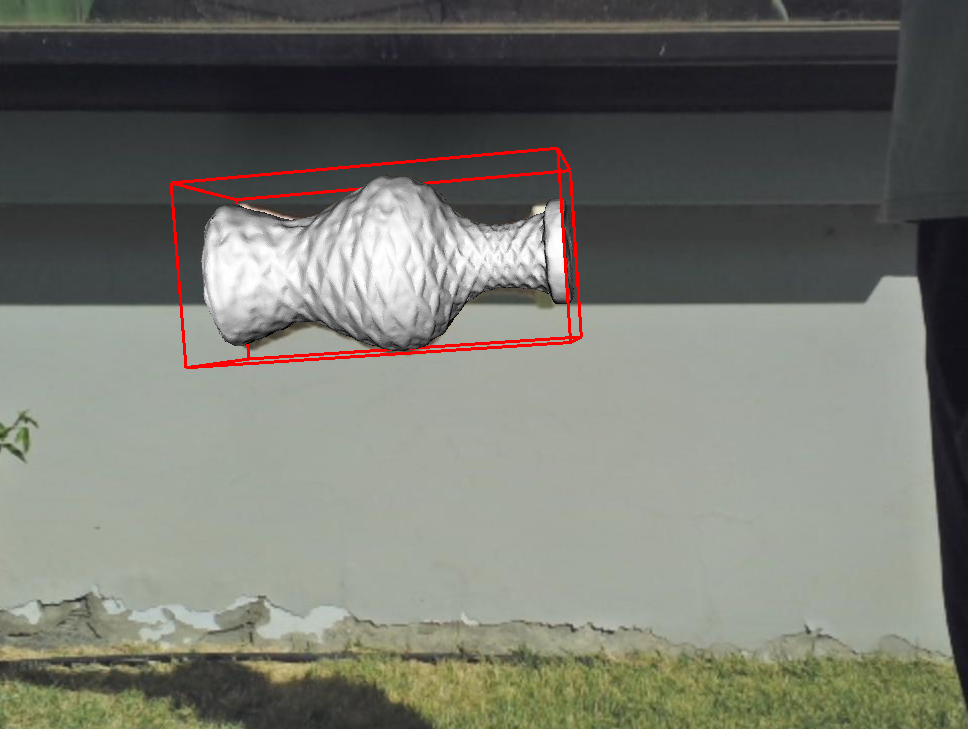 } }
    \subcaptionbox*{}%
    [.09\textwidth]{\includegraphics[width=\linewidth]{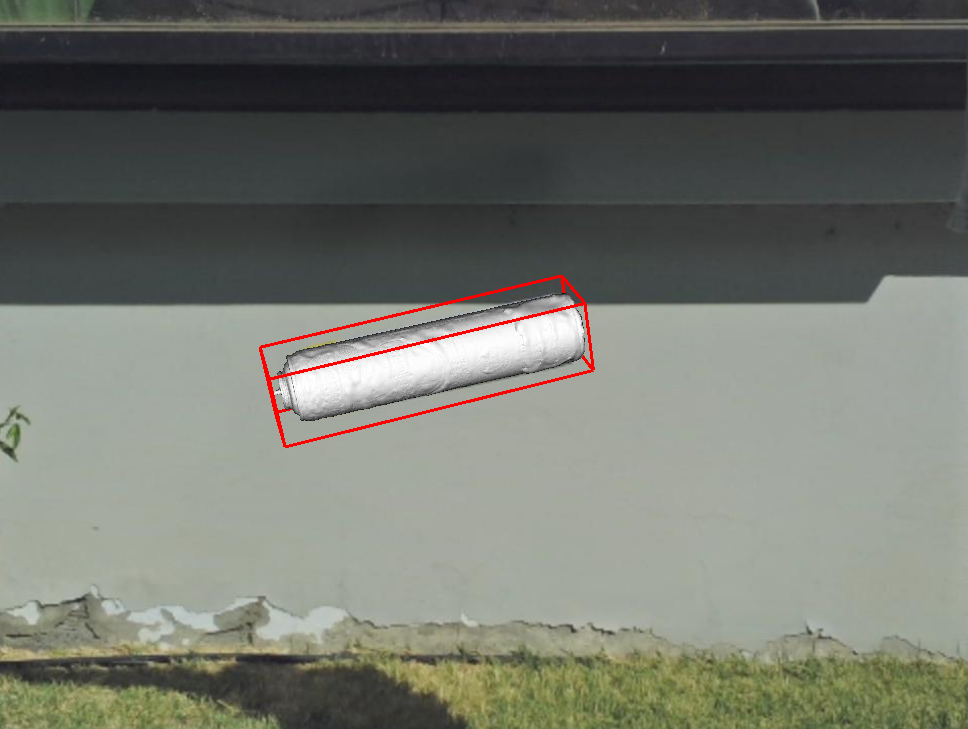 } }
    \subcaptionbox*{}%
    [.09\textwidth]{\includegraphics[width=\linewidth]{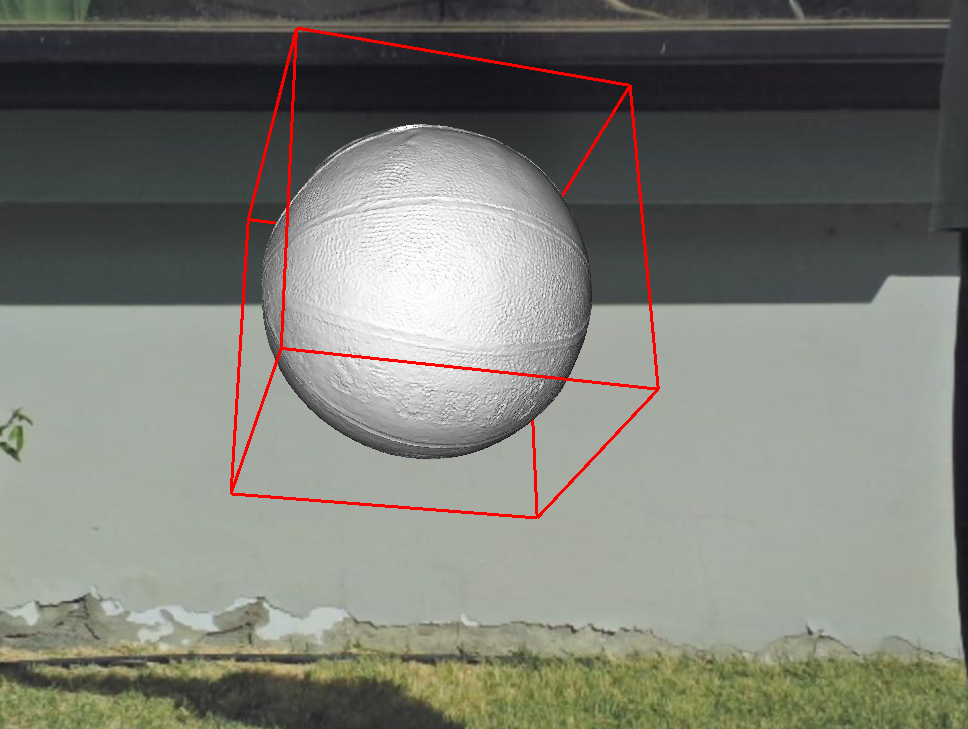 } }
    \subcaptionbox*{}%
    [.09\textwidth]{\includegraphics[width=\linewidth]{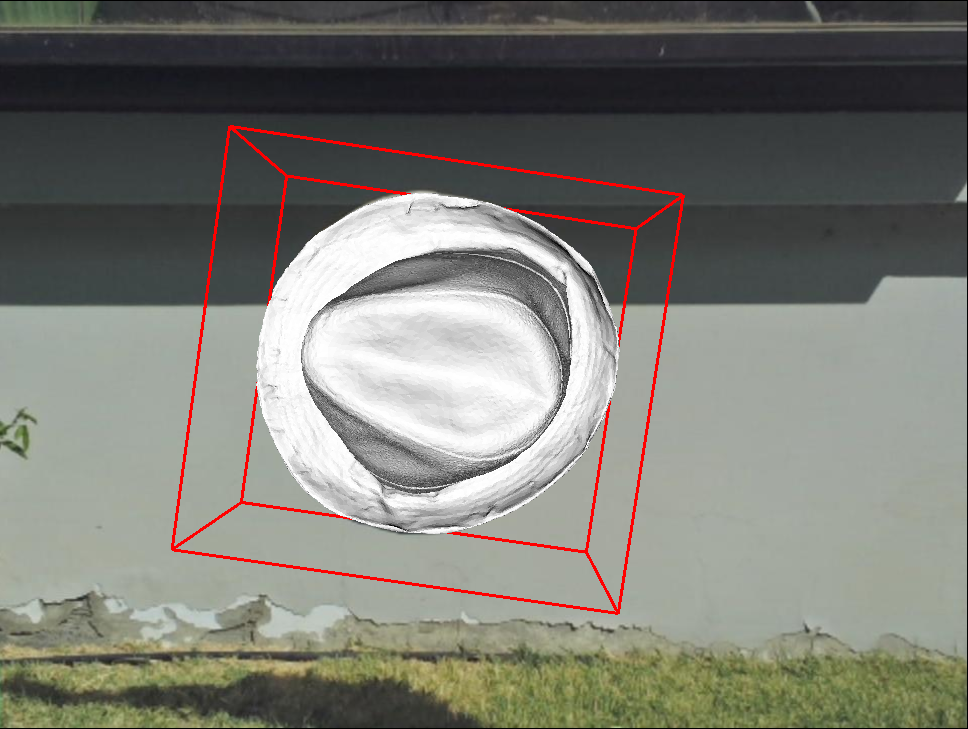 } }
    \subcaptionbox*{}%
    [.09\textwidth]{\includegraphics[width=\linewidth]{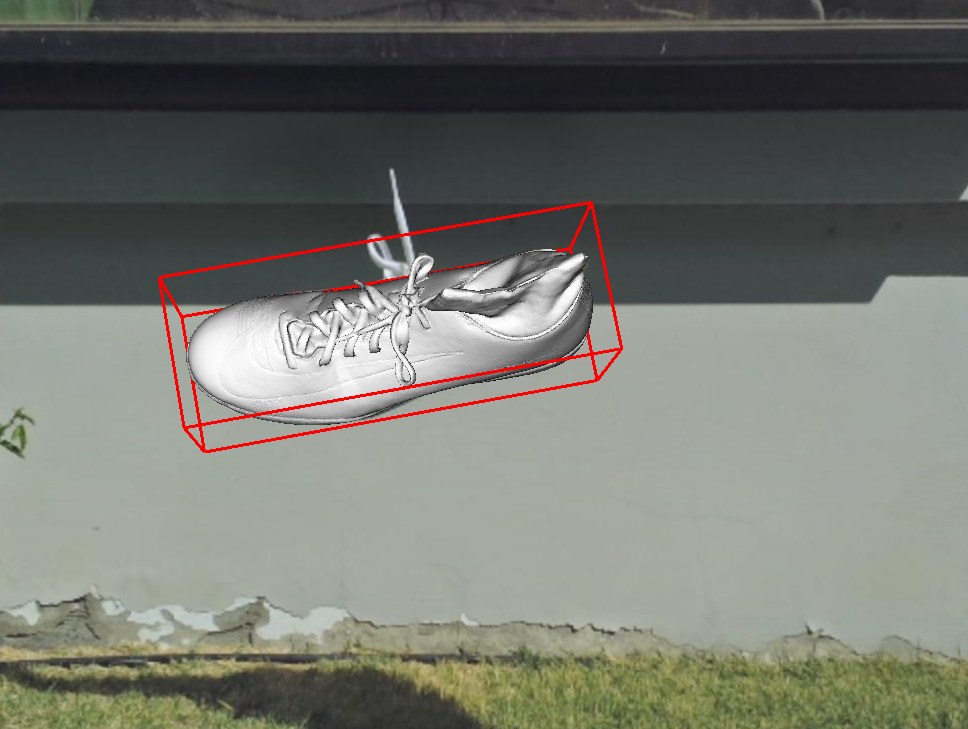 } }

    \vspace{-3mm}

    \subcaptionbox*{}%
    [.09\textwidth]{\includegraphics[width=\linewidth]{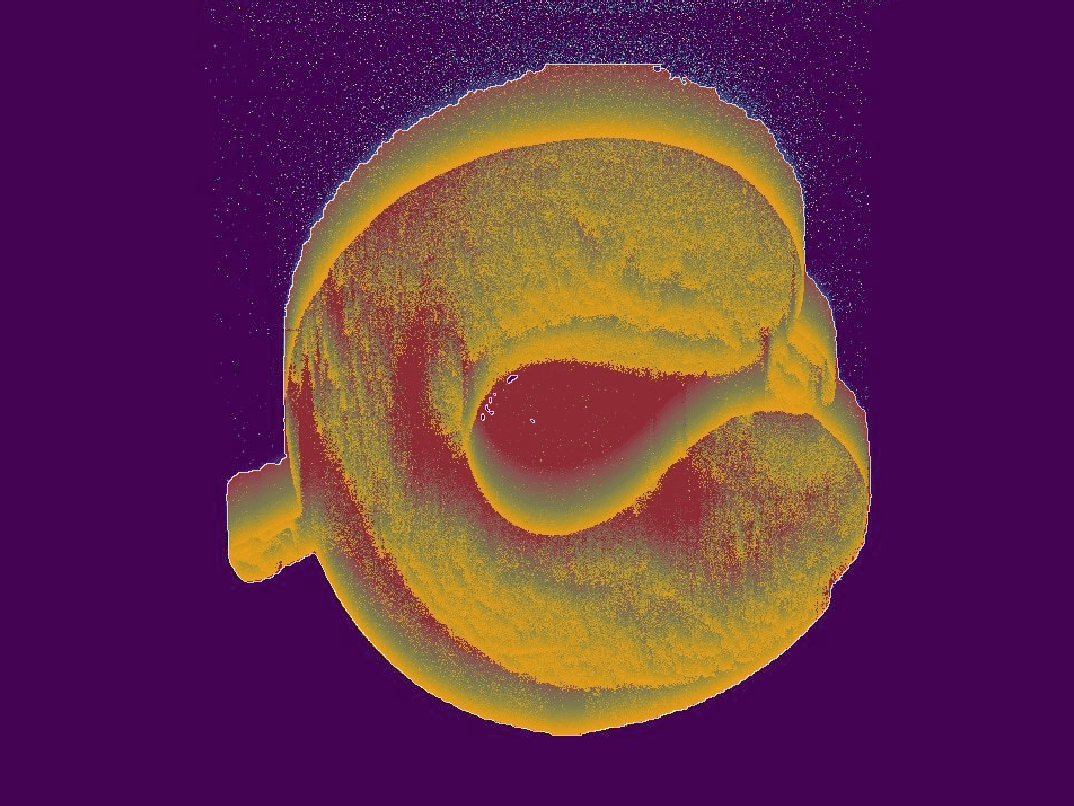 } }
    \subcaptionbox*{}%
    [.09\textwidth]{\includegraphics[width=\linewidth]{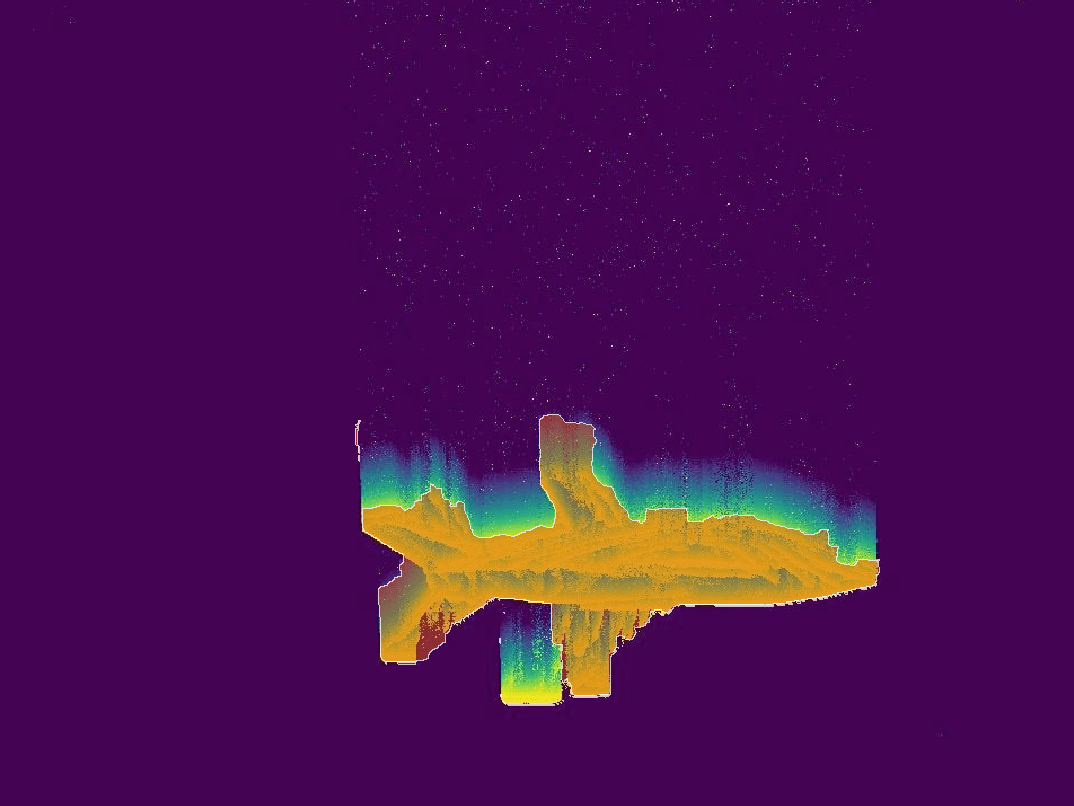 } }
    \subcaptionbox*{}%
    [.09\textwidth]{\includegraphics[width=\linewidth]{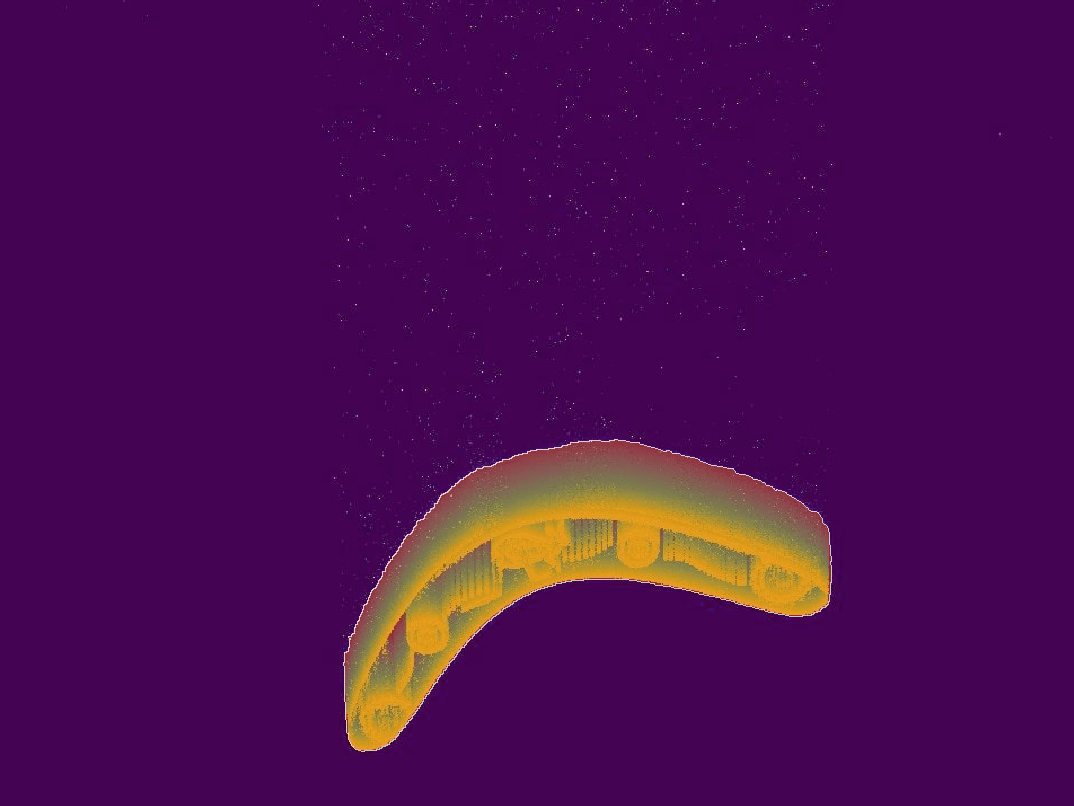 } }
    \subcaptionbox*{}%
    [.09\textwidth]{\includegraphics[width=\linewidth]{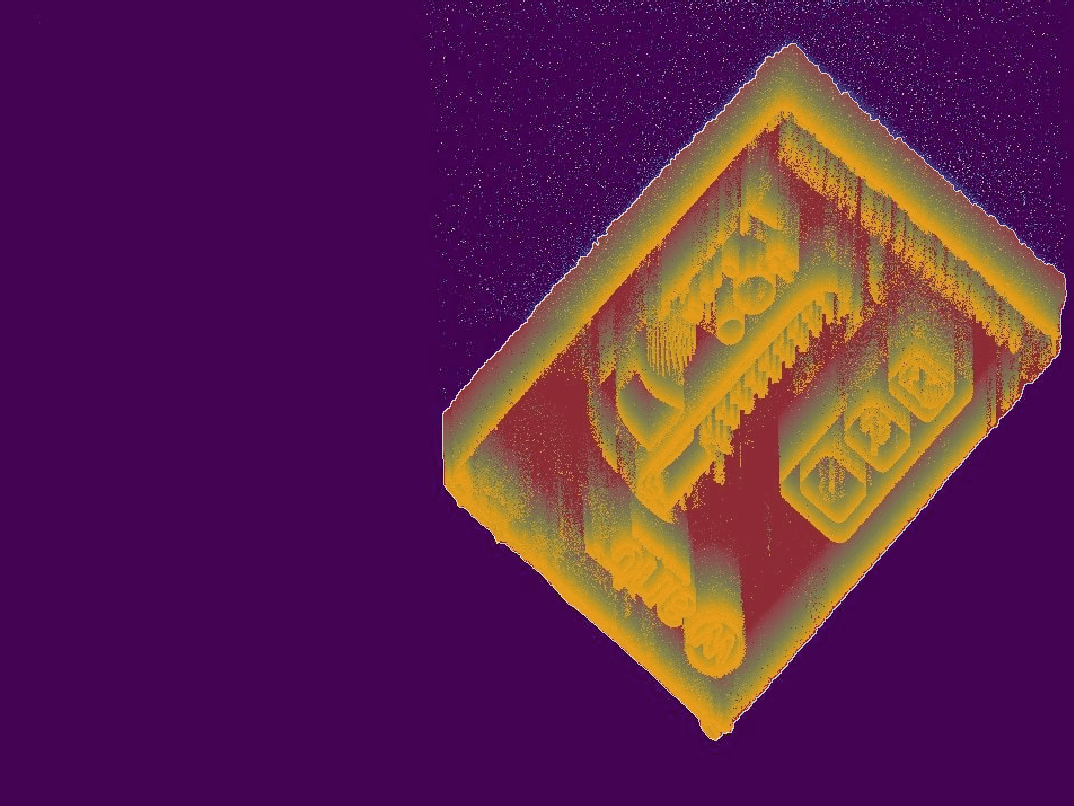 } }
    \subcaptionbox*{}%
    [.09\textwidth]{\includegraphics[width=\linewidth]{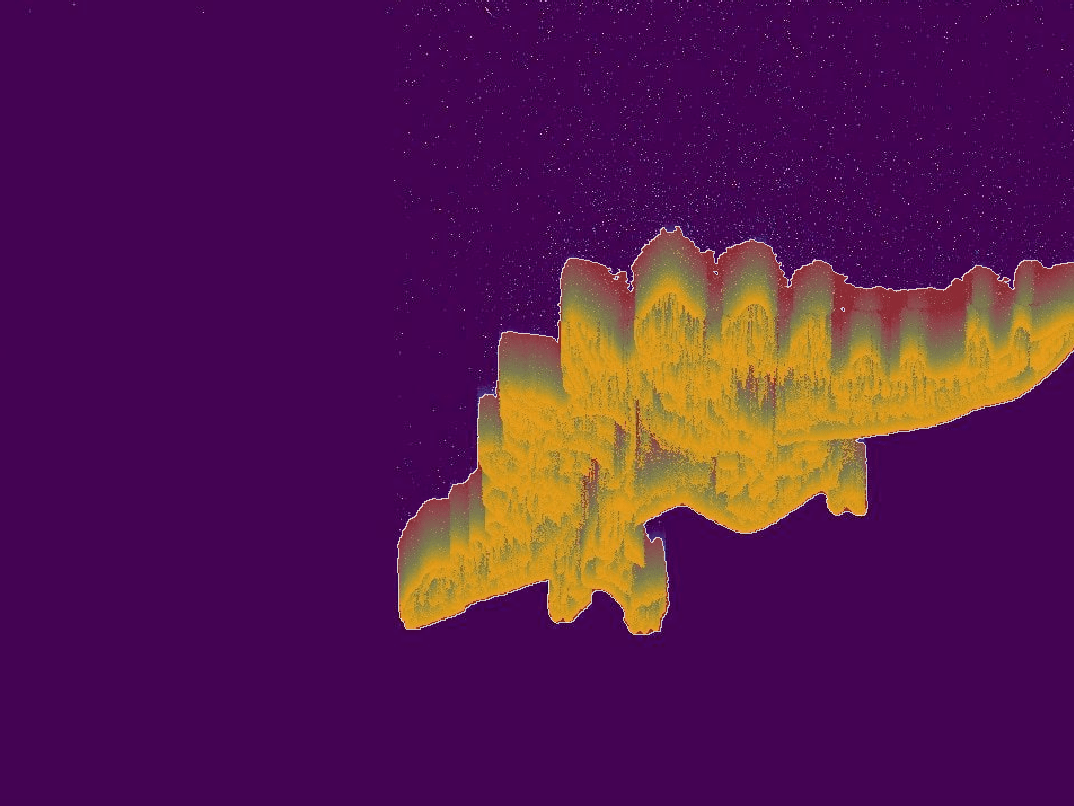 } }
    \subcaptionbox*{}%
    [.09\textwidth]{\includegraphics[width=\linewidth]{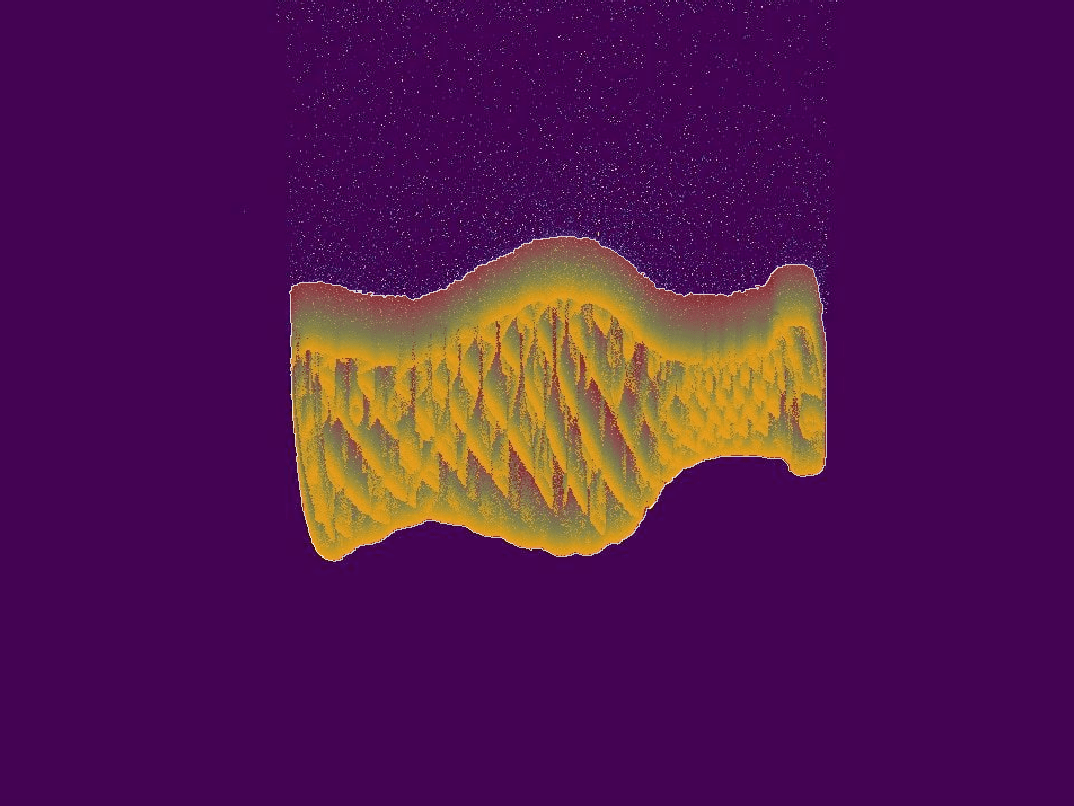 } }
    \subcaptionbox*{}%
    [.09\textwidth]{\includegraphics[width=\linewidth]{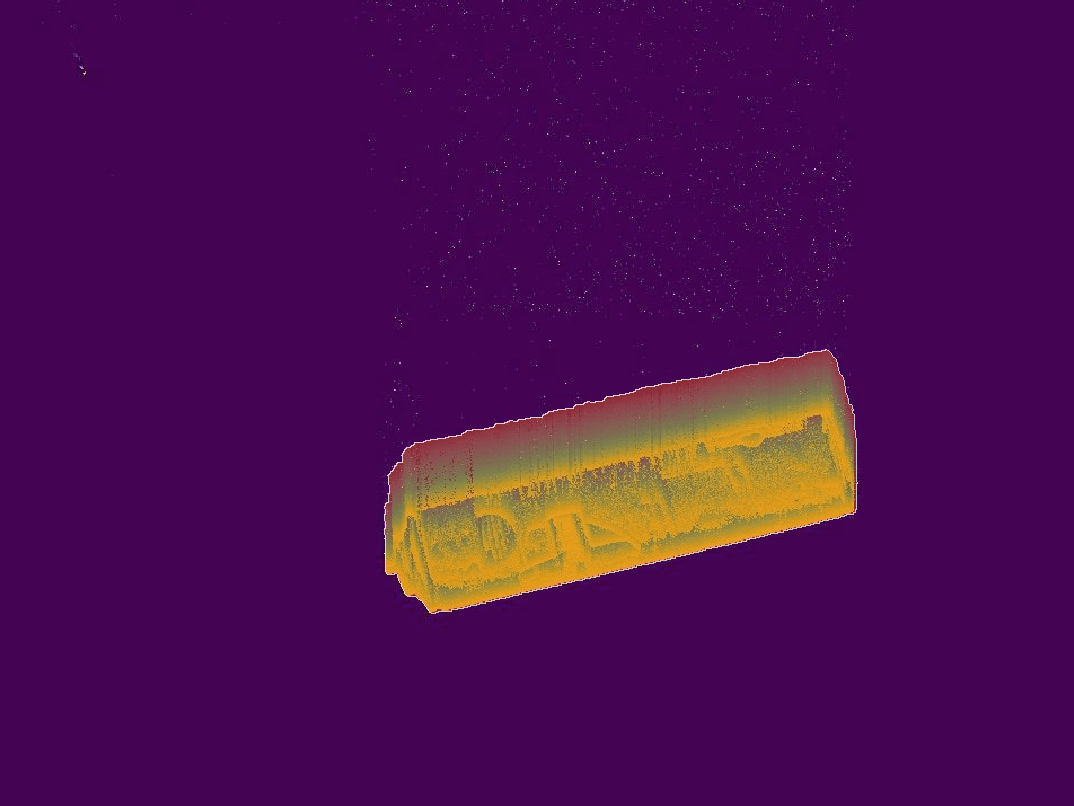 } }
    \subcaptionbox*{}%
    [.09\textwidth]{\includegraphics[width=\linewidth]{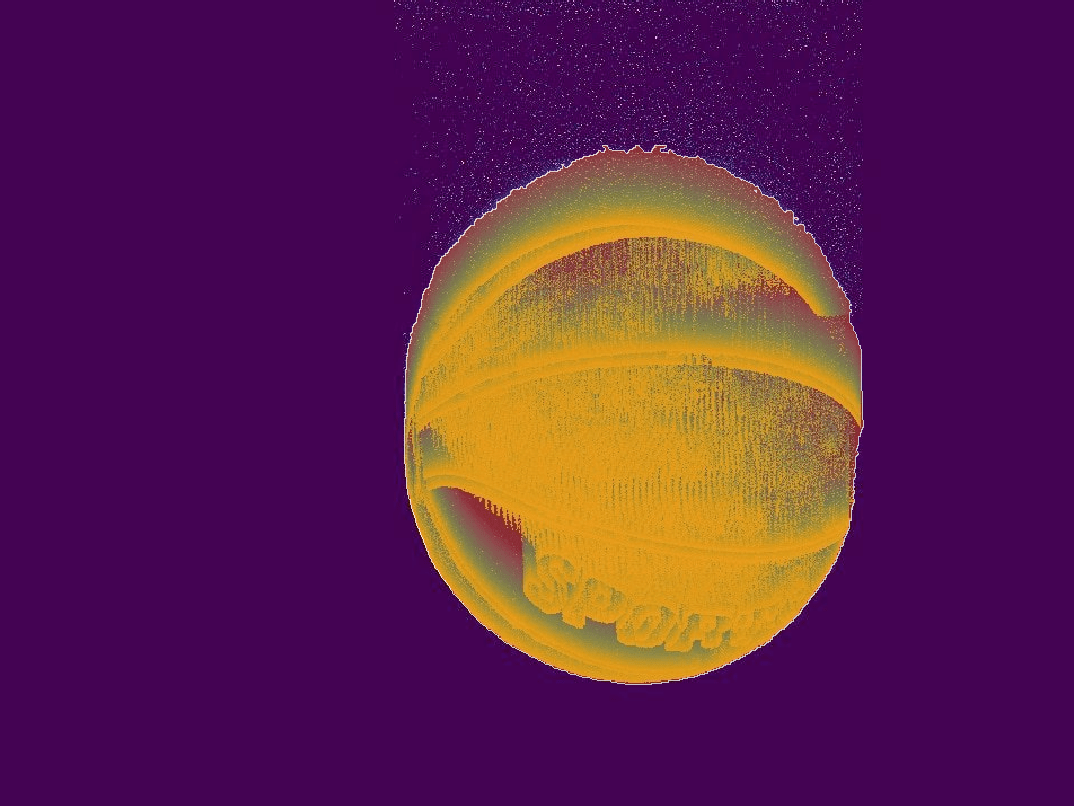 } }
    \subcaptionbox*{}%
    [.09\textwidth]{\includegraphics[width=\linewidth]{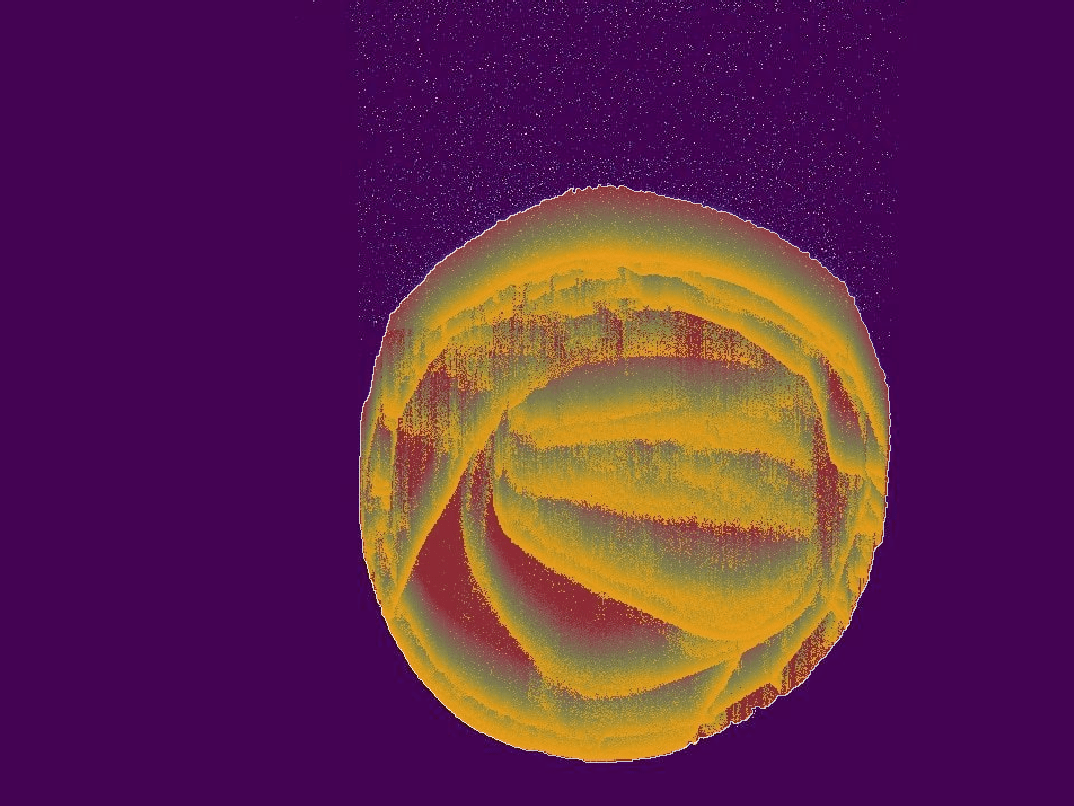 } }
    \subcaptionbox*{}%
    [.09\textwidth]{\includegraphics[width=\linewidth]{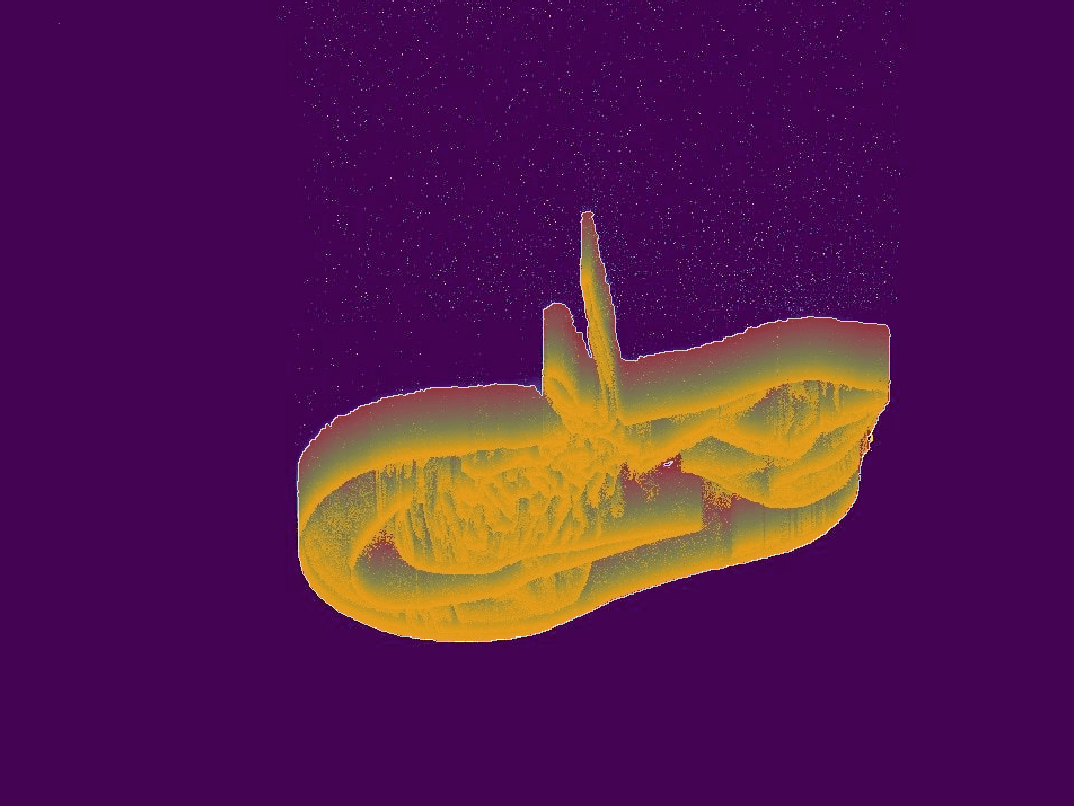 } }

    \vspace{-3mm}

    \subcaptionbox*{}%
    [.09\textwidth]{\includegraphics[width=\linewidth]{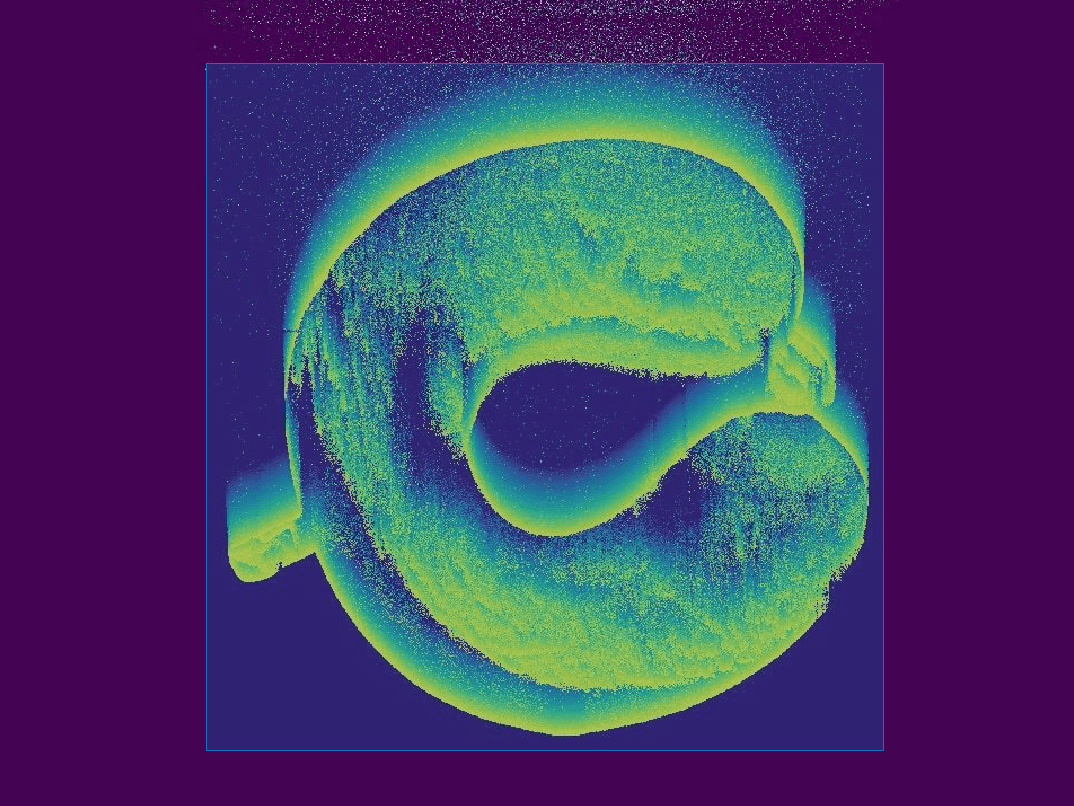 } }
    \subcaptionbox*{}%
    [.09\textwidth]{\includegraphics[width=\linewidth]{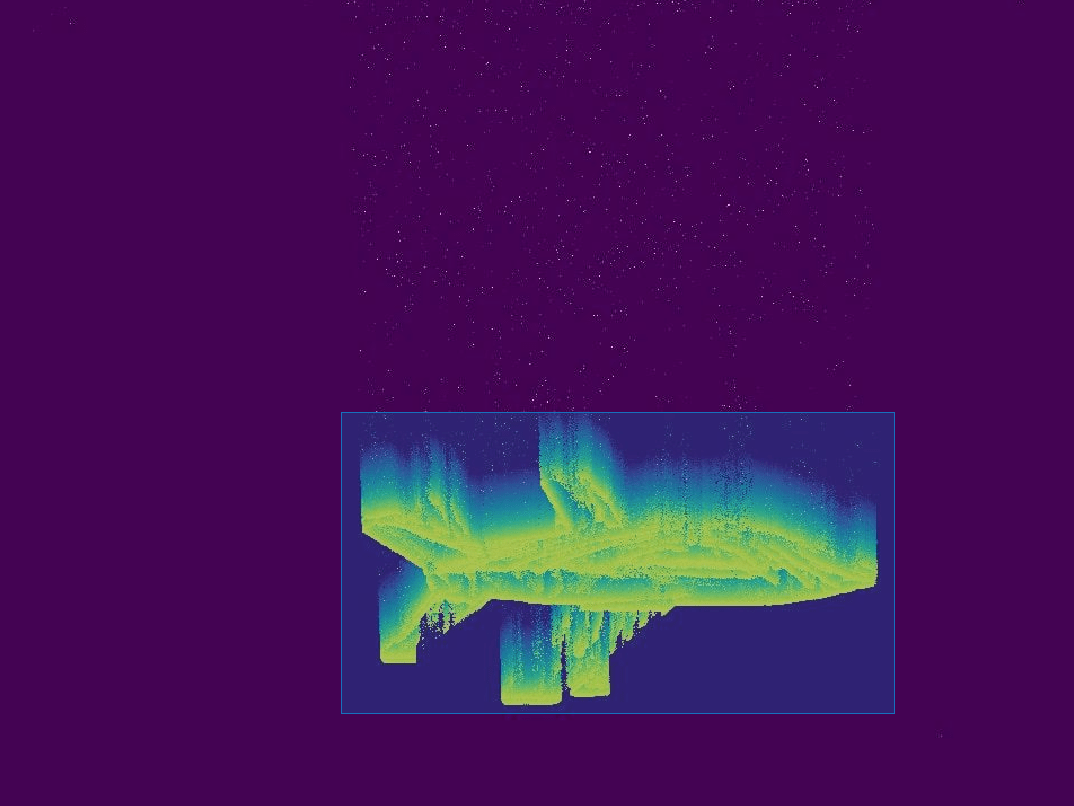 } }
    \subcaptionbox*{}%
    [.09\textwidth]{\includegraphics[width=\linewidth]{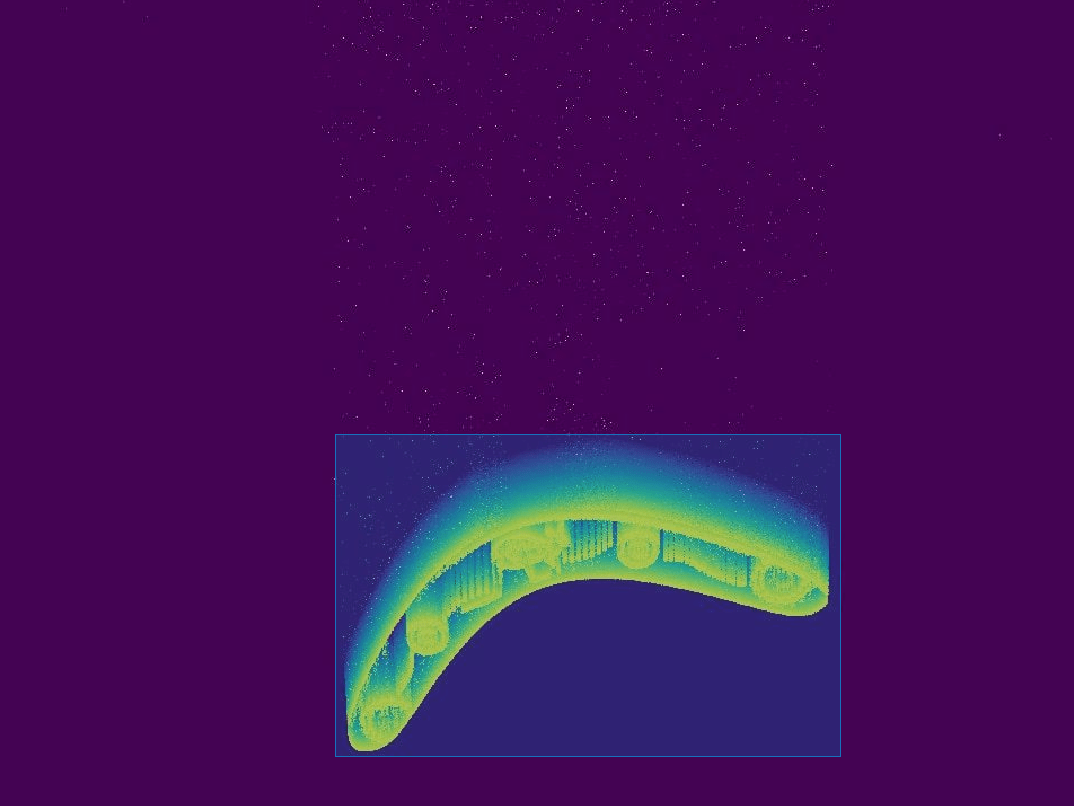 } }
    \subcaptionbox*{}%
    [.09\textwidth]{\includegraphics[width=\linewidth]{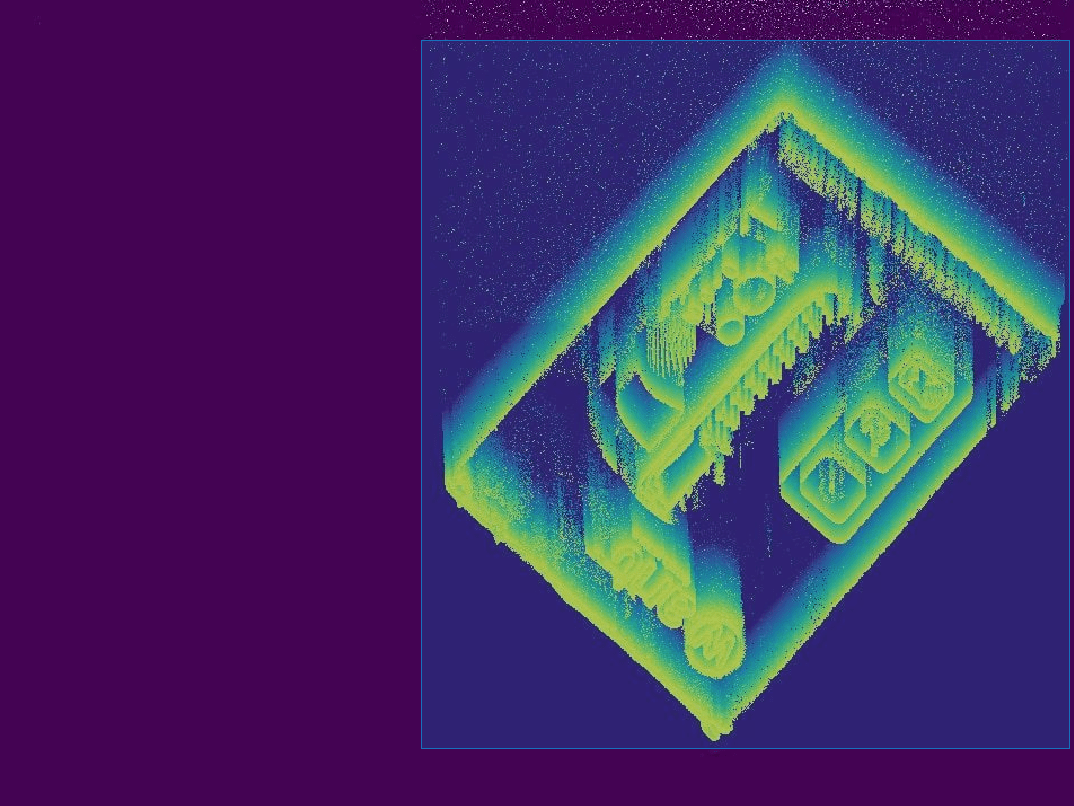 } }
    \subcaptionbox*{}%
    [.09\textwidth]{\includegraphics[width=\linewidth]{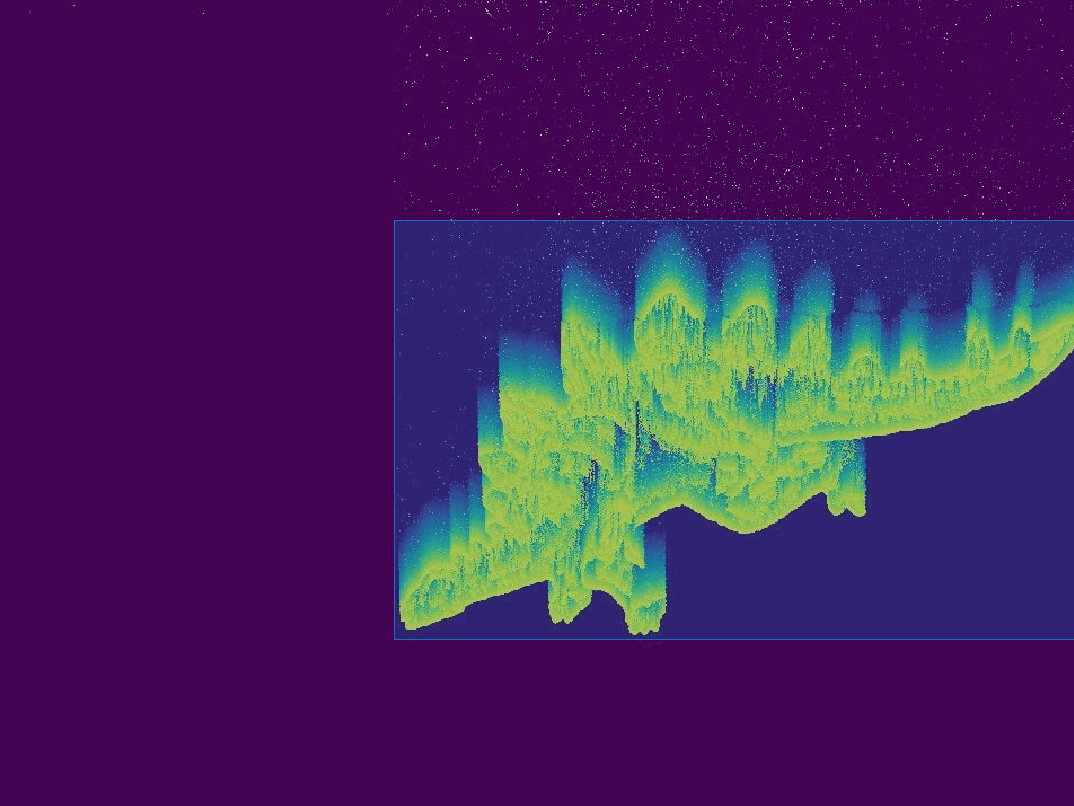 } }
    \subcaptionbox*{}%
    [.09\textwidth]{\includegraphics[width=\linewidth]{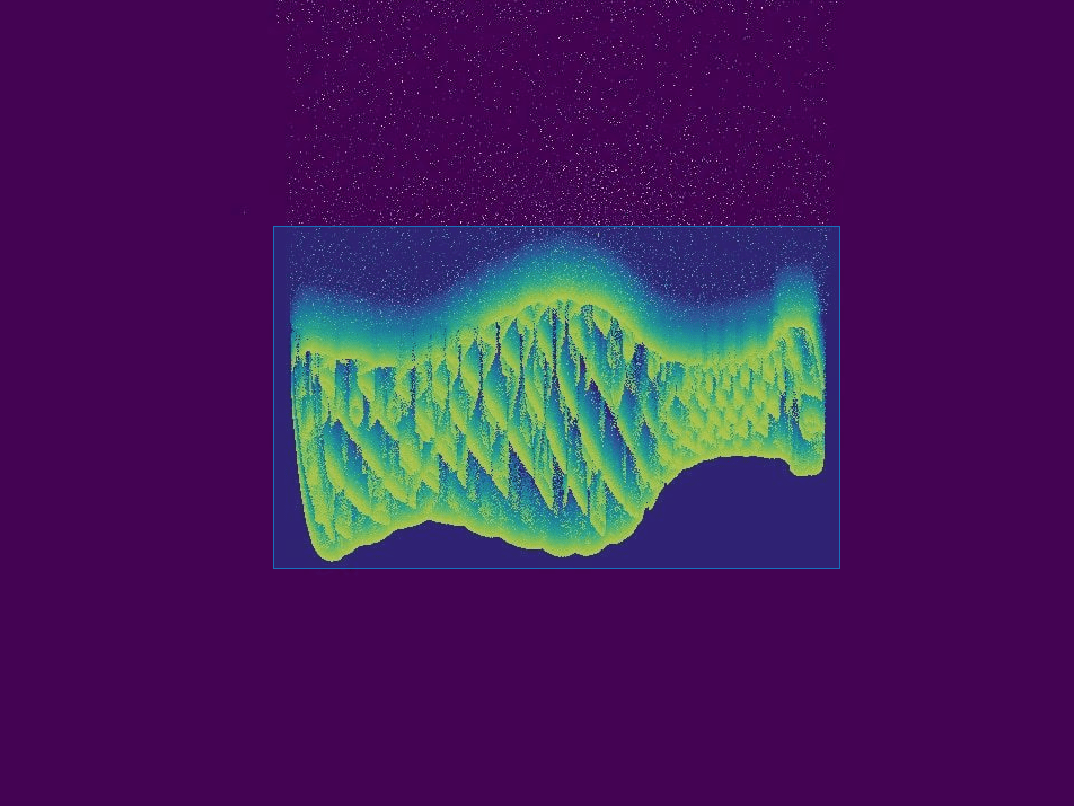 } }
    \subcaptionbox*{}%
    [.09\textwidth]{\includegraphics[width=\linewidth]{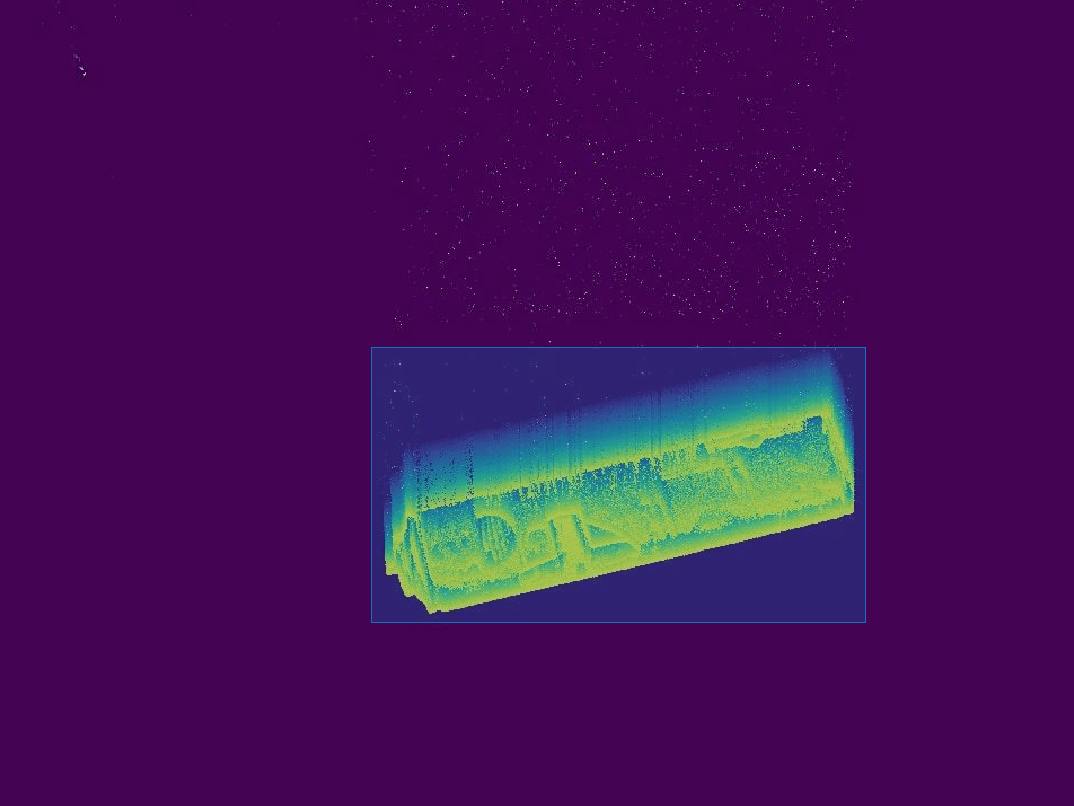 } }
    \subcaptionbox*{}%
    [.09\textwidth]{\includegraphics[width=\linewidth]{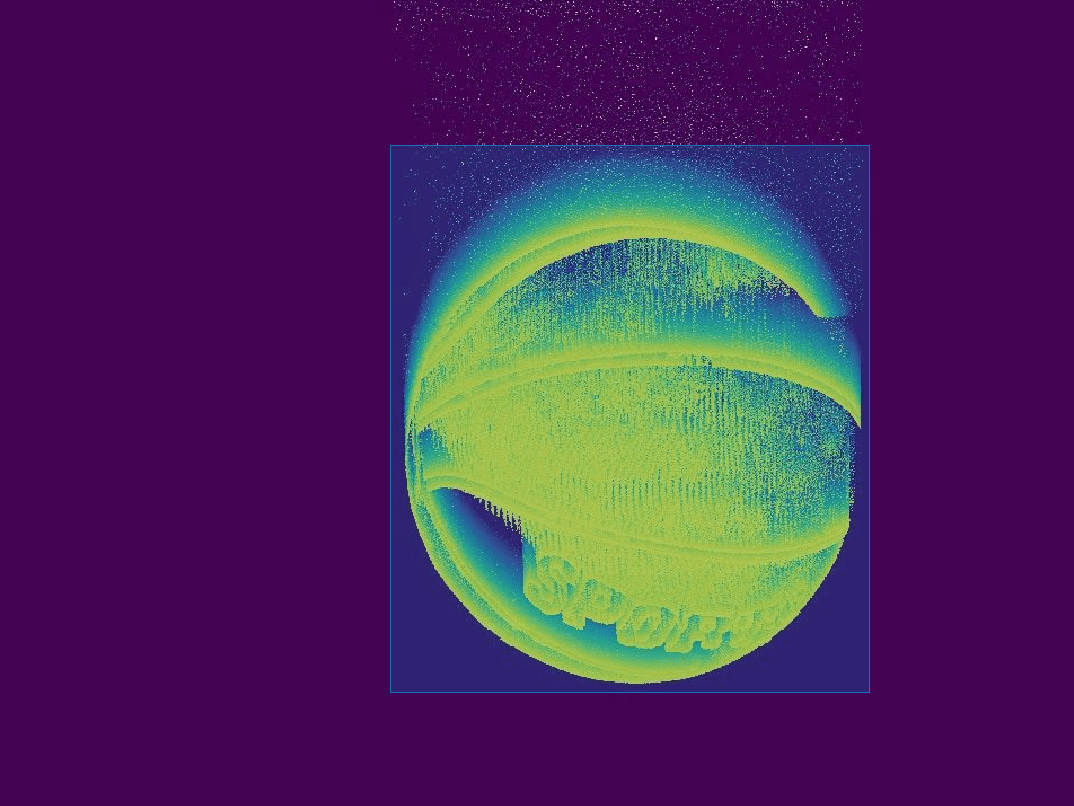 } }
    \subcaptionbox*{}%
    [.09\textwidth]{\includegraphics[width=\linewidth]{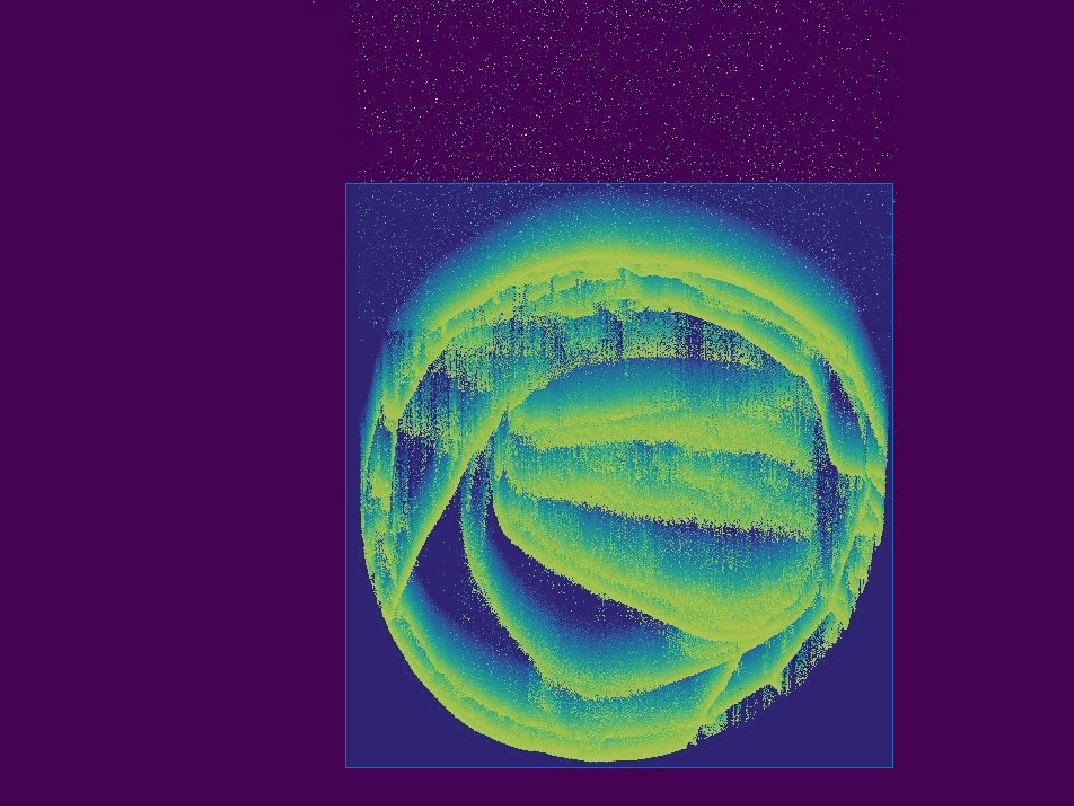 } }
    \subcaptionbox*{}%
    [.09\textwidth]{\includegraphics[width=\linewidth]{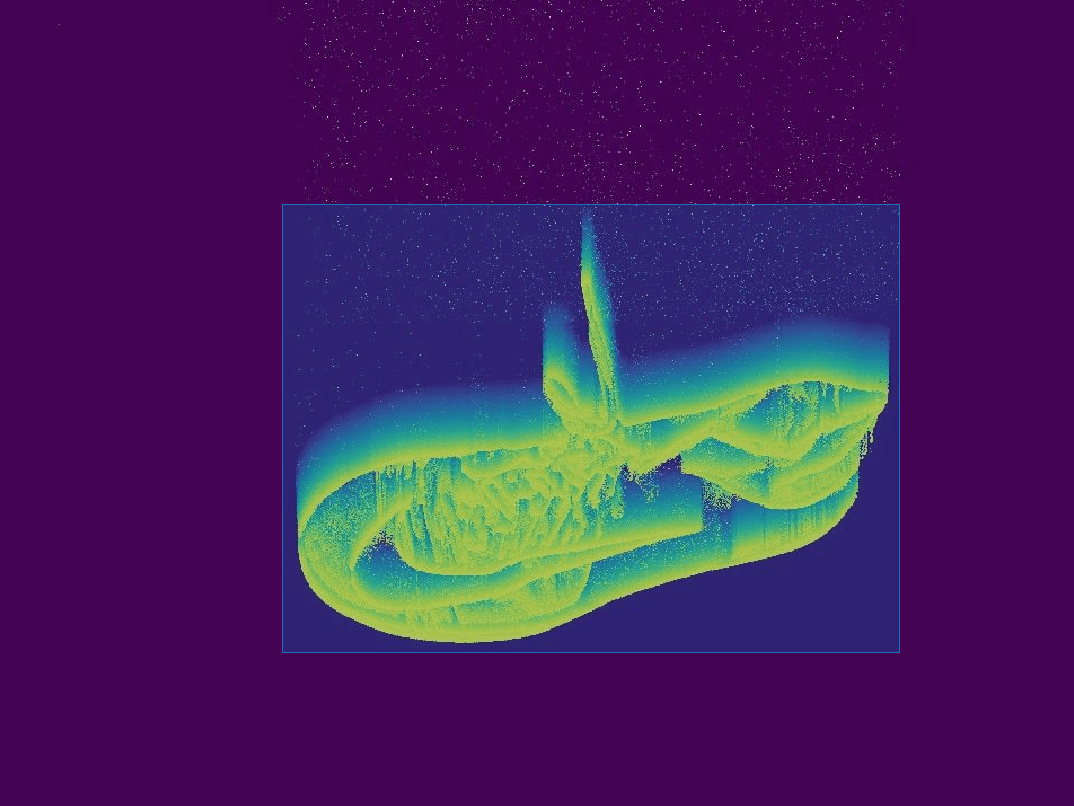 } }

    \vspace{-3mm}

    \subcaptionbox*{}%
    [.09\textwidth]{\includegraphics[width=\linewidth]{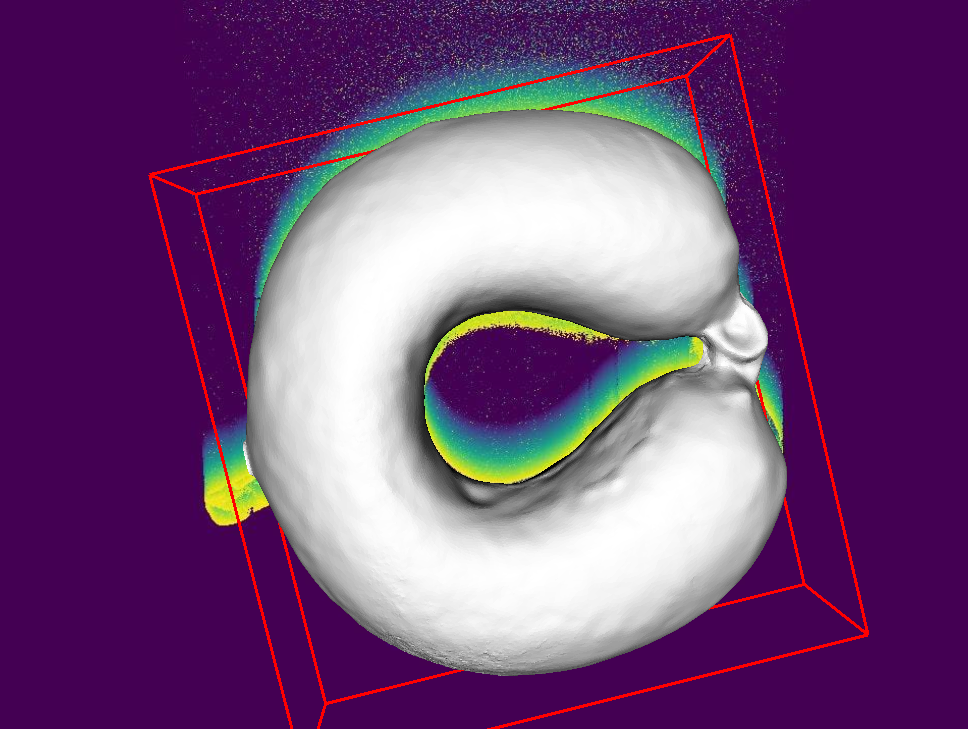 } }
    \subcaptionbox*{}%
    [.09\textwidth]{\includegraphics[width=\linewidth]{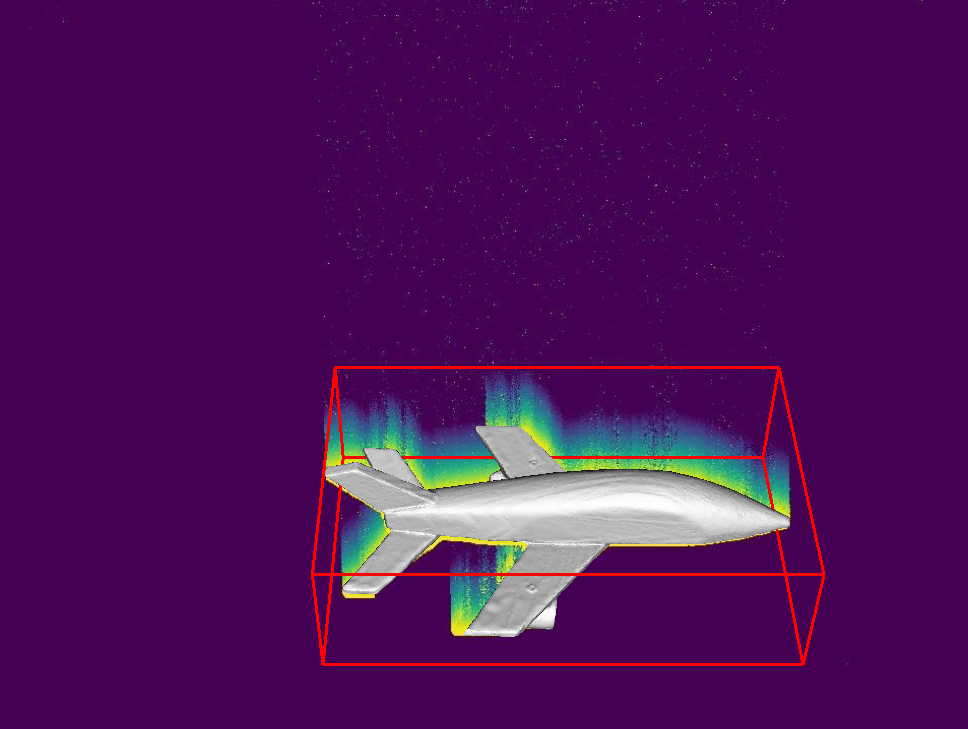 } }
    \subcaptionbox*{}%
    [.09\textwidth]{\includegraphics[width=\linewidth]{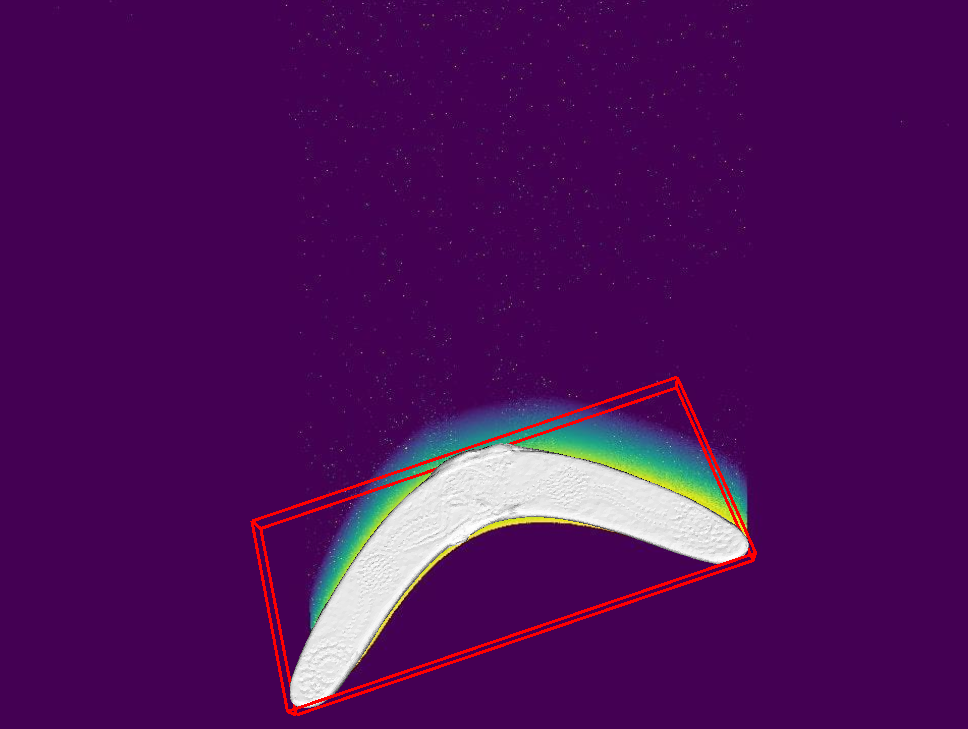 } }
    \subcaptionbox*{}%
    [.09\textwidth]{\includegraphics[width=\linewidth]{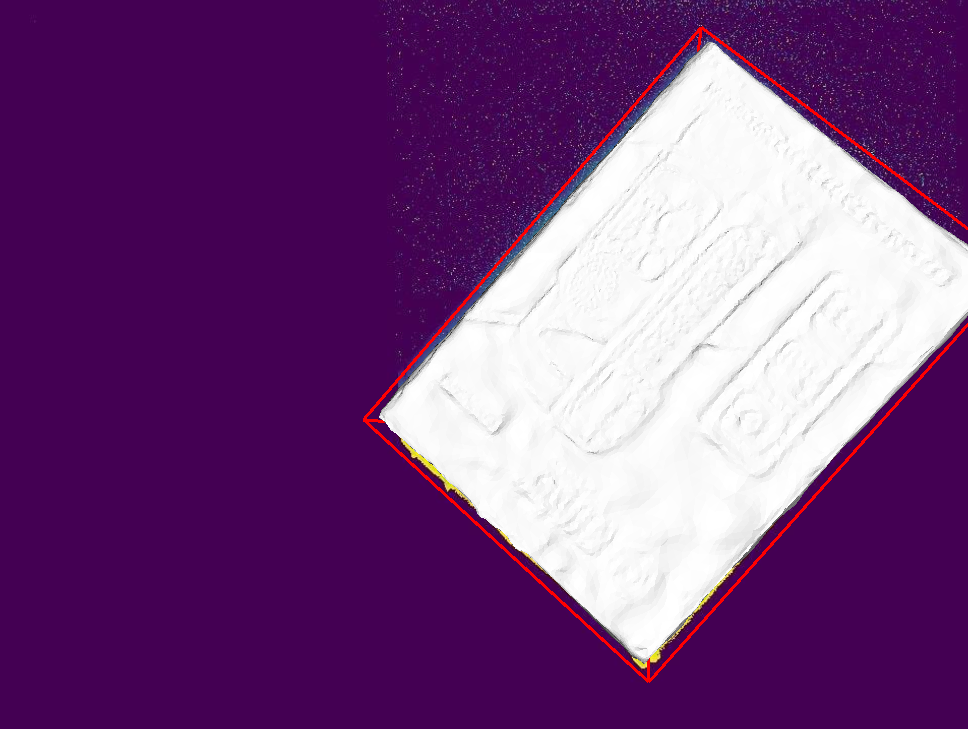 } }
    \subcaptionbox*{}%
    [.09\textwidth]{\includegraphics[width=\linewidth]{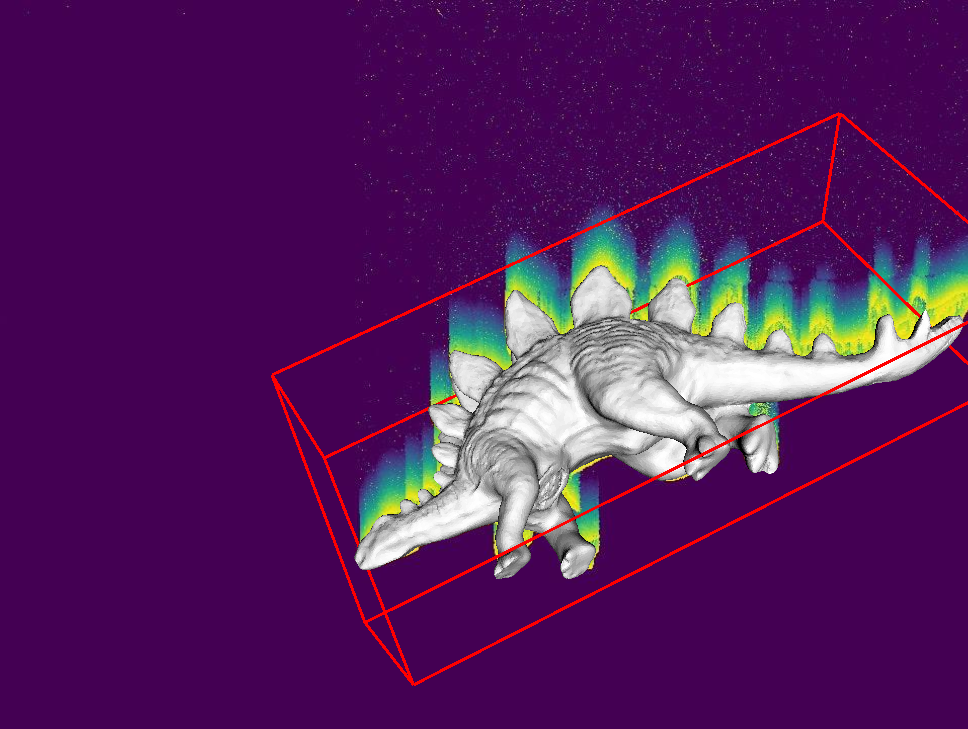 } }
    \subcaptionbox*{}%
    [.09\textwidth]{\includegraphics[width=\linewidth]{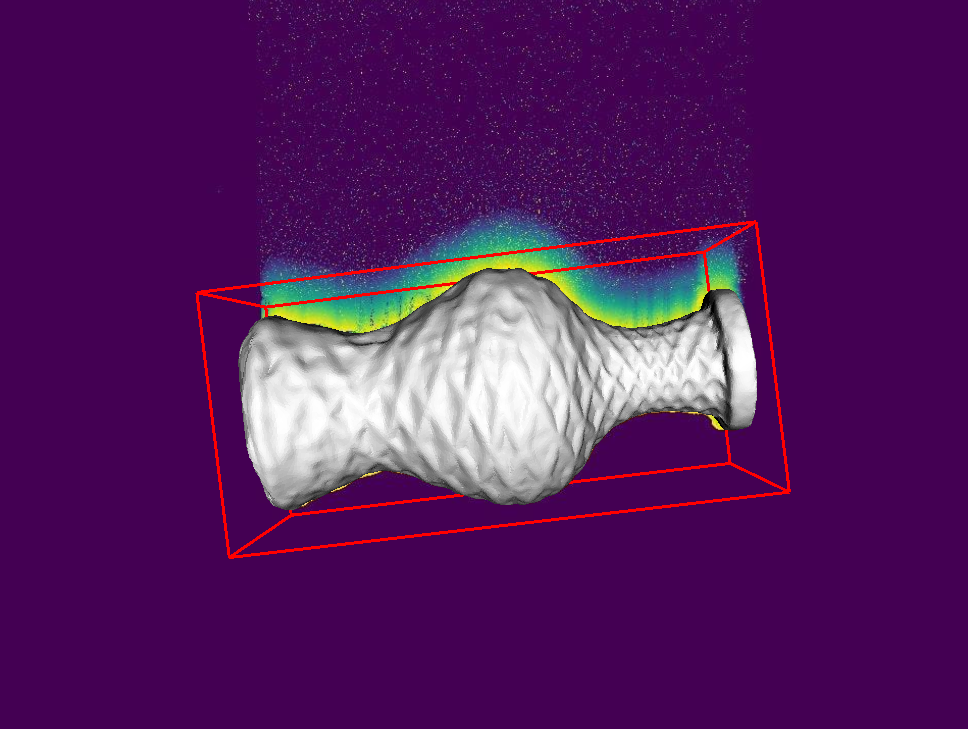 } }
    \subcaptionbox*{}%
    [.09\textwidth]{\includegraphics[width=\linewidth]{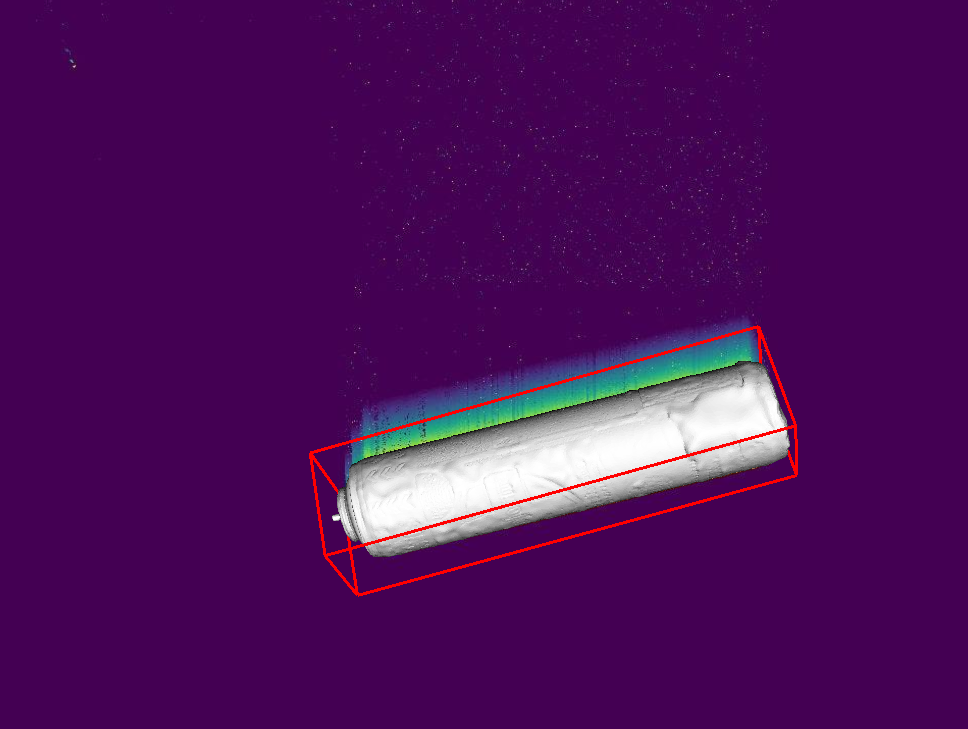 } }
    \subcaptionbox*{}%
    [.09\textwidth]{\includegraphics[width=\linewidth]{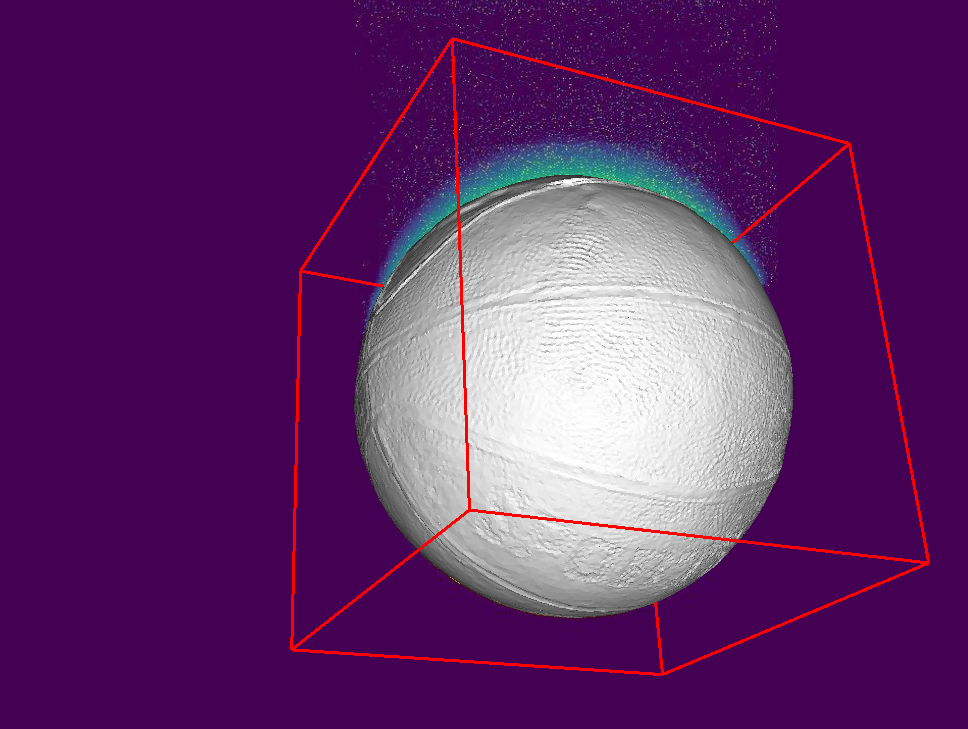 } }
    \subcaptionbox*{}%
    [.09\textwidth]{\includegraphics[width=\linewidth]{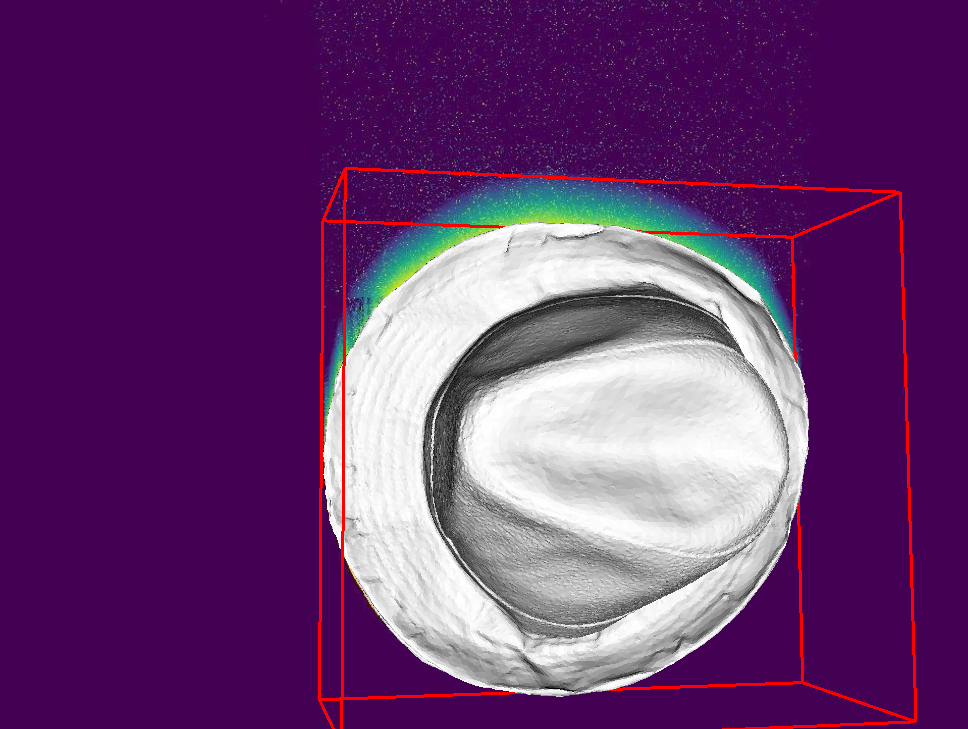 } }
    \subcaptionbox*{}%
    [.09\textwidth]{\includegraphics[width=\linewidth]{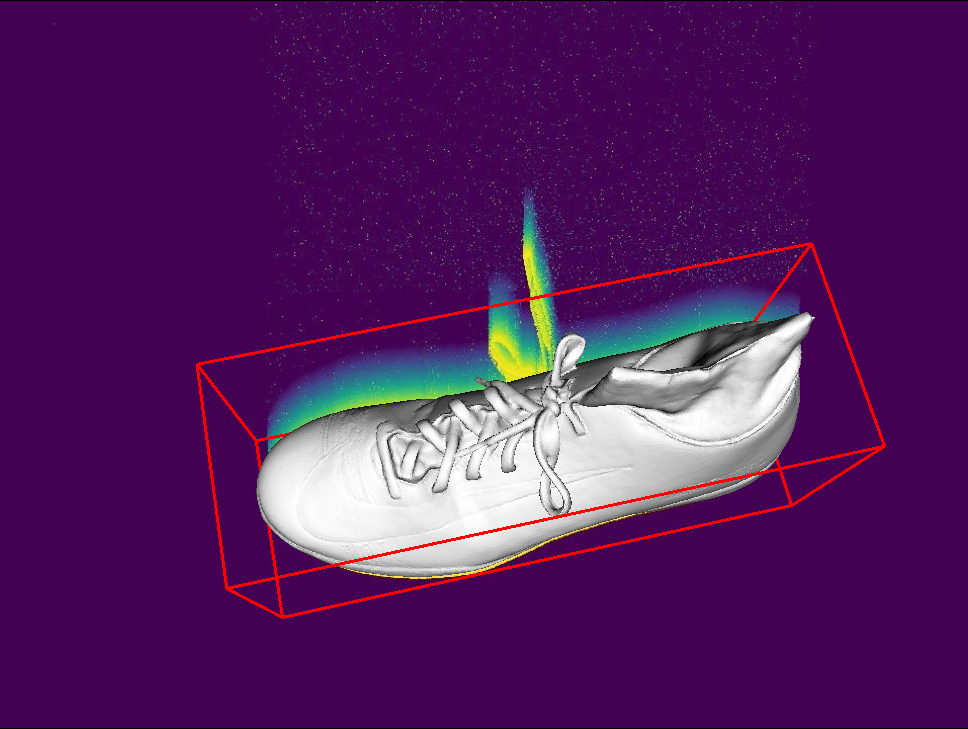 } }

    \caption{Moving6DPoSe-R: real-world moving object sequences captured with the ROG Eye S and Prophesee EVK4. The top three rows show frame-based segmentation, detection, and 6D pose annotations, while the bottom three rows show the event annotations.}    \label{fig:moving6dpose-r}
\end{figure}

For 6D pose annotation, object orientation and initial position were estimated from ZED-2 images using a Perspective-n-Point (PnP) formulation together with the corresponding CAD model. Depth measurements from the ZED-2 were subsequently used to refine object distance estimates and improve pose accuracy. When automatic annotations were insufficient, LabelImg3D \cite{labelimg3d} was used to manually adjust object position and orientation by aligning the CAD model with the image observations. Finally, the resulting 6D poses were transformed into the coordinate systems of the ROG Eye S and event cameras through known intersensor transformations, ensuring consistent annotations across all sensing modalities. Figure~\ref{fig:moving6dpose_annotation} illustrates the pose annotation process, while Figure~\ref{fig:moving6dpose-r} presents representative annotated samples.

\subsection{Moving6DPoSe-S dataset annotation}

\begin{figure*}[!tb]
\centering
    \subcaptionbox*{}%
    [.09\textwidth]{\includegraphics[width=\linewidth]{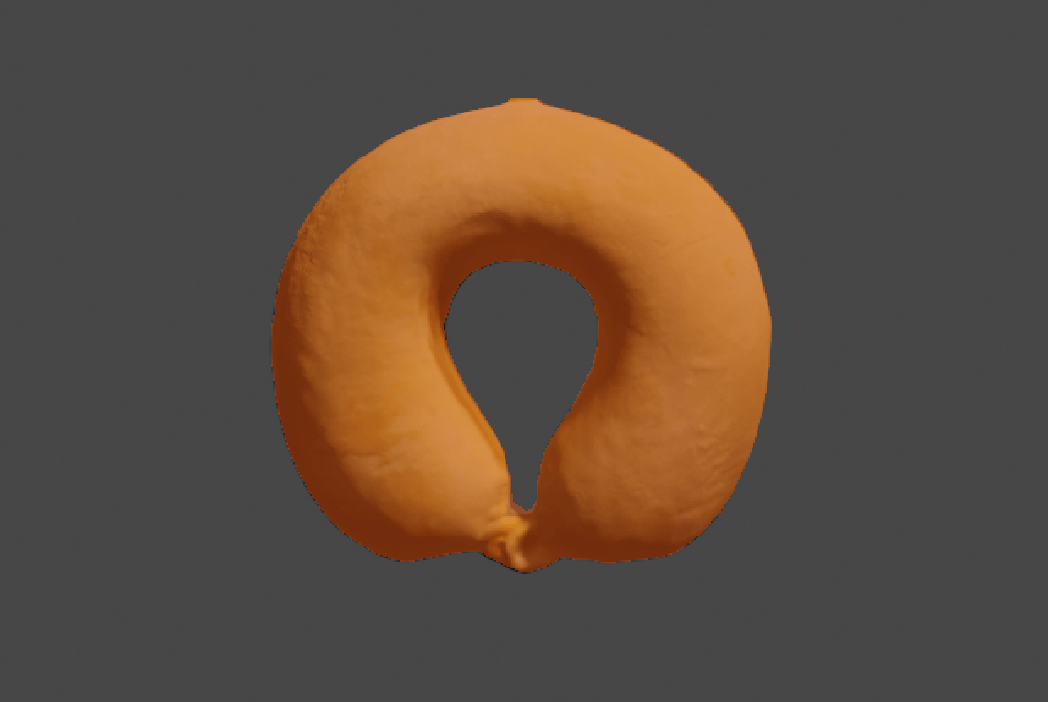 } }
    \subcaptionbox*{}%
    [.09\textwidth]{\includegraphics[width=\linewidth]{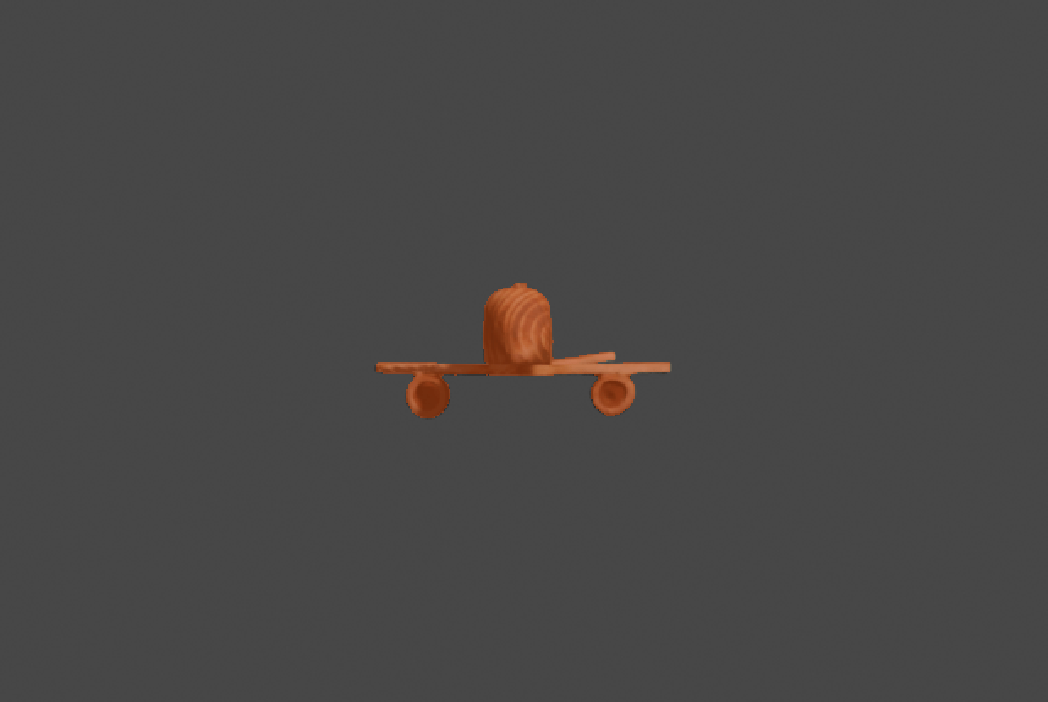 } }
    \subcaptionbox*{}%
    [.09\textwidth]{\includegraphics[width=\linewidth]{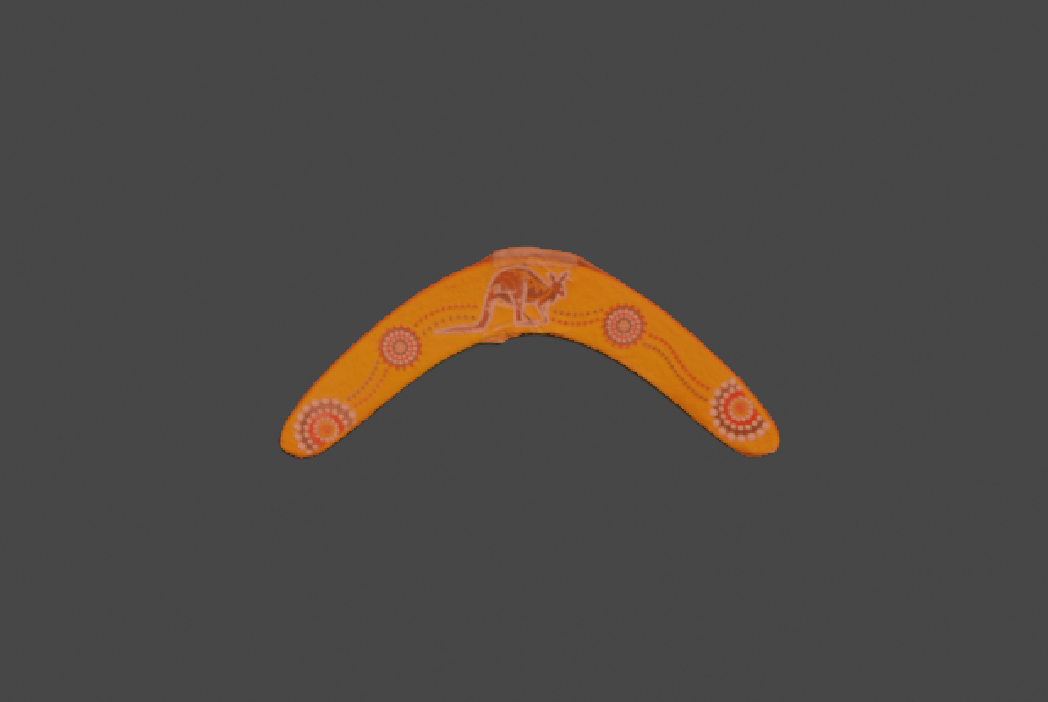 } }
    \subcaptionbox*{}%
    [.09\textwidth]{\includegraphics[width=\linewidth]{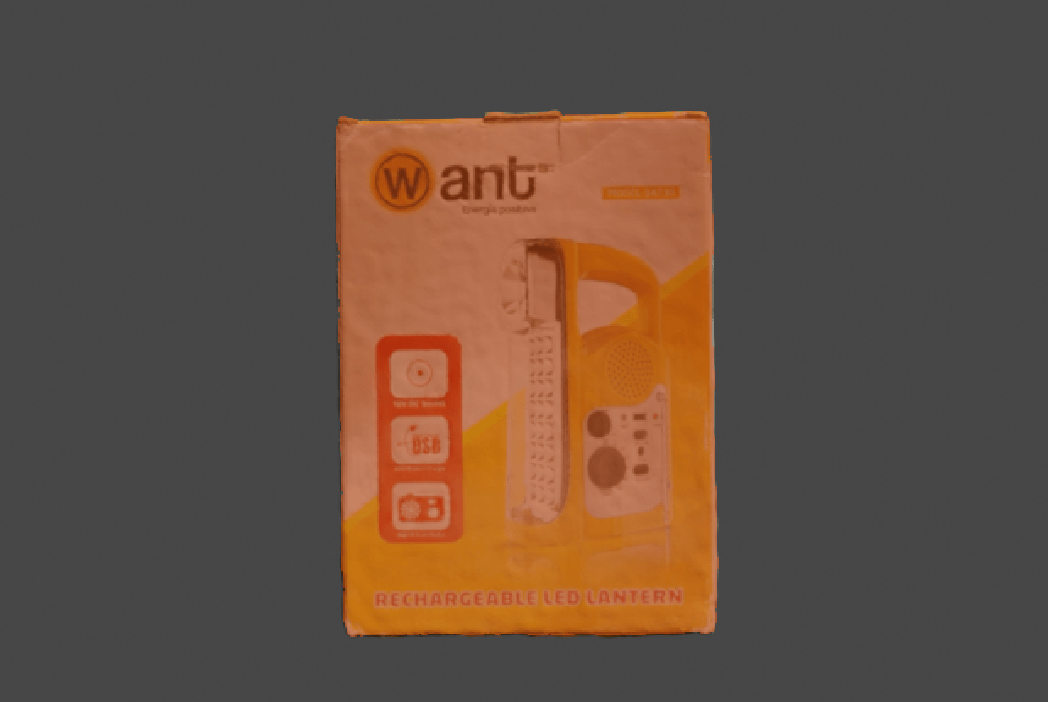 } }
    \subcaptionbox*{}%
    [.09\textwidth]{\includegraphics[width=\linewidth]{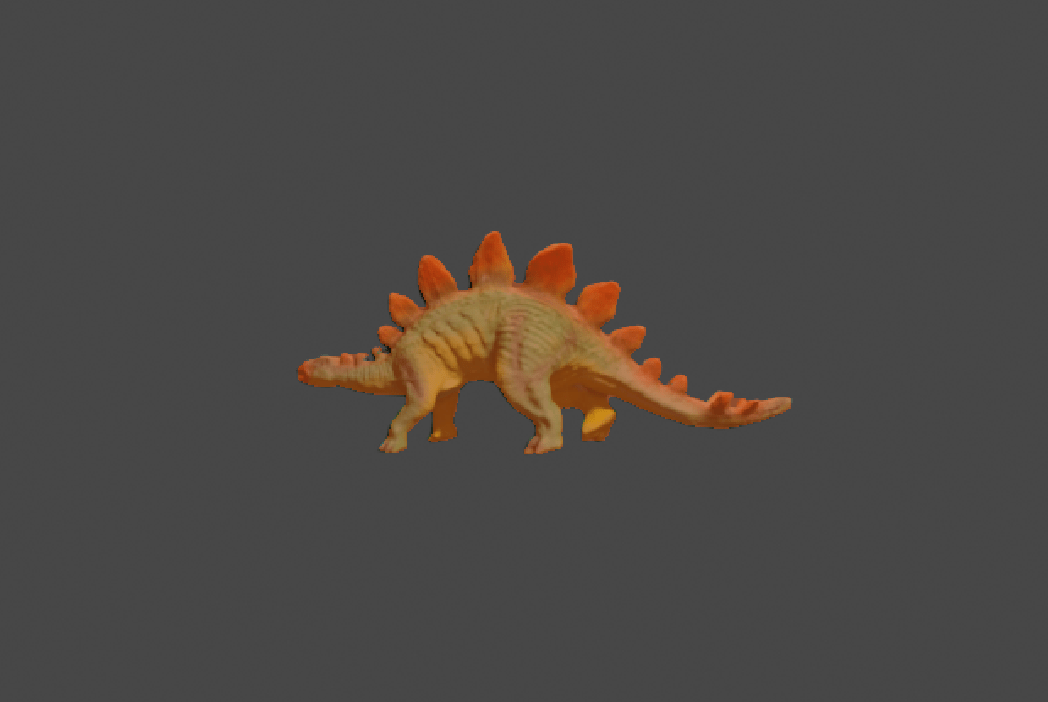 } }
    \subcaptionbox*{}%
    [.09\textwidth]{\includegraphics[width=\linewidth]{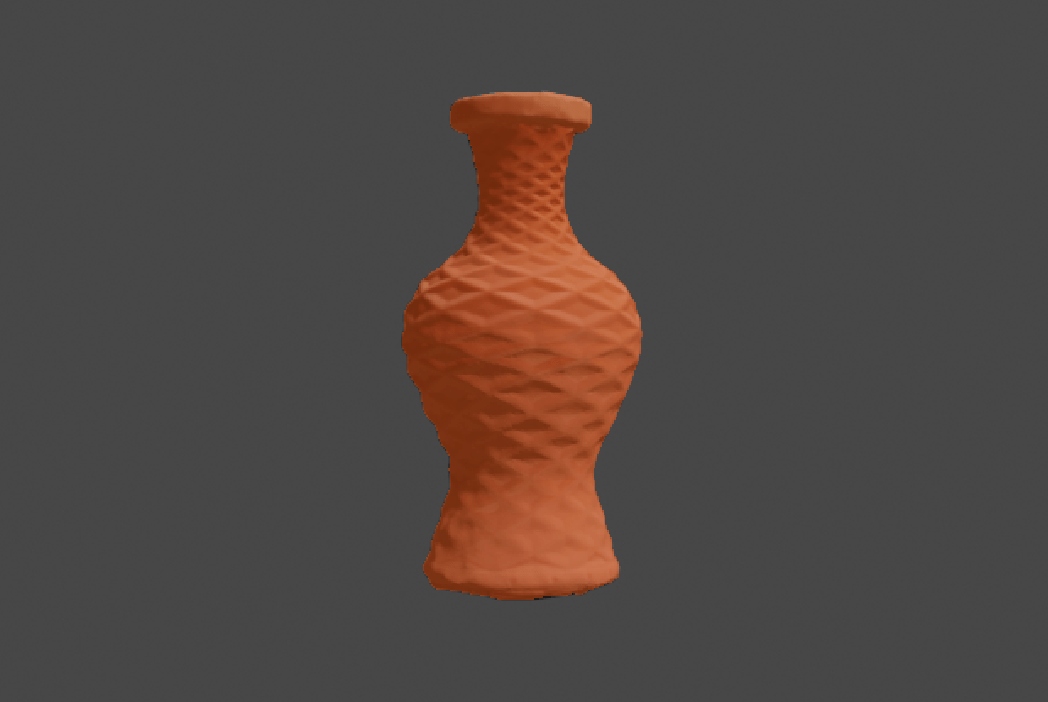 } }
    \subcaptionbox*{}%
    [.09\textwidth]{\includegraphics[width=\linewidth]{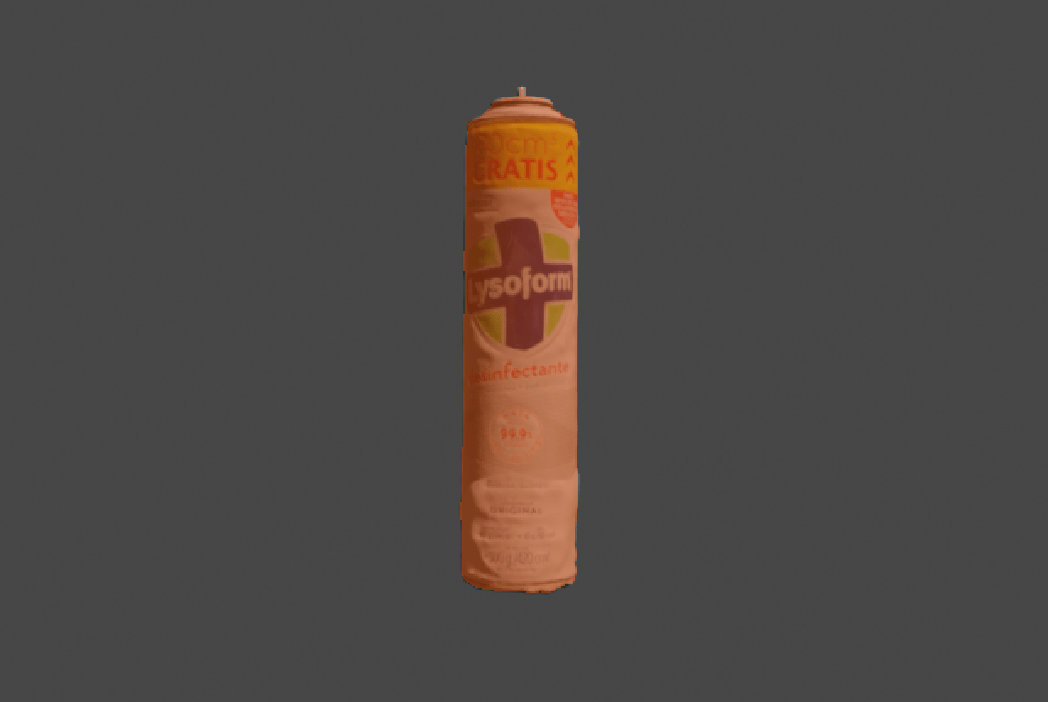 } }
    \subcaptionbox*{}%
    [.09\textwidth]{\includegraphics[width=\linewidth]{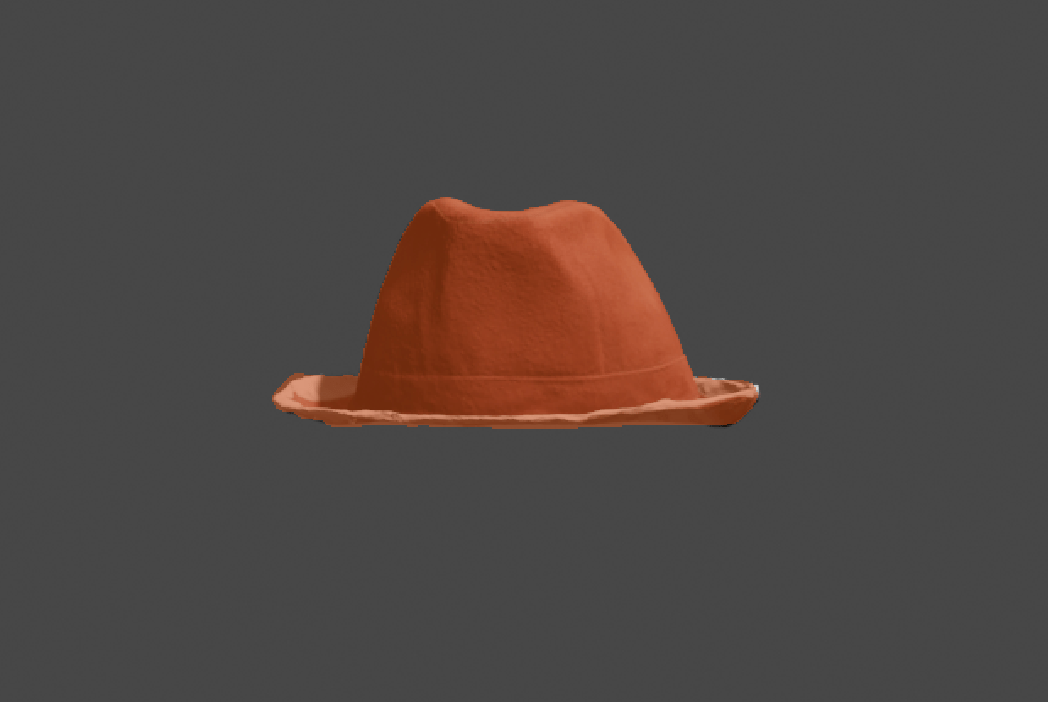 } }
    \subcaptionbox*{}%
    [.09\textwidth]{\includegraphics[width=\linewidth]{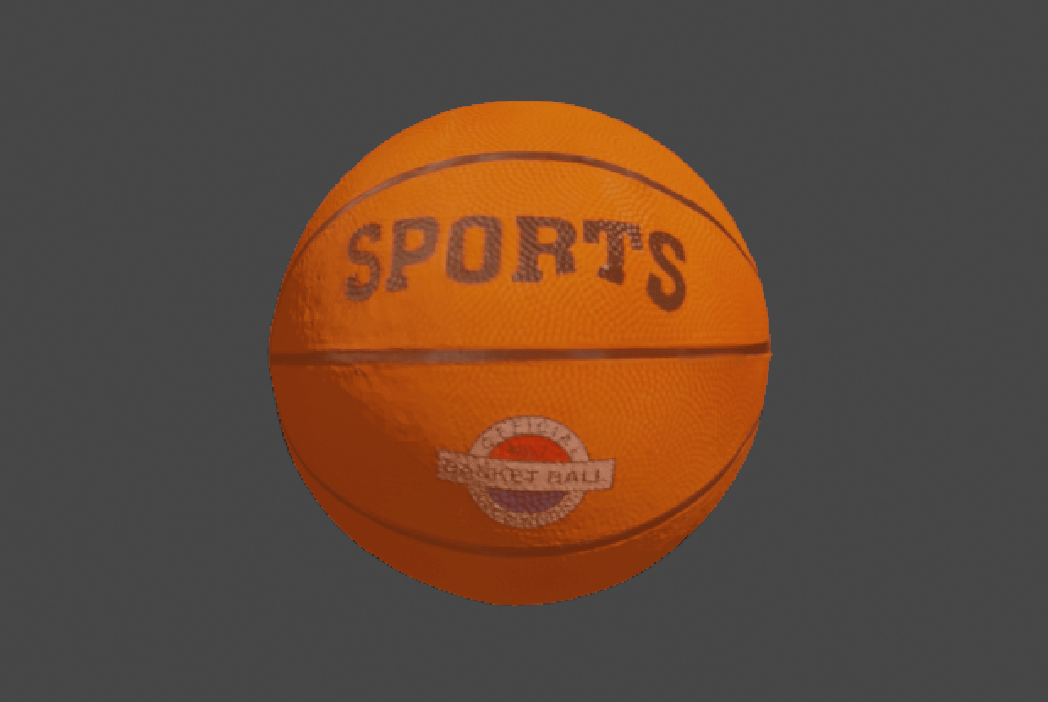 } }
    \subcaptionbox*{}%
    [.09\textwidth]{\includegraphics[width=\linewidth]{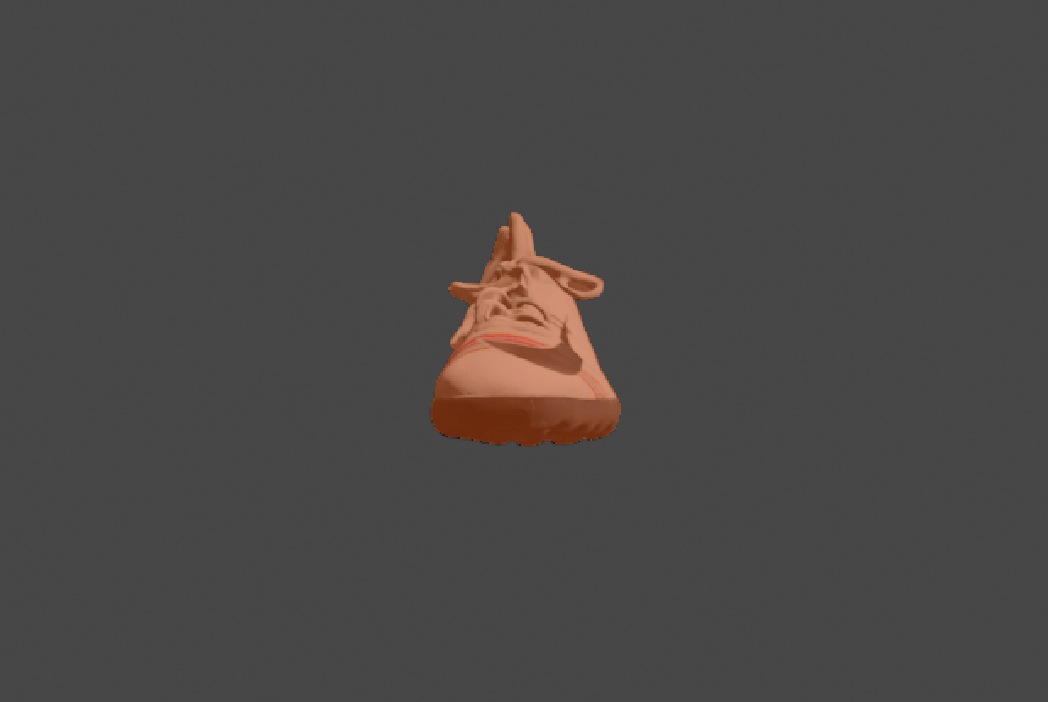 } }

    \vspace{-3mm}

    \subcaptionbox*{}%
    [.09\textwidth]{\includegraphics[width=\linewidth]{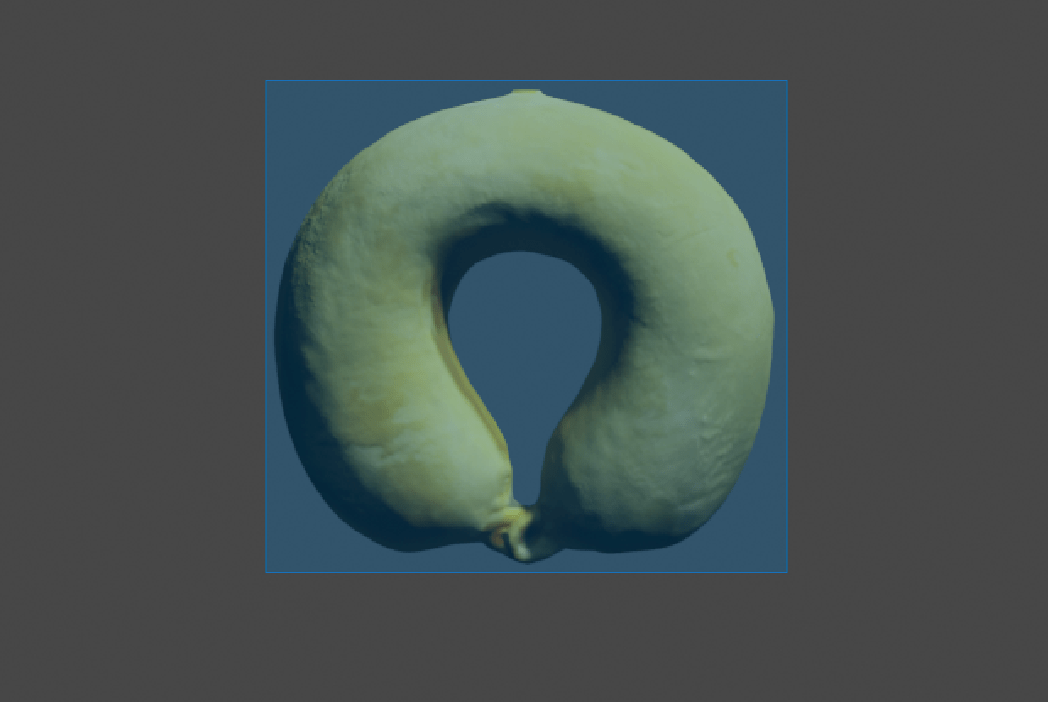 } }
    \subcaptionbox*{}%
    [.09\textwidth]{\includegraphics[width=\linewidth]{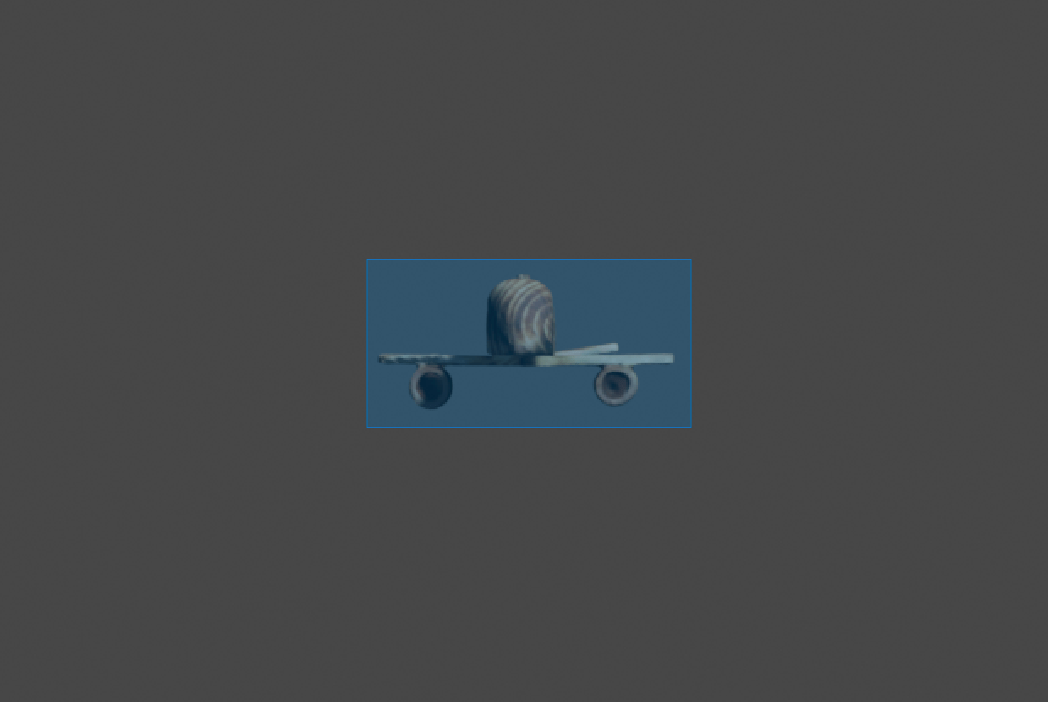 } }
    \subcaptionbox*{}%
    [.09\textwidth]{\includegraphics[width=\linewidth]{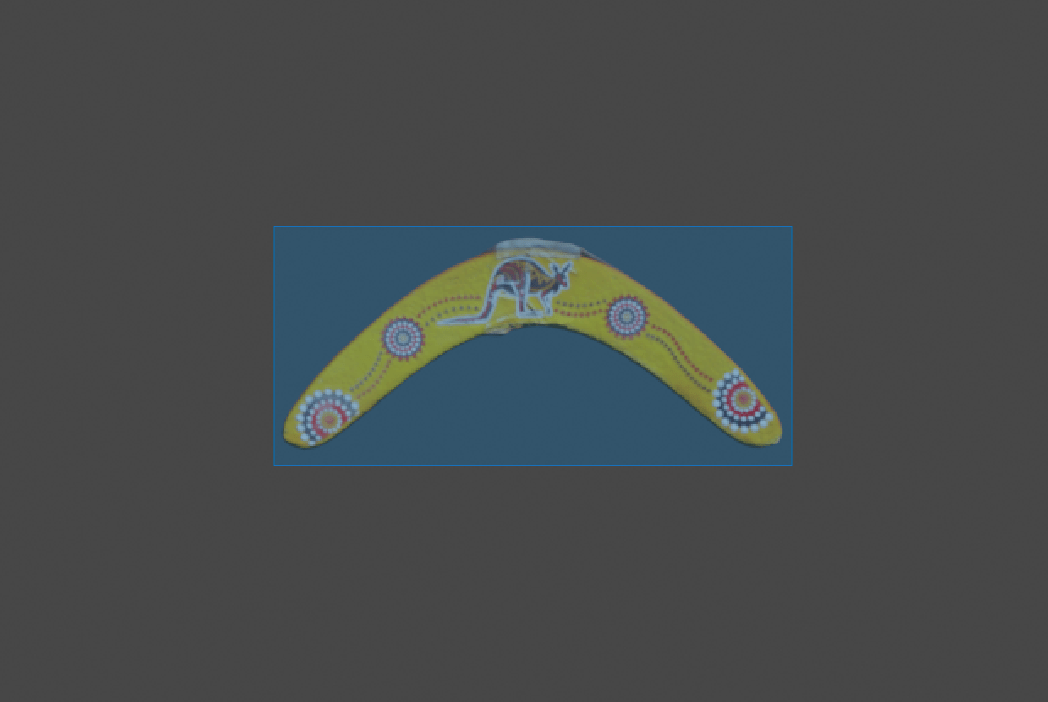 } }
    \subcaptionbox*{}%
    [.09\textwidth]{\includegraphics[width=\linewidth]{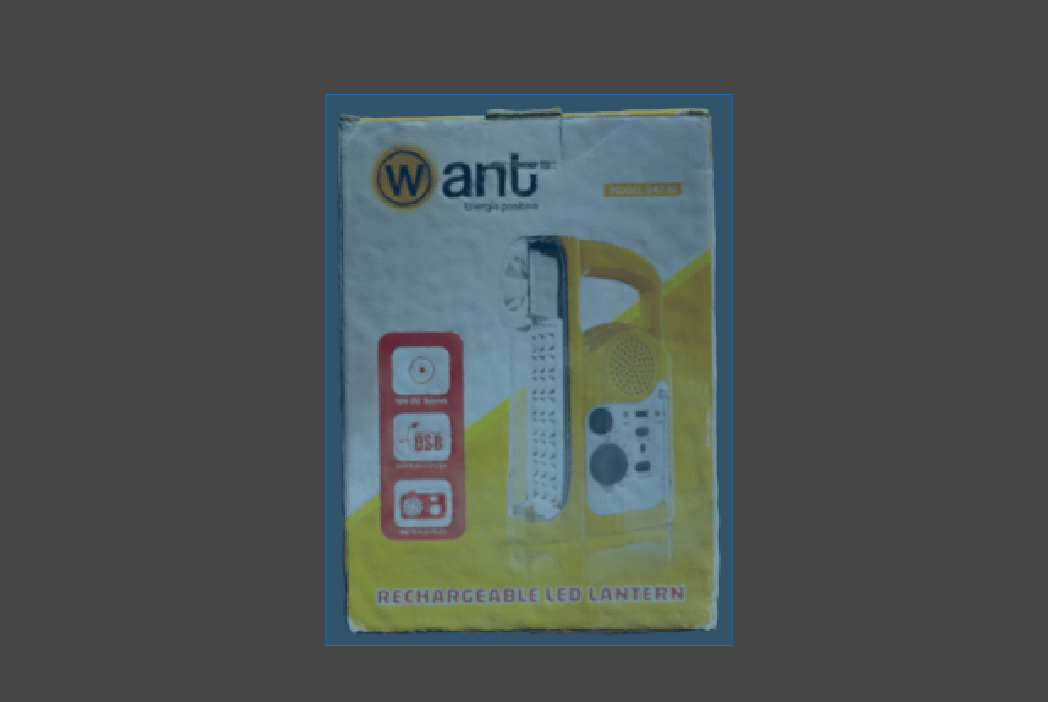 } }
    \subcaptionbox*{}%
    [.09\textwidth]{\includegraphics[width=\linewidth]{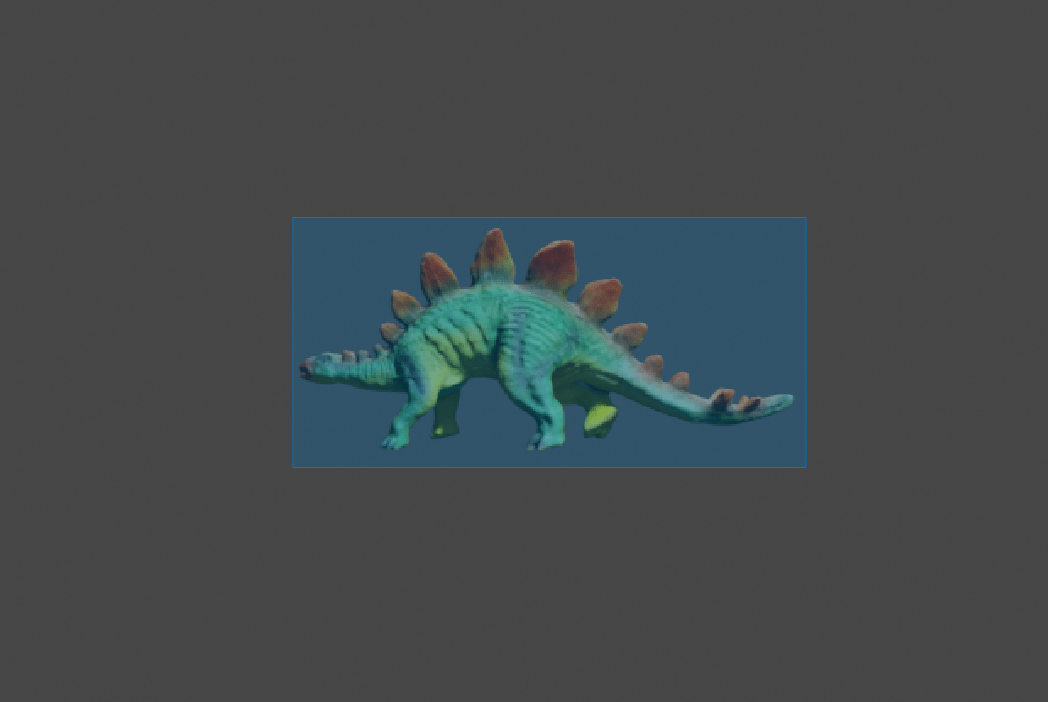 } }
    \subcaptionbox*{}%
    [.09\textwidth]{\includegraphics[width=\linewidth]{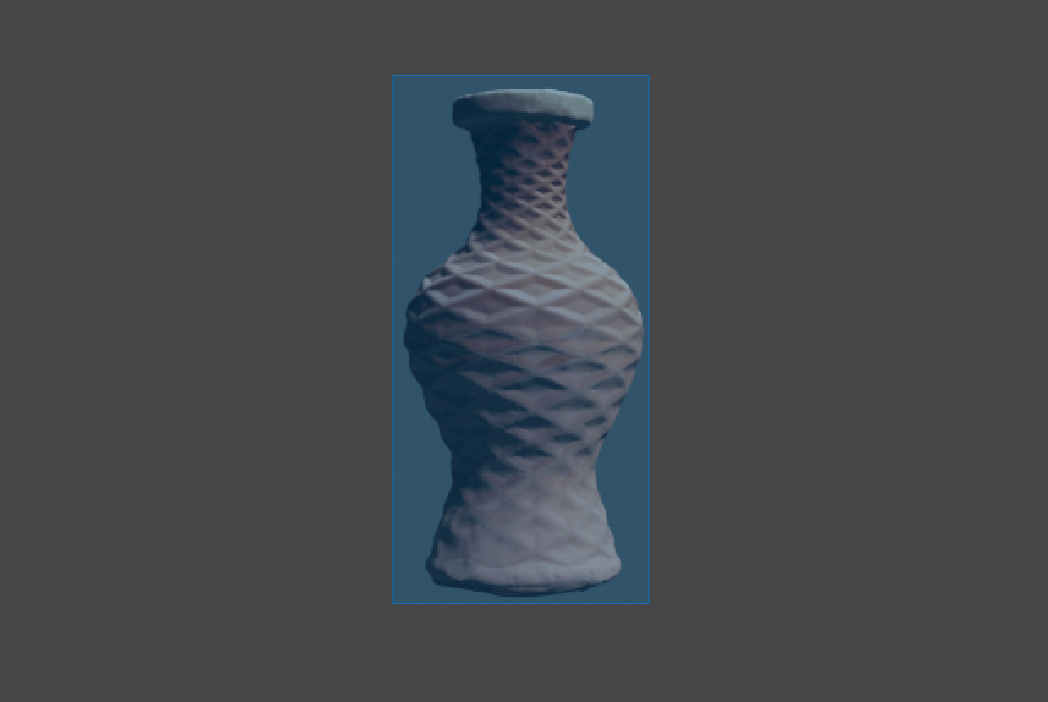 } }
    \subcaptionbox*{}%
    [.09\textwidth]{\includegraphics[width=\linewidth]{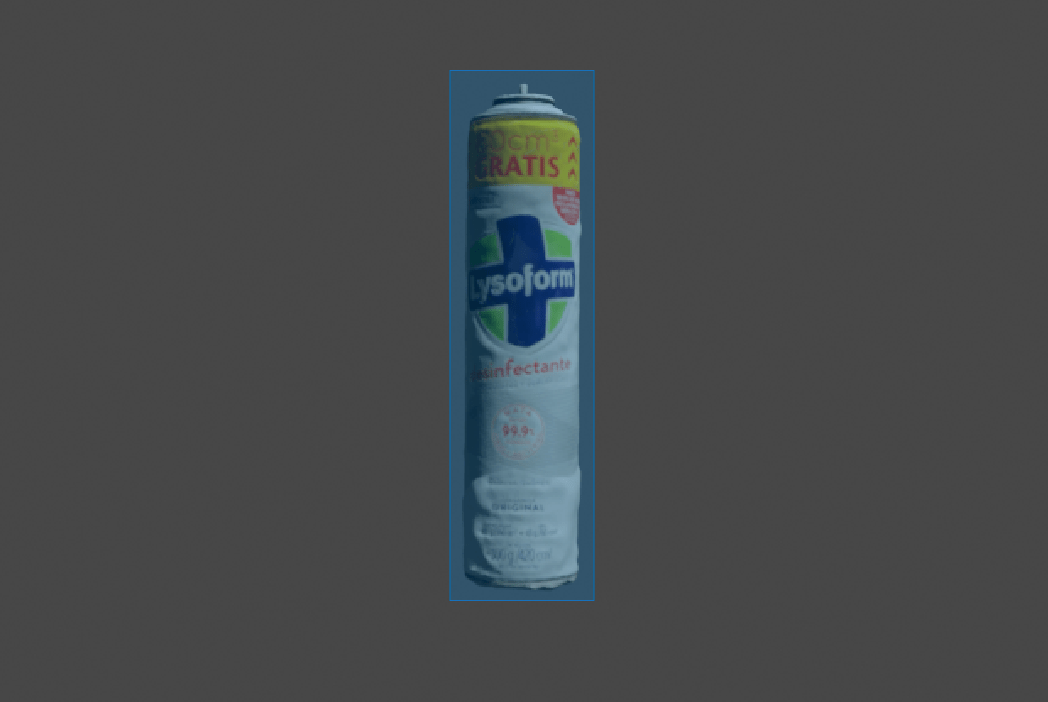 } }
    \subcaptionbox*{}%
    [.09\textwidth]{\includegraphics[width=\linewidth]{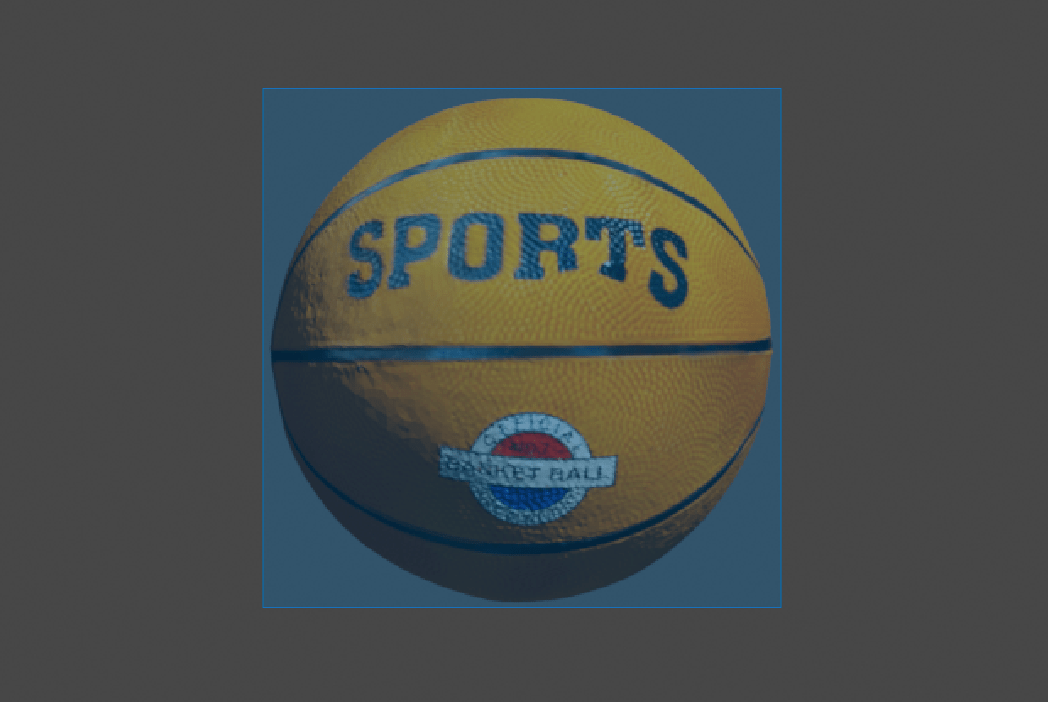 } }
    \subcaptionbox*{}%
    [.09\textwidth]{\includegraphics[width=\linewidth]{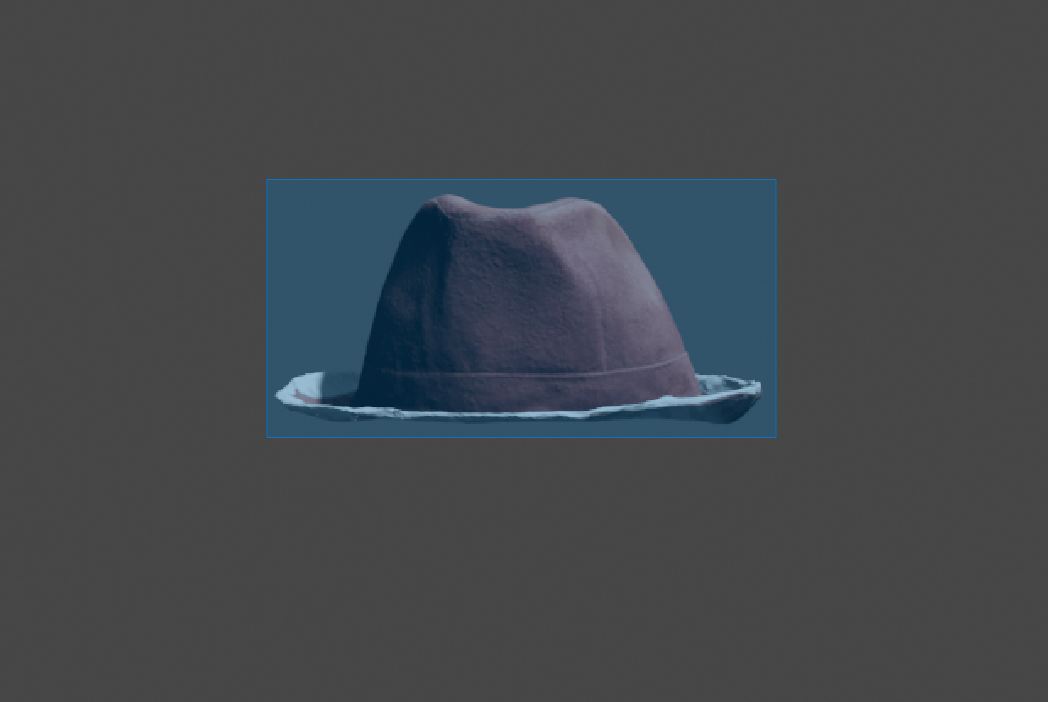 } }
    \subcaptionbox*{}%
    [.09\textwidth]{\includegraphics[width=\linewidth]{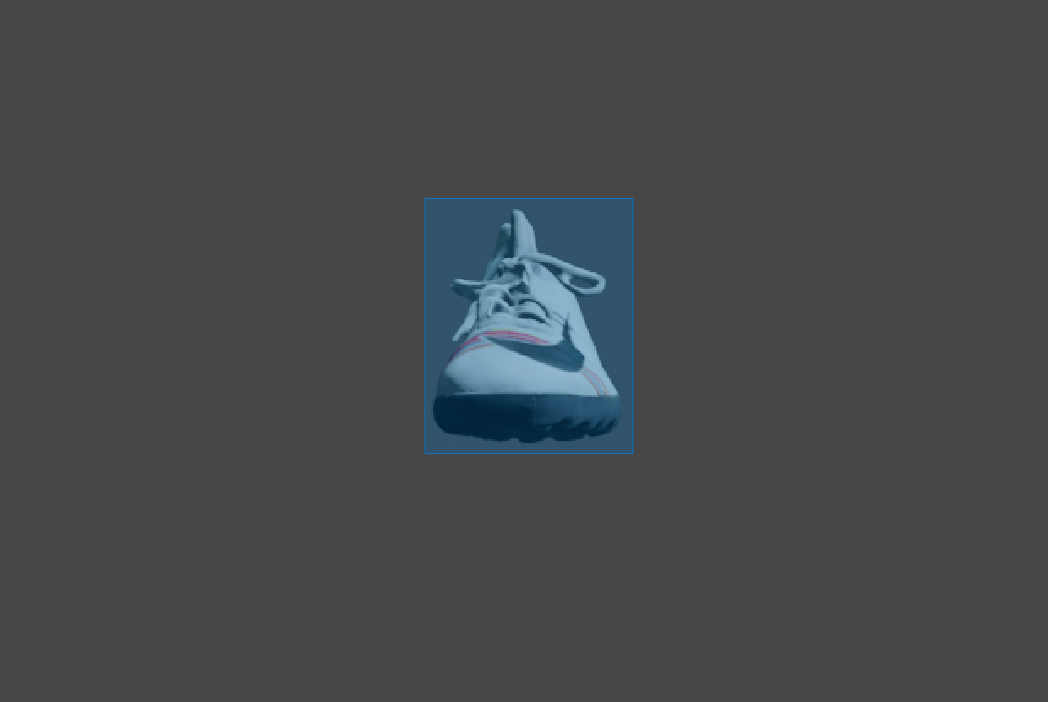 } }

    \vspace{-3mm}

    \subcaptionbox*{}%
    [.09\textwidth]{\includegraphics[width=\linewidth]{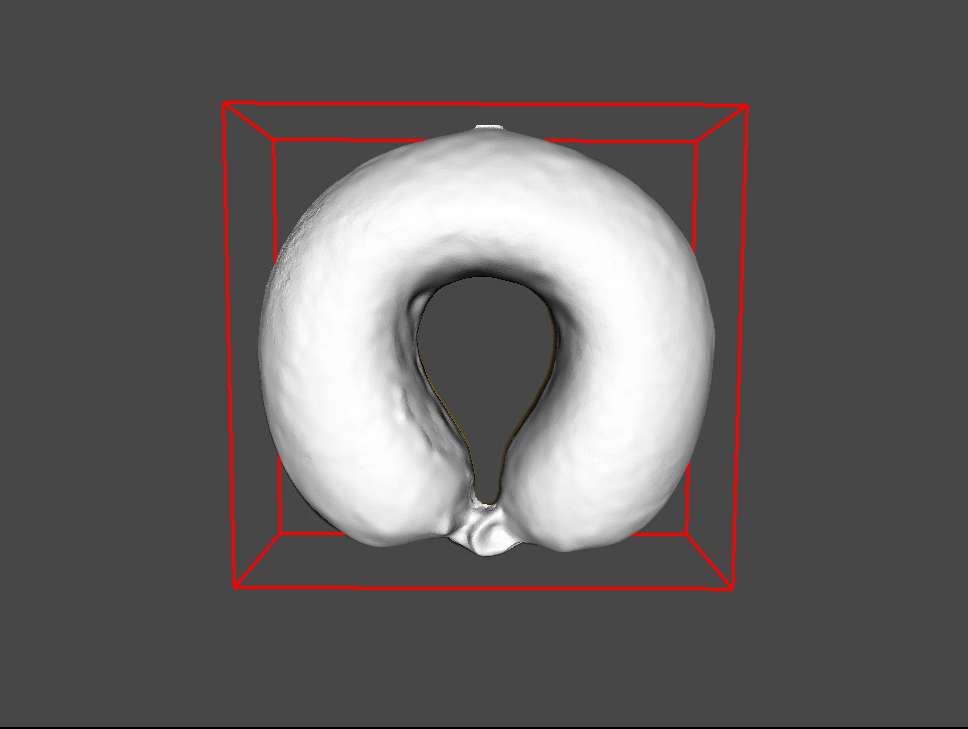} }
    \subcaptionbox*{}%
    [.09\textwidth]{\includegraphics[width=\linewidth]{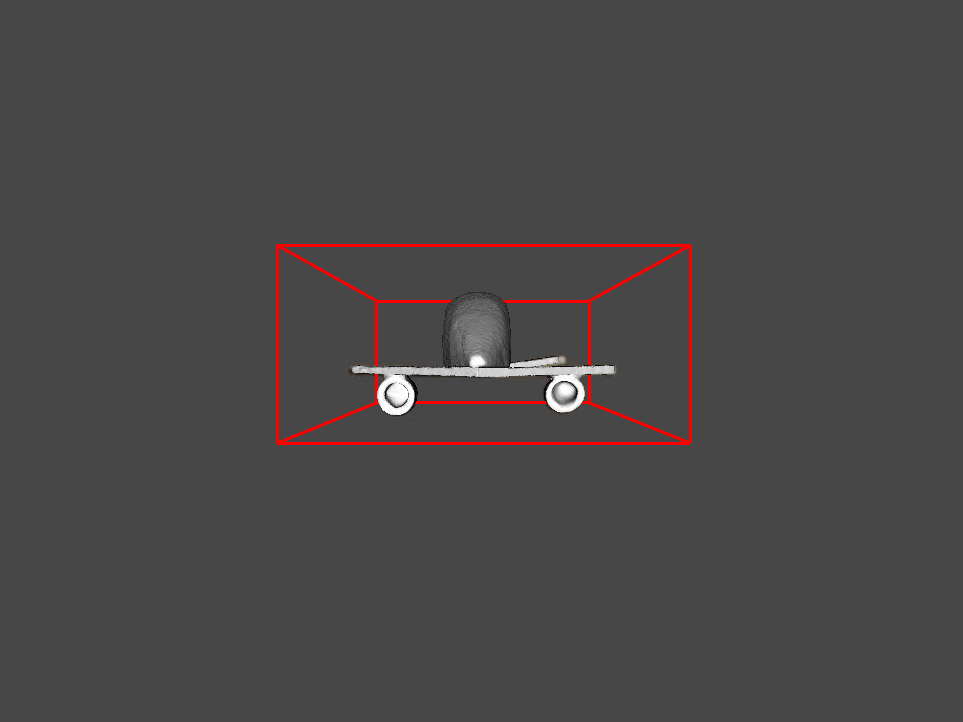 } }
    \subcaptionbox*{}%
    [.09\textwidth]{\includegraphics[width=\linewidth]{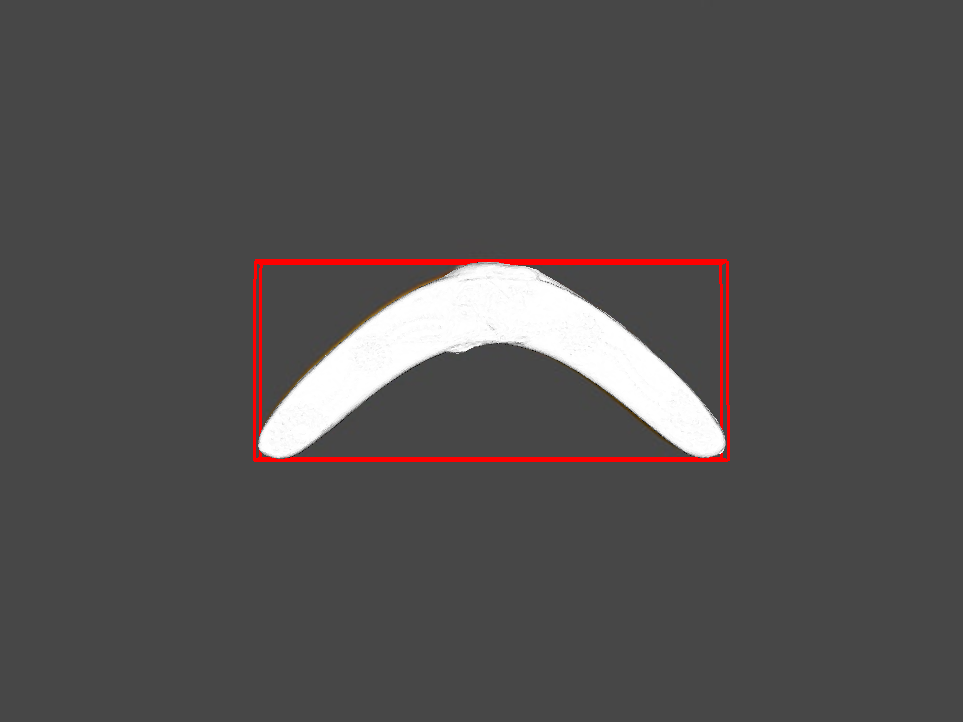 } }
    \subcaptionbox*{}%
    [.09\textwidth]{\includegraphics[width=\linewidth]{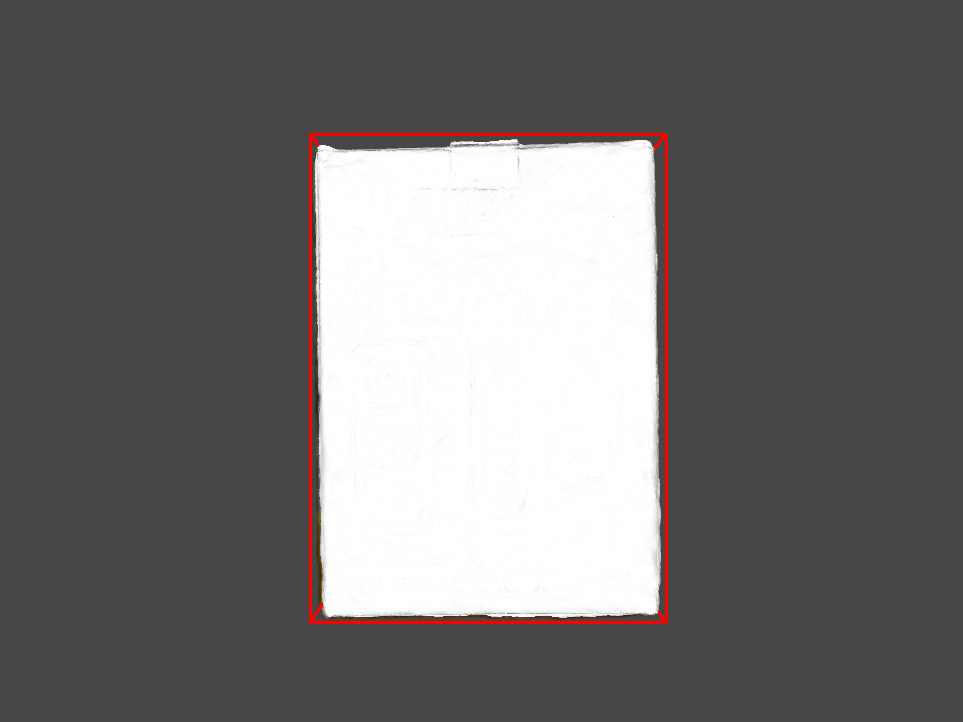 } }
    \subcaptionbox*{}%
    [.09\textwidth]{\includegraphics[width=\linewidth]{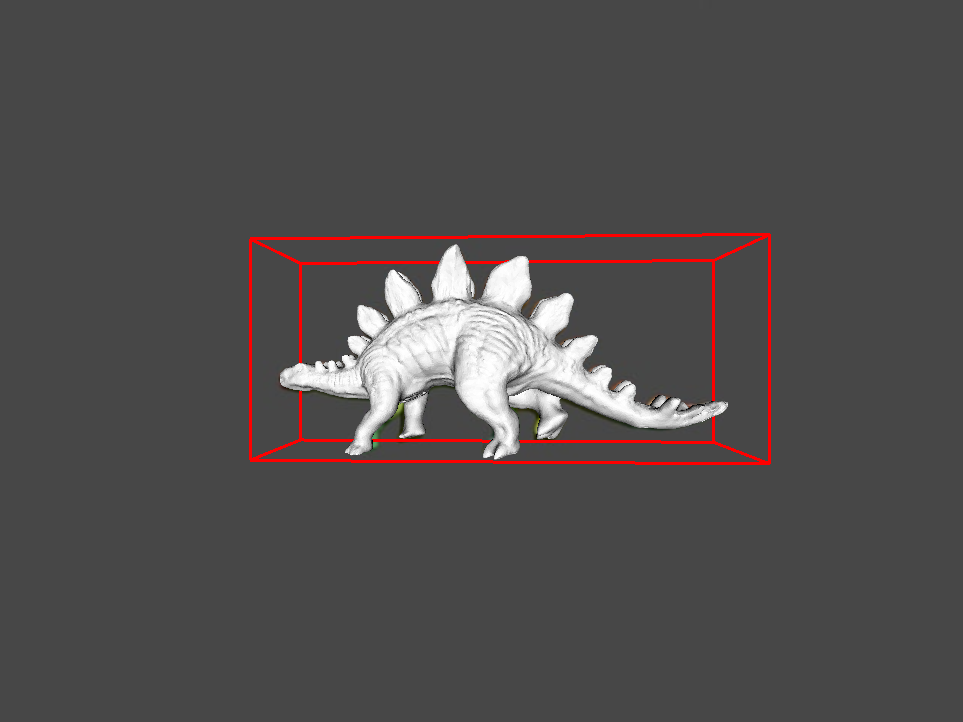 } }
    \subcaptionbox*{}%
    [.09\textwidth]{\includegraphics[width=\linewidth]{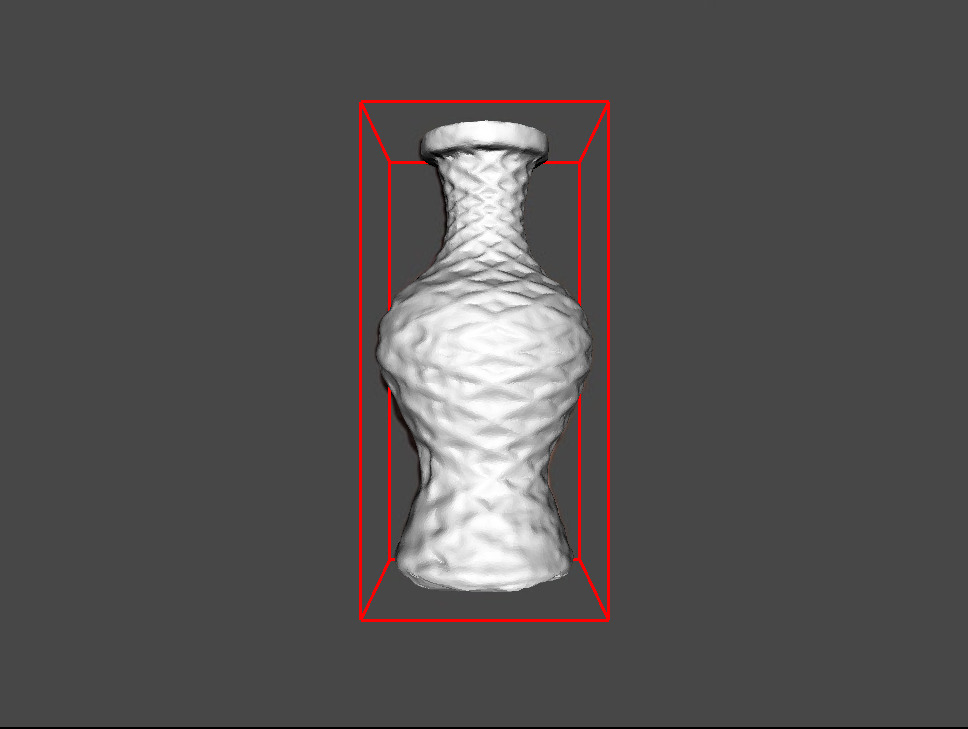 } }
    \subcaptionbox*{}%
    [.09\textwidth]{\includegraphics[width=\linewidth]{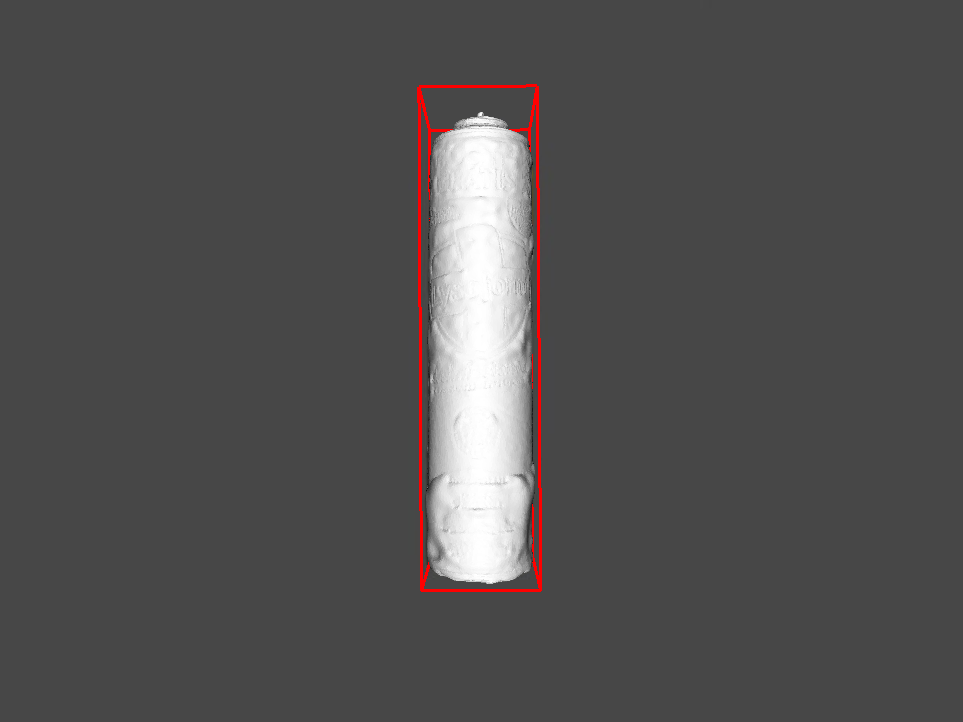 } }
    \subcaptionbox*{}%
    [.09\textwidth]{\includegraphics[width=\linewidth]{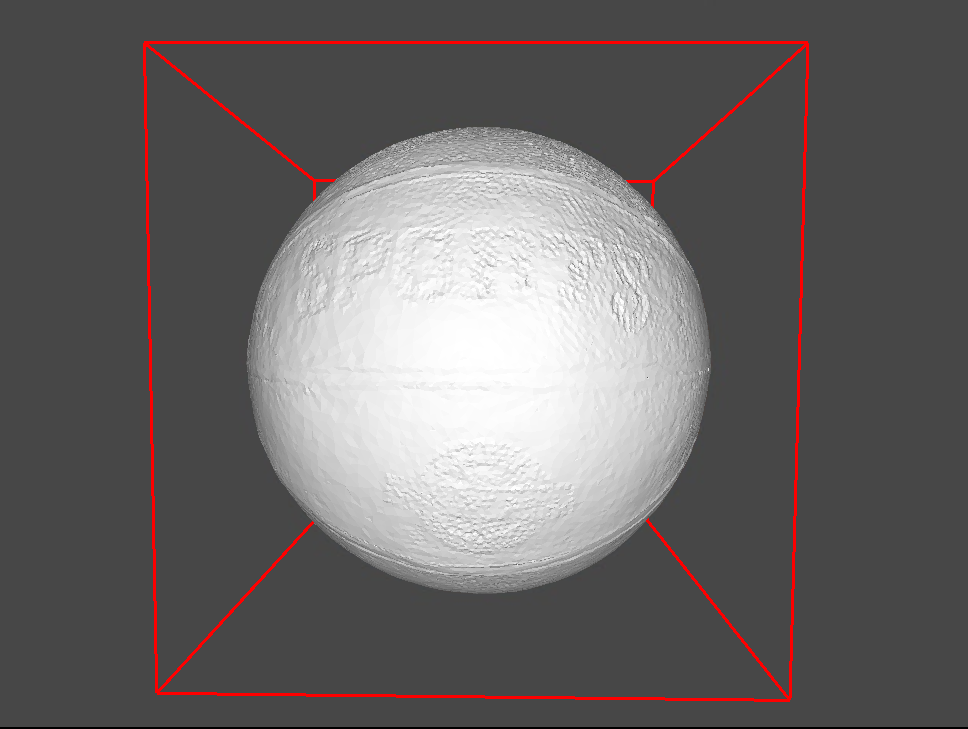 } }
    \subcaptionbox*{}%
    [.09\textwidth]{\includegraphics[width=\linewidth]{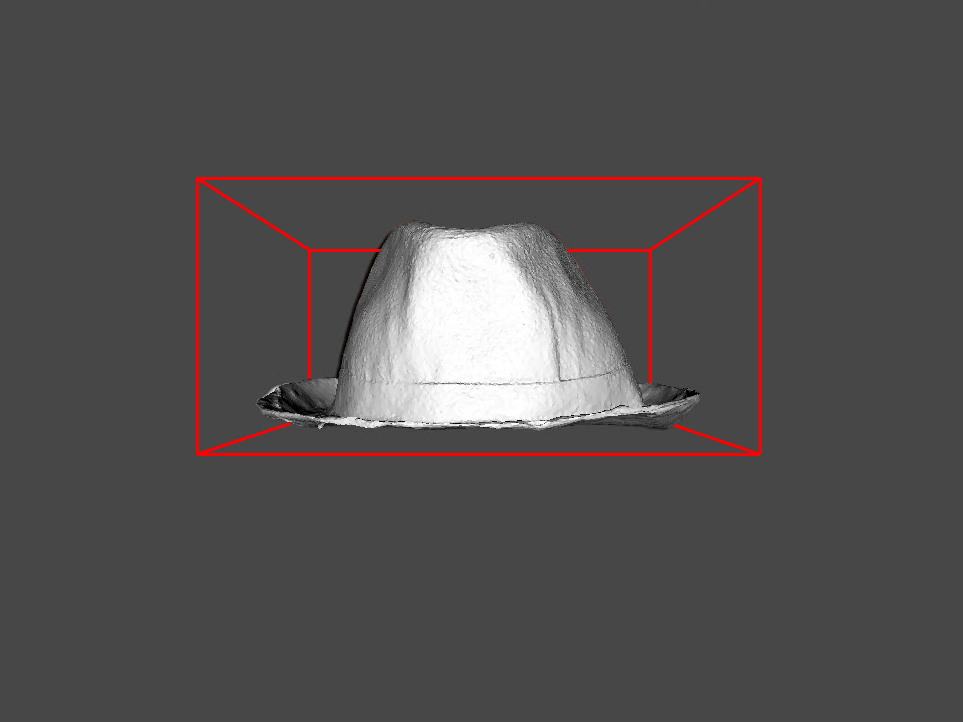 } }
    \subcaptionbox*{}%
    [.09\textwidth]{\includegraphics[width=\linewidth]{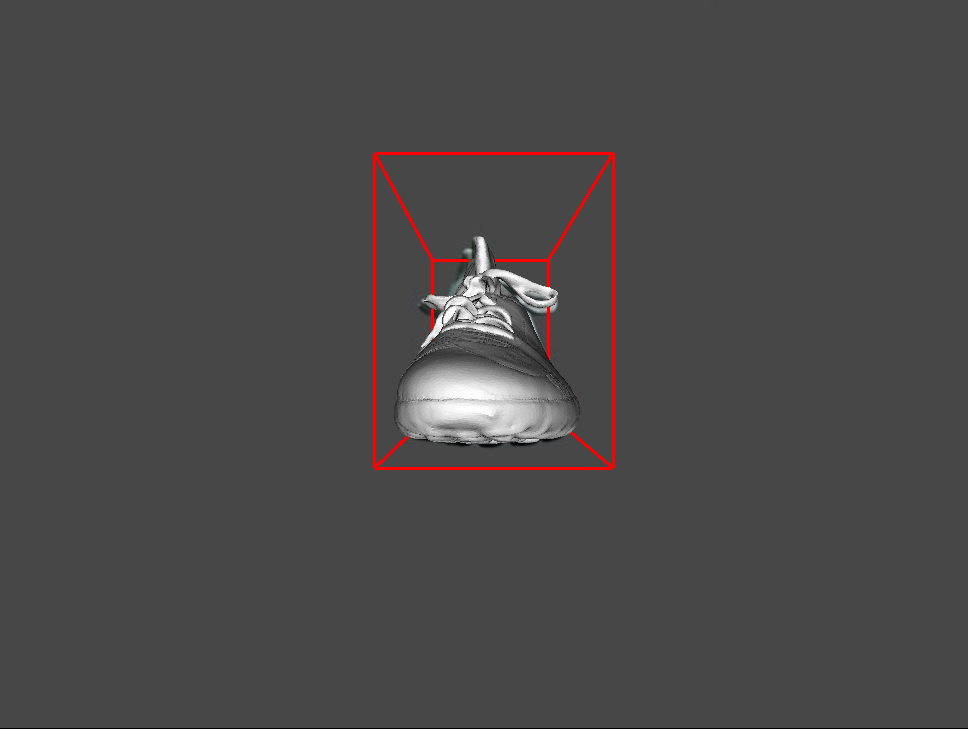 } }

    \vspace{-3mm}

    \subcaptionbox*{}%
    [.09\textwidth]{\includegraphics[width=\linewidth]{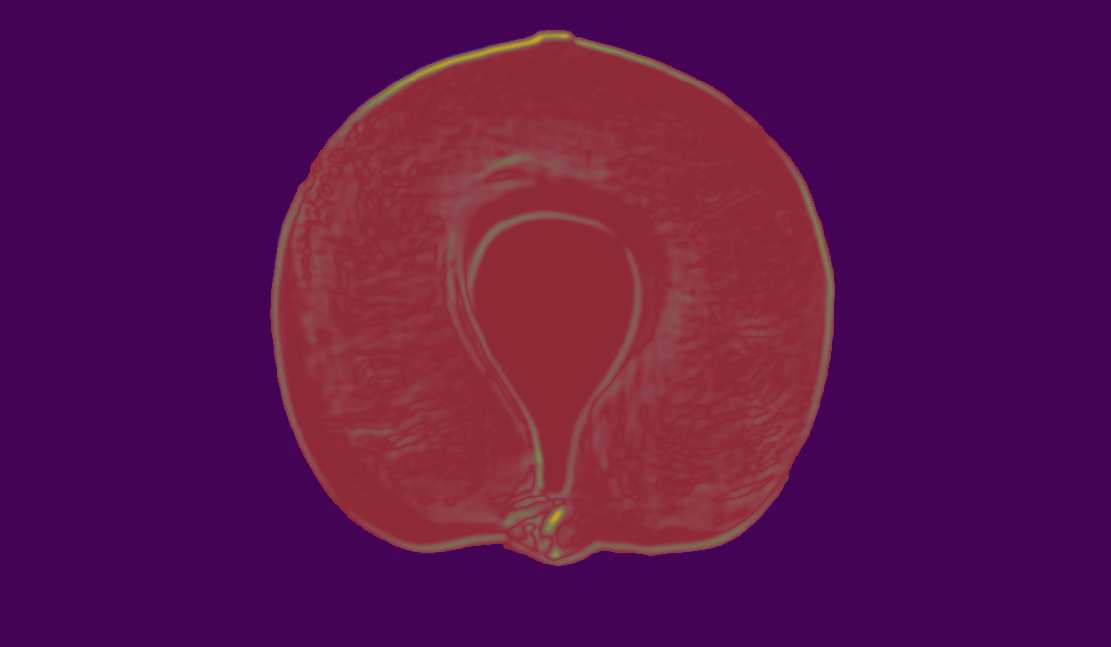 } }
    \subcaptionbox*{}%
    [.09\textwidth]{\includegraphics[width=\linewidth]{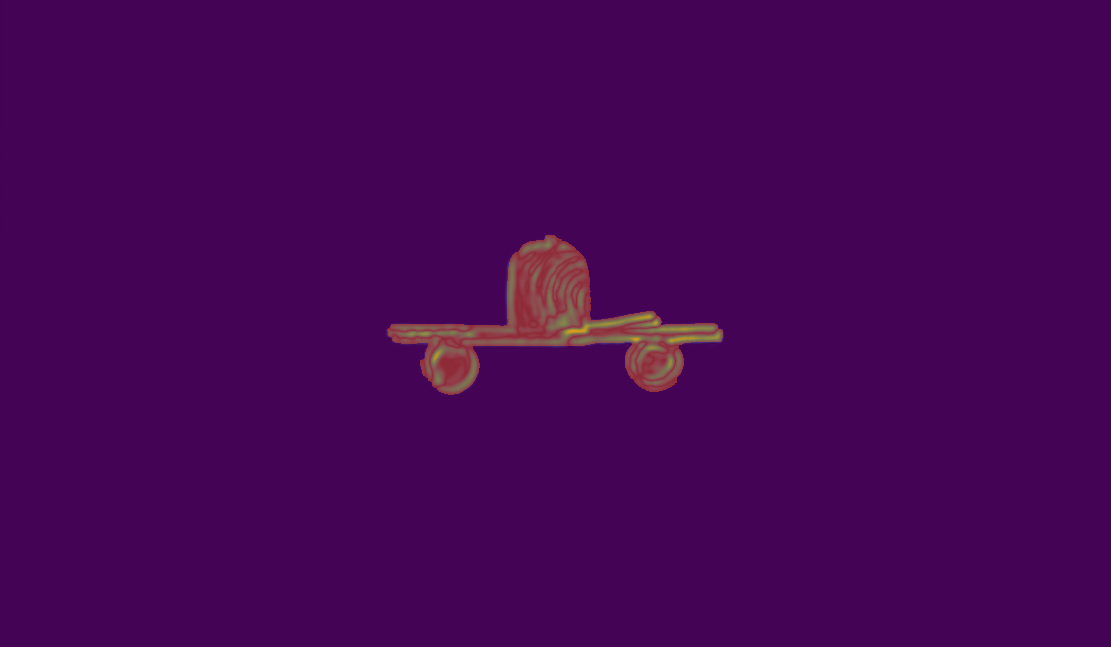 } }
    \subcaptionbox*{}%
    [.09\textwidth]{\includegraphics[width=\linewidth]{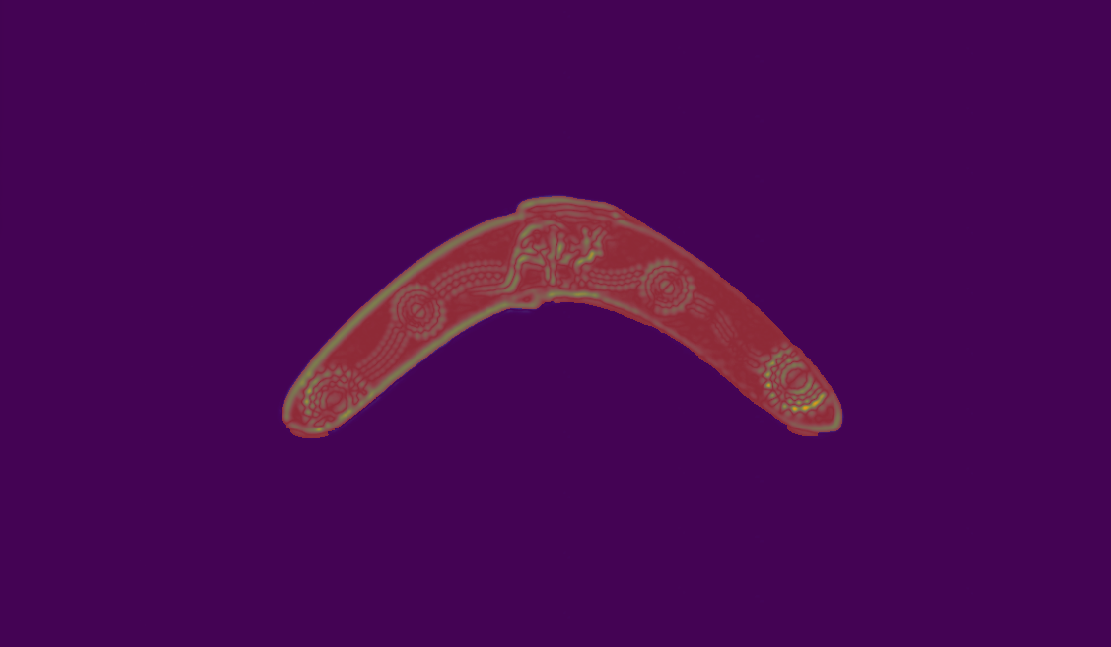 } }
    \subcaptionbox*{}%
    [.09\textwidth]{\includegraphics[width=\linewidth]{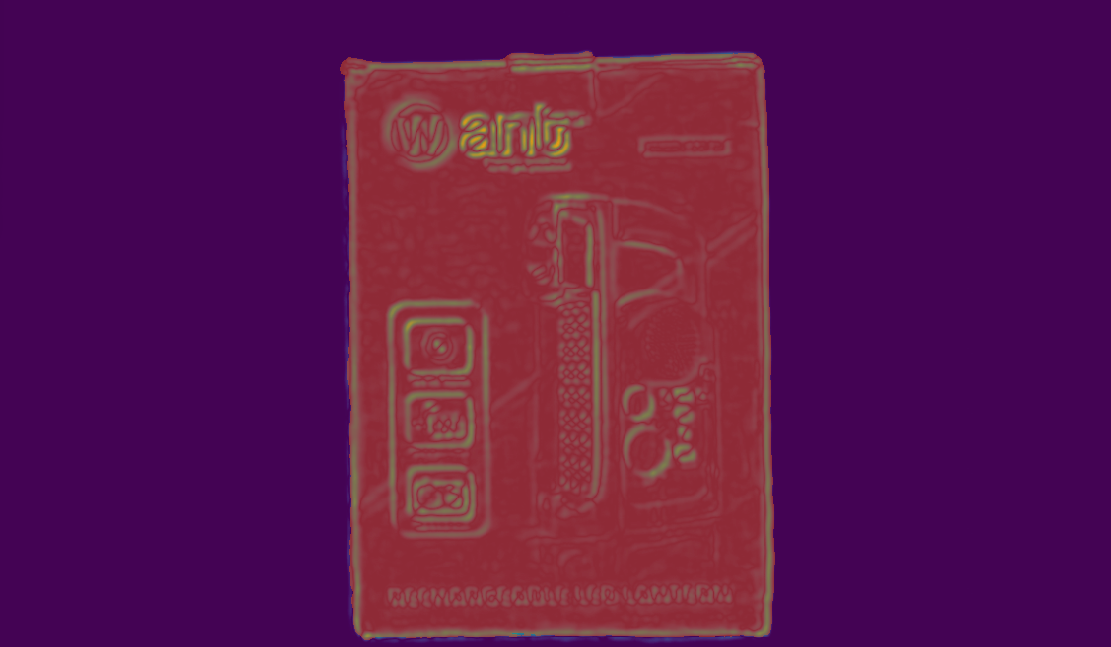 } }
    \subcaptionbox*{}%
    [.09\textwidth]{\includegraphics[width=\linewidth]{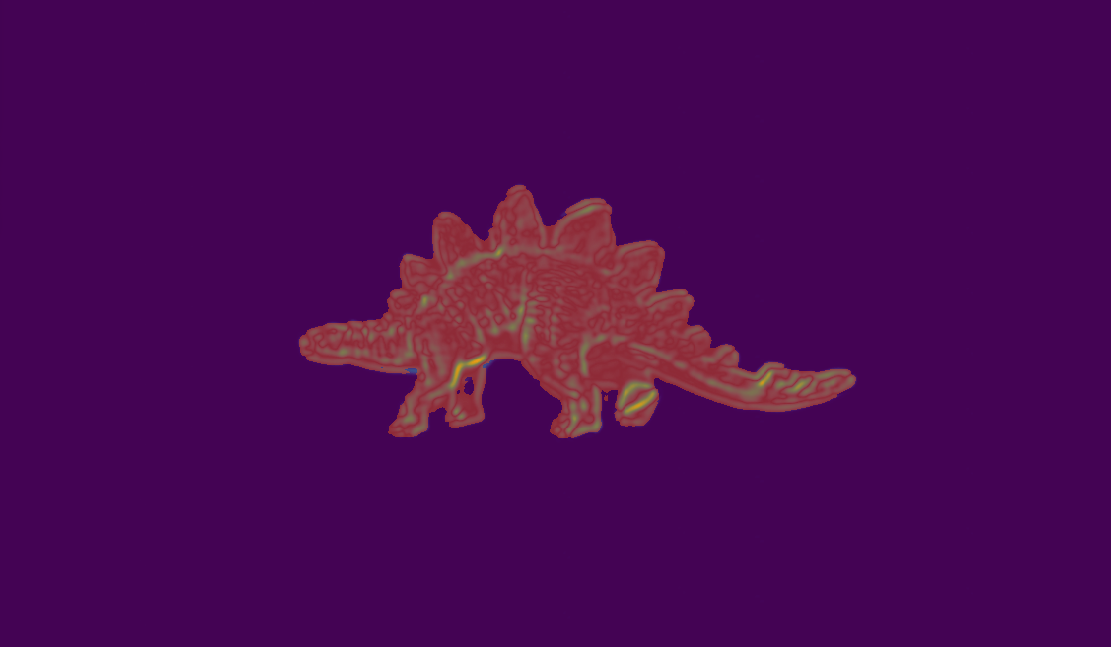 } }
    \subcaptionbox*{}%
    [.09\textwidth]{\includegraphics[width=\linewidth]{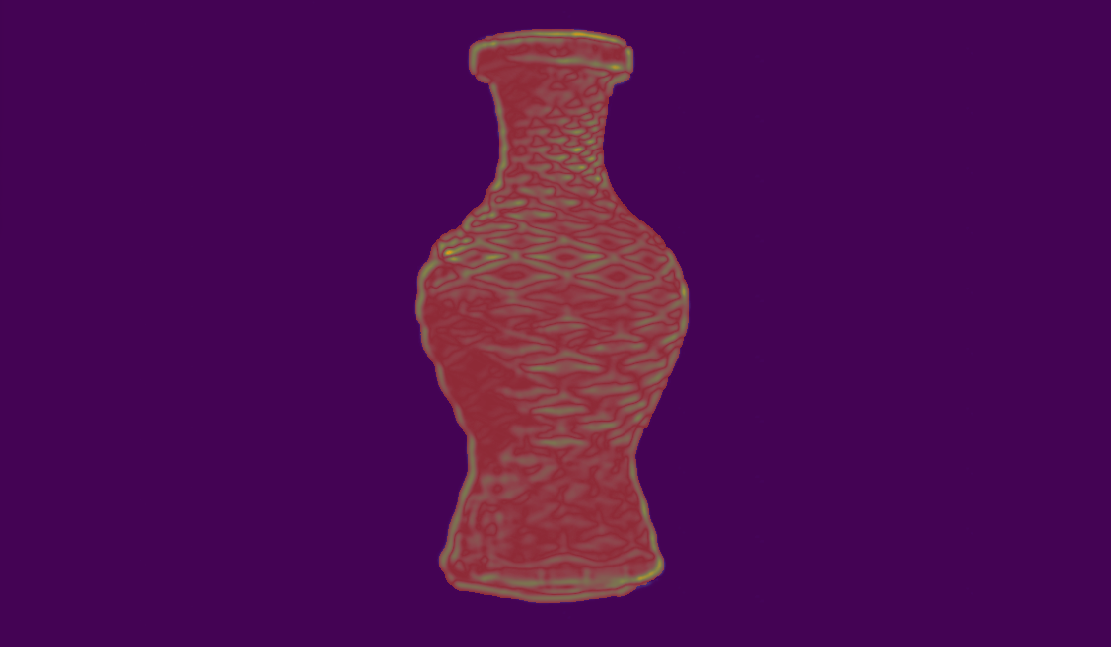 } }
    \subcaptionbox*{}%
    [.09\textwidth]{\includegraphics[width=\linewidth]{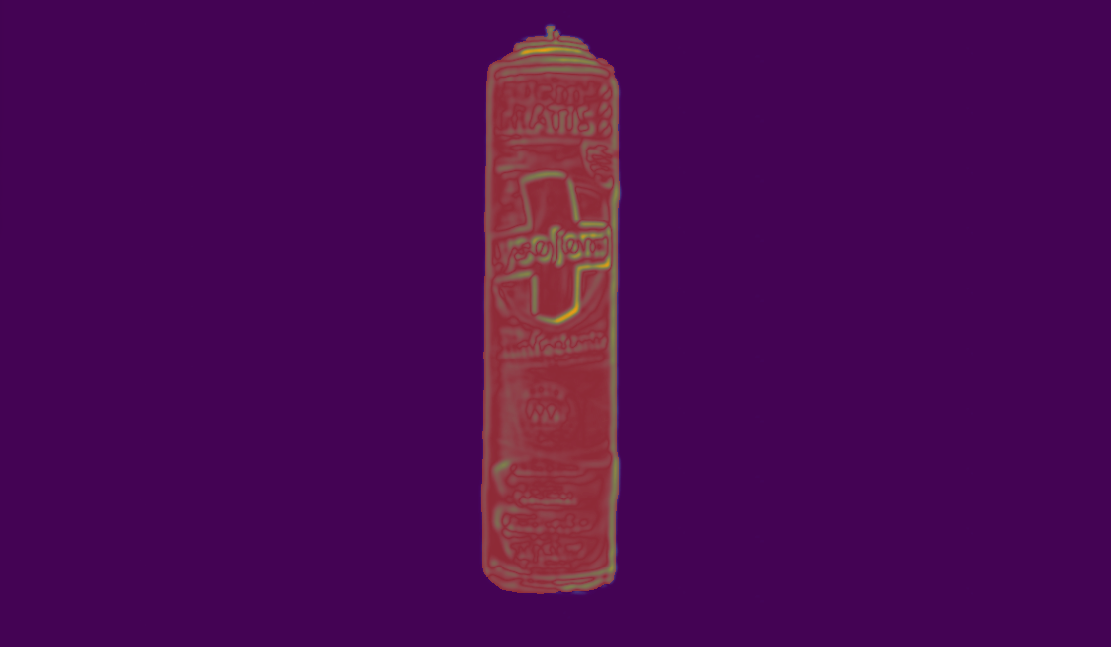 } }
    \subcaptionbox*{}%
    [.09\textwidth]{\includegraphics[width=\linewidth]{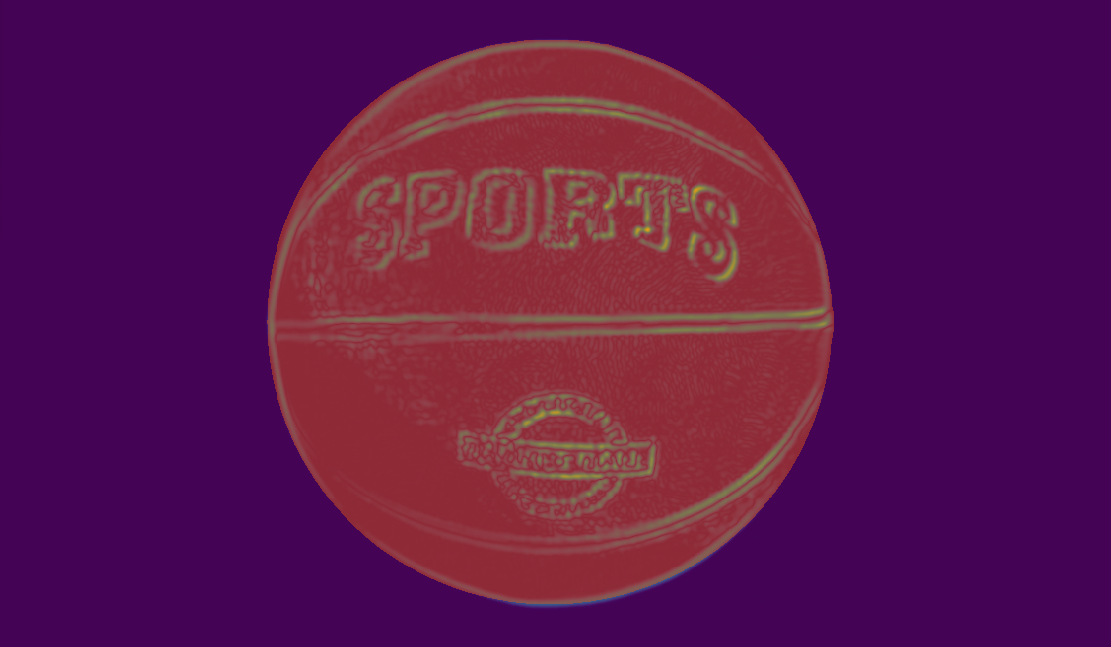 } }
    \subcaptionbox*{}%
    [.09\textwidth]{\includegraphics[width=\linewidth]{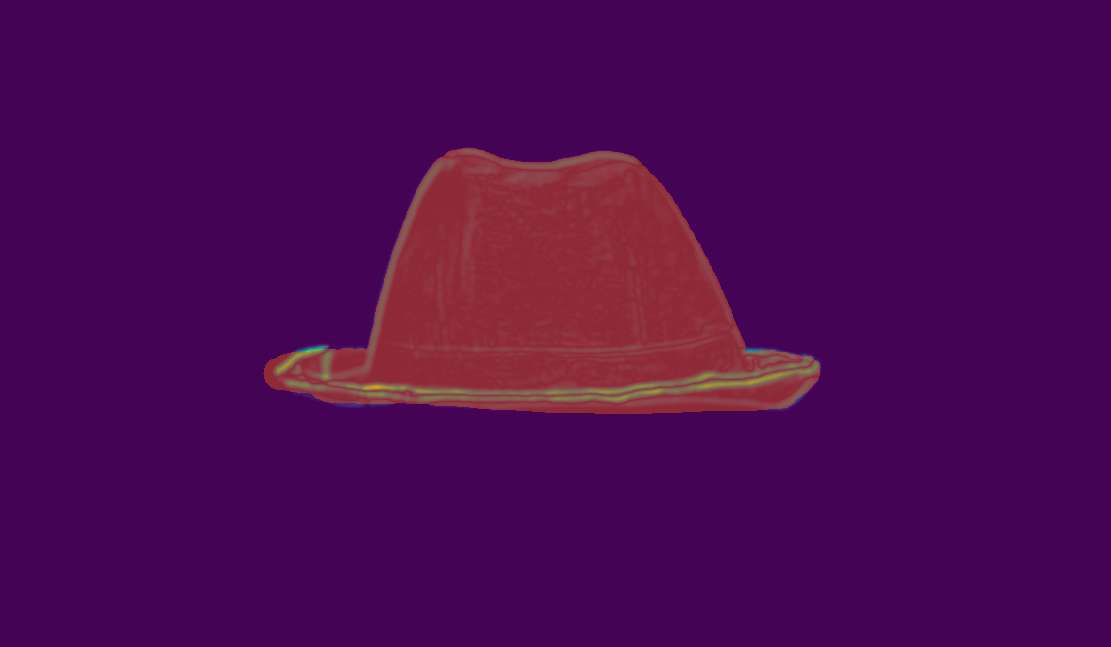 } }
    \subcaptionbox*{}%
    [.09\textwidth]{\includegraphics[width=\linewidth]{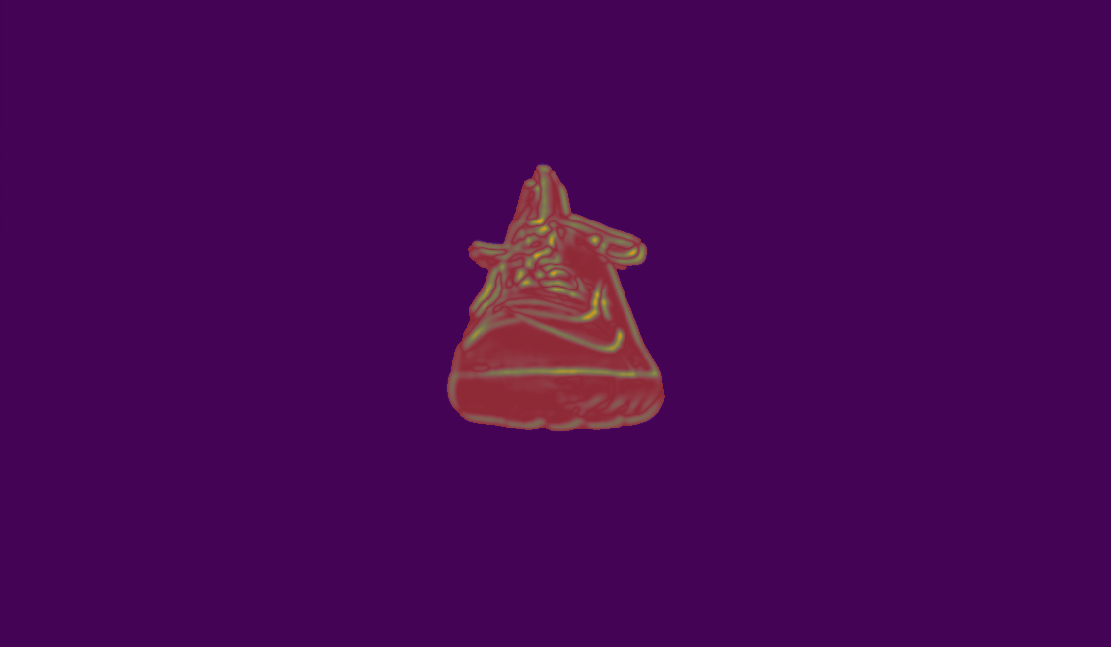 } }

    \vspace{-3mm}

    \subcaptionbox*{}%
    [.09\textwidth]{\includegraphics[width=\linewidth]{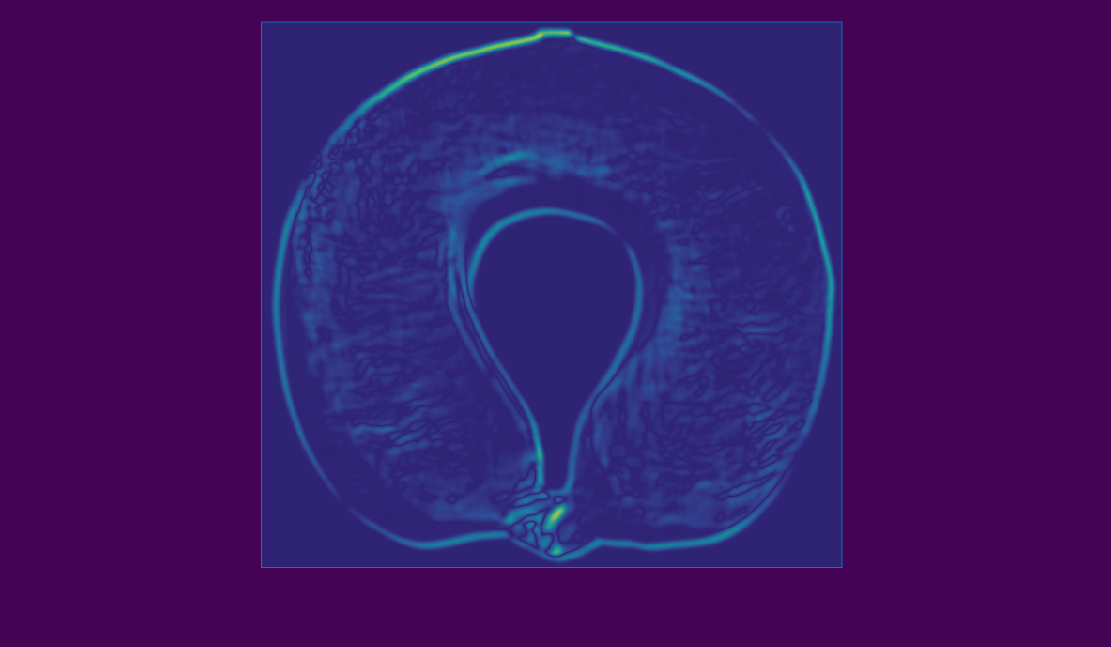 } }
    \subcaptionbox*{}%
    [.09\textwidth]{\includegraphics[width=\linewidth]{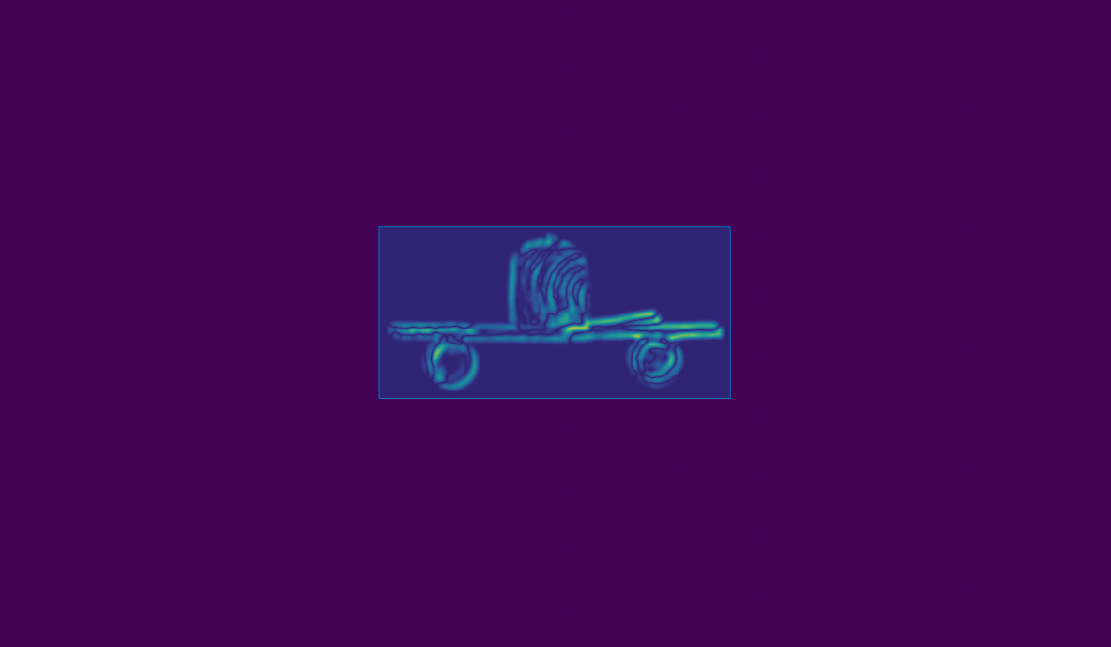 } }
    \subcaptionbox*{}%
    [.09\textwidth]{\includegraphics[width=\linewidth]{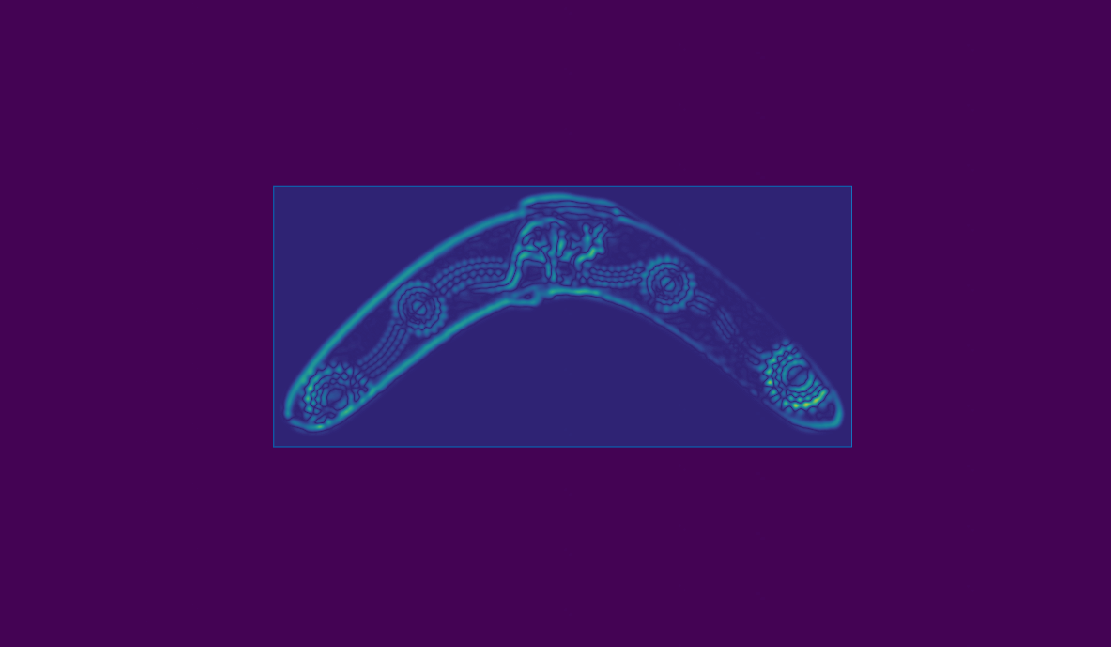 } }
    \subcaptionbox*{}%
    [.09\textwidth]{\includegraphics[width=\linewidth]{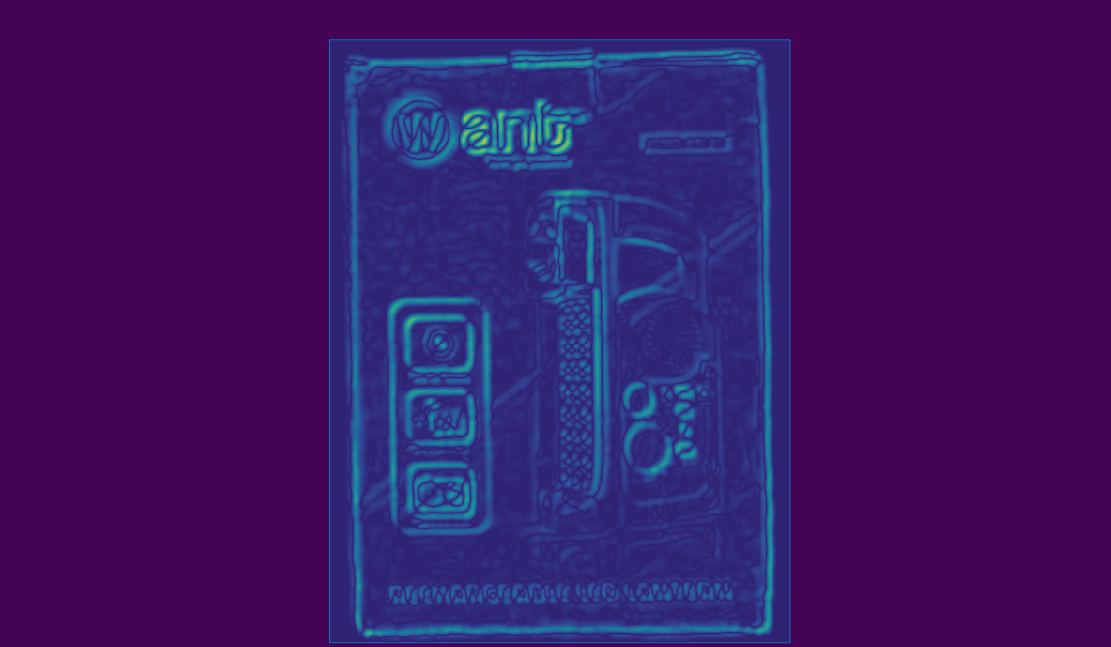 } }
    \subcaptionbox*{}%
    [.09\textwidth]{\includegraphics[width=\linewidth]{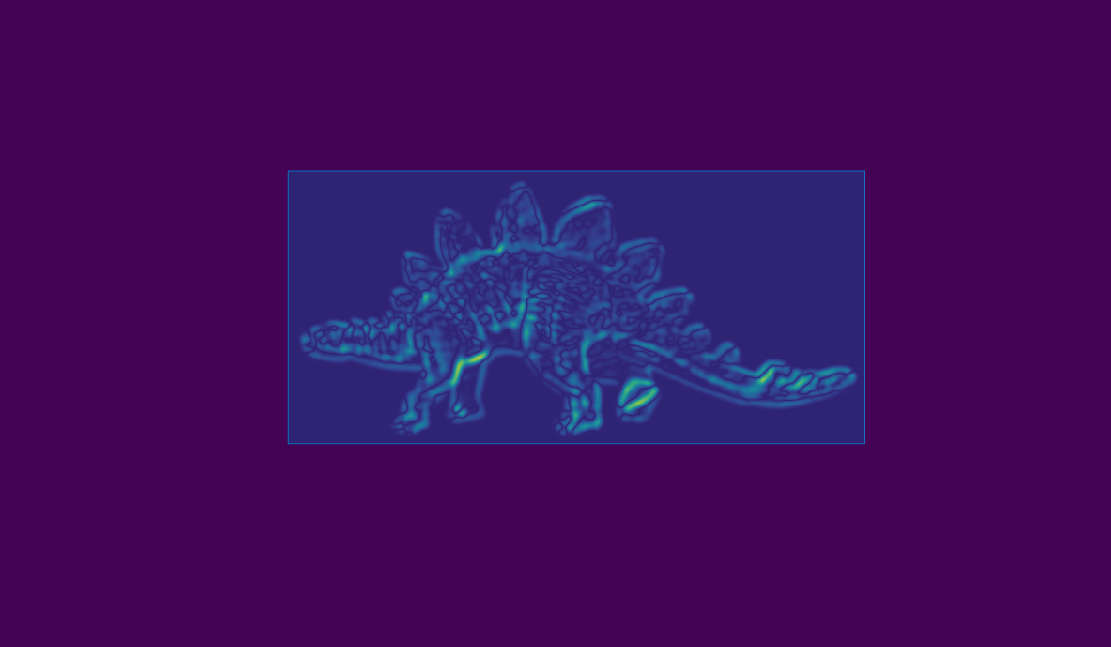 } }
    \subcaptionbox*{}%
    [.09\textwidth]{\includegraphics[width=\linewidth]{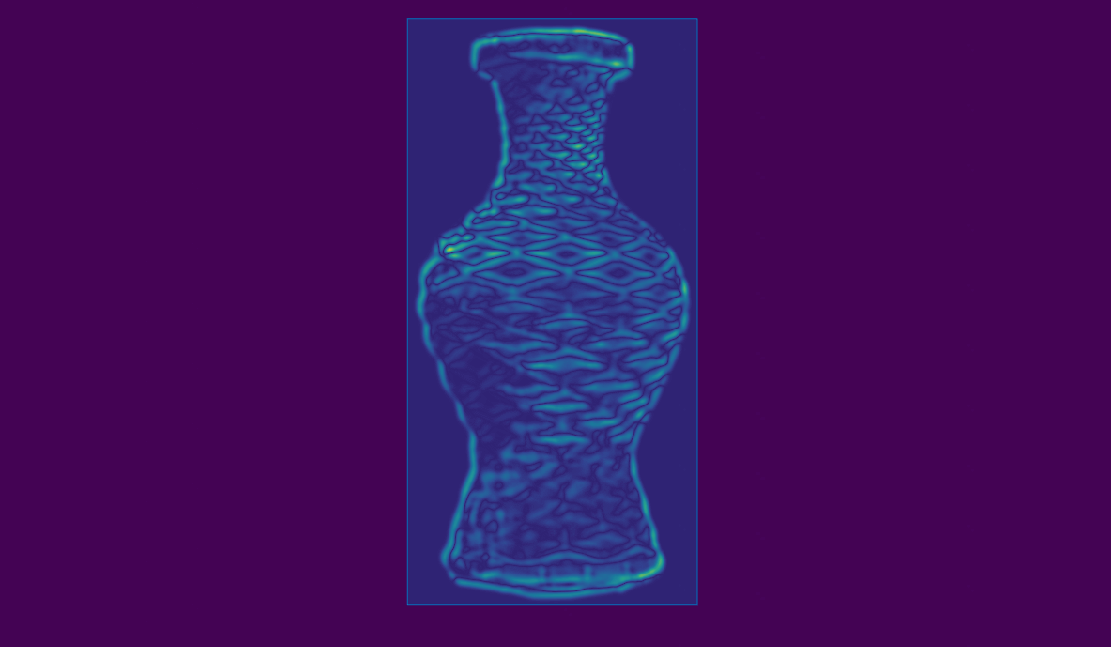 } }
    \subcaptionbox*{}%
    [.09\textwidth]{\includegraphics[width=\linewidth]{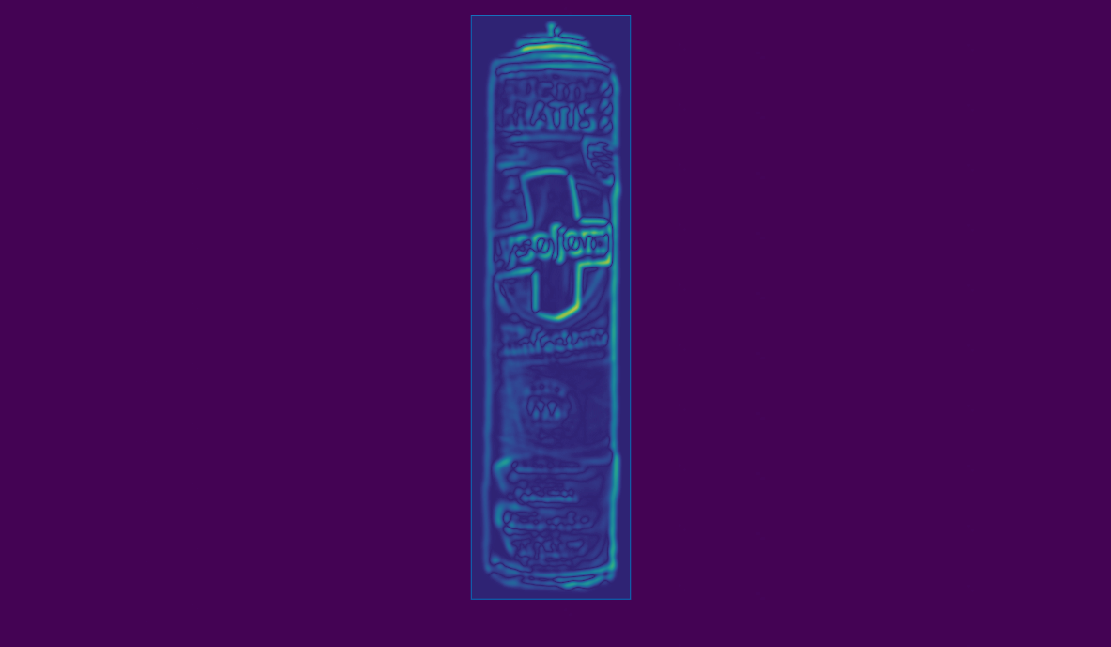 } }
    \subcaptionbox*{}%
    [.09\textwidth]{\includegraphics[width=\linewidth]{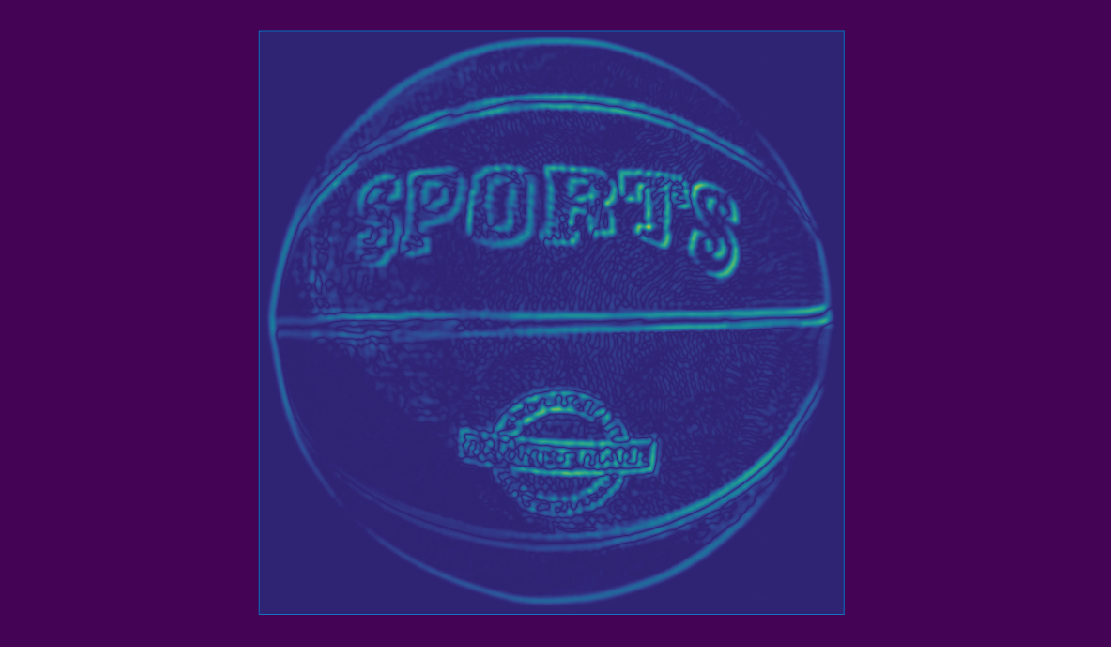 } }
    \subcaptionbox*{}%
    [.09\textwidth]{\includegraphics[width=\linewidth]{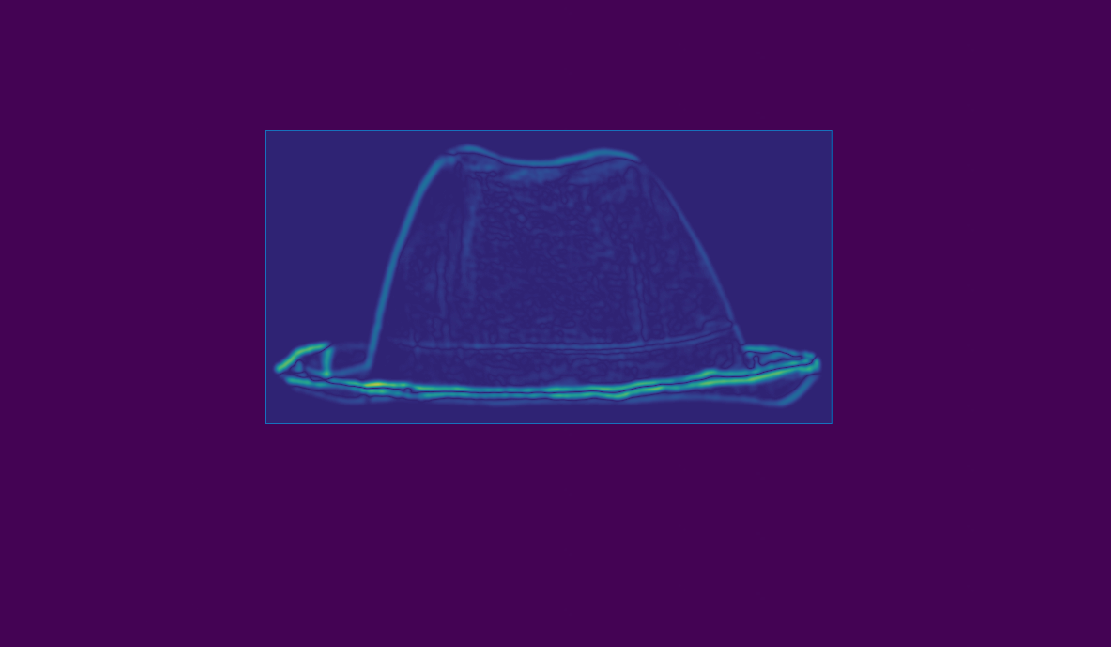 } }
    \subcaptionbox*{}%
    [.09\textwidth]{\includegraphics[width=\linewidth]{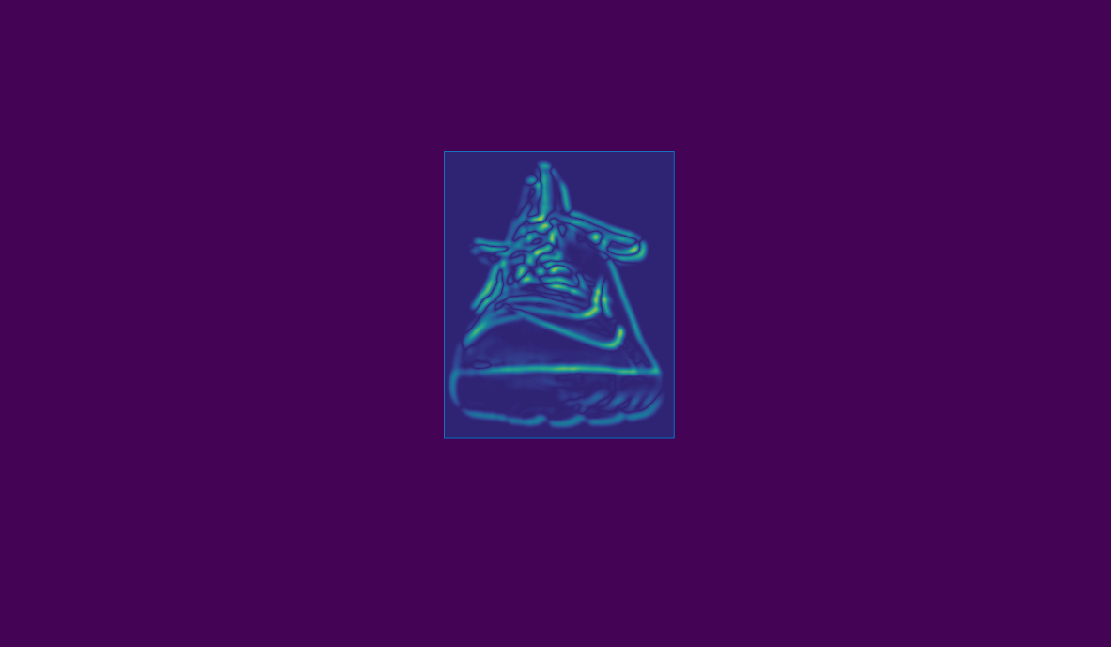 } }

    \vspace{-3mm}

    \subcaptionbox*{}%
    [.09\textwidth]{\includegraphics[width=\linewidth]{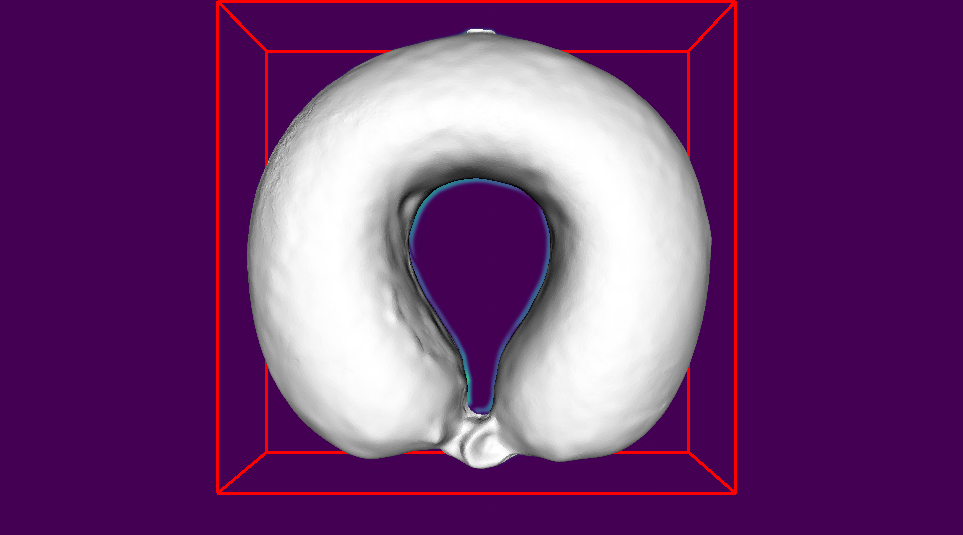 } }
    \subcaptionbox*{}%
    [.09\textwidth]{\includegraphics[width=\linewidth]{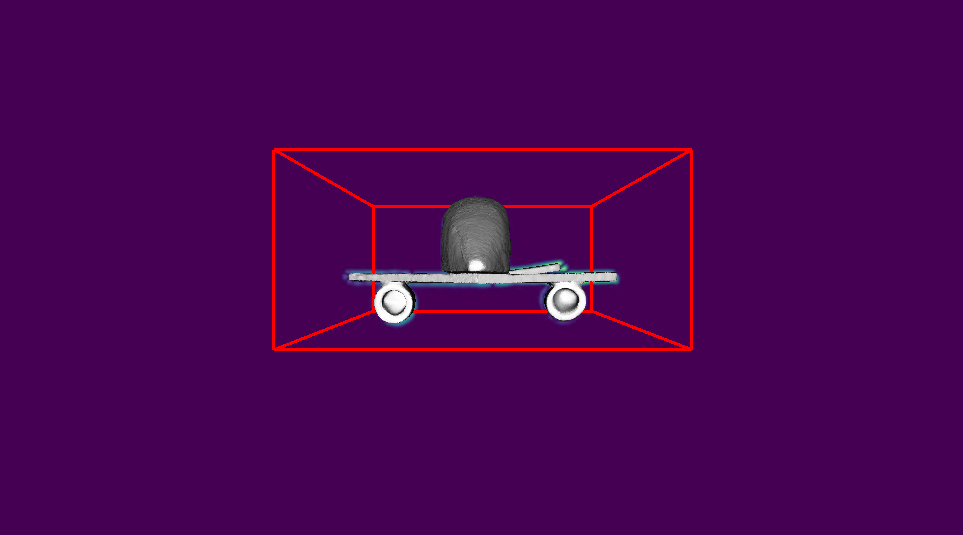 } }
    \subcaptionbox*{}%
    [.09\textwidth]{\includegraphics[width=\linewidth]{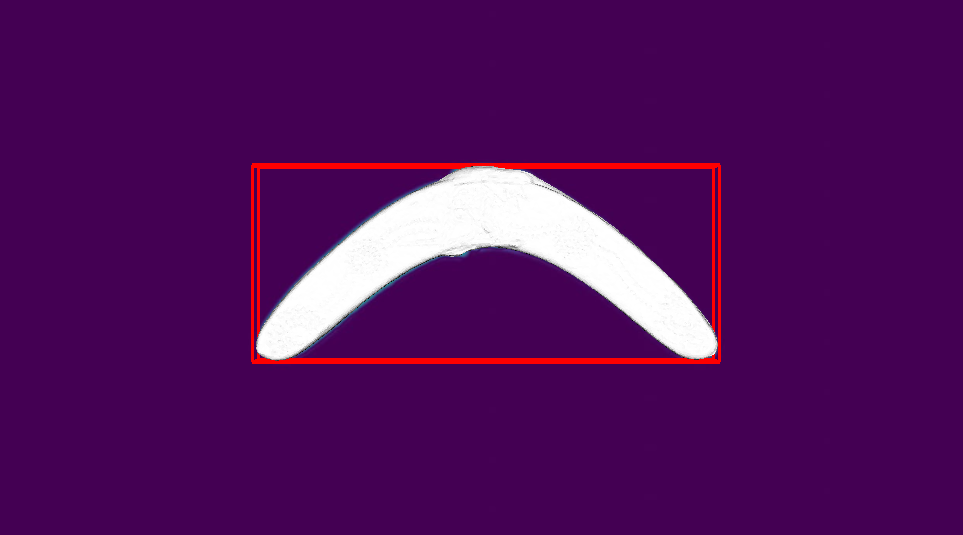 } }
    \subcaptionbox*{}%
    [.09\textwidth]{\includegraphics[width=\linewidth]{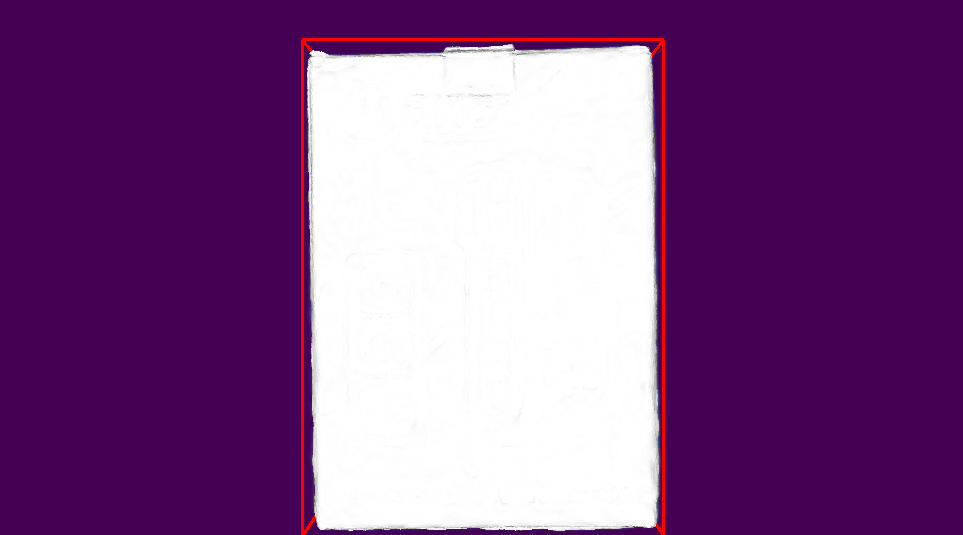 } }
    \subcaptionbox*{}%
    [.09\textwidth]{\includegraphics[width=\linewidth]{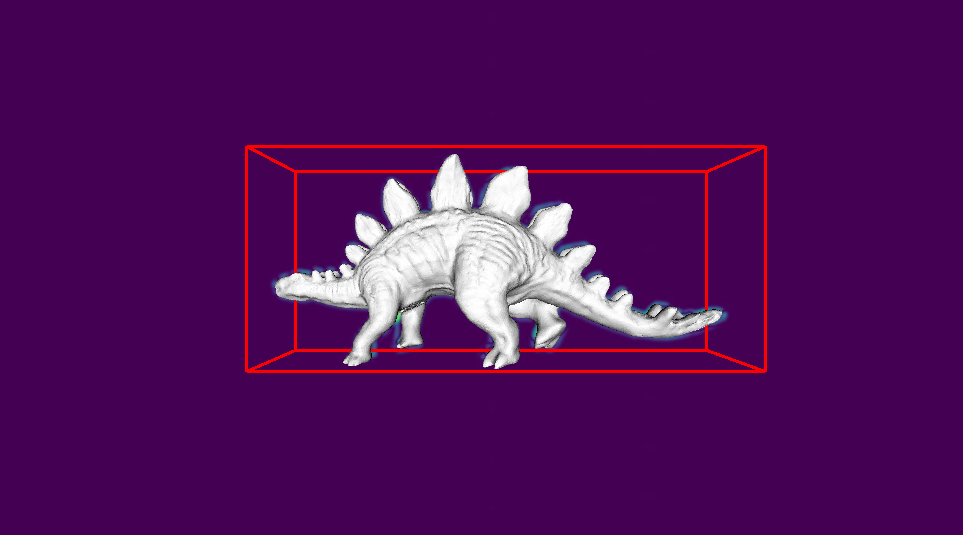 } }
    \subcaptionbox*{}%
    [.09\textwidth]{\includegraphics[width=\linewidth]{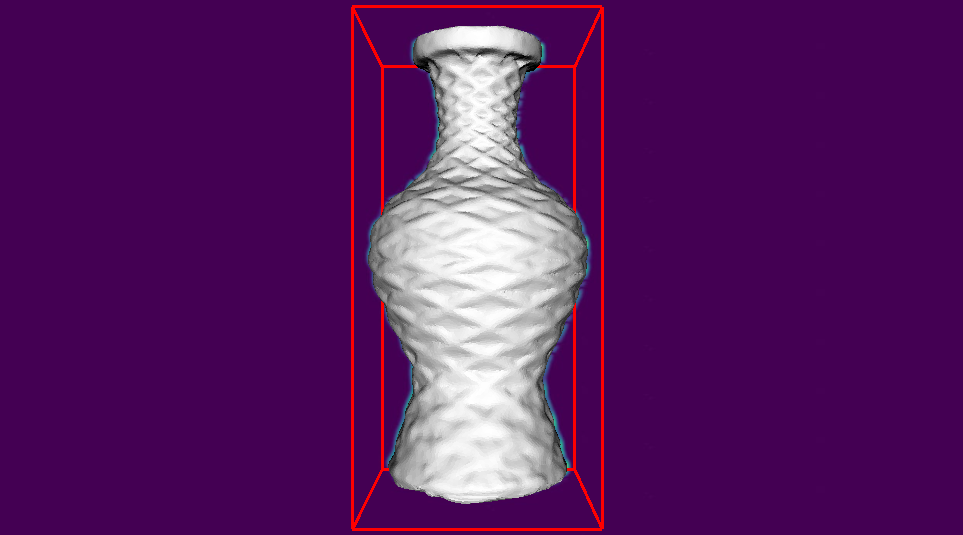 } }
    \subcaptionbox*{}%
    [.09\textwidth]{\includegraphics[width=\linewidth]{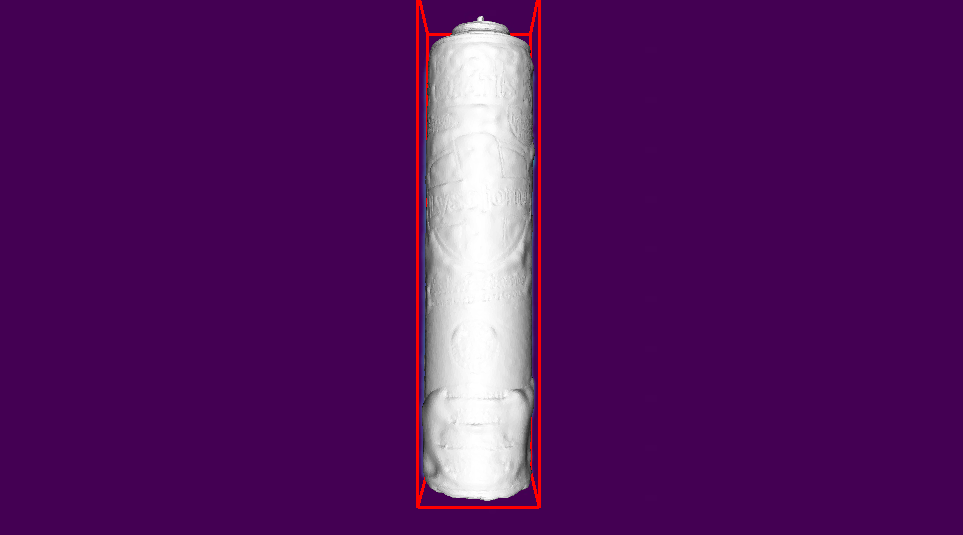 } }
    \subcaptionbox*{}%
    [.09\textwidth]{\includegraphics[width=\linewidth]{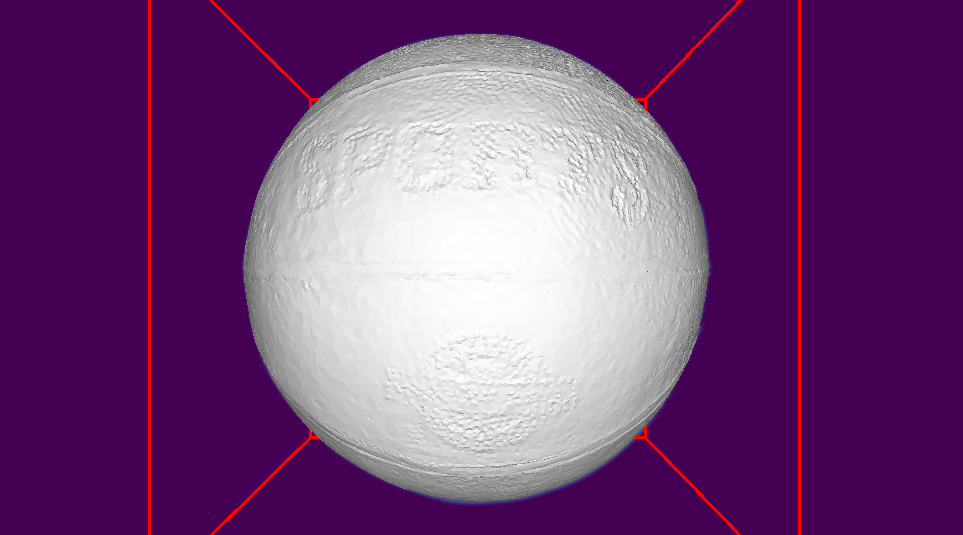 } }
    \subcaptionbox*{}%
    [.09\textwidth]{\includegraphics[width=\linewidth]{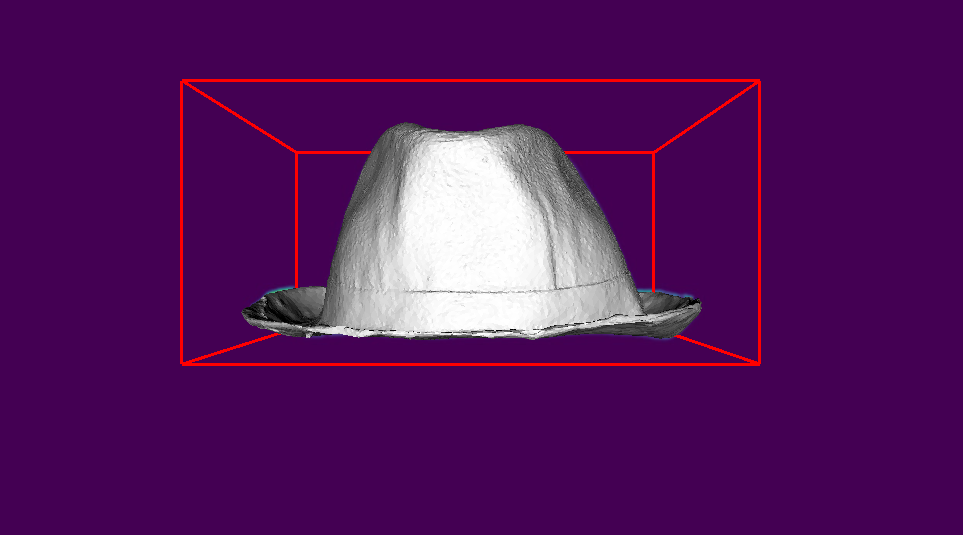 } }
    \subcaptionbox*{}%
    [.09\textwidth]{\includegraphics[width=\linewidth]{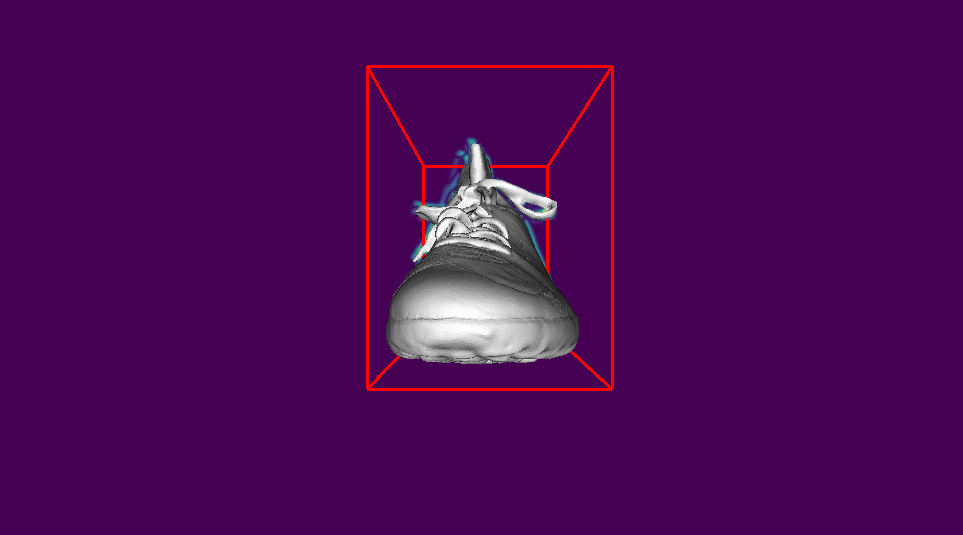 } }

    \caption{Moving6DPoSe-S: synthetic moving object sequences generated with the ROG Eye S and Prophesee EVK4 models. The top three rows show frame-based segmentation, detection, and 6D pose annotations, while the bottom three rows show the event annotations.}
    \label{fig:moving6dpose-s}
\end{figure*}

Ground-truth segmentation and detection annotations were automatically generated from rendered frames in Blender.
For 6D pose annotation, moving objects were rendered in front of a frame-based camera, and their poses relative to the camera were automatically extracted using Blender's Python API. The 6D pose was derived by computing the transformation matrix of the object to the camera’s coordinate system. This was achieved by inverting the camera's world transformation matrix and multiplying it by the object's world matrix, yielding the relative pose. The resulting 6D pose annotations were stored frame-by-frame and then extrapolated to the simulated events, as shown in Figure~\ref{fig:moving6dpose-s}.

\section{Moving6DPoSe Database Format and Usage}
\label{sec:db-format}

The database comprises two subsets, Moving6DPoSe-R and Moving6DPoSe-S, both containing data from the 16 scanned objects. Each subset is hierarchically organized by object, scenario, experiment, and run. Each run stores the data recorded by the different cameras.
The data from the simulated and employed cameras include:
\begin{itemize}
    \item ROG Eye S frames \texttt{.png}.
    \item ZED2: Right and left stereo camera frames \texttt{.png} and depth data \texttt{.npy}.
    \item DAVIS346: Event data timestamp image \texttt{.png} and events \texttt{.npy}.
    \item Prophesee EVK4: Event data timestamp image \texttt{.png} and events \texttt{.npy}.
\end{itemize}
Experiments are also provided in \texttt{.bag} format.
In addition, an annotation file \texttt{.txt} is included for each task in all sensors, and we also provide the segmentation and depth masks.
Finally, intrinsic and extrinsic calibration parameters are provided for each sensor.

\section{Moving6DPoSe baseline evaluation}
\label{sec:baselines}

Moving6DPoSe supports multiple tasks, including segmentation, detection, and 6D pose estimation. 

\subsection{Object Semantic Segmentation}

U-Net \cite{ronneberger2015unet} is adopted as the baseline for object semantic segmentation in Moving6DPoSe. Its encoder-decoder architecture with skip connections enables accurate pixel-level predictions across different sensing modalities. The model performs binary semantic segmentation on RGB images acquired with the ROG Eye S camera and timestamp representations generated from the DAVIS346 and EVK4 event cameras.
Performance was evaluated using mean Intersection over Union (mIoU), Dice coefficient, and Recall. The model was trained with a batch size of 16, a learning rate of $10^{-3}$, for 20 epochs, using four data-loading workers. Table~\ref{tab:segmentation_unet_results} reports the average performance across all motion scenarios for each sensor and dataset.
\begin{table}[!t]
\centering
\setlength{\tabcolsep}{7pt}
\renewcommand{\arraystretch}{1.0}
\caption{Average semantic segmentation results using the U-Net baseline. mIoU, Dice and Recall metrics are averaged across all motion scenarios.}
\label{tab:segmentation_unet_results}
\begin{tabular}{|c|c|c|c|c|c|c|}
\hline
\multirow{2}{*}{\textbf{Sensor}} &
\multicolumn{3}{c|}{\textbf{Moving6DPoSe-S}} &
\multicolumn{3}{c|}{\textbf{Moving6DPoSe-R}} \\
\cline{2-7}
&
\textbf{mIoU} &
\textbf{Dice} &
\textbf{Recall} &
\textbf{mIoU} &
\textbf{Dice} &
\textbf{Recall} \\
\hline

ROG Eye S &
0.5694 &
0.6972 &
0.6008 &
0.4835 &
0.6169 &
0.5322 \\
\hline

DAVIS346 &
0.7798 &
0.8668 &
0.8701 &
0.7686 &
0.8595 &
0.8656 \\
\hline

EVK4 &
0.6672 &
0.7867 &
0.7163 &
0.6277 &
0.7548 &
0.6778 \\
\hline

\end{tabular}
\end{table}
 
\subsection{Object Detection}

Faster R-CNN \cite{ren2017fasterrcnn} is adopted as the baseline for object detection in Moving6DPoSe. The model is applied to RGB images acquired with the ROG Eye S camera and timestamp representations generated from the DAVIS346 and EVK4 event cameras.
Detection performance is evaluated using mAP@95. The detector was trained with a batch size of 16, a learning rate of $10^{-3}$, for 20 epochs, using four data-loading workers. Table~\ref{tab:detection_map95_results} reports the average detection performance across all motion scenarios for each sensor and dataset.
\begin{table}[t]
\centering
\caption{Average object detection results using the Faster R-CNN baseline. Results are averaged over all scenarios.}
\label{tab:detection_map95_results}
\begin{tabular}{|c||c|c|}
\hline
\multirow{2}{*}{\textbf{Sensor}} &
\multicolumn{2}{c|}{\textbf{mAP@95}} \\
\cline{2-3}
& \textbf{Moving6DPoSe-S} & \textbf{Moving6DPoSe-R} \\
\hline
ROG Eye S & 0.9192 & 0.9366 \\
\hline
DAVIS346 & 0.6904 & 0.6852 \\
\hline
EVK4 & 0.8528 & 0.8386 \\
\hline
\end{tabular}
\end{table}

\subsection{6D Object Pose Estimation}

\begin{figure}[!tb]
\centering
\includegraphics[width=0.9\linewidth]{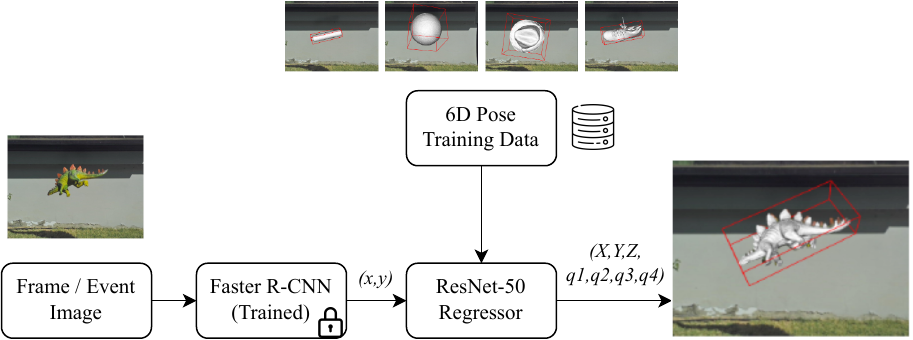}
\caption{Baseline pipeline for 6D pose estimation. A Faster R-CNN detector provides object bounding boxes, and the cropped regions are processed by a ResNet-50 regressor to estimate the object 3D center position and orientation.}
\label{fig:baseline_6dpose}
\end{figure}

Most monocular 6D pose estimation methods rely on object CAD models during training or inference \cite{xiang2018posecnn,maji2024yolo6dpose,melekhov2019dgcnet}. As an initial evaluation of Moving6DPoSe, we adopt a simple image-based baseline that estimates object pose directly from visual observations without explicitly using CAD model information.
Inspired by PoseContrast \cite{xiao2021posecontrast}, we employ a direct regression approach based on a ResNet-50 backbone. The network estimates the object 6D pose from cropped detections using two output heads: one predicts the object depth and 3D center position $(x,y)$ in camera coordinates, while the other predicts the object orientation as a normalized quaternion $(q_1,q_2,q_3,q_4)$.
As illustrated in Figure~\ref{fig:baseline_6dpose}, detected objects are cropped, normalized, resized, and processed by an ImageNet-pretrained ResNet-50. The network is trained end-to-end to jointly estimate object position and orientation from RGB images (ROG Eye S) and time-surface event representations.

The loss function combines mean squared error (MSE) for object center position and depth estimation with a geodesic loss for quaternion-based orientation prediction,
$
\mathcal{L}_{geo}=2\arccos\!\left(|\langle\hat{q},q\rangle|\right),
$
where $\langle\hat{q},q\rangle$ denotes the dot product between unit quaternions. Performance is evaluated using the mean absolute error (MAE) of the estimated object center position ($x,y$) and depth ($z$), together with the mean geodesic orientation error (Rot. Err.). Center-position errors are computed as
$
\Delta x=\frac{(x_{pred}-x_{gt})z_{pred}}{f_x},\qquad
\Delta y=\frac{(y_{pred}-y_{gt})z_{pred}}{f_y},
$
where $f_x$ and $f_y$ are the camera focal lengths.
The model was trained using a batch size of 16, learning rate of $10^{-3}$, 20 epochs, and four data-loading workers. Table~\ref{tab:pose_results} reports the average 6D pose estimation performance across all motion scenarios for each sensor and dataset.

\begin{table*}[!t]
\centering
\caption{Average 6D object pose estimation results using the ResNet-50 regression baseline. MAE$_x$, MAE$_y$, and MAE$_z$ denote the mean absolute errors (cm) in the estimated object center position and depth, respectively. Rot. Err. denotes the mean geodesic orientation error (deg). Results are averaged over all scenarios.}
\label{tab:pose_results}
\begin{tabular}{|c||c|c|c|c||c|c|c|c|}
\hline
\multirow{2}{*}{\textbf{Sensor}}
&
\multicolumn{4}{c||}{\textbf{Moving6DPoSe-S}}
&
\multicolumn{4}{c|}{\textbf{Moving6DPoSe-R}}
\\
\cline{2-9}
&
\textbf{MAE$_x$}
&
\textbf{MAE$_y$}
&
\textbf{MAE$_z$}
&
\textbf{Rot. Err.}
&
\textbf{MAE$_x$}
&
\textbf{MAE$_y$}
&
\textbf{MAE$_z$}
&
\textbf{Rot. Err.}
\\
\hline
ROG Eye S &
0.987 &
1.040 &
10.69 &
72.40 &
1.078 &
1.013 &
8.97 &
76.37
\\ \hline

DAVIS346 &
1.197 &
1.300 &
10.16 &
112.11 &
1.483 &
1.405 &
10.79 &
115.84
\\ \hline

EVK4 &
1.317 &
1.253 &
13.32 &
99.08 &
2.010 &
1.288 &
13.53 &
103.04
\\ \hline

\end{tabular}
\end{table*}

\section{Hardware specifications}

The Moving6DPoSe pipeline was executed on three computing platforms for simulation, annotation, and baseline evaluation. Four Apple Mac Mini M2 Pro servers (10-core CPU, 16-core GPU) were used to render synthetic sequences in Blender. A cluster equipped with NVIDIA Tesla T4 GPUs (16 GB VRAM) was used for dataset annotation and event simulation for the DAVIS346 and EVK4 cameras. Finally, a DGX-1 cluster with 8 NVIDIA V100 GPUs (32 GB VRAM each) was used for event simulation, data generation, and baseline training.

\section{Conclusions and Future Work}
\label{sec}

We presented Moving6DPoSe, a new multimodal dataset of moving objects acquired with frame, depth and event-based vision sensors. The dataset comprises paired real and synthetic subsets (Moving6DPoSe-R and Moving6DPoSe-S) and provides annotations for semantic segmentation, object detection, and monocular 6D pose estimation. 
Baseline experiments show that Moving6DPoSe presents challenging scenarios for semantic segmentation, object detection, and monocular 6D pose estimation. The monocular frame-based camera achieved the lowest orientation errors under the adopted baseline, while the event-based modalities yielded higher errors. Moreover, several objects (e.g., the pillow, ball, and hat) exhibit geometric symmetries, making multiple orientations visually indistinguishable. Consequently, quaternion-based orientation errors are substantially inflated despite similar object appearances, showing the ambiguity of symmetric objects rather than a limitation of the dataset.

Future work will extend both the real and synthetic subsets with additional objects, motion scenarios, and sensing conditions to support more challenging tasks, including 3D reconstruction, robotic manipulation, human-robot interaction, and sim-to-real learning. We also plan to incorporate additional samples, annotations and stronger multimodal baselines to further expand the database.

\section*{Acknowledgment}
This work was partially supported by the FONDEQUIP Project EQM170041.
A special mention to Isao Parra-Tsunekawa (Advanced Mining Technology Center) for the senior advice and Samuel Bugueno-Cordova (University of Chile) for the data collection support. 

%
%
\bibliographystyle{splncs04}
\bibliography{main}
\end{document}